\documentclass[10pt,twocolumn,letterpaper]{article}

\usepackage{arxiv}
\usepackage{times}
\usepackage{epsfig}
\usepackage{graphicx}
\usepackage{amsmath}
\usepackage{amssymb}
\usepackage{bm}
\usepackage{multirow}
\usepackage{xcolor}
\usepackage{tabulary}
\usepackage{amssymb}%
\usepackage{pifont}%
\newcommand{\cmark}{\ding{52}}%
\newcommand{\xmark}{\ding{56}}%

\newcommand{\bd}[1]{\textbf{#1}}
\newcommand{\app}{\raise.17ex\hbox{$\scriptstyle\sim$}}

\newcommand{\mypar}[1]{\smallskip\noindent {\bf #1}\enskip}

\newcolumntype{x}[1]{>{\centering\arraybackslash}p{#1pt}}

\newlength\savewidth\newcommand\shline{\noalign{\global\savewidth\arrayrulewidth
  \global\arrayrulewidth 1pt}\hline\noalign{\global\arrayrulewidth\savewidth}}
\newcommand{\tablestyle}[2]{\setlength{\tabcolsep}{#1}\renewcommand{\arraystretch}{#2}\centering\footnotesize}
\makeatletter\renewcommand\paragraph{\@startsection{paragraph}{4}{\z@}
  {.5em \@plus1ex \@minus.2ex}{-.5em}{\normalfont\normalsize\bfseries}}\makeatother

\usepackage[pagebackref=true,breaklinks=true,letterpaper=true,colorlinks,bookmarks=false]{hyperref}

\begin{document}

\title{Fast Online Object Tracking and Segmentation: A Unifying Approach}

\author{Qiang Wang\thanks{Equal contribution.}\\
CASIA\\
{\tt\small qiang.wang@nlpr.ia.ac.cn}
\and
Li Zhang$^*$\\
University of Oxford\\
{\tt\small lz@robots.ox.ac.uk}
\and
Luca Bertinetto$^*$\\
FiveAI\\
{\tt\small luca@robots.ox.ac.uk}
\and
Weiming Hu\\
CASIA\\
{\tt\small wmhu@nlpr.ia.ac.cn}
\and
Philip H.S. Torr\\
University of Oxford\\
{\tt\small philip.torr@eng.ox.ac.uk}
}

\maketitle
\begin{abstract}
\noindent 
In this paper we illustrate how to perform both visual object tracking and semi-supervised video object segmentation, in real-time, with a single simple approach.
Our method, dubbed SiamMask, improves the offline training procedure of popular fully-convolutional Siamese approaches for object tracking by augmenting their loss with a binary segmentation task.
Once trained, SiamMask solely relies on a single bounding box initialisation and operates online, producing class-agnostic object segmentation masks and rotated bounding boxes at 55 frames per second.
Despite its simplicity, versatility and fast speed, our strategy allows us to establish a new state of the art among real-time trackers on VOT-2018, while at the same time demonstrating competitive performance and the best speed for the semi-supervised video object segmentation task on DAVIS-2016 and DAVIS-2017. 
The project website is \url{http://www.robots.ox.ac.uk/~qwang/SiamMask}.
\end{abstract}

\section{Introduction}
Tracking is a fundamental task in any video application requiring some degree of reasoning about objects of interest, as it allows to establish object correspondences between frames~\cite{makovski2008visual}. 
It finds use in a wide range of scenarios such as automatic surveillance, vehicle navigation, video labelling, human-computer interaction and activity recognition.
Given the location of an arbitrary target of interest in the first frame of a video, the aim of \emph{visual object tracking} is to estimate its position in all the subsequent frames with the best possible accuracy~\cite{smeulders2014visual}.

\begin{figure}[t]
\centering
\setlength{\tabcolsep}{0.25ex}

\begin{tabular}
{ccc ccccc}
\includegraphics[trim={2cm 0cm 2cm 0cm},clip,width = 0.61in]{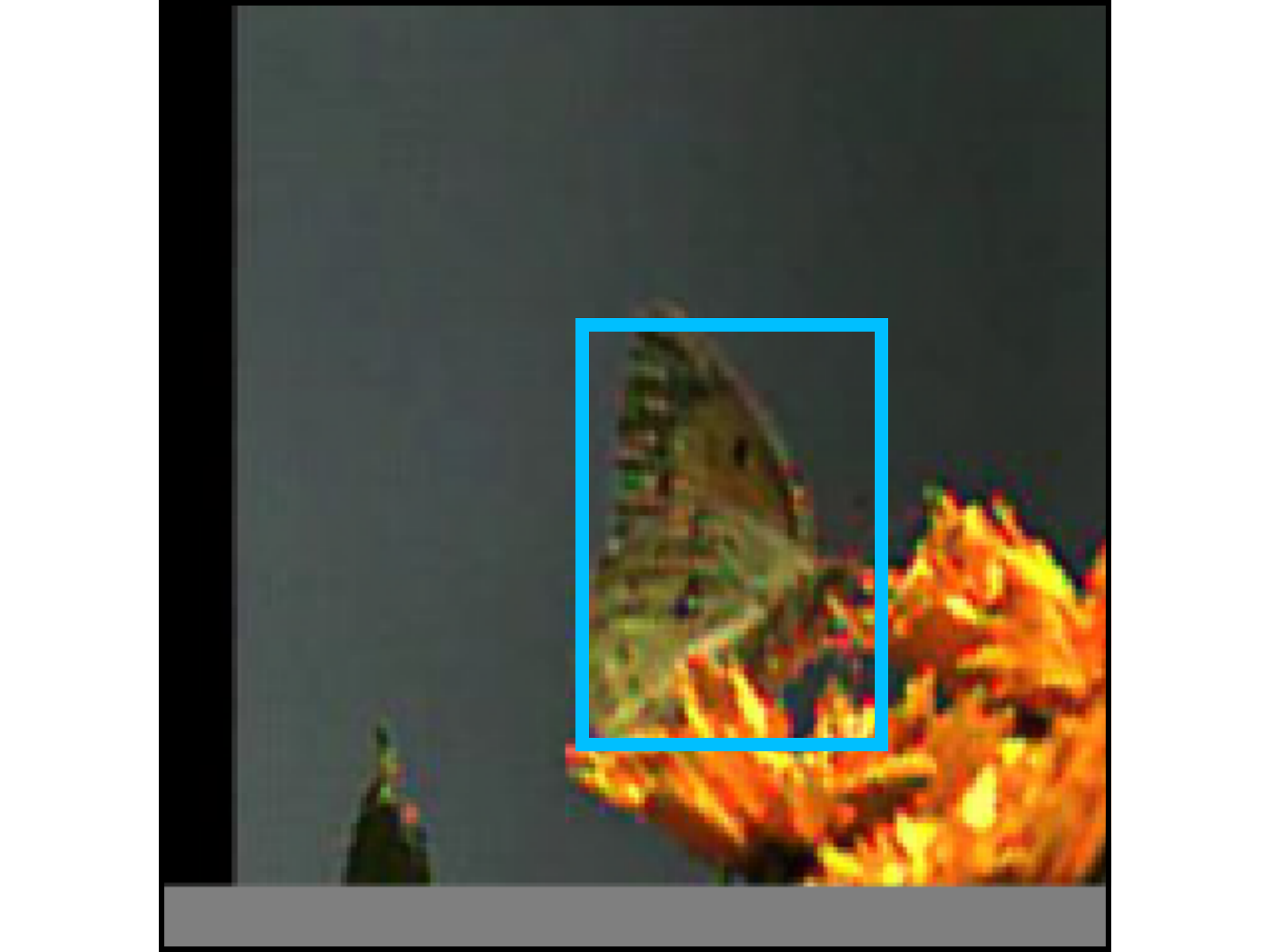}
&&&
&\includegraphics[trim={2cm 0cm 2cm 0cm},clip,width = 0.61in]{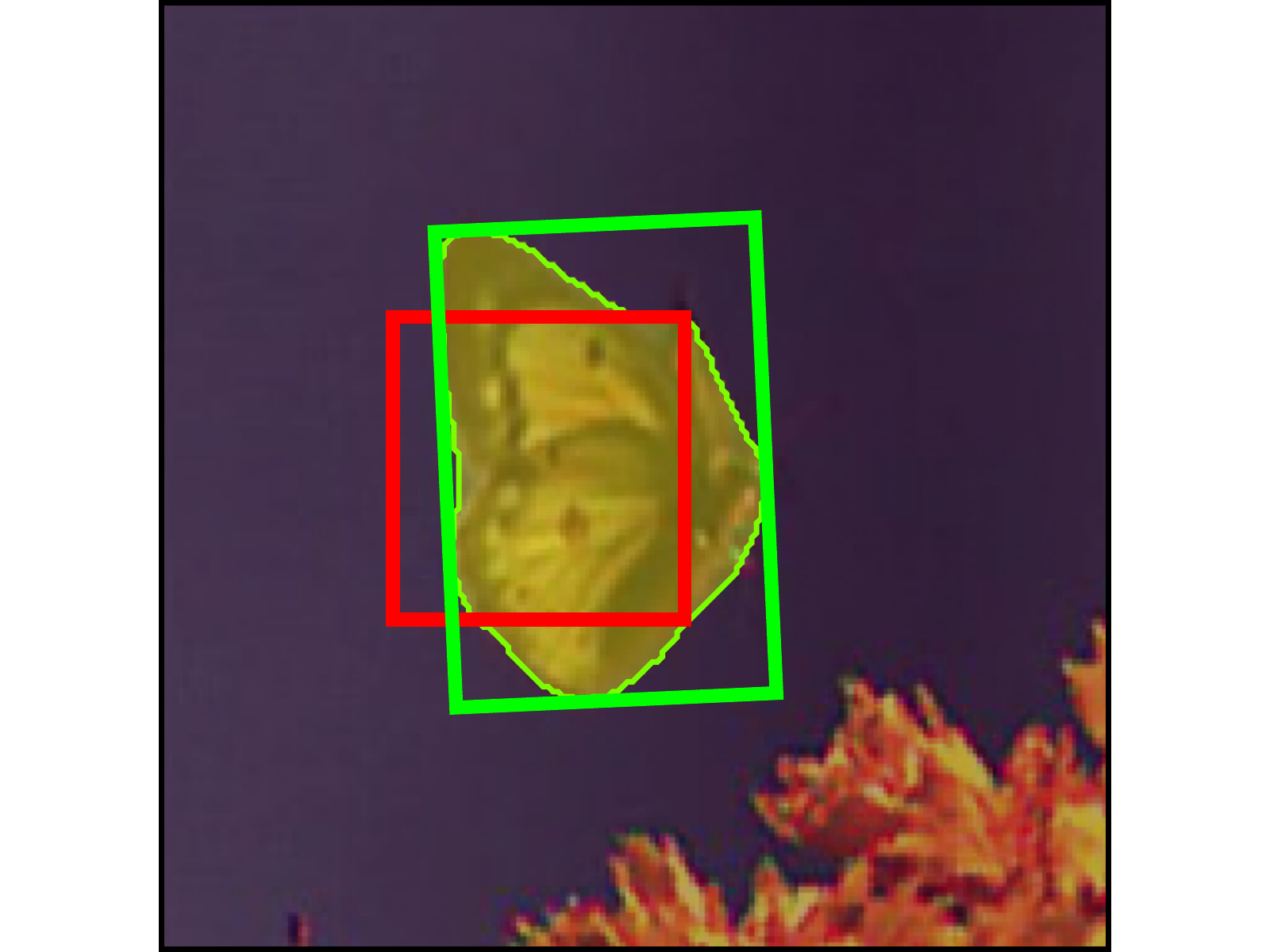}
& \includegraphics[trim={2cm 0cm 2cm 0cm},clip,width = 0.61in]{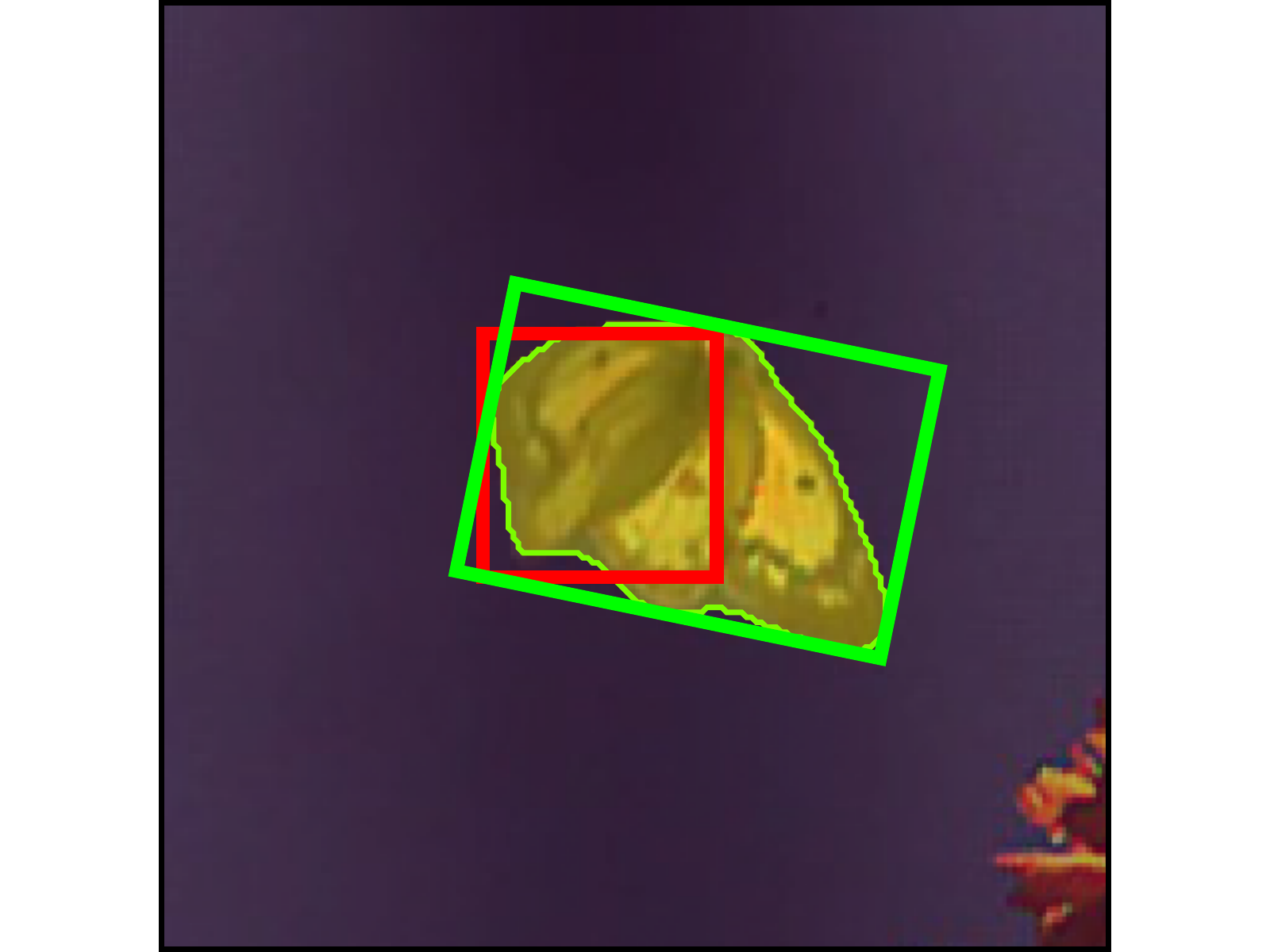}
& \includegraphics[trim={2cm 0cm 2cm 0cm},clip,width = 0.61in]{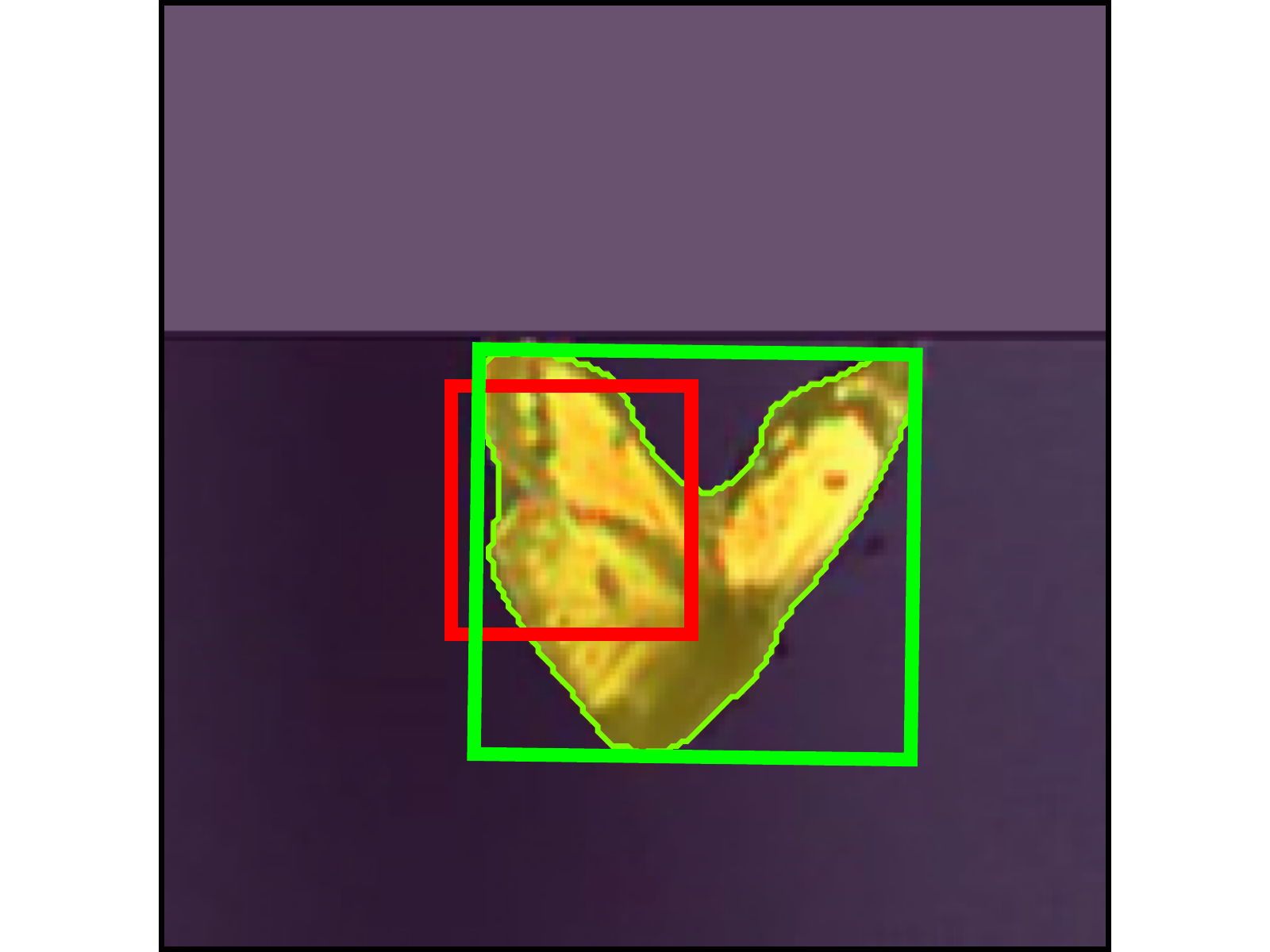}
& \includegraphics[trim={2cm 0cm 2cm 0cm},clip,width = 0.61in]{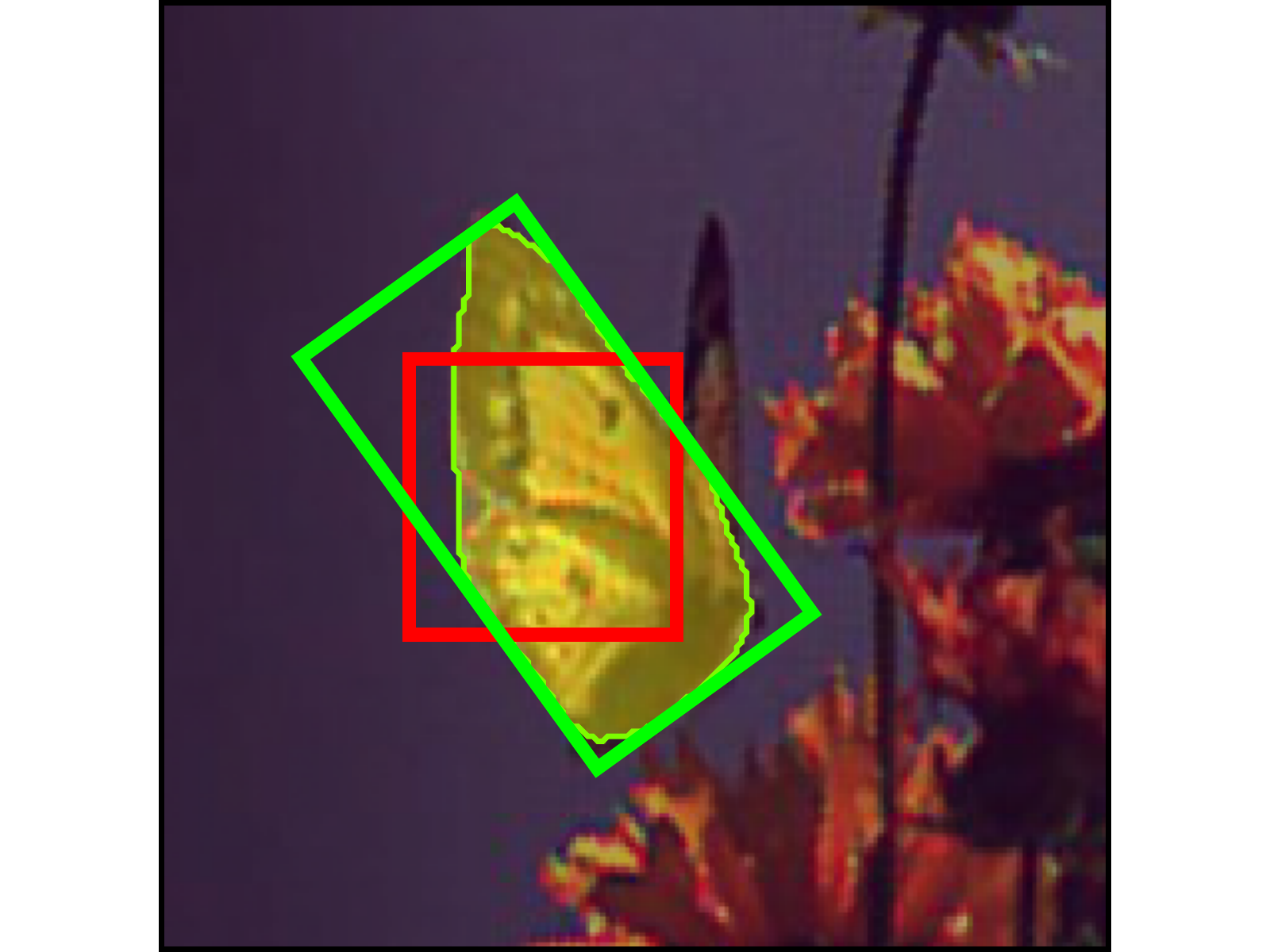}
\\

\includegraphics[trim={2cm 0cm 2cm 0cm},clip,width = 0.61in]{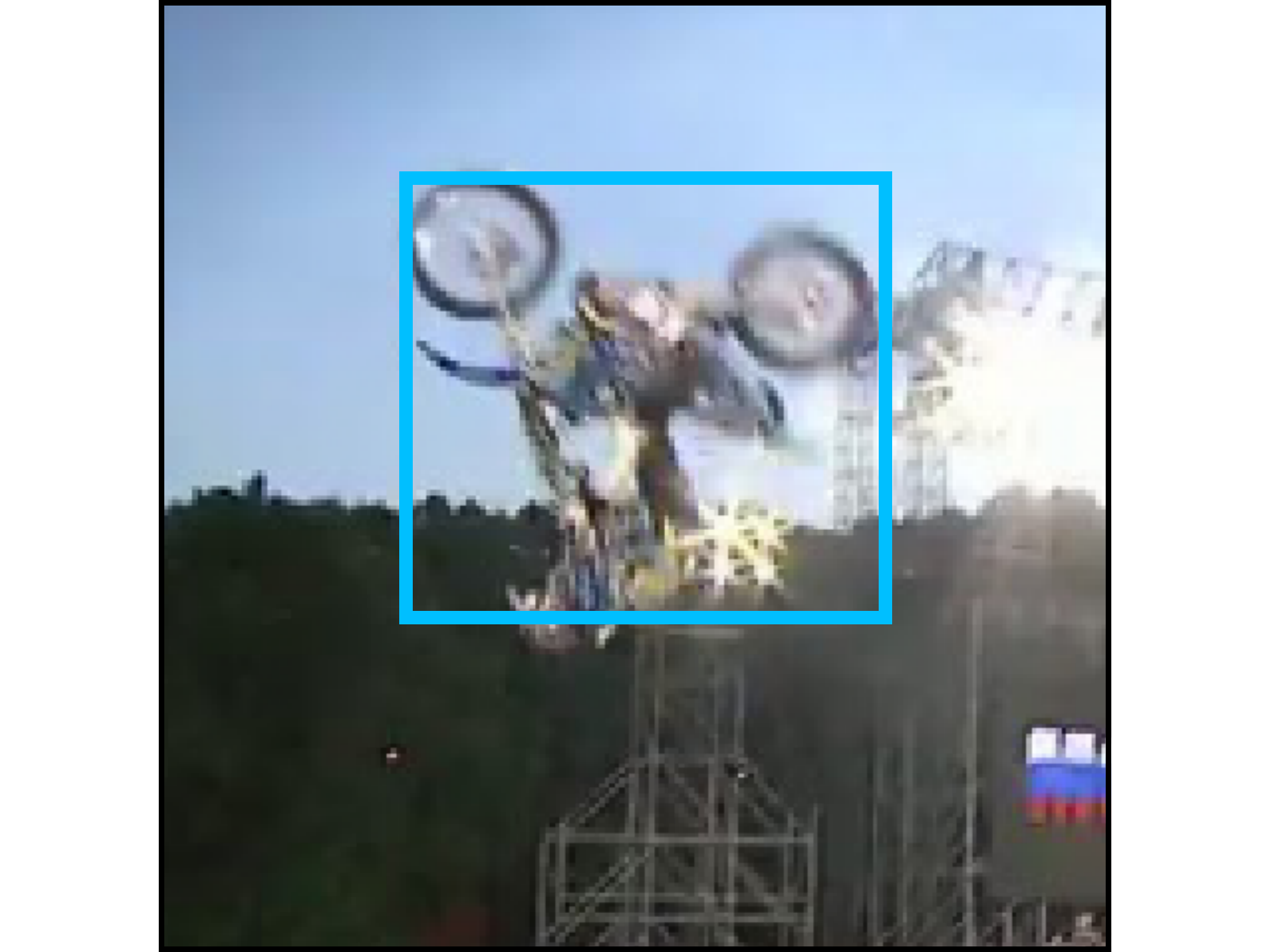}
&&&

&\includegraphics[trim={2cm 0cm 2cm 0cm},clip,width = 0.61in]{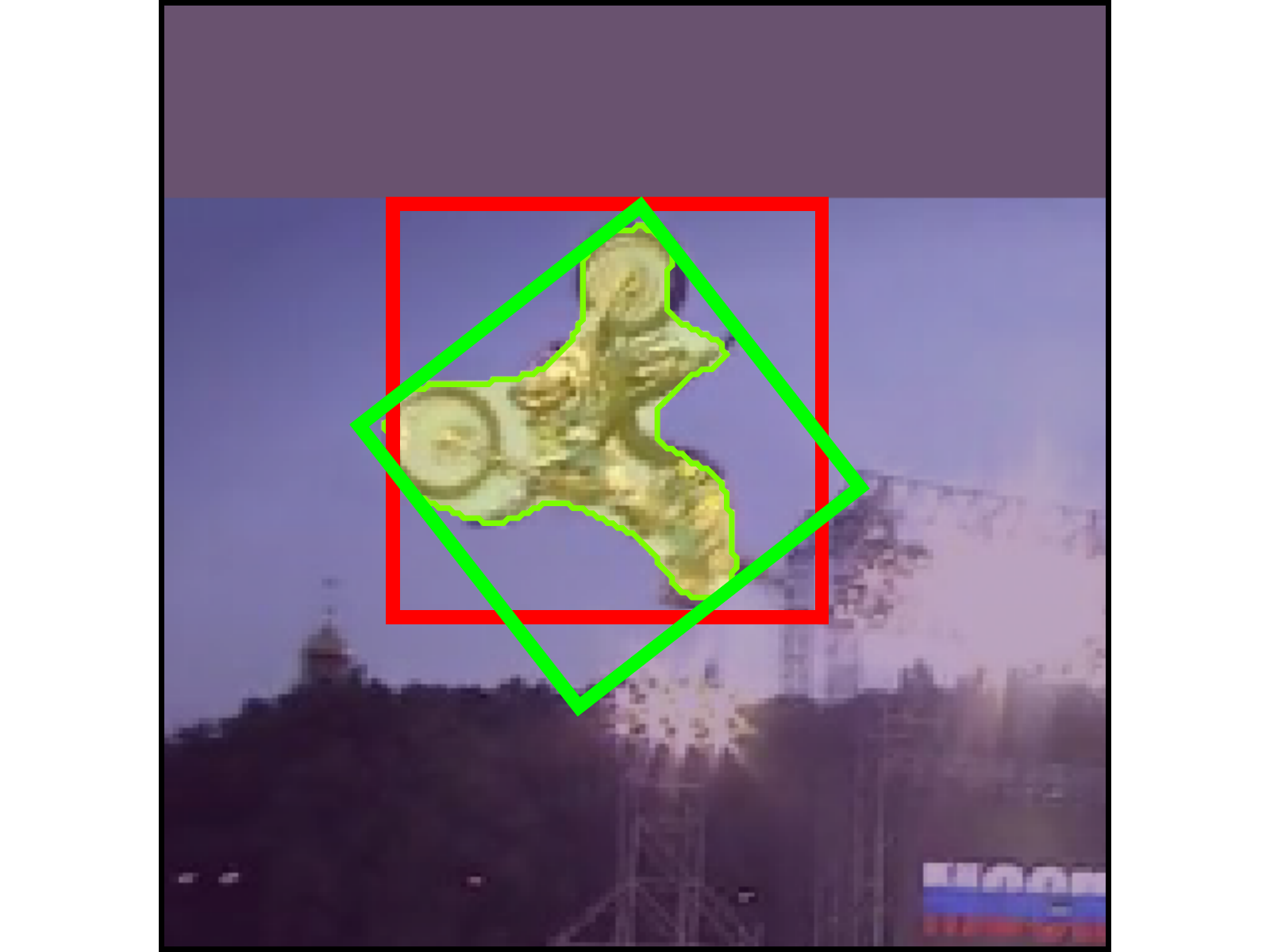}
& \includegraphics[trim={2cm 0cm 2cm 0cm},clip,width = 0.61in]{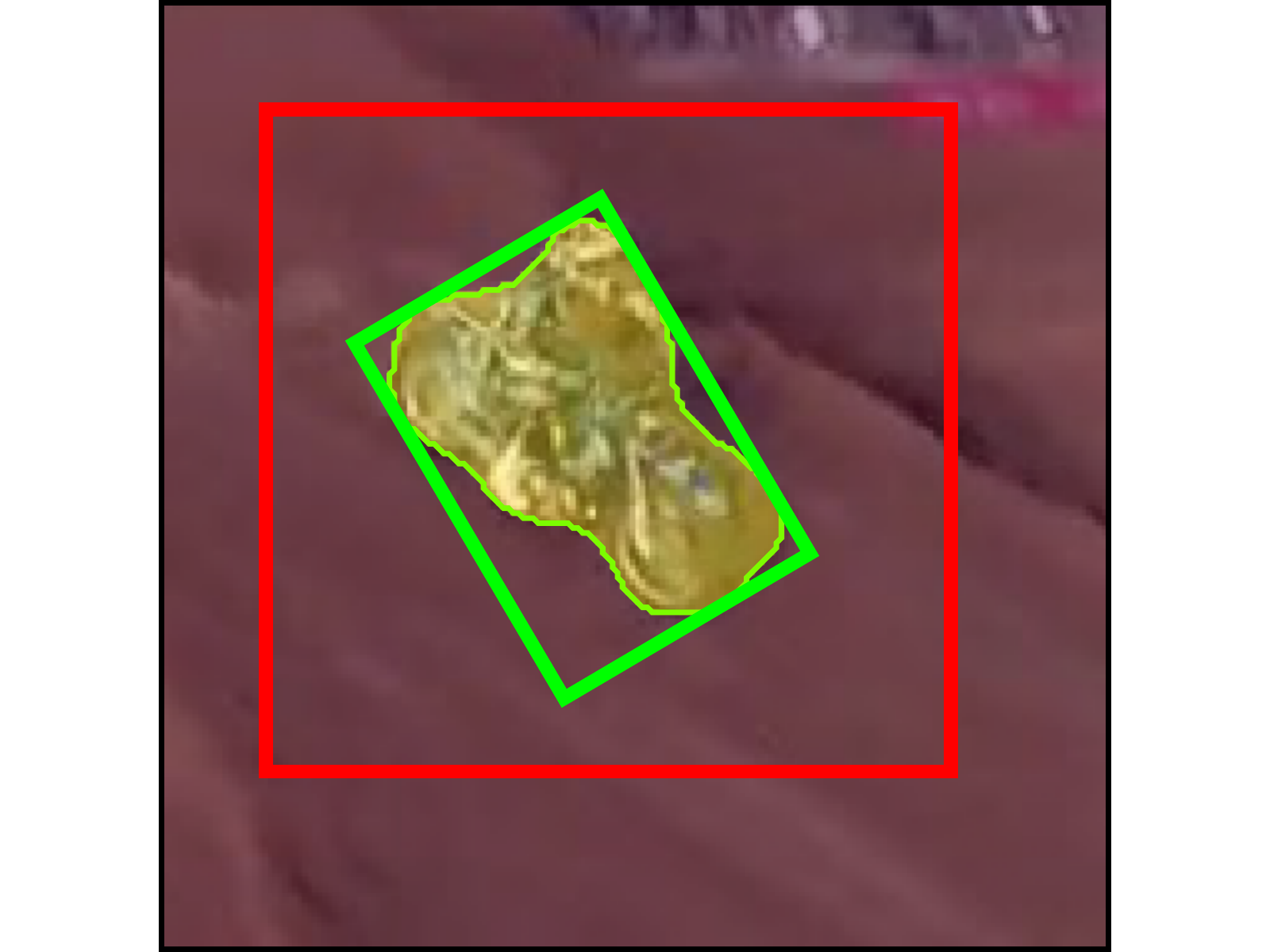}
& \includegraphics[trim={2cm 0cm 2cm 0cm},clip,width = 0.61in]{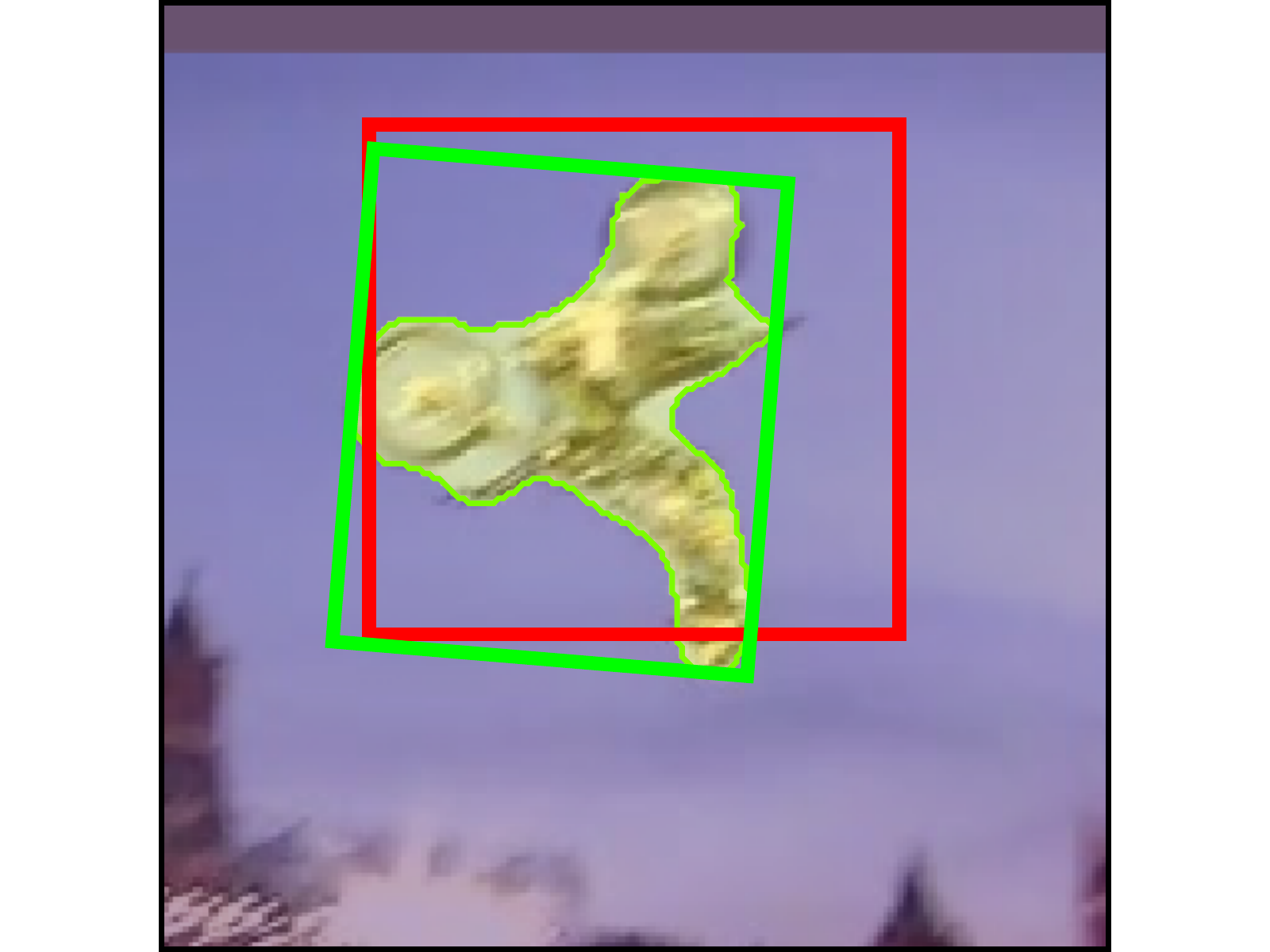}
& \includegraphics[trim={2cm 0cm 2cm 0cm}, clip,width = 0.61in]{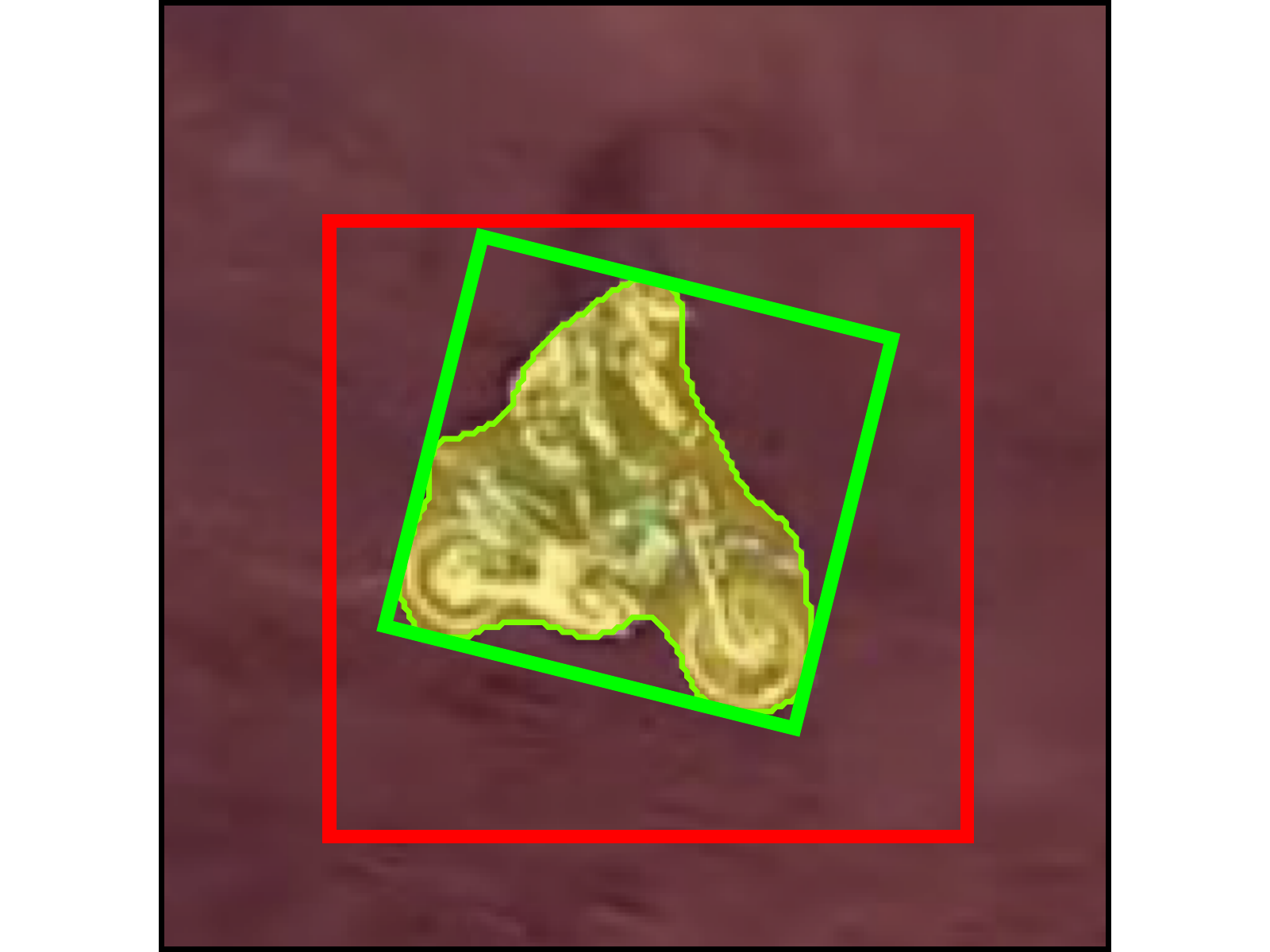}

\\

\includegraphics[trim={2cm 0cm 2cm 0cm},clip,width = 0.61in]{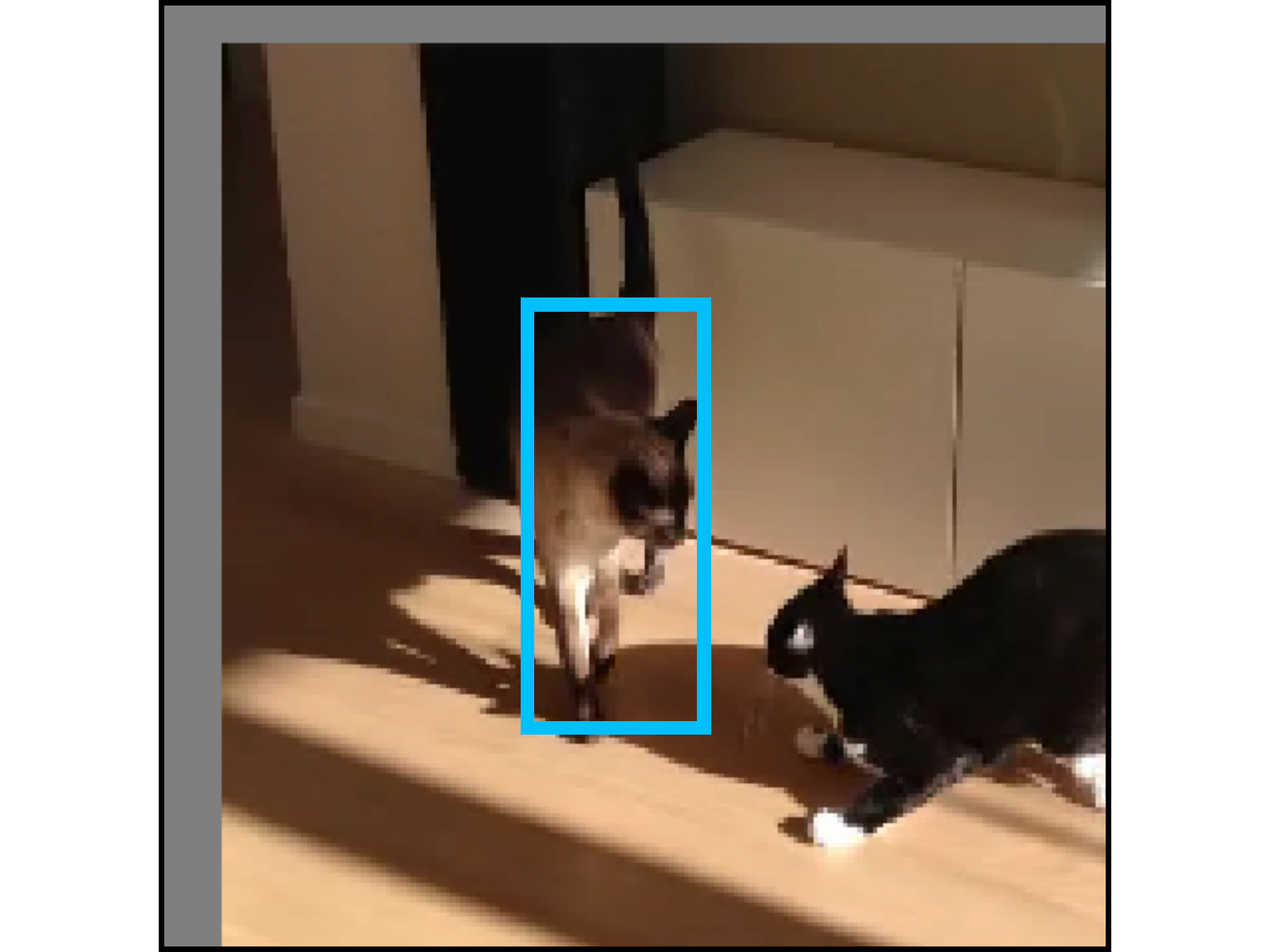}
&&&
&\includegraphics[trim={2cm 0cm 2cm 0cm},clip,width = 0.61in]{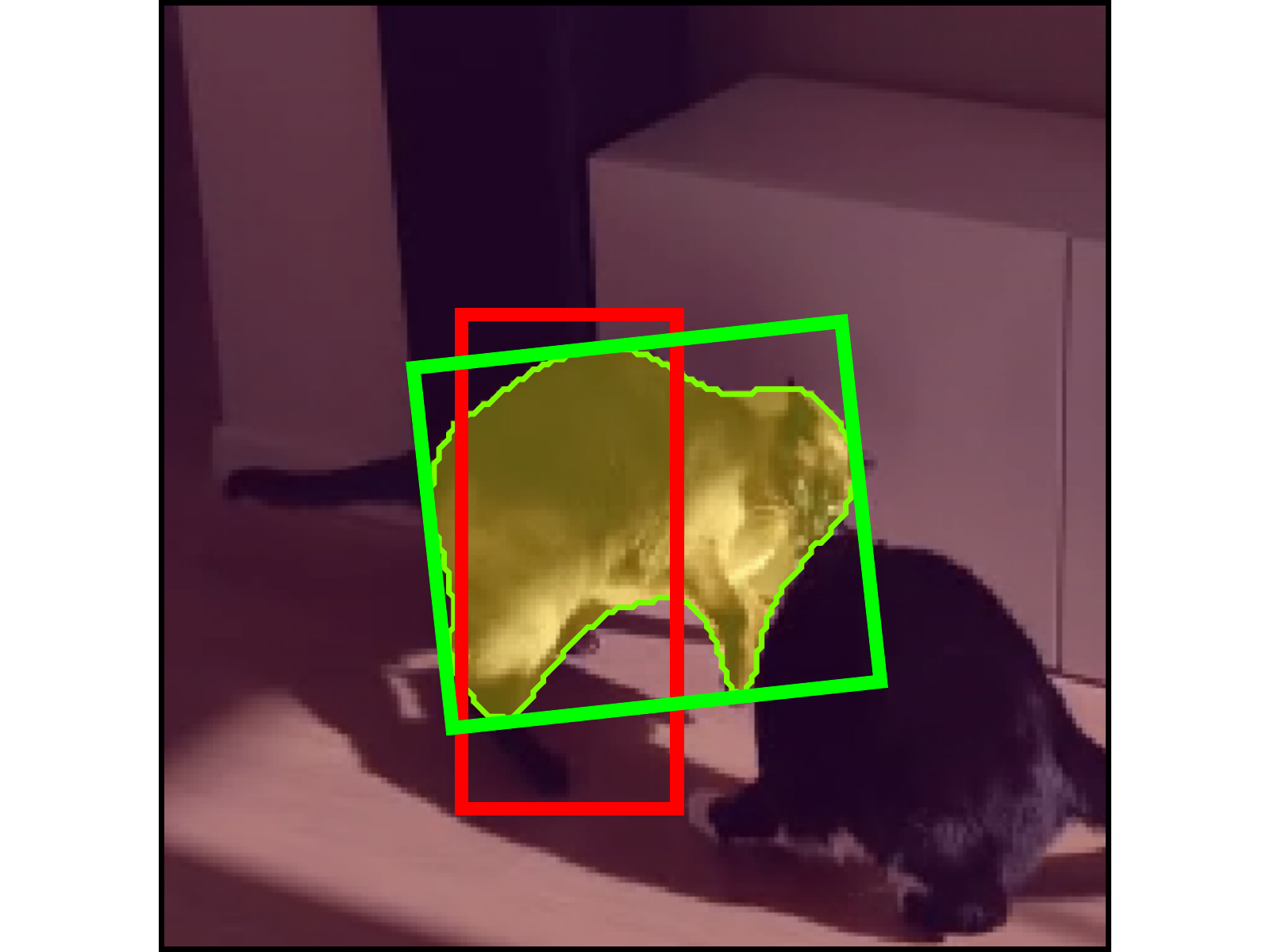}
& \includegraphics[trim={2cm 0cm 2cm 0cm},clip,width = 0.61in]{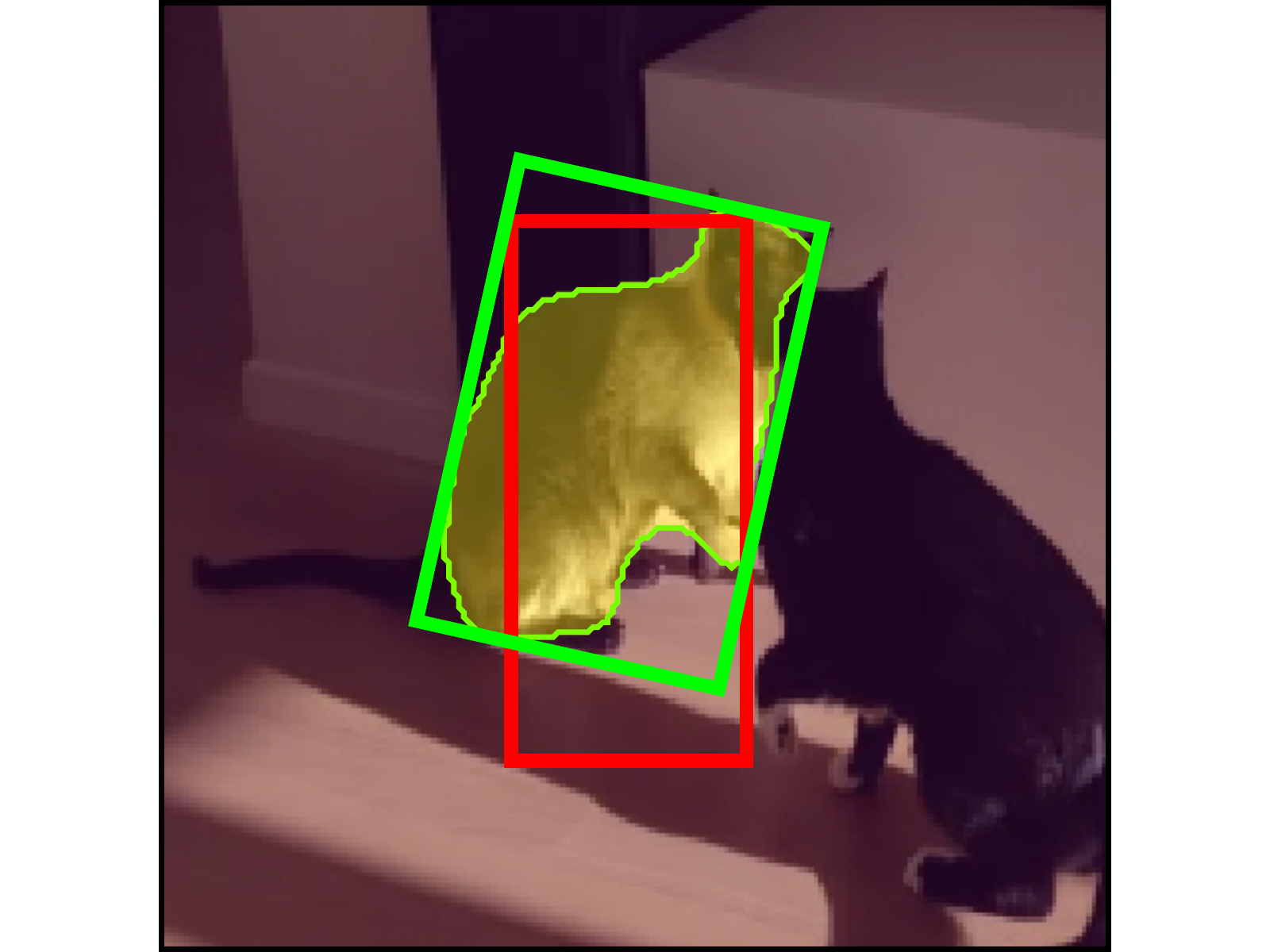}
& \includegraphics[trim={2cm 0cm 2cm 0cm},clip,width = 0.61in]{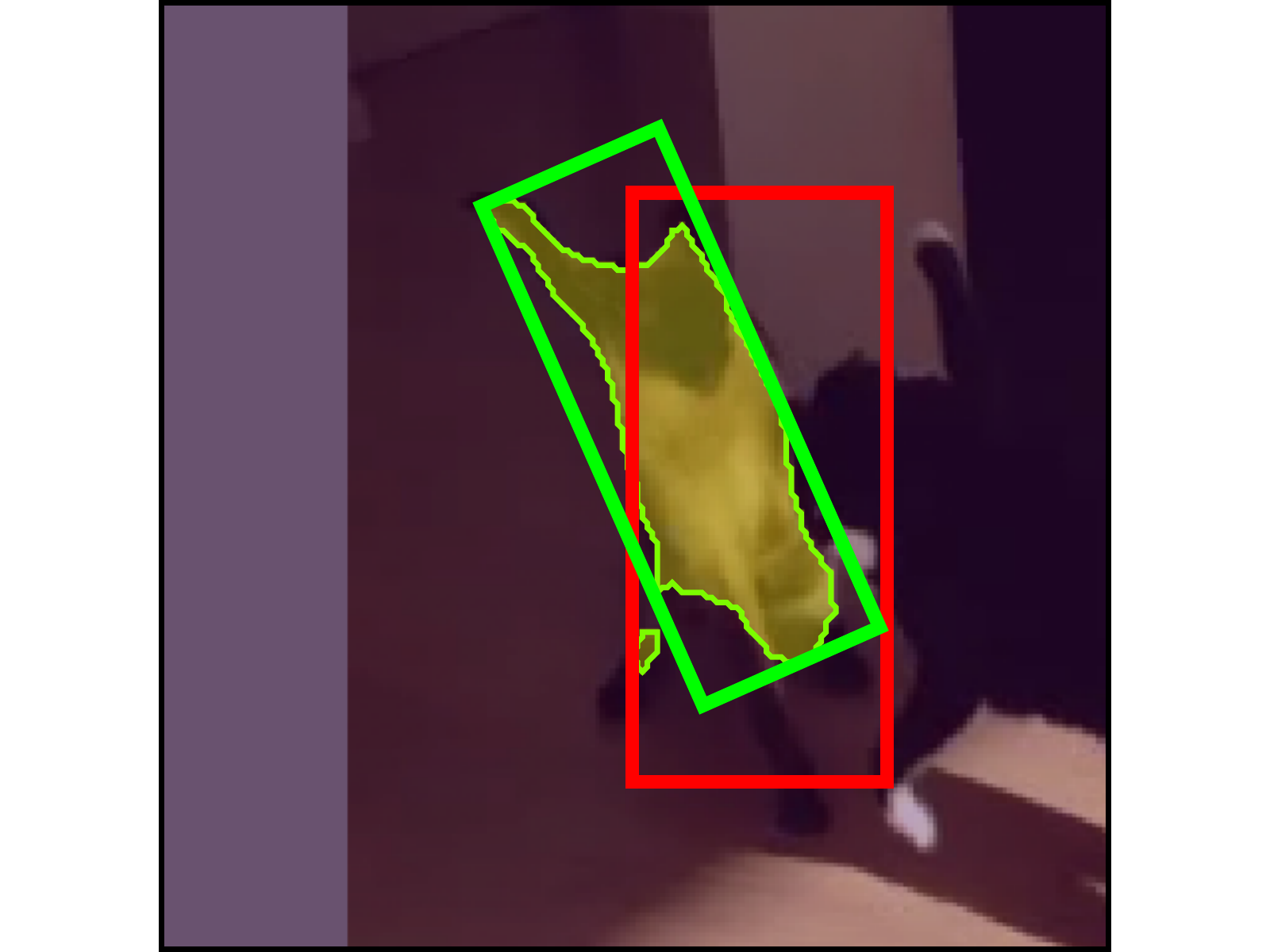}
& \includegraphics[trim={2cm 0cm 2cm 0cm},clip,width = 0.61in]{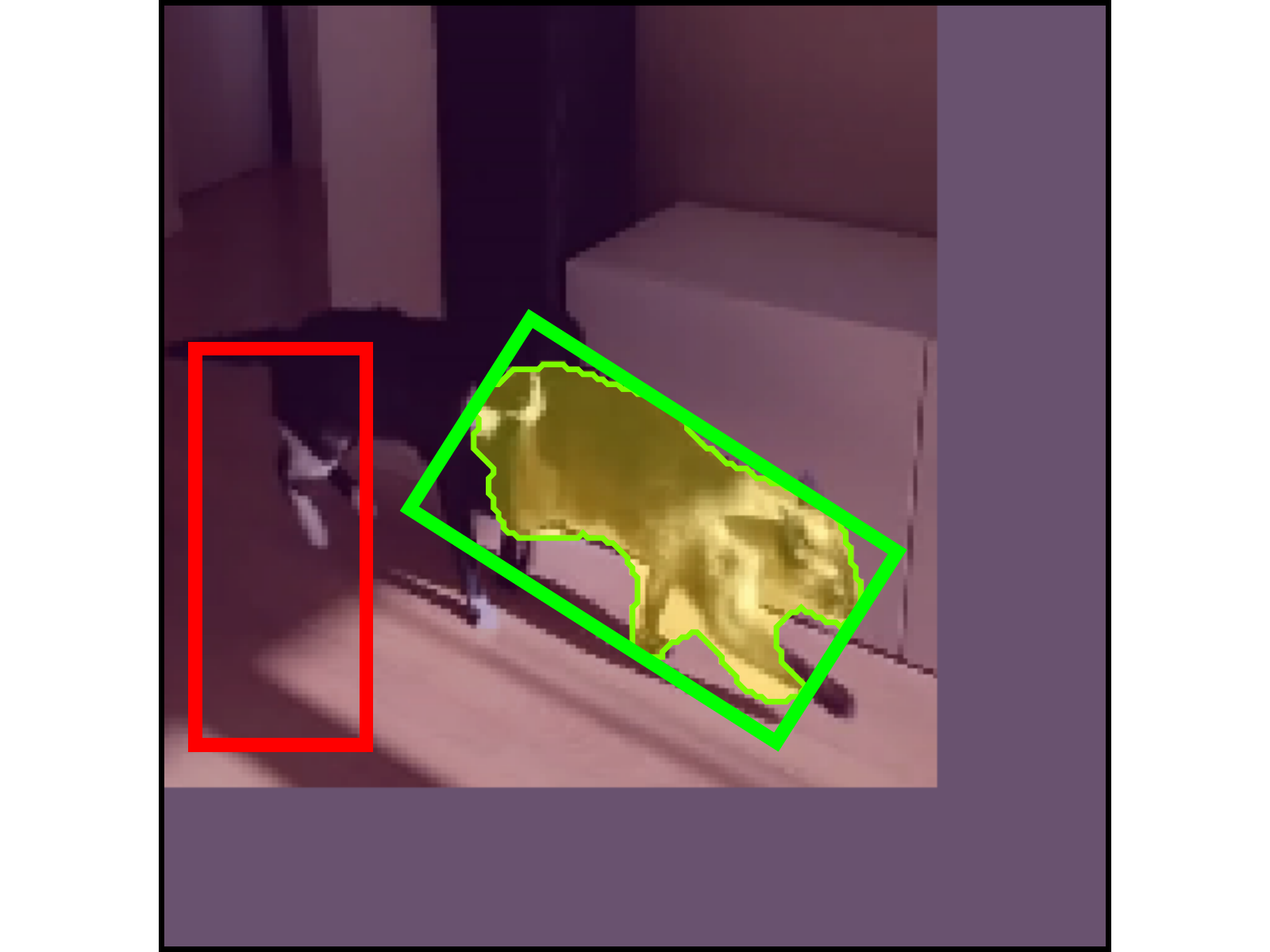}
\\

Init &
\multicolumn{7}{c}{Estimates}

\end{tabular}

\caption{
Our method aims at the intersection between the tasks of visual tracking and video object segmentation to achieve high practical convenience.
Like conventional object trackers, it relies on a simple bounding box initialisation (blue) and operates online.
Differently from state-of-the-art trackers such as ECO~\cite{danelljan2017eco} (red), SiamMask (green) is able to produce binary segmentation masks, which can more accurately describe the target object.
}
\label{fig:video_page1}
\vspace{-0.5cm}
\end{figure}

For many applications, it is important that tracking can be performed \emph{online}, while the video is streaming.
In other words, the tracker should not make use of future frames to reason about the current position of the object~\cite{kristan2016visual}.
This is the scenario portrayed by visual object tracking benchmarks, which represent the target object with a simple axis-aligned (\eg~\cite{wu2013online,valmadre2018long}) or rotated~\cite{kristan2016visual,VOT2018} bounding box.
Such a simple annotation helps to keep the cost of data labelling low; what is more, it allows a user to perform a quick and simple initialisation of the target.

Similar to object tracking, the task of semi-supervised \emph{video object segmentation} (VOS) requires estimating the position of an arbitrary target specified in the first frame of a video.
However, in this case the object representation consists of a binary segmentation mask which expresses whether or not a pixel belongs to the target~\cite{perazzi2016benchmark}.
Such a detailed representation is more desirable for applications that require pixel-level information, like video editing~\cite{perazzi2017video} and rotoscoping~\cite{miksik2017roam}.
Understandably, producing pixel-level estimates requires more computational resources than a simple bounding box.
As a consequence, VOS methods have been traditionally slow, often requiring several seconds per frame (\eg~\cite{wen2015jots,tsai2016video,perazzi2017learning,bao2018cnn}).
Very recently, there has been a surge of interest in faster approaches~\cite{Yang_2018_CVPR,marki2016bilateral,wug2018fast,cheng2018fast,chen2018blazingly,jampani2017video,hu2018videomatch}.
However, even the fastest still cannot operate in real-time.

In this paper, we aim at narrowing the gap between arbitrary object tracking and VOS by proposing \textit{SiamMask}, a simple multi-task learning approach that can be used to address \emph{both} problems.
Our method is motivated by the success of fast tracking approaches based on fully-convolutional Siamese networks~\cite{bertinetto2016fully} trained offline on millions of pairs of video frames (\eg \cite{SiamRPN,zhu2018distractor,he2018towards,yang2018learning}) and by the very recent availability of YouTube-VOS~\cite{xu2018youtube}, a large video dataset with pixel-wise annotations. 
We aim at retaining the offline trainability and online speed of these methods while at the same time significantly refining their representation of the target object, which is limited to a simple axis-aligned bounding box.

To achieve this goal, we simultaneously train a Siamese network on three tasks, each corresponding to a different strategy to establish correspondances between the target object and candidate regions in the new frames.
As in the fully-convolutional approach of Bertinetto \etal~\cite{bertinetto2016fully}, one task is to learn a measure of similarity between the target object and multiple candidates in a sliding window fashion.
The output is a dense response map which only indicates the location of the object, without providing any information about its spatial extent.
To refine this information, we simultaneously learn two further tasks: bounding box regression using a Region Proposal Network~\cite{ren2015faster,SiamRPN} and class-agnostic binary segmentation~\cite{DeepMask}.
Notably, binary labels are only required during offline training to compute the segmentation loss and \emph{not} online during segmentation/tracking. 
In our proposed architecture, each task is represented by a different branch departing from a shared CNN and contributes towards a final loss, which sums the three outputs together.

Once trained, SiamMask solely relies on a single bounding box initialisation, operates online without updates and produces object segmentation masks and rotated bounding boxes at 55 frames per second.
Despite its simplicity and fast speed, SiamMask establishes a new state-of-the-art on VOT-2018 for the problem of real-time object tracking.
Moreover, the \emph{same method} is also very competitive against recent semi-supervised VOS approaches on DAVIS-2016 and DAVIS-2017, while being the fastest by a large margin.
This result is achieved with a simple bounding box initialisation (as opposed to a mask) and without adopting costly techniques often used by VOS approaches such as fine-tuning~\cite{maninis2017video,perazzi2017learning,bao2018cnn,voigtlaender2017online}, data augmentation~\cite{LucidDataDreaming_CVPR17_workshops,li2018video} and optical flow~\cite{tsai2016video,bao2018cnn,perazzi2017learning,li2018video,cheng2018fast}.

The rest of this paper is organised as follows.
Section~\ref{sec:related} briefly outlines some of the most relevant prior work in visual object tracking and semi-supervised VOS; Section~\ref{sec:method} describes our proposal; Section~\ref{sec:experiments} evaluates it on four benchmarks and illustrates several ablative studies; Section~\ref{sec:conclusion} concludes the paper.

\section{ Related Work}
\label{sec:related}
In this section, we briefly cover the most representative techniques for the two problems tackled in this paper.

\mypar{Visual object tracking.}
Arguably, until very recently, the most popular paradigm for tracking arbitrary objects has been to train online a discriminative classifier exclusively from the ground-truth information provided in the first frame of a video (and then update it online).

In the past few years, the Correlation Filter, a simple algorithm that allows to discriminate between the template of an arbitrary target and its 2D translations, rose to prominence as particularly fast and effective strategy for tracking-by-detection thanks to the pioneering work of Bolme \etal~\cite{bolme2010visual}.
Performance of Correlation Filter-based trackers has then been notably improved with the adoption of multi-channel formulations~\cite{kiani2013multi,henriques2015tracking}, spatial constraints~\cite{kiani2015correlation,danelljan2015learning,lukezic2017discriminative,li2018learning} and deep features (\eg~\cite{danelljan2017eco,valmadre2017end}). 

Recently, a radically different approach has been introduced~\cite{bertinetto2016fully,held2016learning,tao2016siamese}. 
Instead of learning a discrimative classifier online, these methods train offline a similarity function on pairs of video frames.
At test time, this function can be simply evaluated on a new video, once per frame.
In particular, evolutions of the fully-convolutional Siamese approach~\cite{bertinetto2016fully} considerably improved tracking performance by making use of region proposals~\cite{SiamRPN}, hard negative mining~\cite{zhu2018distractor}, ensembling~\cite{he2018towards} and memory networks~\cite{yang2018learning}.

Most modern trackers, including all the ones mentioned above, use a rectangular bounding box both to initialise the target \emph{and} to estimate its position in the subsequent frames.
Despite its convenience, a simple rectangle often fails to properly represent an object, as it is evident in the examples of Figure~\ref{fig:video_page1}.
This motivated us to propose a tracker able to produce binary segmentation masks while still only relying on a bounding box initialisation.

Interestingly, in the past it was not uncommon for trackers to produce a coarse binary mask of the target object (\eg~\cite{comaniciu2000real,perez2002color}).
However, to the best of our knowledge, the only recent tracker that, like ours, is able to operate online and produce a binary mask starting from a bounding box initialisation is the superpixel-based approach of Yeo \etal~\cite{yeo2017superpixel}.
However, at 4 frames per seconds (fps), its fastest variant is significantly slower than our proposal.
Furthermore, when using CNN features, its speed is affected by a 60-fold decrease, plummeting below 0.1 fps.
Finally, it has not demonstrated to be competitive on modern tracking or VOS benchmarks.
Similar to us, the methods of Perazzi \etal~\cite{perazzi2017learning} and Ci \etal~\cite{ci2018video} can also start from a rectangle and output per-frame masks.
However, they require fine-tuning at test time, which makes them slow.

\mypar{Semi-supervised video object segmentation.}
Benchmarks for arbitrary object tracking (\eg~\cite{smeulders2014visual,kristan2016visual,wu2013online}) assume that trackers receive input frames in a sequential fashion.
This aspect is generally referred to with the attributes \emph{online} or \emph{causal}~\cite{kristan2016visual}.
Moreover, methods are often focused on achieving a speed that exceeds the ones of typical video framerates~\cite{VOT2018}.
Conversely, semi-supervised VOS algorithms have been traditionally more concerned with an accurate representation of the object of interest~\cite{perazzi2017video,perazzi2016benchmark}.

In order to exploit consistency between video frames, several methods propagate the supervisory segmentation mask of the first frame to the temporally adjacent ones via graph labeling approaches (\eg~\cite{wen2015jots,perazzi2015fully,tsai2016video,marki2016bilateral,bao2018cnn}). 
In particular, Bao \etal~\cite{bao2018cnn} recently proposed a very accurate method that makes use of a spatio-temporal MRF in which temporal dependencies are modelled by optical flow, while spatial dependencies are expressed by a CNN.

Another popular strategy is to process video frames independently (\eg~\cite{maninis2017video,perazzi2017learning,voigtlaender2017online}), similarly to what happens in most tracking approaches.
For example, in OSVOS-S Maninis \etal~\cite{maninis2017video} do not make use of any temporal information.
They rely on a fully-convolutional network pre-trained for classification and then, at test time, they fine-tune it using the ground-truth mask provided in the first frame.
MaskTrack~\cite{perazzi2017learning} instead is trained from scratch on individual images, but it does exploit some form of temporality at test time by using the latest mask prediction and optical flow as additional input to the network.

Aiming towards the highest possible accuracy, at test time VOS methods often feature computationally intensive techniques such as fine-tuning~\cite{maninis2017video,perazzi2017learning,bao2018cnn,voigtlaender2017online}, data augmentation~\cite{LucidDataDreaming_CVPR17_workshops,li2018video} and optical flow~\cite{tsai2016video,bao2018cnn,perazzi2017learning,li2018video,cheng2018fast}.
Therefore, these approaches are generally characterised by low framerates and the inability to operate online.
For example, it is not uncommon for methods to require minutes~\cite{perazzi2017learning,cheng2017segflow} or even hours~\cite{tsai2016video,bao2018cnn} for videos that are just a few seconds long, like the ones of DAVIS.

Recently, there has been an increasing interest in the VOS community towards \emph{faster} methods~\cite{marki2016bilateral,wug2018fast,cheng2018fast,chen2018blazingly,jampani2017video,hu2018videomatch}.
To the best of our knowledge, the fastest approaches with a performance competitive with the state of the art are the ones of Yang \etal~\cite{Yang_2018_CVPR} and Wug \etal~\cite{wug2018fast}.
The former uses a meta-network ``modulator'' to quickly adapt the parameters of a segmentation network during test time, while the latter does not use any fine-tuning and adopts an encoder-decoder Siamese architecture trained in multiple stages.
Both these methods run below 10 frames per second, while we are more than six times faster and only rely on a bounding box initialisation.

\section{Methodology}
\label{sec:method}
To allow online operability and fast speed, we adopt the fully-convolutional Siamese framework~\cite{bertinetto2016fully}.
Moreover, to illustrate that our approach is agnostic to the specific fully-convolutional method used as a starting point (\eg~\cite{bertinetto2016fully,SiamRPN,zhu2018distractor,yang2018learning,he2018twofold}), we consider the popular SiamFC~\cite{bertinetto2016fully} and SiamRPN~\cite{SiamRPN} as two representative examples.
We first introduce them in Section~\ref{sec:fc} and then describe our approach in Section~\ref{sec:siammask}.

\subsection{Fully-convolutional Siamese networks}
\label{sec:fc}
\mypar{SiamFC.}
Bertinetto \etal~\cite{bertinetto2016fully} propose to use, as a fundamental building block of a tracking system, an offline-trained fully-convolutional Siamese network that compares an exemplar image $z$ against a (larger) search image $x$ to obtain a dense response map.
$z$ and $x$ are, respectively, a $w{\times}h$ crop centered on the target object  and a larger crop centered on the last estimated position of the target.
The two inputs are processed by the same CNN $f_{\theta}$, yielding two feature maps that are cross-correlated:
\begin{equation}\label{eq:cross}
g_{\theta}(z,~x) = f_{\theta}(z) \star f_{\theta}(x).
\end{equation}
In this paper, we refer to each spatial element of the response map (left-hand side of Eq.~\ref{eq:cross}) as \emph{response of a candidate window} (\textbf{RoW}).
For example, ~$g_{\theta}^{n}(z,~x)$, ~encodes a \textit{similarity} between the examplar $z$ and $n$-th candidate window in $x$.
For SiamFC, the goal is for the maximum value of the response map to correspond to the target location in the search area $x$.
Instead, in order to allow each RoW to encode richer information about the target object, we replace the simple cross-correlation of Eq.~\ref{eq:cross} with depth-wise cross-correlation~\cite{bertinetto2016learning} and produce a multi-channel response map.
SiamFC is trained offline on millions of video frames with the logistic loss~\cite[Section 2.2]{bertinetto2016fully}, which we refer to as $\mathcal{L}_{sim}$.

\mypar{SiamRPN.}
Li \etal~\cite{SiamRPN} considerably improve the performance of SiamFC by relying on a region proposal network (RPN)~\cite{ren2015faster, feichtenhofer2017detect}, which allows to estimate the target location with a bounding box of variable aspect ratio.
In particular, in SiamRPN each RoW encodes a set of $k$ anchor box proposals and corresponding object/background scores.
Therefore, SiamRPN outputs box predictions in parallel with classification scores.
The two output branches are trained using the smooth $L_{1}$ and the cross-entropy losses~\cite[Section 3.2]{SiamRPN}.
In the following, we refer to them as $\mathcal{L}_{box}$ and $\mathcal{L}_{score}$ respectively.

\begin{figure*}
\begin{center}
\includegraphics[width=1.0 \textwidth]{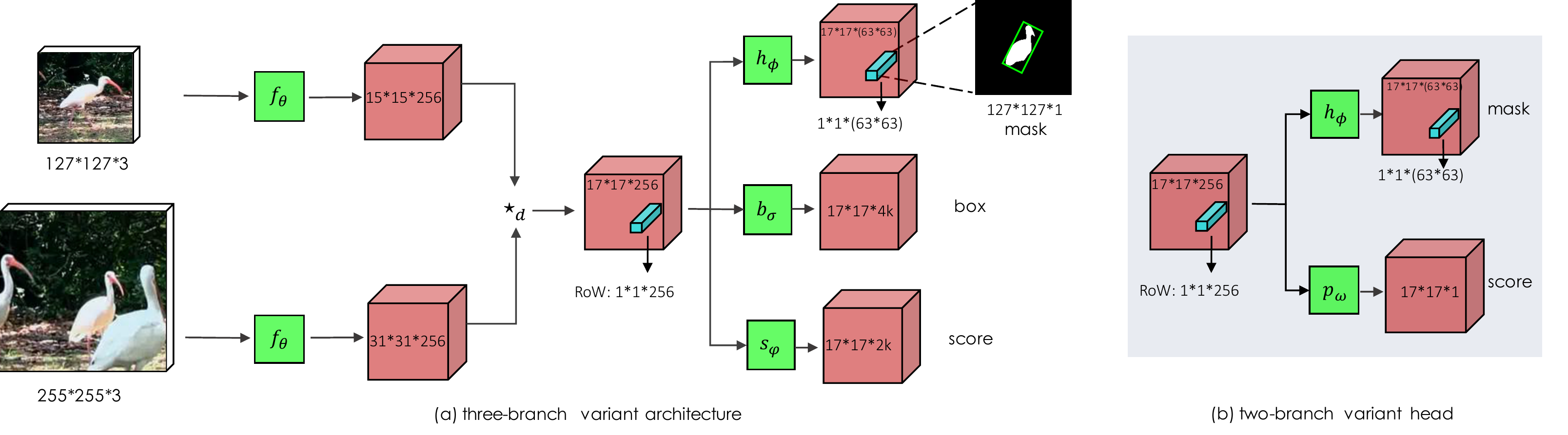}
\end{center}
\caption{Schematic illustration of SiamMask variants: (a) \textit{three-branch} architecture (full), (b) \textit{two-branch} architecture (head).  $\star_d$ denotes depth-wise cross correlation. For simplicity, upsampling layer and mask refinement module are omitted here and detailed in Appendix~\ref{sec:appendix_architecture}.}
\label{fig:schematic}
\vspace{-0.2cm}
\end{figure*}

\subsection{SiamMask}
\label{sec:siammask}
Unlike existing tracking methods that rely on low-fidelity object representations, we argue the importance of producing per-frame binary segmentation masks.
To this aim we show that, besides similarity scores and bounding box coordinates, it is possible for the RoW of a fully-convolutional Siamese network to also encode the information necessary to produce a pixel-wise binary mask.
This can be achieved by extending existing Siamese trackers with an extra branch and loss.

We predict $w{\times}h$ binary masks (one for each RoW) using a simple two-layers neural network $h_{\phi}$ with learnable parameters $\phi$.
Let $m_{n}$ denote the predicted mask corresponding to the $n$-th RoW,
\begin{equation}\label{eq:mask}
m_{n} = h_{\phi}(g_{\theta}^{n}(z,~x)).
\end{equation}
From Eq.~\ref{eq:mask} we can see that the mask prediction is a function of \emph{both} the image to segment $x$ and the target object in $z$.
In this way, $z$ can be used as a reference to guide the segmentation process:
given a different reference image, the network will produce a different segmentation mask for $x$.

\mypar{Loss function.}
During training, each RoW is labelled with a ground-truth binary label $y_{n} \in \{\pm 1\}$ and also associated with a pixel-wise ground-truth mask $c_n$ of size $w{\times}h$.
Let $c^{ij}_{n} \in \{\pm 1\}$ denote the label corresponding to pixel $(i,j)$ of the object mask in the $n$-th candidate RoW.
The loss function $\mathcal{L}_{mask}$ (Eq.~\ref{eq:loss}) for the mask prediction task is a binary logistic regression loss over all RoWs:

\begin{equation}\label{eq:loss}
		\mathcal{L}_{mask}(\theta,~\phi) =  \sum_{n} (\frac{1+y_{n}}{2wh}\sum_{ij} \log (1 + e^{-c^{ij}_{n}m_{n}^{ij}}
		)).
\end{equation}
Thus, the classification layer of $h_{\phi}$ consists of $w{\times}h$ classifiers, each indicating whether a given pixel belongs to the object in the candidate window or not.
Note that $\mathcal{L}_{mask}$ is considered only for positive RoWs (\ie with $y_{n}=1$).

\mypar{Mask representation.}
In contrast to semantic segmentation methods in the style of FCN~\cite{long2015fully} and Mask R-CNN~\cite{maskrcnn}, which maintain explicit spatial information throughout the network, our approach follows the spirit of~\cite{DeepMask,SharpMask} and generates masks starting from a flattened representation of the object.
In particular, in our case this representation corresponds to one of the ($17{\times}17$) RoWs produced by the depth-wise cross-correlation between $f_\theta(z)$ and $f_\theta(x)$.
Importantly, the network $h_\phi$ of the segmentation task is composed of two $1{\times}1$ convolutional layers, one with 256 and the other with $63^2$ channels (Figure~\ref{fig:schematic}).
This allows every pixel classifier to utilise information contained in the entire RoW and thus to have a complete view of its corresponding candidate window in $x$, which is critical to disambiguate between instances that look like the target (\eg last row of Figure~\ref{fig:davis16}), often referred to as distractors.
With the aim of producing a more accurate object mask, we follow the strategy of~\cite{SharpMask}, which merges low and high resolution features using multiple~\textit{refinement} modules made of upsampling layers and skip connections (see Appendix~\ref{sec:appendix_architecture}).

\mypar{Two variants.}
For our experiments, we augment the architectures of SiamFC~\cite{bertinetto2016fully} and SiamRPN~\cite{SiamRPN} with our segmentation branch and the loss $\mathcal{L}_{mask}$, obtaining what we call the \emph{two-branch} and \emph{three-branch} variants of SiamMask.
These respectively optimise the multi-task losses $\mathcal{L}_{2B}$ and $\mathcal{L}_{3B}$, defined as:
\begin{equation}
\label{eq:2b}
\mathcal{L}_{2B} = \lambda_{1} \cdot \mathcal{L}_{mask} + \lambda_{2} \cdot \mathcal{L}_{sim},
\end{equation}
\begin{equation}
\label{eq:3b}
\mathcal{L}_{3B} = \lambda_{1} \cdot \mathcal{L}_{mask} + \lambda_{2} \cdot \mathcal{L}_{score}+ \lambda_{3} \cdot \mathcal{L}_{box}.
\end{equation}
We refer the reader to~\cite[Section 2.2]{bertinetto2016fully} for $\mathcal{L}_{sim}$ and to~\cite[Section 3.2]{SiamRPN} for $\mathcal{L}_{box}$ and $\mathcal{L}_{score}$. 
For $\mathcal{L}_{3B}$, a RoW is considered positive ($y_{n} = 1$) if one of its anchor boxes has IOU with the ground-truth box of at least 0.6 and negative ($y_{n} = -1$) otherwise.
For $\mathcal{L}_{2B}$, we adopt the same strategy of~\cite{bertinetto2016fully} to define positive and negative samples.
We did not search over the hyperparameters of Eq.~\ref{eq:2b} and Eq.~\ref{eq:3b} and simply set $\lambda_1=32$ like in~\cite{DeepMask} and $\lambda_{2}=\lambda_{3}=1$.
The task-specific branches for the box and score outputs are constituted by two $1{\times}1$ convolutional layers.
Figure~\ref{fig:schematic} illustrates the two variants of SiamMask.

\begin{figure}[t]
\centering
\setlength{\tabcolsep}{0.25ex}

\begin{tabular}
{c cccc}
& \includegraphics[trim={2cm 0cm 2cm 0cm},clip,width = 0.79in]{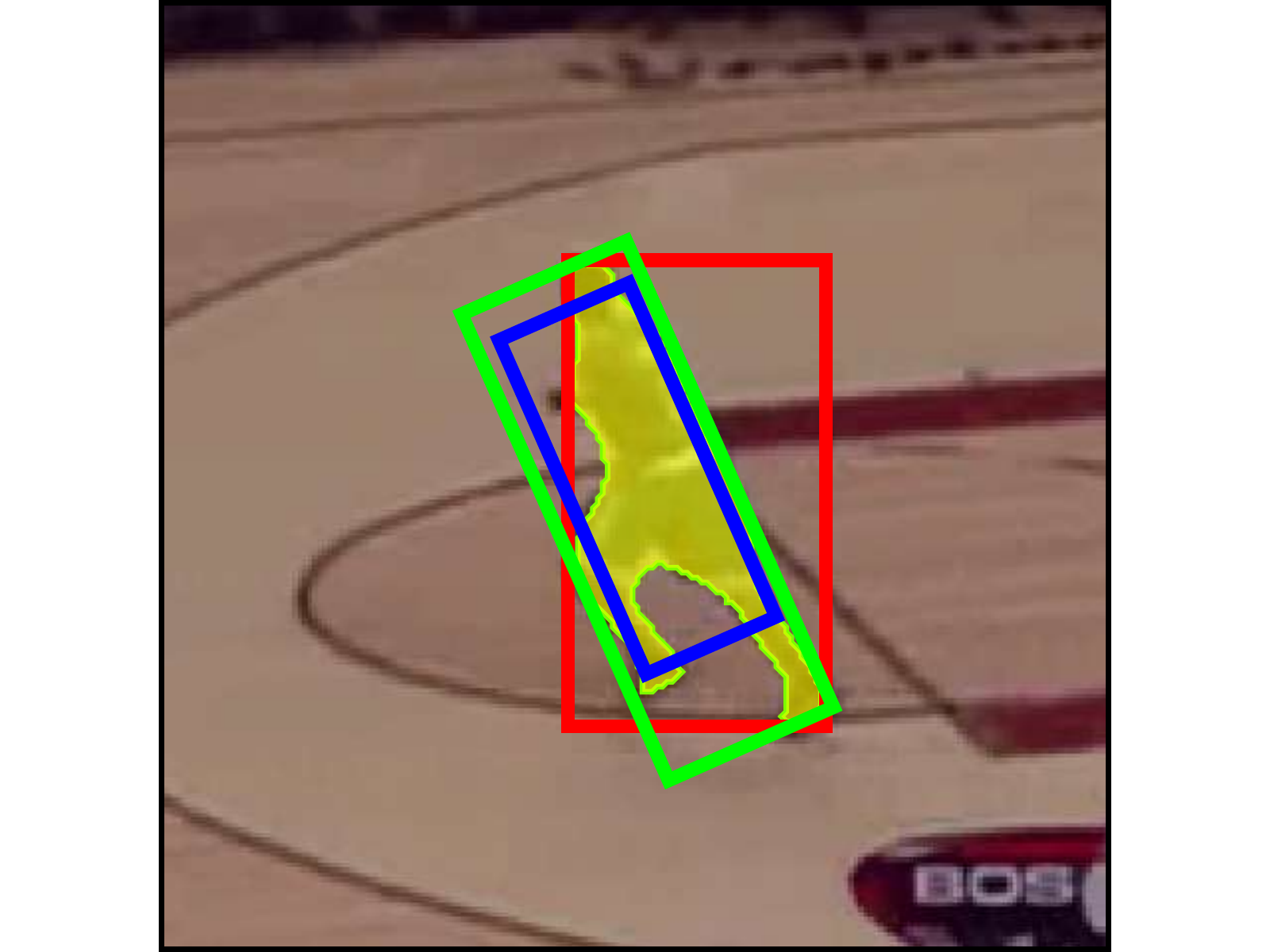}
& \includegraphics[trim={2cm 0cm 2cm 0cm},clip,width = 0.79in]{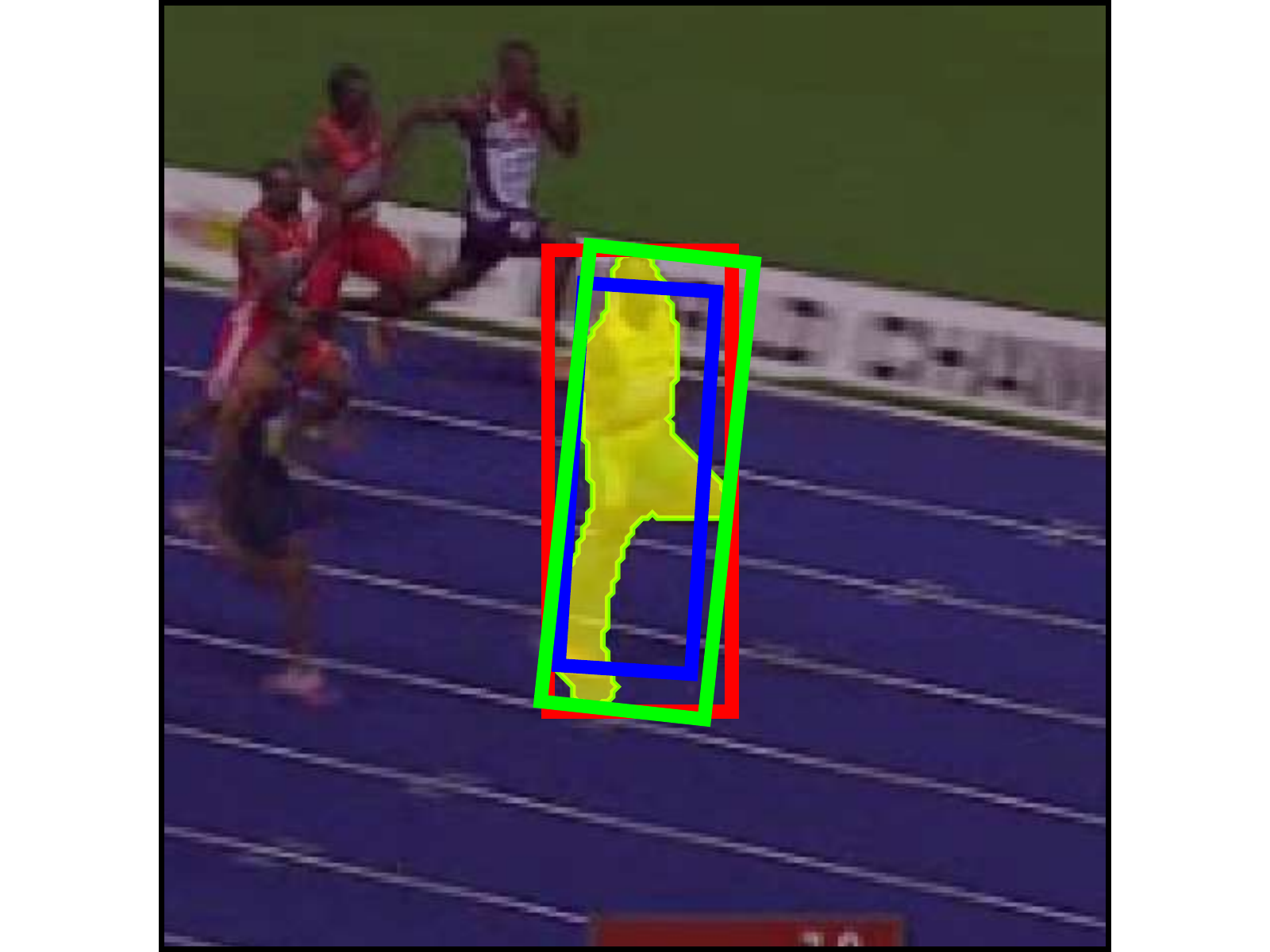}
& \includegraphics[trim={2cm 0cm 2cm 0cm},clip,width = 0.79in]{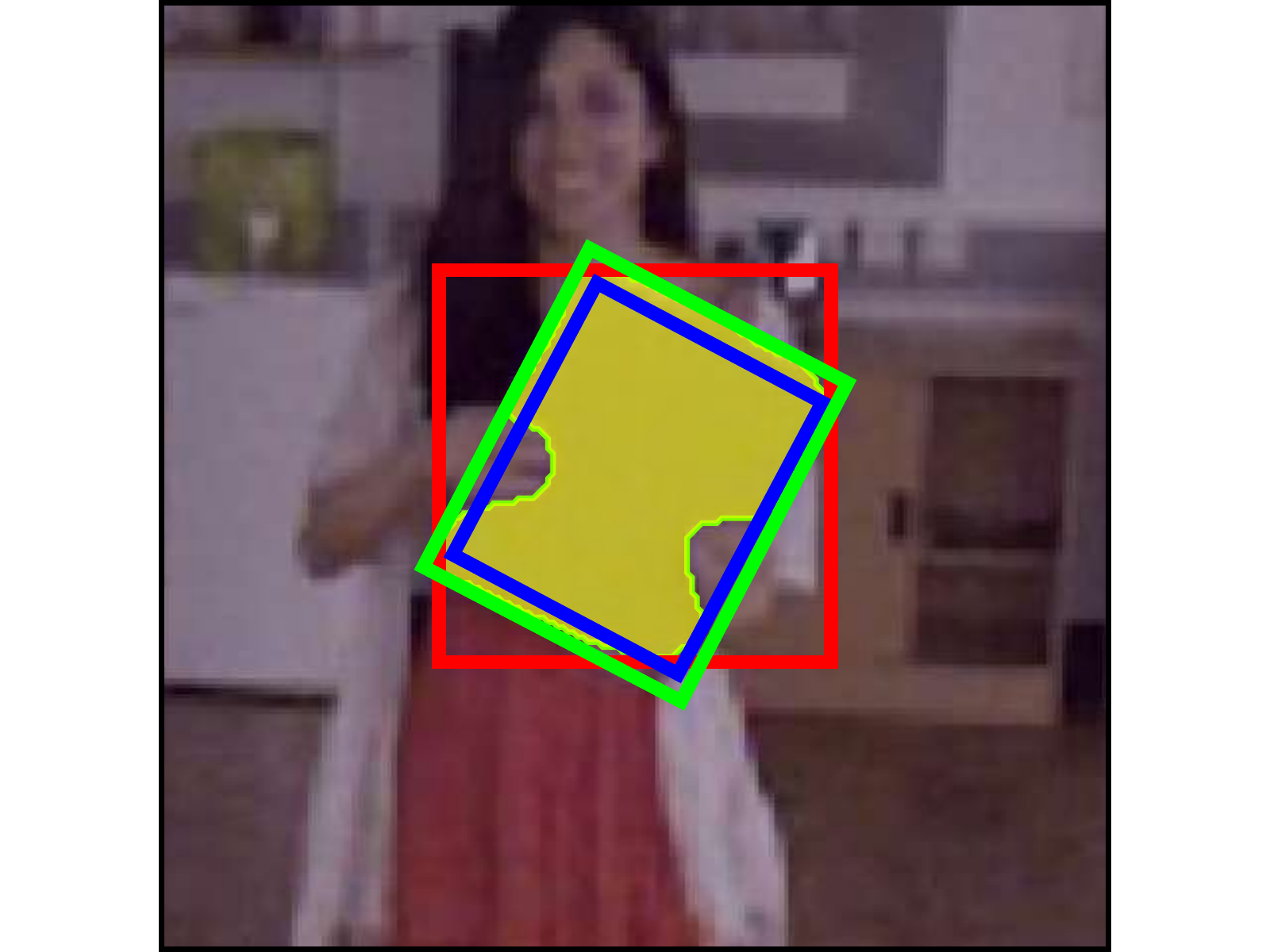}
& \includegraphics[trim={2cm 0cm 2cm 0cm},clip,width = 0.79in]{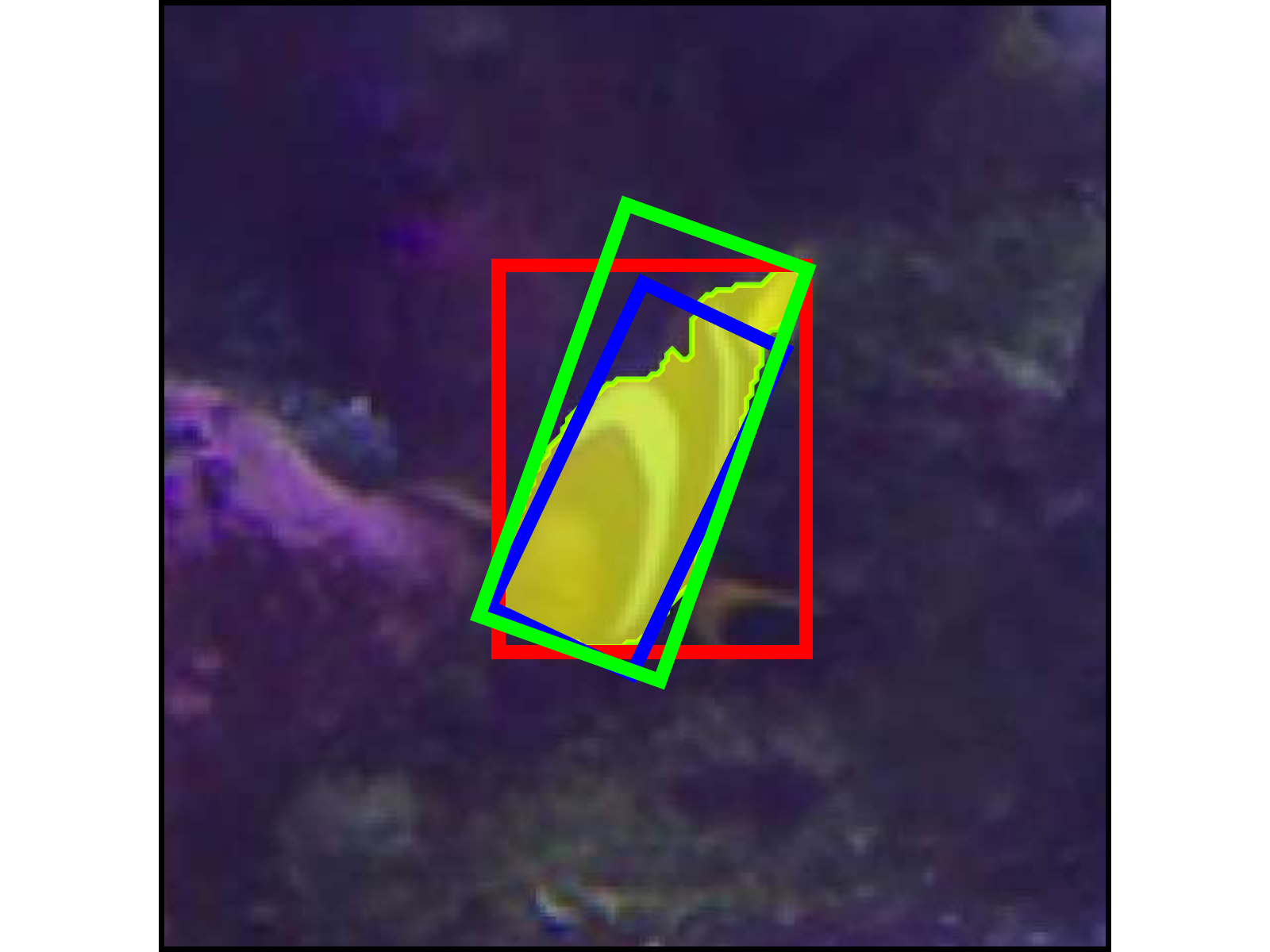}
\\

\end{tabular}

\caption{
In order to generate a bounding box from a binary mask (in yellow), we experiment with three different methods.
\textit{Min-max}: the axis-aligned rectangle containing the object (red); \textit{MBR}: the minimum bounding rectangle (green); \textit{Opt}: the rectangle obtained via the optimisation strategy proposed in VOT-2016~\cite{kristan2016visual} (blue).}
\vspace{-1em}
\label{fig:bbox}
\end{figure}

\mypar{Box generation.}
Note that, while VOS benchmarks require binary masks, typical tracking benchmarks such as VOT~\cite{kristan2016visual,VOT2018} require a bounding box as final representation of the target object.
We consider three different strategies to generate a bounding box from a binary mask (Figure~\ref{fig:bbox}):
(1) axis-aligned bounding rectangle (\emph{Min-max}), (2) rotated minimum bounding rectangle (\emph{MBR}) and (3) the optimisation strategy used for the automatic bounding box generation proposed in VOT-2016~\cite{kristan2016visual} (\emph{Opt}).
We empirically evaluate these alternatives in Section~\ref{sec:experiments} (Table~\ref{tab:iou}).

\subsection{Implementation details}
\mypar{Network architecture.}
For both our variants, we use a ResNet-50~\cite{he2016deep} until the final convolutional layer of the \mbox{$4$-th} stage as our backbone $f_\theta$.
In order to obtain a high spatial resolution in deeper layers, we reduce the output stride to $8$ by using convolutions with stride 1.
Moreover, we increase the receptive field by using dilated convolutions~\cite{chen2018deeplab}.
In our model, we add to the shared backbone $f_{\theta}$ an unshared \emph{adjust} layer ($1{\times}1$ \textit{conv} with $256$ outputs). 
For simplicity, we omit it in Eq.~\ref{eq:cross}.
We describe the network architectures in more detail in Appendix~\ref{sec:appendix_architecture}.

\mypar{Training.}
Like SiamFC~\cite{bertinetto2016fully}, we use examplar and search image patches of $127{\times}127$ and $255{\times}255$ pixels respectively.
During training, we randomly jitter examplar and search patches.
Specifically, we consider random translations (up to $\pm 8$ pixels) and rescaling (of $2^{\pm 1/8}$ and $2^{\pm 1/4}$ for examplar and search respectively).

The network backbone is pre-trained on the \mbox{ImageNet-$1k$} classification task.
We use SGD with a first \emph{warmup} phase in which the learning rate increases linearly from $10^{-3}$ to $5{\times}10^{-3}$ for the first 5 epochs and then descreases logarithmically until $5{\times}10^{-4}$ for 15 more epochs.
We train all our models using COCO~\cite{lin2014microsoft}, ImageNet-VID~\cite{russakovsky2015imagenet} and YouTube-VOS~\cite{xu2018youtube}. 

\mypar{Inference.}
During tracking, SiamMask is simply evaluated once per frame, without any adaptation.
In both our variants, we select the output mask using the location attaining the maximum score in the classification branch.
Then, after having applied a per-pixel sigmoid, we binarise the output of the mask branch at the threshold of $0.5$.
In the \textit{two-branch} variant, for each video frame after the first one, we fit the output mask with the \emph{Min-max} box and use it as reference to crop the next frame search region.
Instead, in the \textit{three-branch} variant, we find more effective to exploit the highest-scoring output of the box branch as reference.

\section{Experiments}
\label{sec:experiments}
In this section, we evaluate our approach on two related tasks: visual object tracking (on VOT-2016 and VOT-2018) and semi-supervised video object segmentation (on DAVIS-2016 and DAVIS-2017).
We refer to our \emph{two-branch} and \emph{three-branch} variants with SiamMask-2B and SiamMask respectively.

\subsection{Evaluation for visual object tracking}
\label{sec:exp_track}

\mypar{Datasets and settings.}
We adopt two widely used benchmarks for the evaluation of the object tracking task: VOT-2016~\cite{kristan2016visual} and VOT-2018~\cite{VOT2018}, both annotated with rotated bounding boxes.
We use VOT-2016 to understand how different types of representation affect the performance.
For this first experiment, we use mean intersection over union (IOU) and Average Precision (AP)@$\{0.5,0.7\}$ IOU.
We then compare against the state-of-the-art on VOT-2018, using the official VOT toolkit and the Expected Average Overlap (EAO), a measure that considers both accuracy and robustness of a tracker~\cite{VOT2018}.

\mypar{How much does the object representation matter?}
Existing tracking methods typically predict axis-aligned bounding boxes with a fixed~\cite{bertinetto2016fully,henriques2015tracking,danelljan2015learning,lukezic2017discriminative} or variable~\cite{SiamRPN,held2016learning,zhu2018distractor} aspect ratio.
We are interested in understanding to which extent producing a per-frame binary mask can improve tracking.
In order to focus on representation accuracy, for this experiment only we ignore the temporal aspect and sample video frames at random.
The approaches described in the following paragraph are tested on randomly cropped search patches (with random shifts within $\pm16$ pixels and scale deformations up to $2^{1\pm0.25}$) from the sequences of VOT-2016.

In Table~\ref{tab:iou}, we compare our \emph{three-branch} variant using the \emph{Min-max}, \emph{MBR} and \emph{Opt} approaches (described at the end of Section~\ref{sec:siammask} and in Figure~\ref{fig:bbox}).
For reference, we also report results for SiamFC and SiamRPN as representative of the fixed and variable aspect-ratio approaches, together with three \emph{oracles} that have access to per-frame ground-truth information and serve as upper bounds for the different representation strategies.
(1) The fixed aspect-ratio oracle uses the per-frame ground-truth area and center location, but fixes the aspect reatio to the one of the first frame and produces an axis-aligned bounding box.
(2) The \emph{Min-max} oracle uses the minimal enclosing rectangle of the rotated ground-truth bounding box to produce an axis-aligned bounding box.
(3) Finally, the \textit{MBR} oracle uses the rotated minimum bounding rectangle of the ground-truth.
Note that (1), (2) and (3) can be considered, respectively, the performance upper bounds for the representation strategies of SiamFC, SiamRPN and SiamMask.

Table~\ref{tab:iou} shows that our method achieves the best mIOU, no matter the box generation strategy used (Figure~\ref{fig:bbox}).
Albeit SiamMask-\textit{Opt} offers the highest IOU and mAP, it requires significant computational resources due to its slow optimisation procedure~\cite{vojir2017pixel}. 
SiamMask-\textit{MBR} achieves a mAP@0.5 IOU  of $85.4$, with a respective improvement of $+29$ and $+9.2$ points w.r.t. the two fully-convolutional baselines.
Interestingly, the gap significantly widens when considering mAP at the higher accuracy regime of 0.7 IOU: $+41.6$ and $+18.4$ respectively.
Notably, our accuracy results are not far from the fixed aspect-ratio oracle.
Moreover, comparing the upper bound performance represented by the oracles, it is possible to notice how, by simply changing the bounding box representation, there is a great room for improvement (\eg $+10.6\%$ mIOU improvement between the fixed aspect-ratio and the \textit{MBR} oracles).

Overall, this study shows how the \textit{MBR} strategy to obtain a rotated bounding box from a binary mask of the object offers a significant advantage over popular strategies that simply report axis-aligned bounding boxes.

\begin{table}[t]
\tablestyle{3.5pt}{1.1}
\begin{tabular}{l|x{42}|x{52}x{52}}
 & mIOU  ($\%$) &  mAP@0.5 IOU  &  mAP@0.7 IOU   \\
\shline
Fixed a.r. Oracle  & 73.43 & 90.15  & 62.52   \\
\textit{Min-max} Oracle  & 77.70 & 88.84  & 65.16  \\
\textit{MBR} Oracle & 84.07 & 97.77 & 80.68 \\
\hline
SiamFC~\cite{bertinetto2016fully}  & 50.48 & 56.42  & 9.28   \\
SiamRPN~\cite{zhu2018distractor}  & 60.02 & 76.20  & 32.47  \\
\hline
\bd{SiamMask}-\textit{Min-max} & 65.05 & 82.99  & 43.09 \\
\bd{SiamMask}-\textit{MBR} & 67.15 & 85.42  & 50.86  \\
\bd{SiamMask}-\textit{Opt} & \bd{71.68} & \bd{90.77} & \bd{60.47}
\end{tabular}
\vspace{1mm}
\caption{Performance for different bounding box representation strategies on VOT-2016.}
\label{tab:iou}
\end{table}

\mypar{Results on VOT-2018 and VOT-2016.}
In Table~\ref{tab:vot18} we compare the two variants of SiamMask with \textit{MBR} strategy and SiamMask--\textit{Opt} against five recently published state-of-the-art trackers on the VOT-2018 benchmark.
Unless stated otherwise, SiamMask refers to our \textit{three-branch} variant with \textit{MBR} strategy.
Both variants achieve outstanding performance and run in real-time.
In particular, our \textit{three-branch} variant significantly outperforms the very recent and top performing DaSiamRPN~\cite{zhu2018distractor}, achieving a EAO of $0.380$ while running at 55 frames per second.
Even without box regression branch, our simpler \textit{two-branch} variant (SiamMask-2B) achieves a high EAO of $0.334$, which is in par with SA\_Siam\_R~\cite{he2018towards} and superior to any other real-time method in the published literature.
Finally, in SiamMask--\textit{Opt}, the strategy proposed in~\cite{vojir2017pixel} to find the optimal rotated rectangle from a binary mask brings the best overall performance (and a particularly high accuracy), but comes at a significant computational cost.

Our model is particularly strong under the accuracy metric, showing a significant advantage with respect to the Correlation Filter-based trackers CSRDCF~\cite{lukezic2017discriminative}, STRCF~\cite{li2018learning}.
This is not surprising, as SiamMask relies on a richer object representation, as outlined in Table~\ref{tab:iou}.
Interestingly, similarly to us, He \etal (SA\_Siam\_R)~\cite{he2018towards} are motivated to achieve a more accurate target representation by considering multiple rotated and rescaled bounding boxes.
However, their representation is still constrained to a fixed aspect-ratio box.

Table~\ref{tab:vot18and16} gives further results of SiamMask with different box generation strategies on VOT-2018 and -2016.
SiamMask-box means the box branch of SiamMask is adopted for inference despite the mask branch has been trained. We can observe clear improvements on all evaluation metrics by using the mask branch for box generation.

\begin{table*}[t]
\tablestyle{2pt}{1}
\begin{tabular}{l|x{52}x{42}x{48}|x{52}x{52}x{52}x{42}x{42}x{42}x{42}}
& SiamMask-\textit{Opt} & SiamMask & SiamMask-2B & DaSiamRPN~\cite{zhu2018distractor} & SiamRPN~\cite{SiamRPN} & SA\_Siam\_R~\cite{he2018towards}
 &CSRDCF~\cite{lukezic2017discriminative}  &STRCF~\cite{li2018learning} \\
\shline
\texttt{EAO $\uparrow$ }& \bd{0.387} & \bd{0.380} &  0.334 & 0.326 & 0.244 & 0.337 & 0.263  & 0.345 \\
\texttt{Accuracy $\uparrow$}& \bd{0.642} & \bd{0.609} & 0.575 & 0.569 & 0.490 & 0.566 & 0.466  & 0.523 \\
\texttt{Robustness $\downarrow$}& 0.295 & 0.276 &  0.304 & 0.337 &  0.460 & 0.258& 0.318  & \bd{0.215}   \\
\hline
\texttt{Speed} (\texttt{fps}) $\uparrow$ & 5 & 55  & 60  &160   &\bd{200}   &32.4  & 48.9    & 2.9 \\
\end{tabular}
\vspace{1mm}
\caption{Comparison with the state-of-the-art under the \texttt{EAO}, \texttt{Accuracy}, and \texttt{Robustness} metrics on VOT-2018.}
\label{tab:vot18}
\end{table*}

\begin{table}[t]
\tablestyle{1.5pt}{1.2}
\begin{tabular}{l|x{22}x{22}x{22}|x{22}x{22}x{22}|x{32}}
& \multicolumn{3}{c}{VOT-2018} & \multicolumn{3}{c}{VOT-2016} \\
& \texttt{EAO} $\uparrow$ & \texttt{A} $\uparrow$ & \texttt{R} $\downarrow$
		& \texttt{EAO} $\uparrow$ &  \texttt{A} $\uparrow$ & \texttt{R} $\downarrow$
 & \texttt{Speed} \\[.1em]
\shline
SiamMask-box & 0.363 & 0.584 &0.300 & 0.412 & 0.623 &0.233 & \textbf{76}    \\ 
SiamMask  & 0.380 & 0.609 &\textbf{0.276} & 0.433 & 0.639 &\textbf{0.214} & 55    \\ 
SiamMask-\textit{Opt}  & \textbf{0.387} & \textbf{0.642} &0.295 & \textbf{0.442} & \textbf{0.670} &0.233 & 5    \\ 
\end{tabular}
\vspace{1mm}
\caption{Results on VOT-2016 and VOT-2018.}
\label{tab:vot18and16}
\end{table}

\begin{table}[t]
\tablestyle{1.2pt}{1.2}
\begin{tabular}{l|x{12}x{12}|x{20}x{20}x{20}|x{20}x{20}x{20}|x{30}}
		& \texttt{FT}& \texttt{M} &  $\mathcal{J}_{\mathcal{M\uparrow}}$ & $\mathcal{J}_{\mathcal{O\uparrow}}$  & $\mathcal{J}_{\mathcal{D\downarrow}}$ & $\mathcal{F}_{\mathcal{M\uparrow}}$ & $\mathcal{F}_{\mathcal{O\uparrow}}$  & $\mathcal{F}_{\mathcal{D\downarrow}}$  & \texttt{Speed}\\[.1em]
\shline
OnAVOS~\cite{voigtlaender2017online}& \cmark & \cmark & \textbf{86.1} & \textbf{96.1} & 5.2 & \textbf{84.9} & \textbf{89.7} & 5.8 & 0.08 \\
MSK~\cite{perazzi2017learning} & \cmark & \cmark & 79.7 & 93.1 & 8.9 & 75.4 & 87.1 & 9.0 & 0.1 \\
MSK$_{b}$~\cite{perazzi2017learning} & \cmark & \xmark & 69.6 & - & - & - & - & - & 0.1 \\
SFL~\cite{cheng2017segflow} & \cmark & \cmark & 76.1 & 90.6 & 12.1 & 76.0 & 85.5 & 10.4 & 0.1 \\ \hline
FAVOS~\cite{cheng2018fast} & \xmark & \cmark & 82.4 & 96.5 & 4.5 & 79.5 & 89.4 & 5.5 & 0.8 \\
RGMP~\cite{wug2018fast} & \xmark & \cmark & 81.5 & 91.7 & 10.9 & 82.0 & 90.8 & 10.1 & 8 \\
PML~\cite{chen2018blazingly} & \xmark & \cmark & 75.5 & 89.6 & 8.5 & 79.3 & 93.4 & 7.8 & 3.6 \\ 
OSMN~\cite{Yang_2018_CVPR} & \xmark & \cmark & 74.0 & 87.6 & 9.0 & 72.9 & 84.0 & 10.6 & 8.0\\
PLM~\cite{yoon2017pixel} & \xmark & \cmark & 70.2 & 86.3 & 11.2 & 62.5 & 73.2 & 14.7 & 6.7 \\
VPN~\cite{jampani2017video} & \xmark & \cmark & 70.2 & 82.3 & 12.4 & 65.5 & 69.0 & 14.4 & 1.6 \\
 \hline
\textbf{SiamMask} & \xmark & \xmark & 71.7 & 86.8 & \textbf{3.0} & 67.8 & 79.8 & \textbf{2.1} & \textbf{55} \\
\end{tabular}
\vspace{1mm}
\caption{Results on DAVIS 2016 (validation set). \texttt{FT} and \texttt{M} respectively denote if the method requires fine-tuning and whether it is initialised with a mask (\cmark) or a bounding box (\xmark).}
\label{tab:davis16}
\end{table}

\subsection{Evaluation for semi-supervised VOS}
\label{sec:exp_seg}
Our model, once trained, can also be used for the task of VOS to achieve competitive performance without requiring any adaptation at test time.
Importantly, differently to typical VOS approaches, ours can operate online, runs in real-time and only requires a simple bounding box initialisation.

\mypar{Datasets and settings.}
We report the performance of SiamMask on DAVIS-2016~\cite{perazzi2016benchmark}, DAVIS-2017~\cite{pont2017davis} and YouTube-VOS~\cite{xu2018youtube} benchmarks.
For both DAVIS datasets, we use the official performance measures: the Jaccard index ($\mathcal{J}$) to express region similarity and the F-measure ($\mathcal{F}$) to express contour accuracy.
For each measure $\mathcal{C} \in \{\mathcal{J}, \mathcal{F}\}$, three statistics are considered: mean $\mathcal{C}_{\mathcal{M}}$, recall $\mathcal{C}_{\mathcal{O}}$, and decay $\mathcal{C}_{\mathcal{D}}$, which informs us about the gain/loss of performance over time~\cite{perazzi2016benchmark}.
Following Xu \etal~\cite{xu2018youtube}, for YouTube-VOS we report the mean Jaccard index and F-measure for both seen ($\mathcal{J}_{\mathcal{S}}$, $\mathcal{F}_{\mathcal{S}}$) and unseen categories ($\mathcal{J}_{\mathcal{U}}$, $\mathcal{F}_{\mathcal{U}}$).
$\mathcal{O}$ is the average of these four measures.

To initialise SiamMask, we extract the axis-aligned bounding box from the mask provided in the first frame (\emph{Min-max} strategy, see Figure~\ref{fig:bbox}).
Similarly to most VOS methods, in case of multiple objects in the same video (DAVIS-2017) we simply perform multiple inferences.

\begin{table}[t]
\tablestyle{1.2pt}{1.2}
\begin{tabular}{l|x{12}x{12}|x{20}x{20}x{20}|x{20}x{20}x{20}|x{30}}
& \texttt{FT}& \texttt{M} & $\mathcal{J}_{\mathcal{M\uparrow}}$ & $\mathcal{J}_{\mathcal{O\uparrow}}$  & $\mathcal{J}_{\mathcal{D\downarrow}}$ & $\mathcal{F}_{\mathcal{M\uparrow}}$ & $\mathcal{F}_{\mathcal{O\uparrow}}$  & $\mathcal{F}_{\mathcal{D\downarrow}}$ & \texttt{Speed} \\[.1em]
\shline
OnAVOS~\cite{voigtlaender2017online} & \cmark & \cmark & \textbf{61.6} & \textbf{67.4} & 27.9 & \textbf{69.1} & \textbf{75.4} & 26.6 & 0.1 \\
OSVOS~\cite{caelles2017one} & \cmark & \cmark & 56.6 & 63.8 & 26.1 & 63.9 & 73.8 & 27.0 & 0.1 \\
FAVOS~\cite{cheng2018fast} & \xmark & \cmark & 54.6 & 61.1 & \textbf{14.1} & 61.8 & 72.3 & \textbf{18.0} & 0.8 \\
OSMN~\cite{Yang_2018_CVPR} & \xmark & \cmark & 52.5 & 60.9 & 21.5 & 57.1 & 66.1 & 24.3 & 8.0 \\\hline
\textbf{SiamMask} & \xmark & \xmark & 54.3 & 62.8 & 19.3 & 58.5 & 67.5 & 20.9 & \textbf{55} \\
\end{tabular}
\vspace{1mm}
\caption{Results on DAVIS 2017 (validation set).
}
\label{tab:davis17}
\end{table}

\begin{table}[t]
\tablestyle{1.2pt}{1.2}
\begin{tabular}{l|x{12}x{12}|x{20}x{20}x{20}x{20}| x{20}|x{30}}
& \texttt{FT}& \texttt{M} & $\mathcal{J}_{\mathcal{S\uparrow}}$ & $\mathcal{J}_{\mathcal{U\uparrow}}$ & $\mathcal{F}_{\mathcal{S\uparrow}}$ & $\mathcal{F}_{\mathcal{U\uparrow}}$  & $\mathcal{O\uparrow}$ & \texttt{Speed} \\[.1em]
\shline
OnAVOS~\cite{voigtlaender2017online} & \cmark & \cmark & 60.1 & 46.6 & \textbf{62.7} & 51.4 & 55.2 &  0.1 \\
OSVOS~\cite{caelles2017one} & \cmark & \cmark & 59.8 & \textbf{54.2} & 60.5 & \textbf{60.7} & \textbf{58.8} & 0.1 \\
OSMN~\cite{Yang_2018_CVPR} & \xmark & \cmark & 60.0 & 40.6 & 60.1 & 44.0 & 51.2 & 8.0 \\\hline
\textbf{SiamMask} & \xmark & \xmark & \textbf{60.2} & 45.1 & 58.2 & 47.7& 52.8 & \textbf{55} \\
\end{tabular}
\vspace{1mm}
\caption{Results on YouTube-VOS (validation set).
}
\label{tab:youtubevos}
\end{table}

\mypar{Results on DAVIS and YouTube-VOS.}
In the semi-supervised setting, VOS methods are initialised with a binary mask~\cite{perazzi2017video} and many of them require computationally intensive techniques at test time such as fine-tuning~\cite{maninis2017video,perazzi2017learning,bao2018cnn,voigtlaender2017online}, data augmentation~\cite{LucidDataDreaming_CVPR17_workshops,li2018video}, inference on MRF/CRF~\cite{wen2015jots,tsai2016video,marki2016bilateral,bao2018cnn} and optical flow~\cite{tsai2016video,bao2018cnn,perazzi2017learning,li2018video,cheng2018fast}.
As a consequence, it is not uncommon for VOS techniques to require several minutes to process a short sequence.
Clearly, these strategies make the online applicability (which is our focus) impossible.
For this reason, in our comparison we mainly concentrate on \emph{fast} state-of-the-art approaches.

Table~\ref{tab:davis16}, \ref{tab:davis17} and~\ref{tab:youtubevos} show how SiamMask can be considered as a strong baseline for online VOS.
First, it is almost two orders of magnitude faster than accurate approaches such as OnAVOS~\cite{voigtlaender2017online} or SFL~\cite{cheng2017segflow}.
Second, it is competitive with recent VOS methods that do not employ fine-tuning, while being four times more efficient than the fastest ones (\ie OSMN~\cite{Yang_2018_CVPR} and RGMP~\cite{wug2018fast}).
Interestingly, we note that SiamMask achieves a very low \emph{decay}~\cite{perazzi2016benchmark} for both region similarity ($\mathcal{J}_{\mathcal{D}}$,) and contour accuracy ($\mathcal{F}_{\mathcal{D}}$).
This suggests that our method is robust over time and thus it is indicated for particularly long sequences.

\begin{figure*}[t]
\centering
\setlength{\tabcolsep}{0.25ex}

\begin{tabular}
{cccccc cccccc}
\mbox{\centering\rotatebox[x=-0.55cm]{90}{\small{Basketball}}}
\includegraphics[trim={2.5cm 1cm 2.5cm 1cm},clip,width = 1.1in]{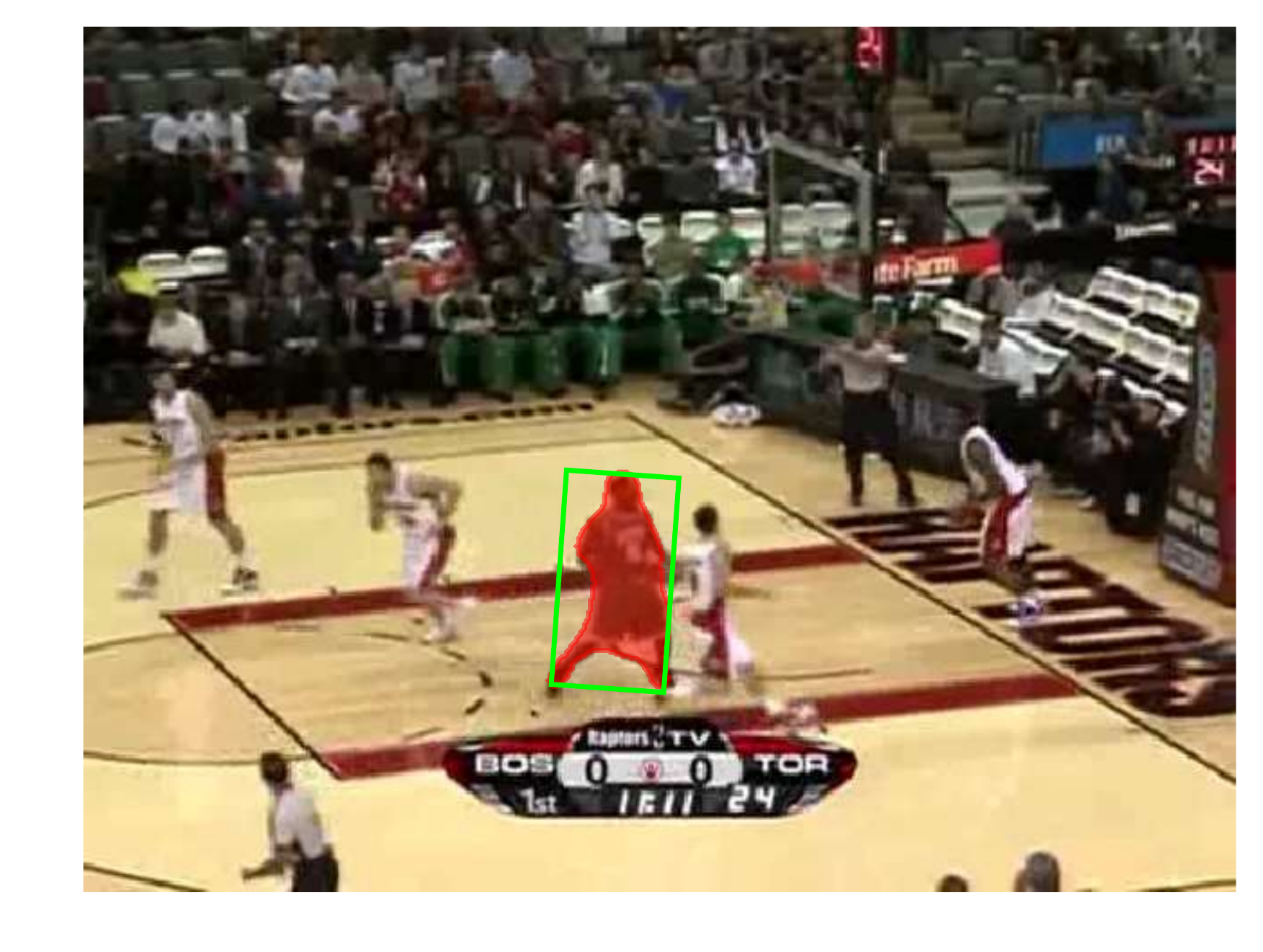}
&\includegraphics[trim={2.5cm 1cm 2.5cm 1cm},clip,width = 1.1in]{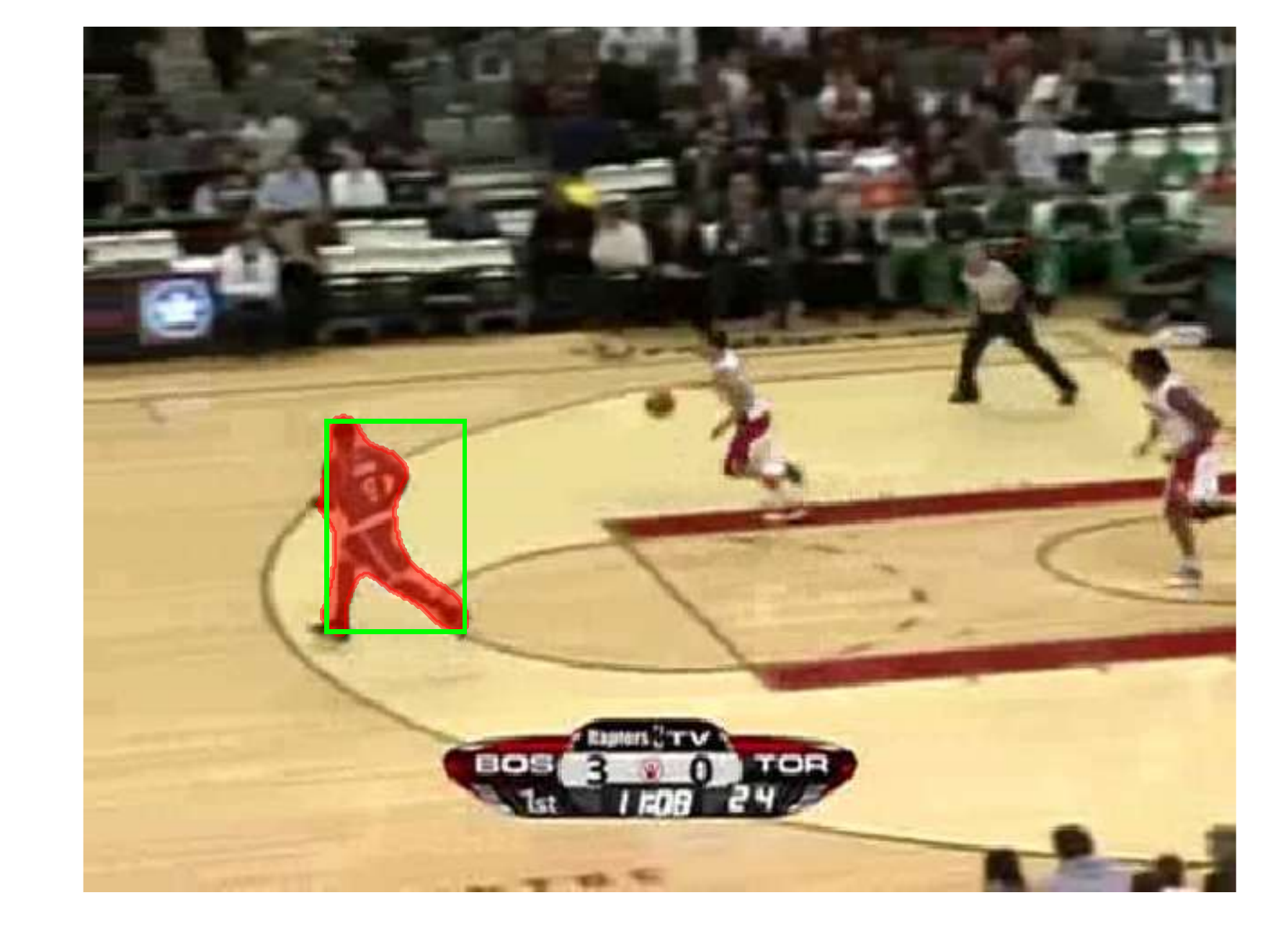}
& \includegraphics[trim={2.5cm 1cm 2.5cm 1cm},clip,width = 1.1in]{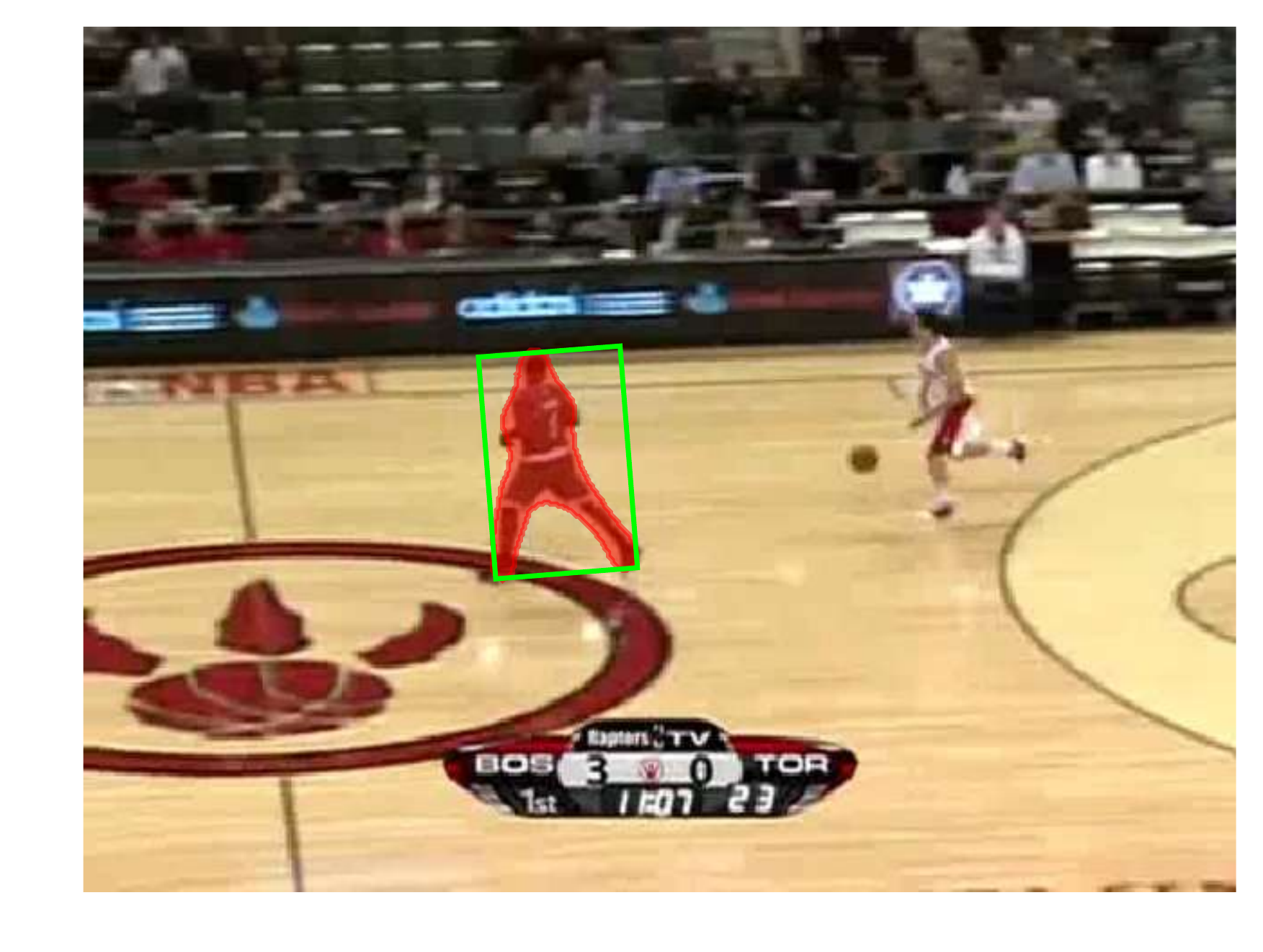}
& \includegraphics[trim={2.5cm 1cm 2.5cm 1cm},clip,width = 1.1in]{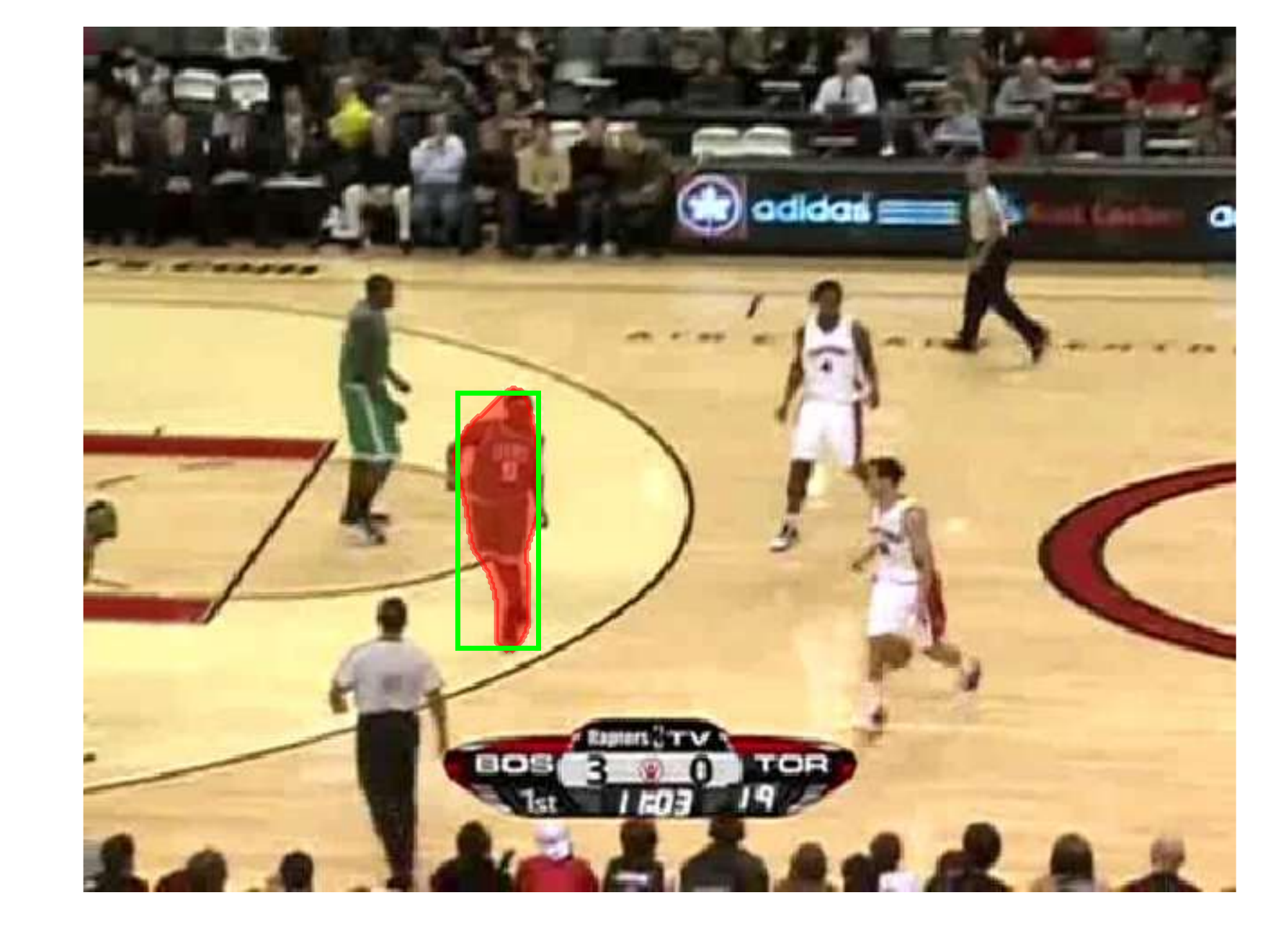}
& \includegraphics[trim={2.5cm 1cm 2.5cm 1cm},clip,width = 1.1in]{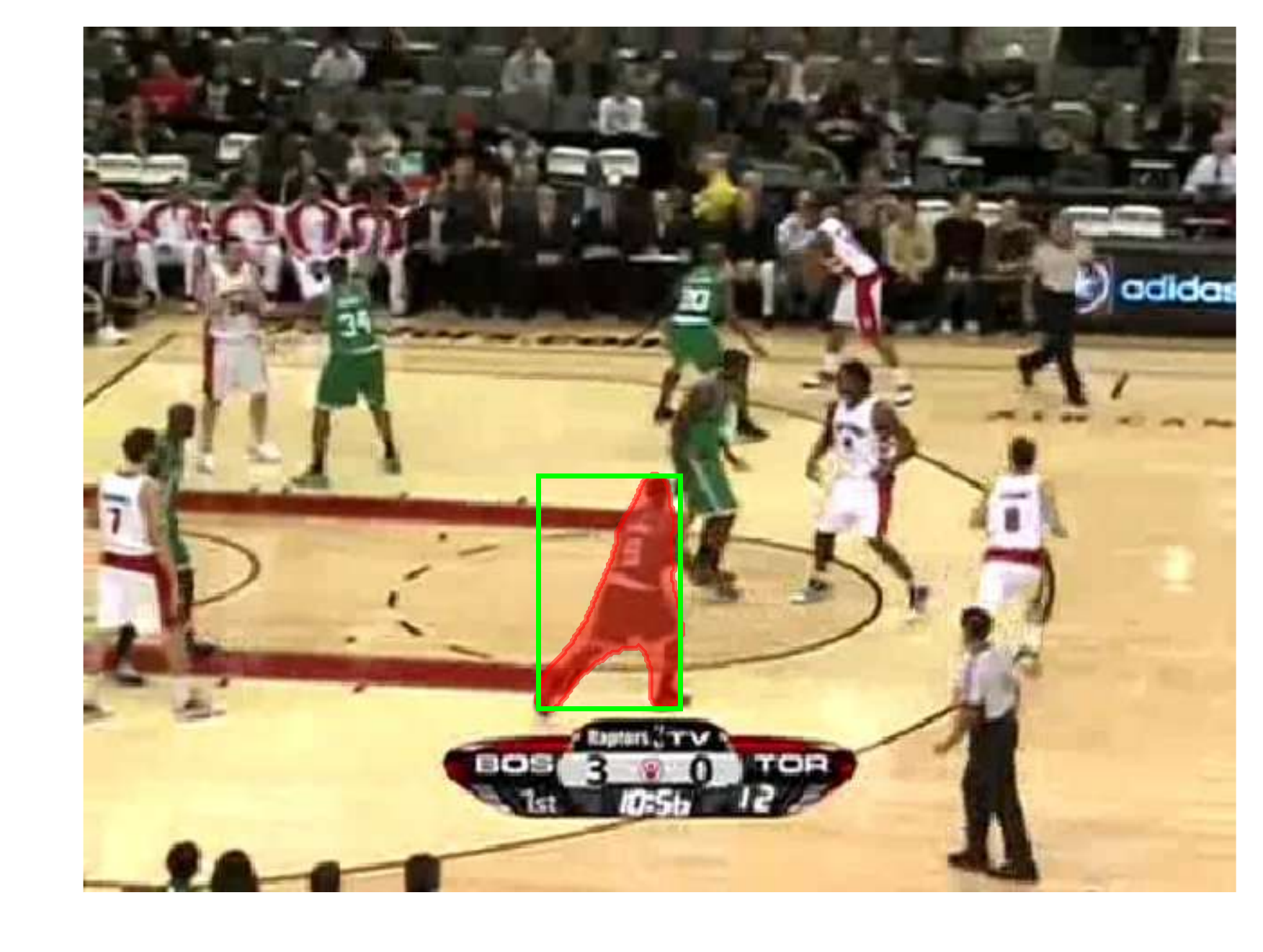}
& \includegraphics[trim={2.5cm 1cm 2.5cm 1cm},clip,width = 1.1in]{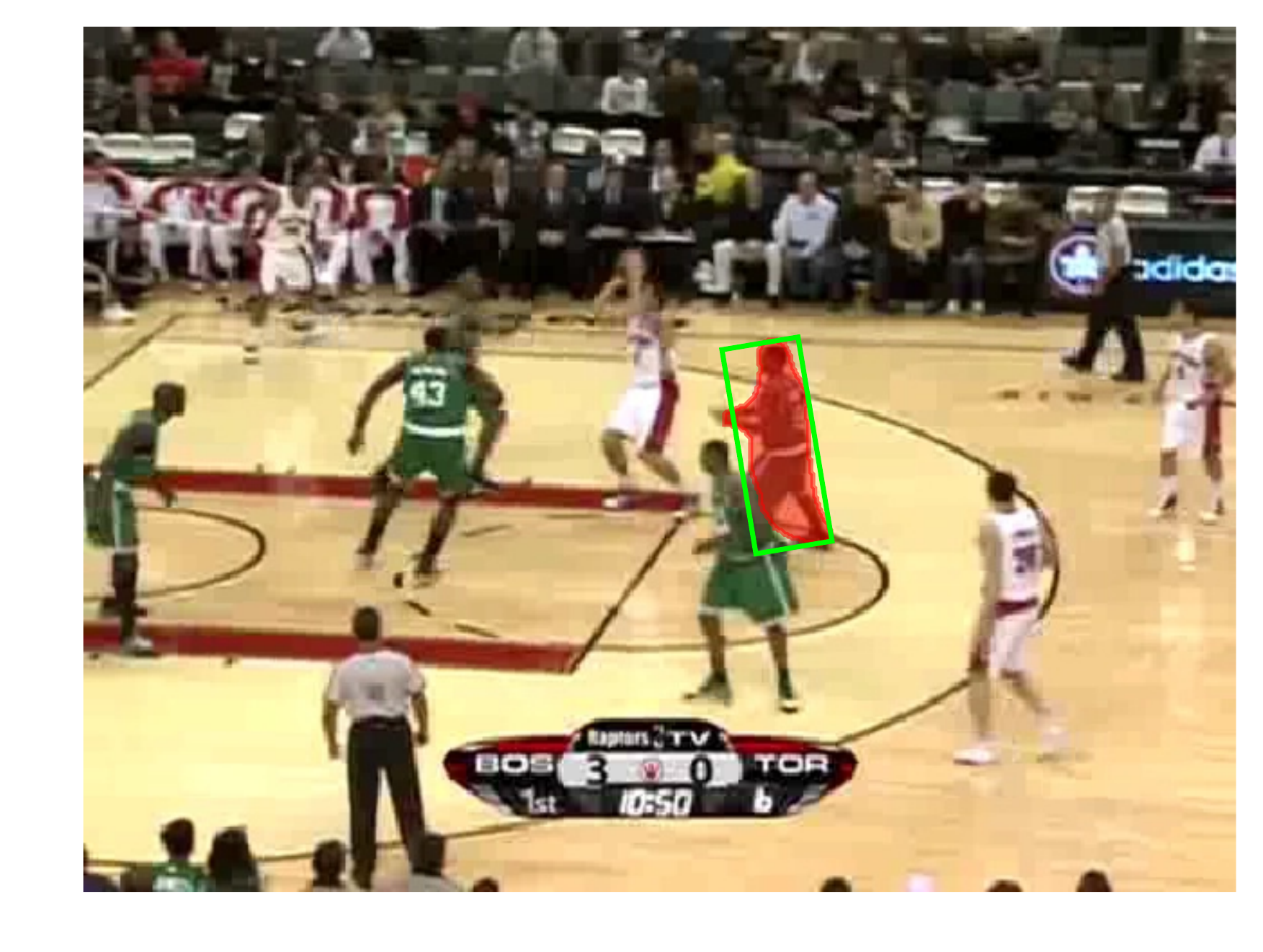}
\\
\mbox{\rotatebox[x=-0.55cm]{90}{\small{Nature}}}
\includegraphics[trim={2.5cm 1cm 2.5cm 1cm},clip,width = 1.1in]{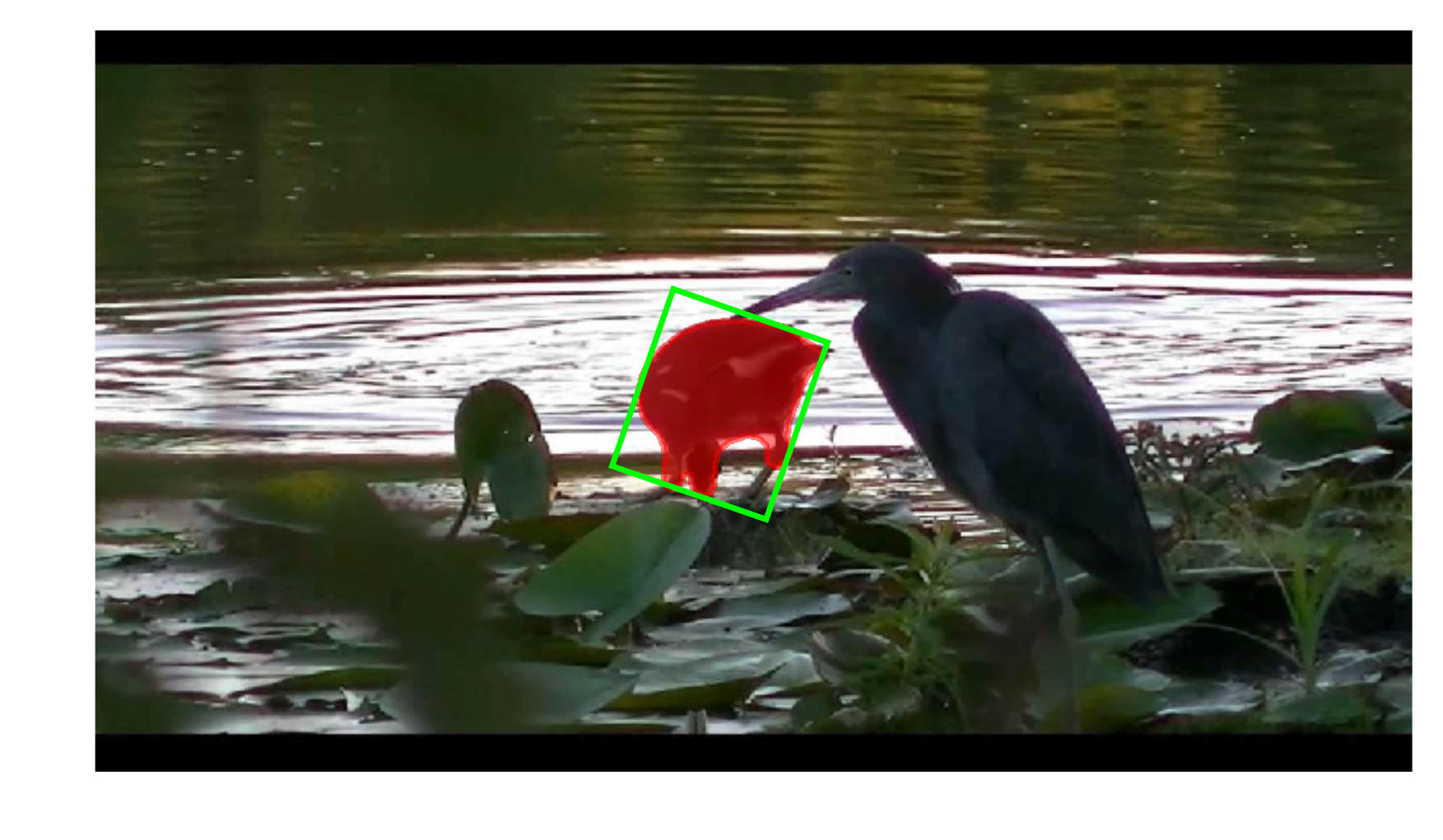}
&\includegraphics[trim={2.5cm 1cm 2.5cm 1cm},clip,width = 1.1in]{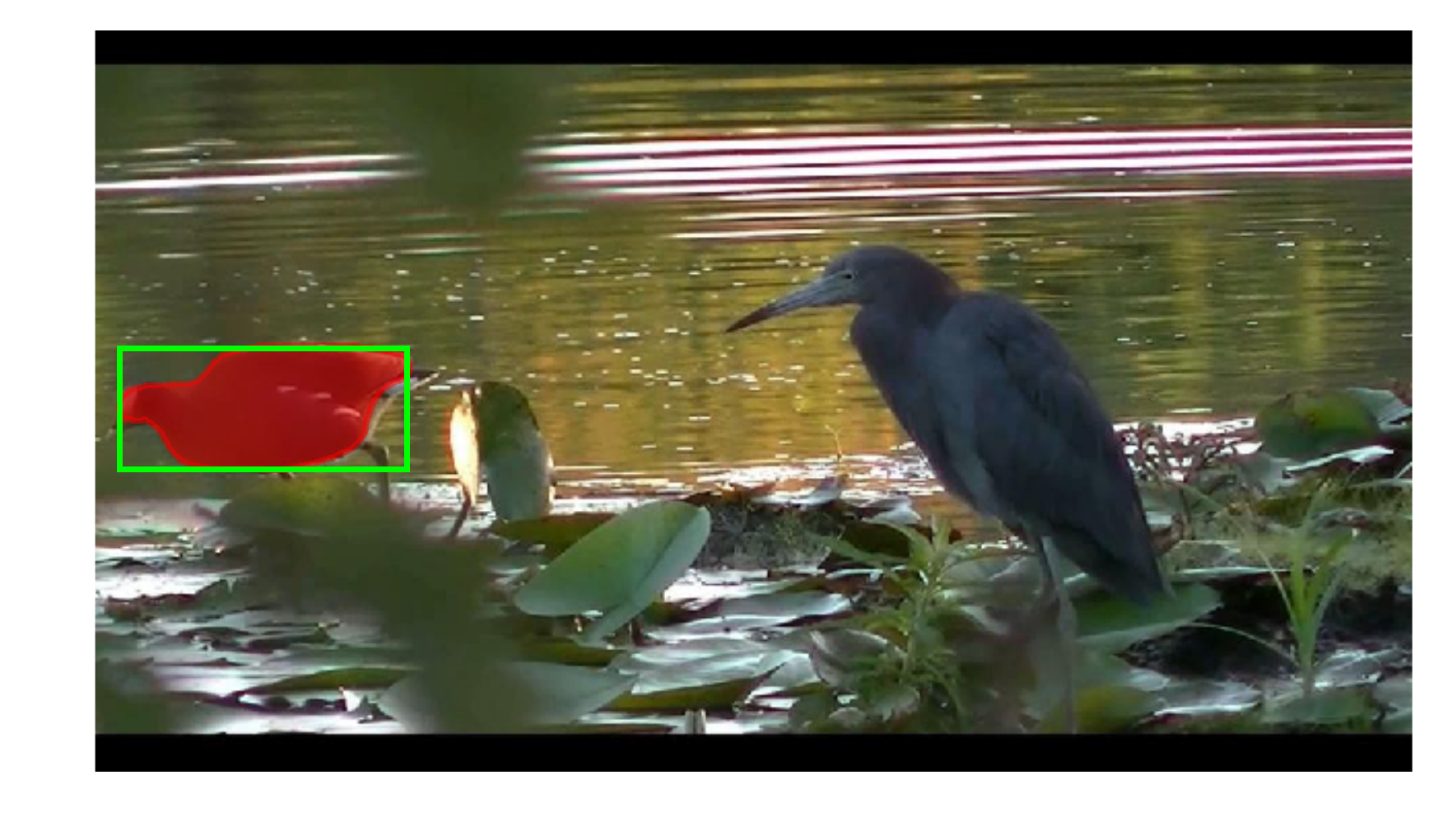}
& \includegraphics[trim={2.5cm 1cm 2.5cm 1cm},clip,width = 1.1in]{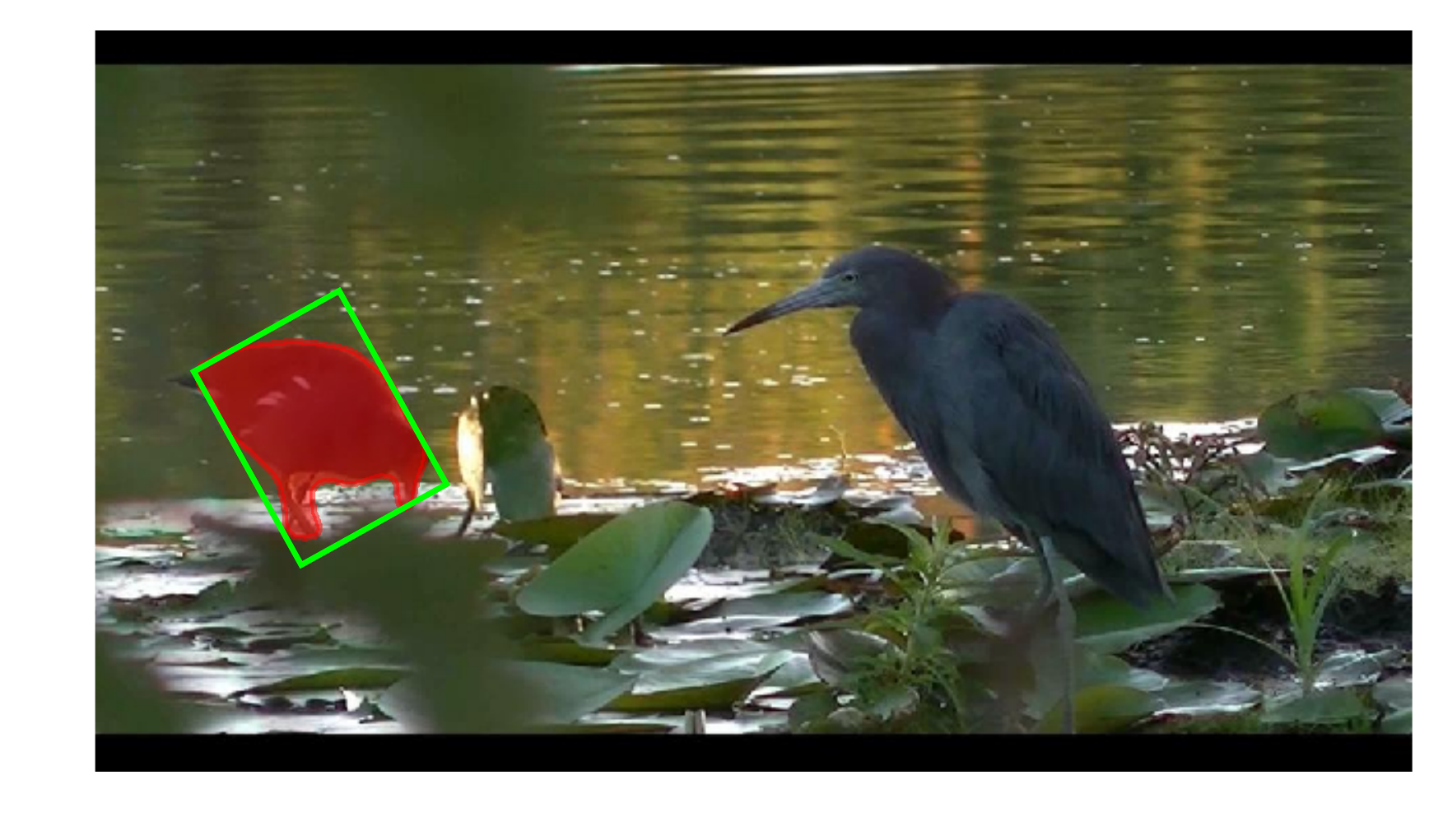}
& \includegraphics[trim={2.5cm 1cm 2.5cm 1cm},clip,width = 1.1in]{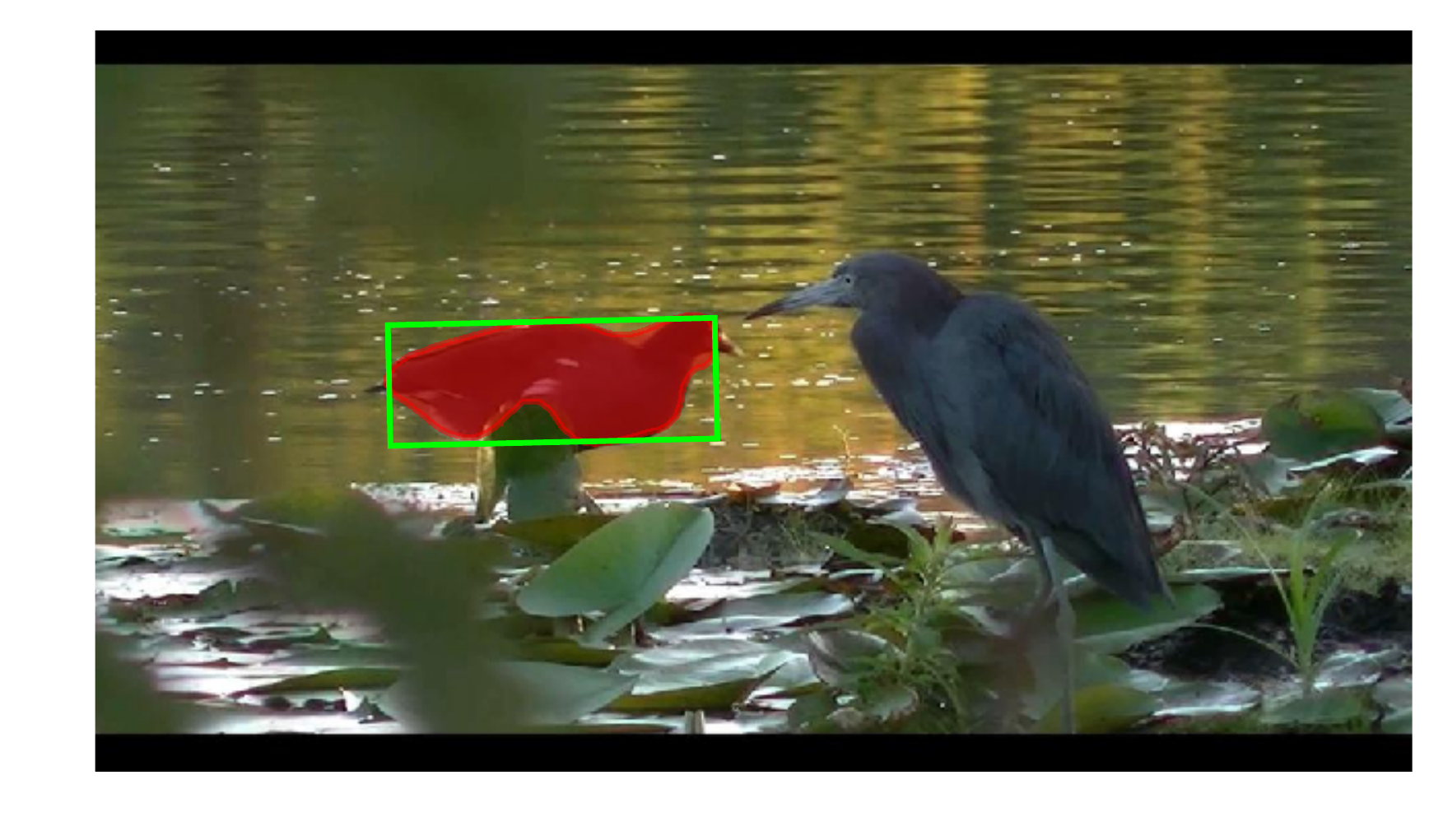}
& \includegraphics[trim={2.5cm 1cm 2.5cm 1cm},clip,width = 1.1in]{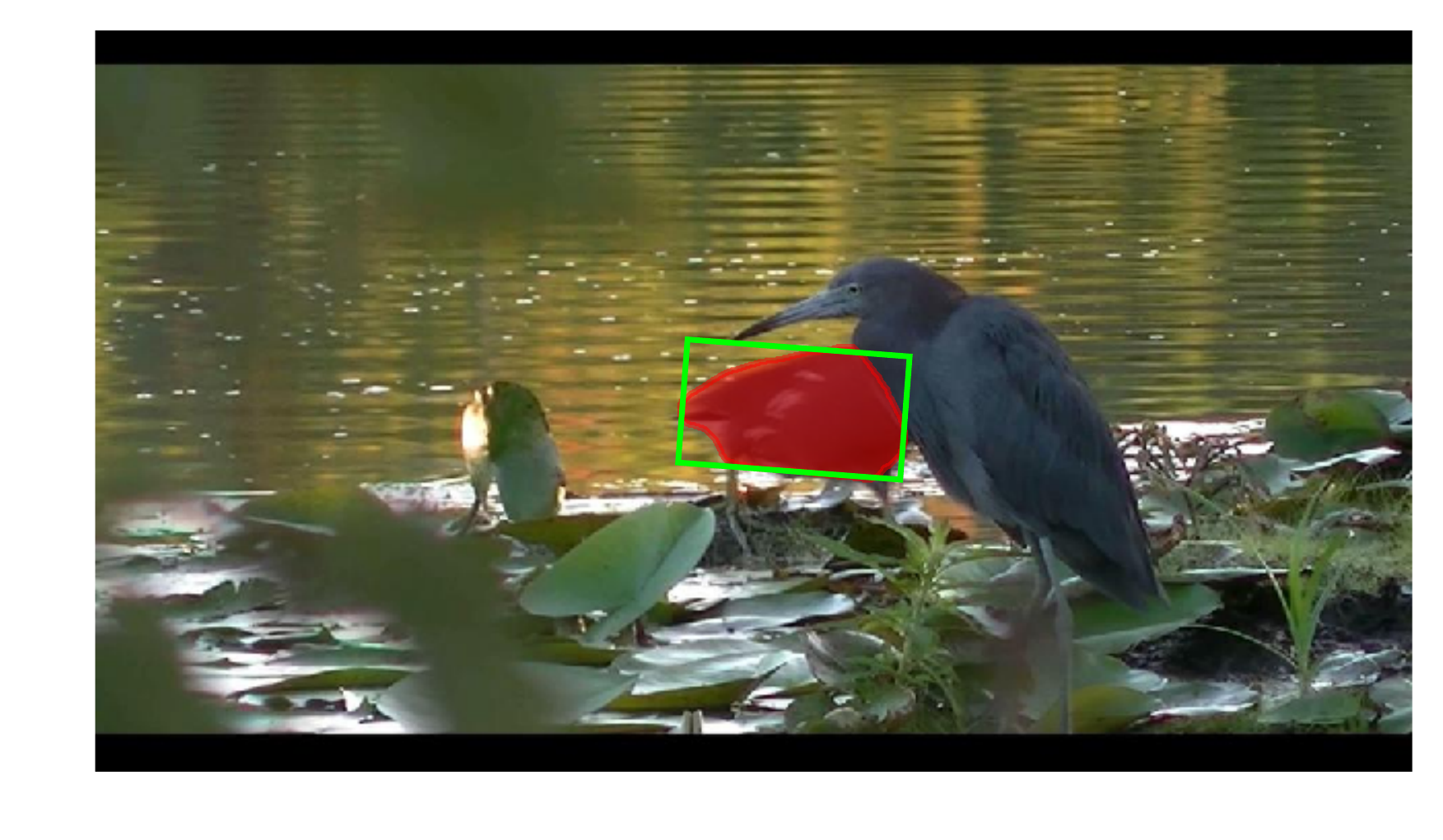}
& \includegraphics[trim={2.5cm 1cm 2.5cm 1cm},clip,width = 1.1in]{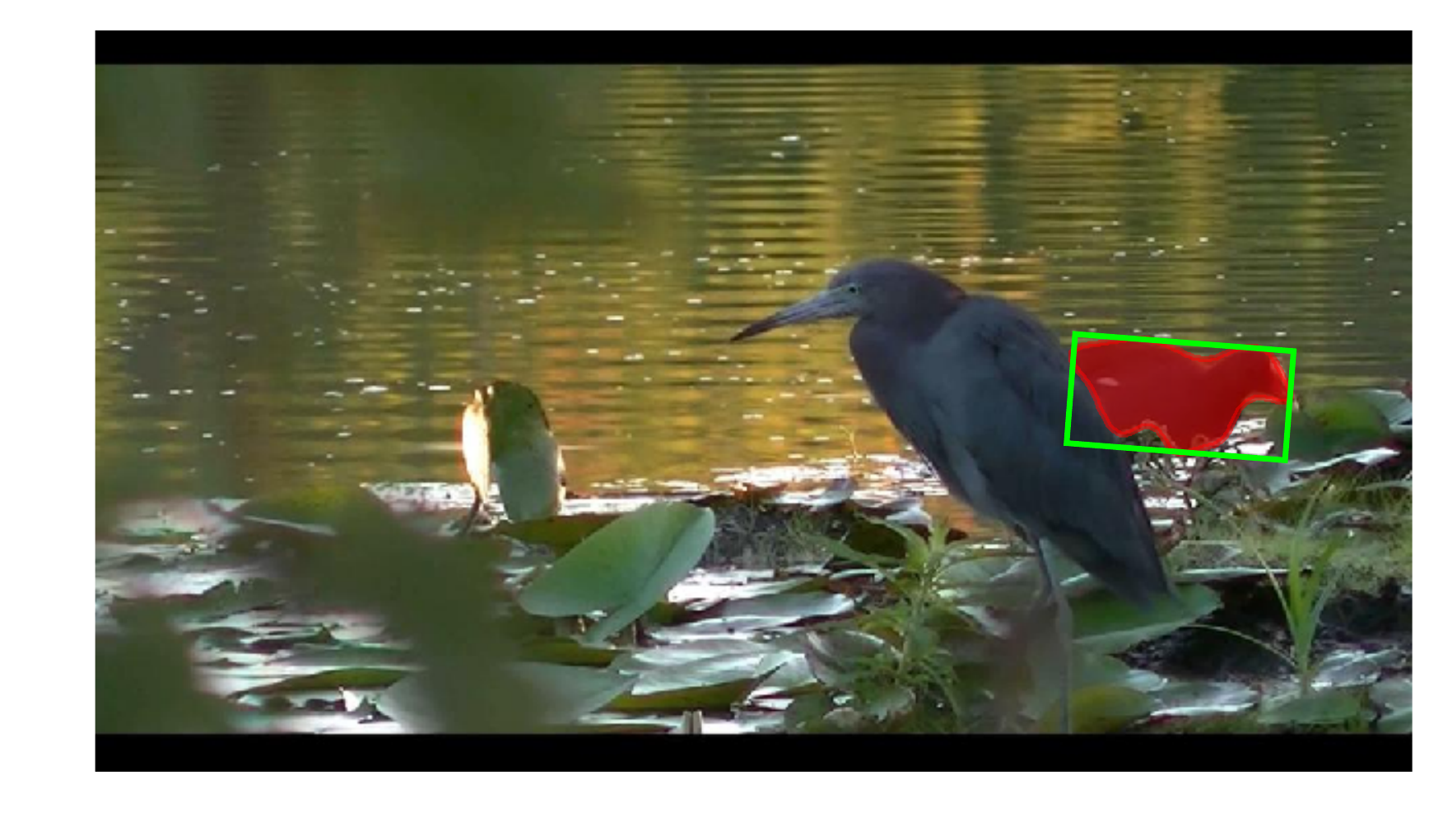}
\\

\mbox{\rotatebox[x=-0.0cm]{90}{\small{Car-Shadow}}}
\includegraphics[trim={2.5cm 1cm 2.5cm 1cm},clip,width = 1.1in]{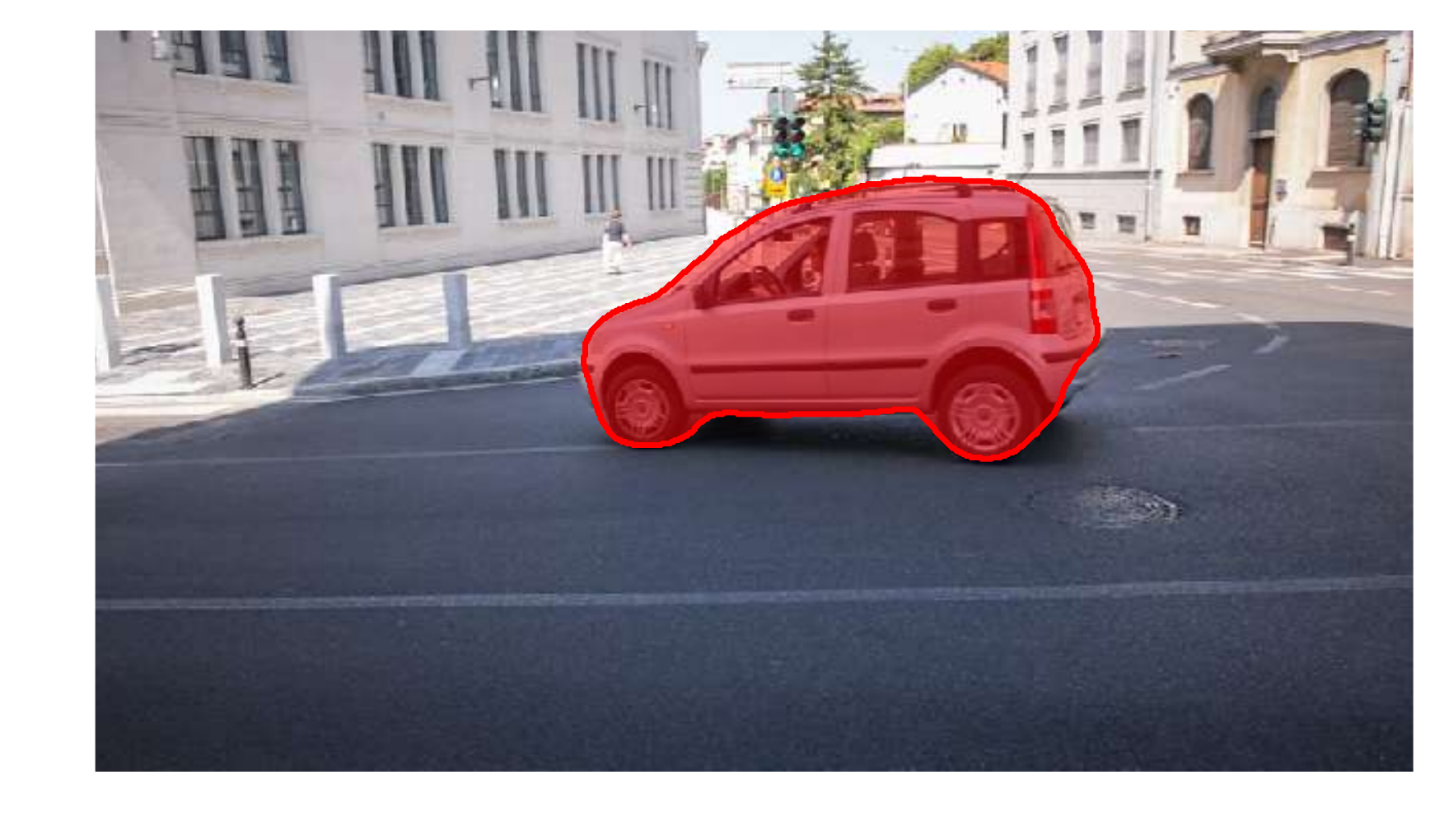}
&\includegraphics[trim={2.5cm 1cm 2.5cm 1cm},clip,width = 1.1in]{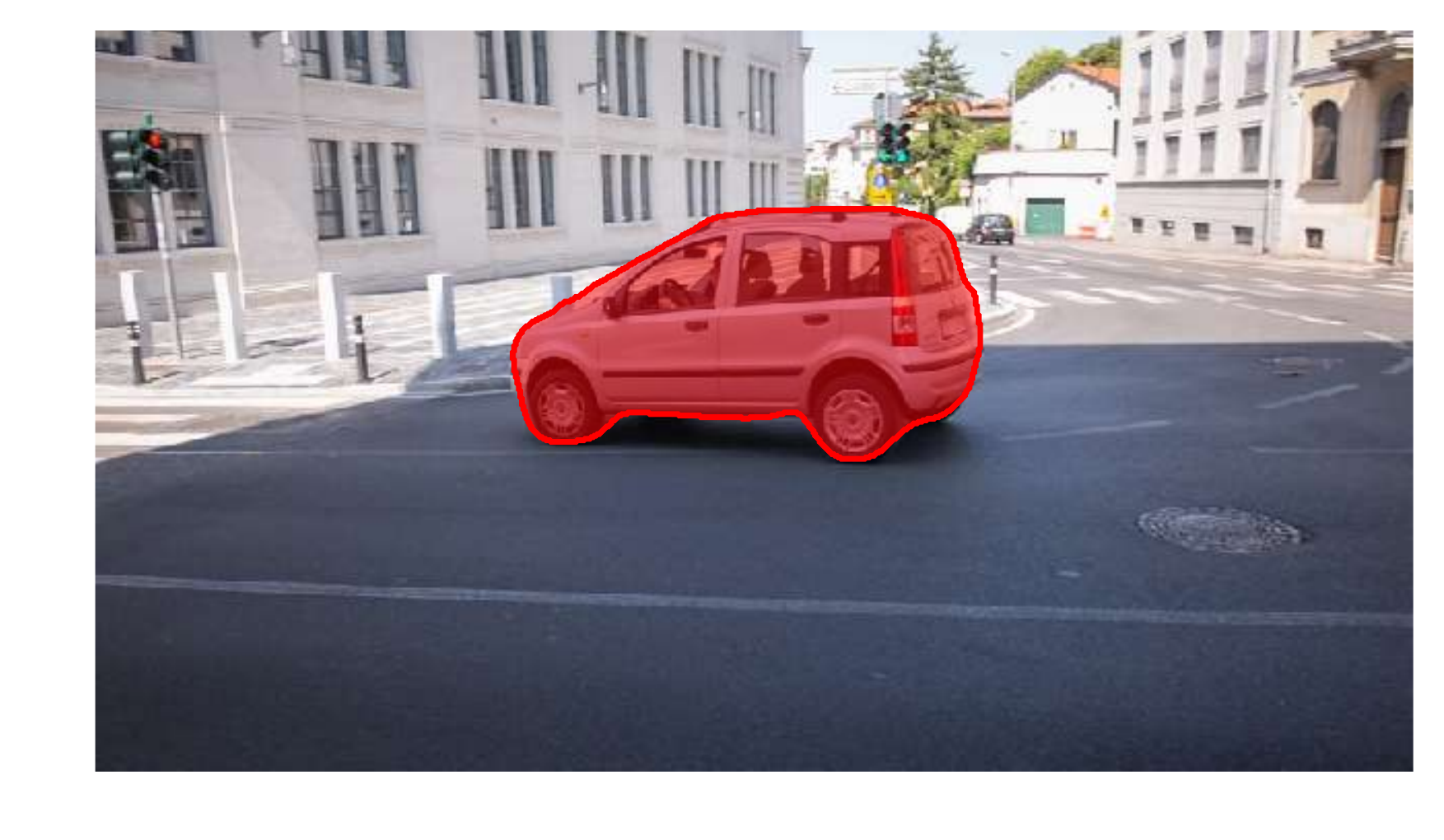}
& \includegraphics[trim={2.5cm 1cm 2.5cm 1cm},clip,width = 1.1in]{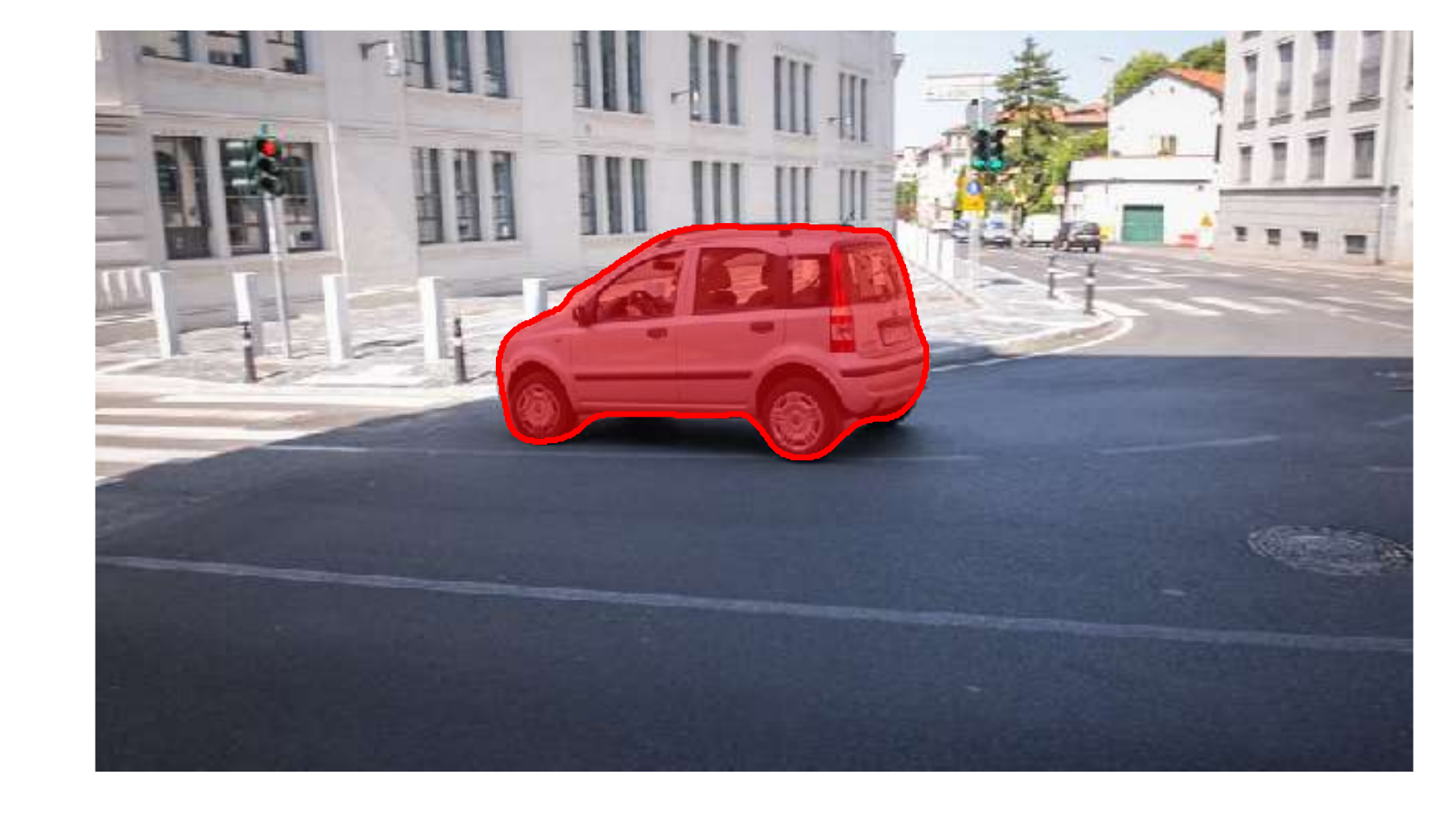}
& \includegraphics[trim={2.5cm 1cm 2.5cm 1cm},clip,width = 1.1in]{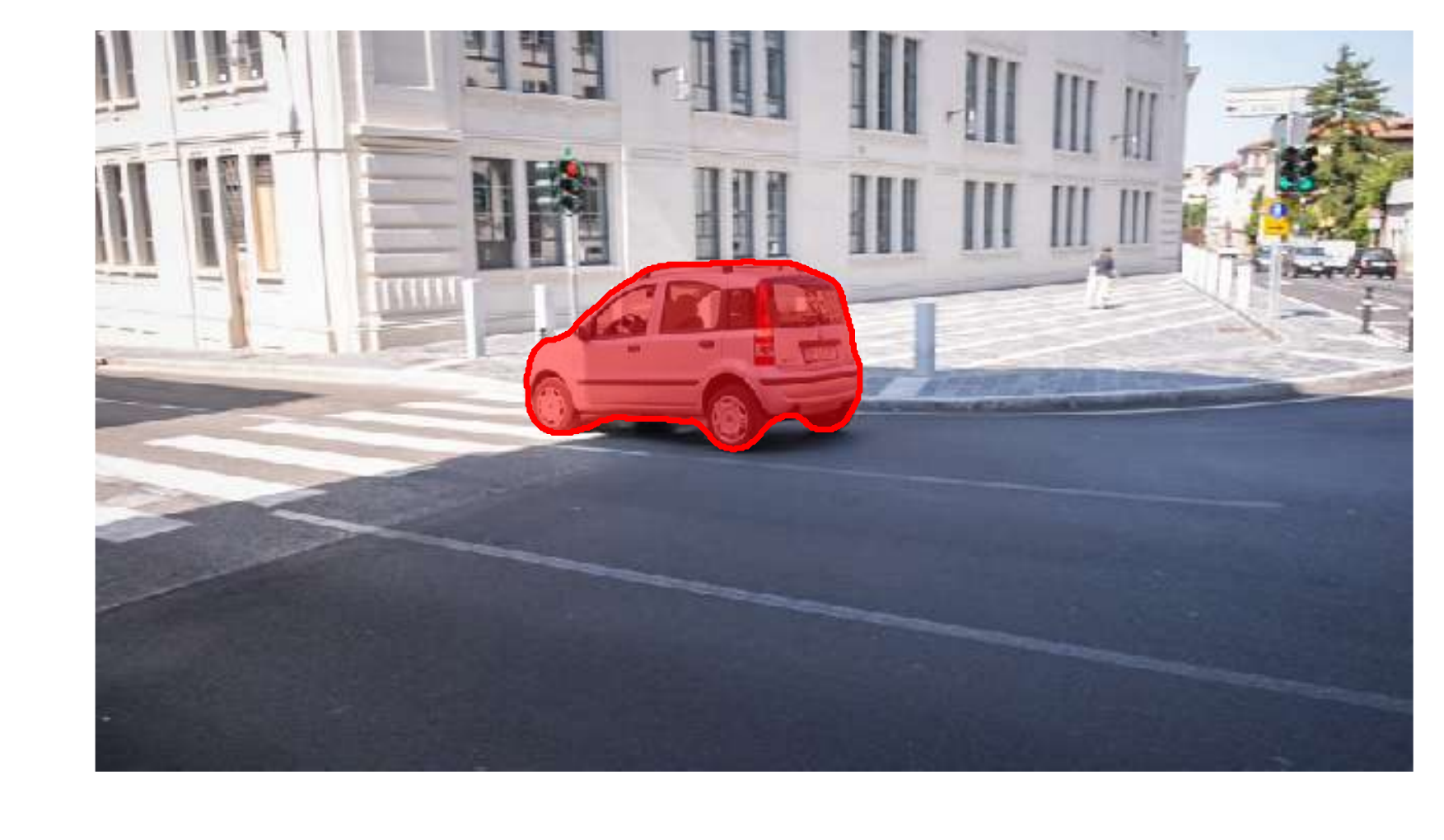}
& \includegraphics[trim={2.5cm 1cm 2.5cm 1cm},clip,width = 1.1in]{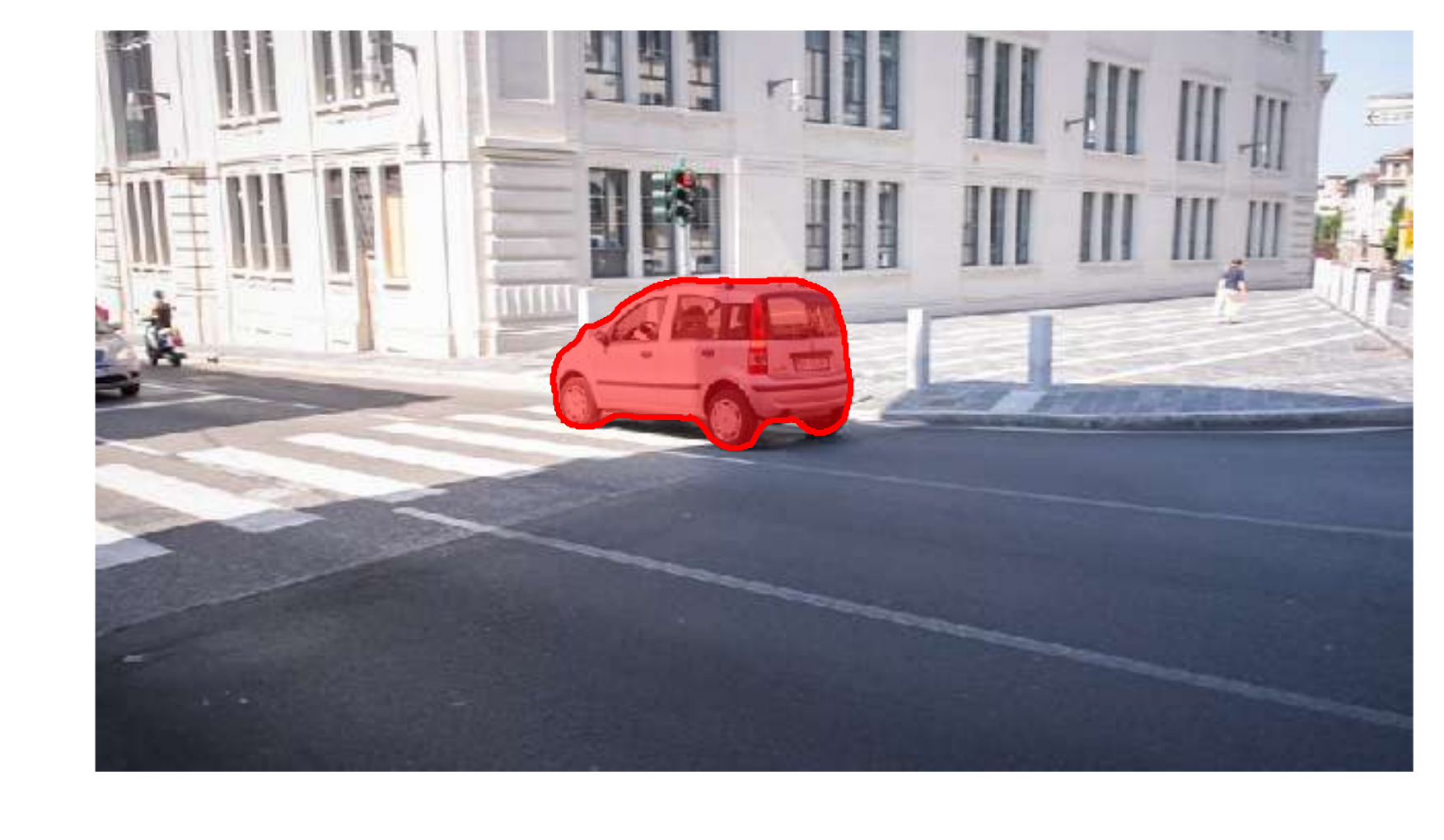}
& \includegraphics[trim={2.5cm 1cm 2.5cm 1cm},clip,width = 1.1in]{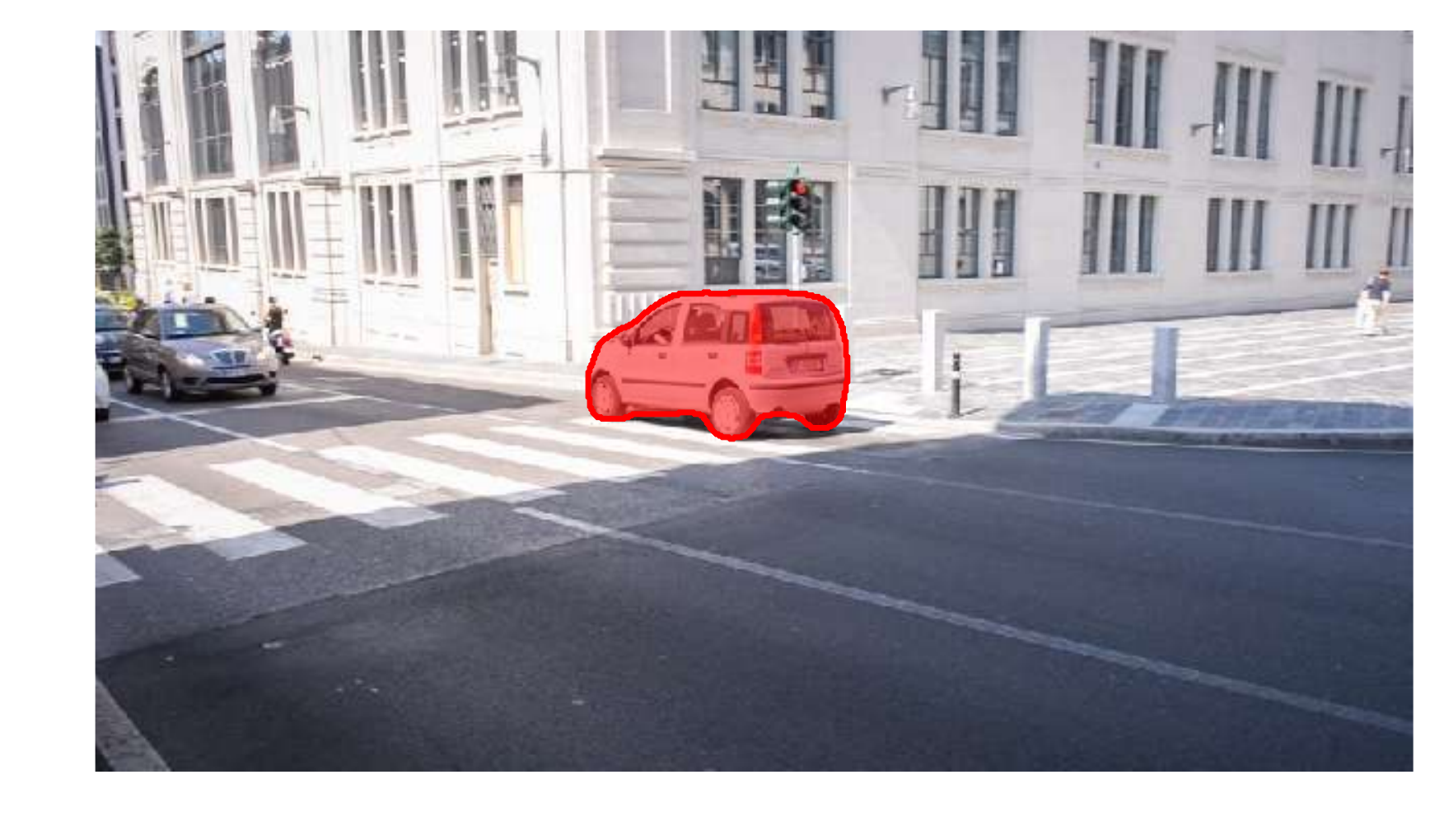}
\\

\mbox{\rotatebox[x=-0.15cm]{90}{\small{Dogs-Jump}}}
\includegraphics[trim={2.5cm 1cm 2.5cm 1cm},clip,width = 1.1in]{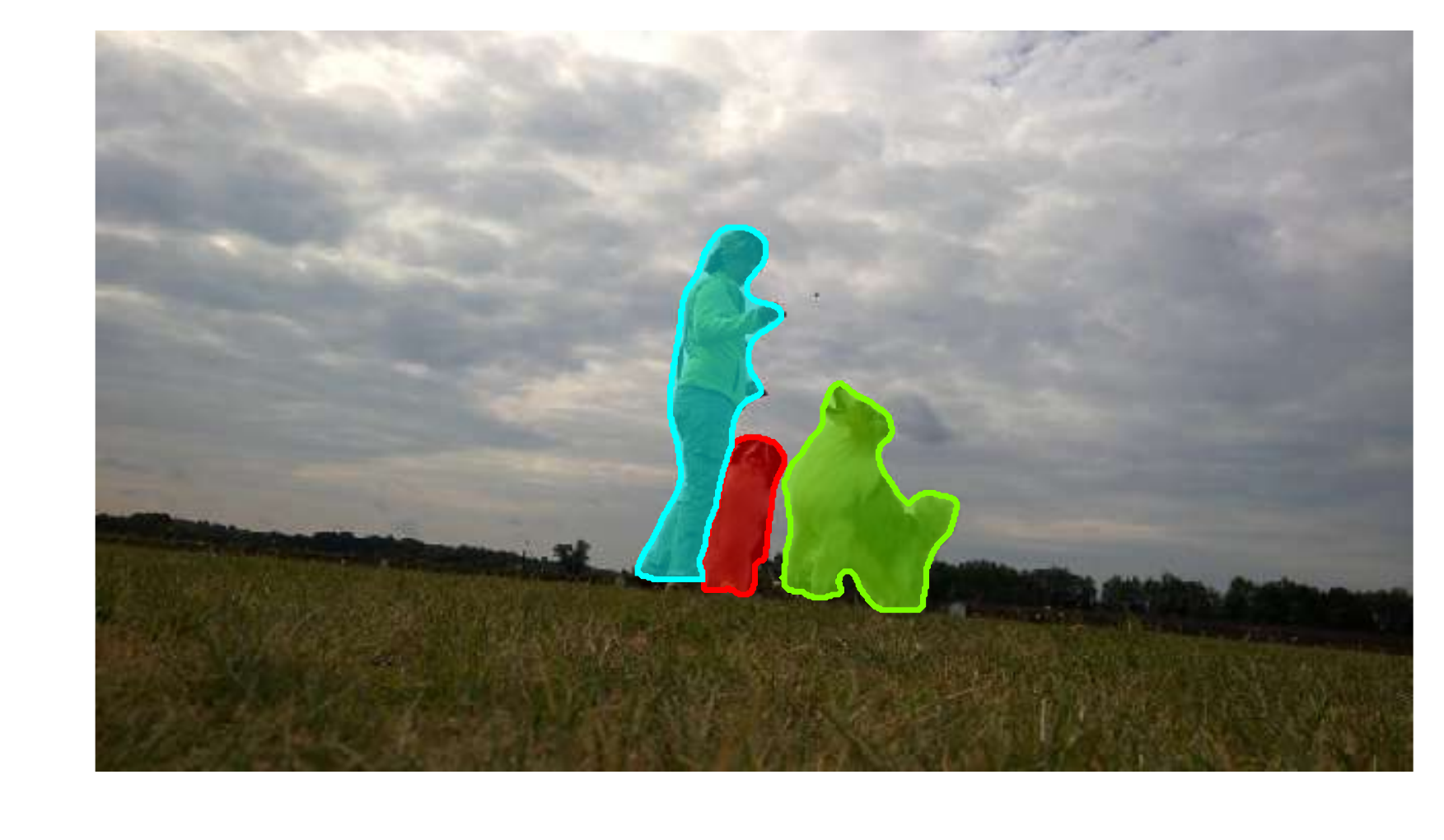}
&\includegraphics[trim={2.5cm 1cm 2.5cm 1cm},clip,width = 1.1in]{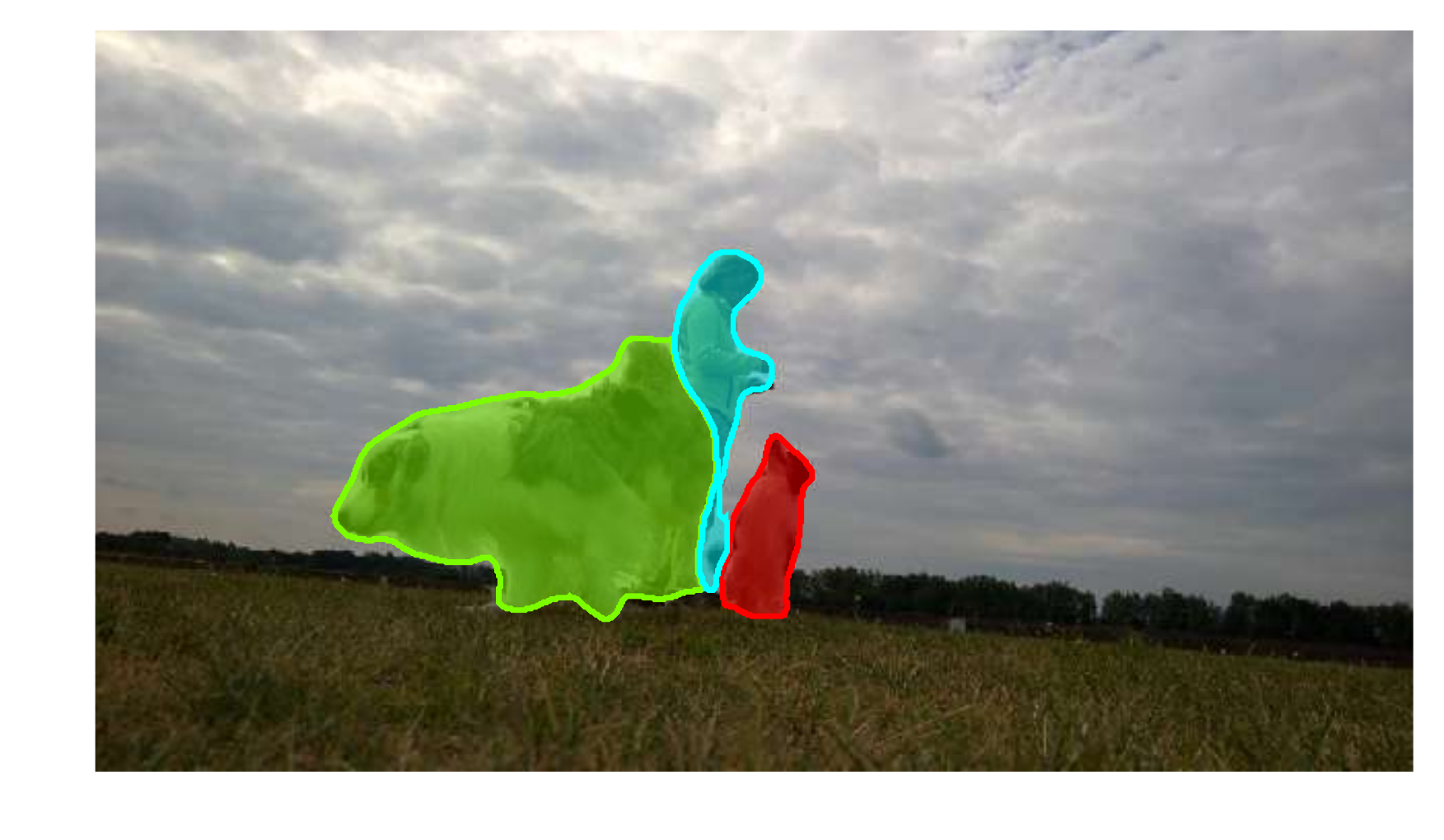}
& \includegraphics[trim={2.5cm 1cm 2.5cm 1cm},clip,width = 1.1in]{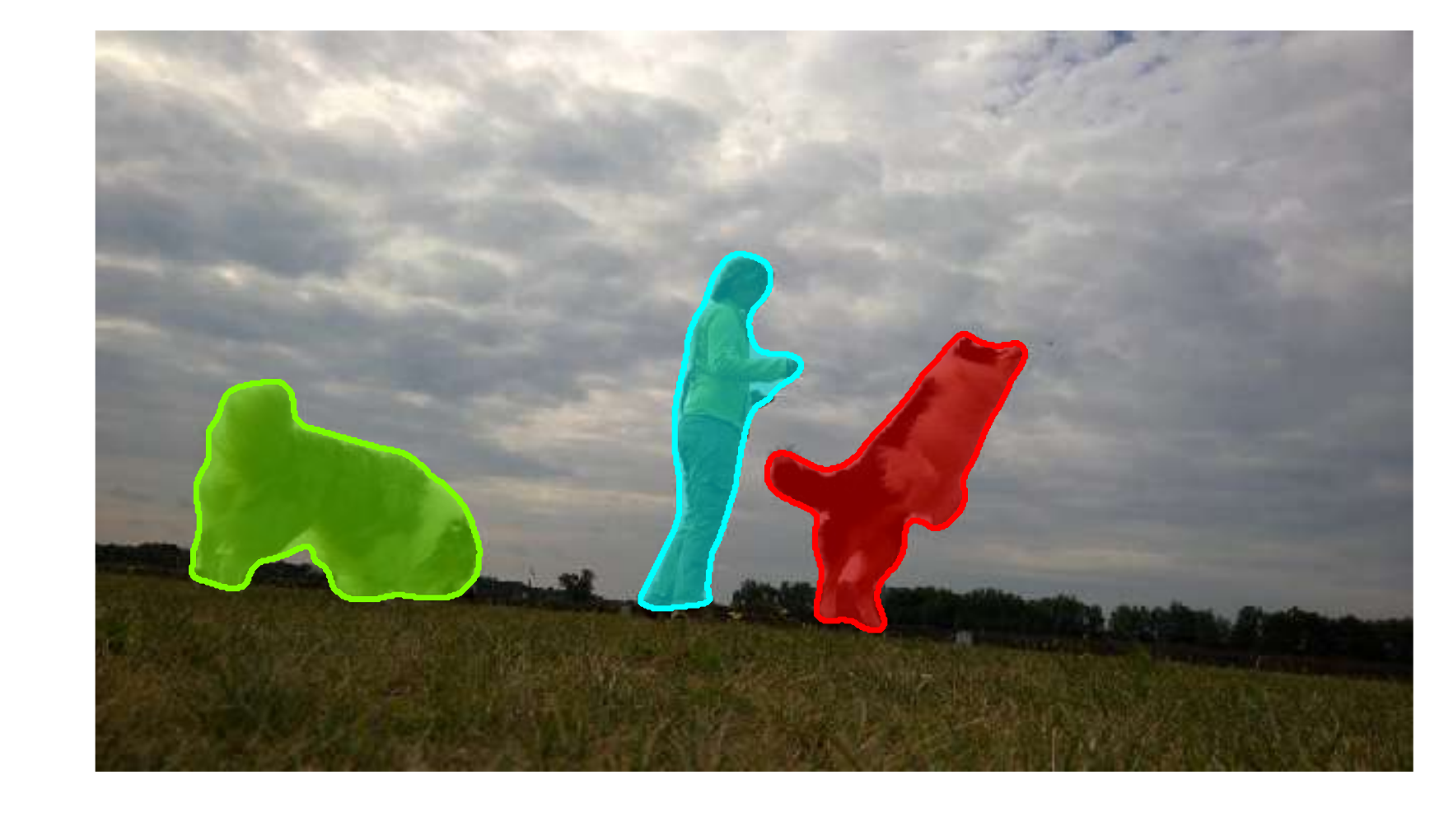}
& \includegraphics[trim={2.5cm 1cm 2.5cm 1cm},clip,width = 1.1in]{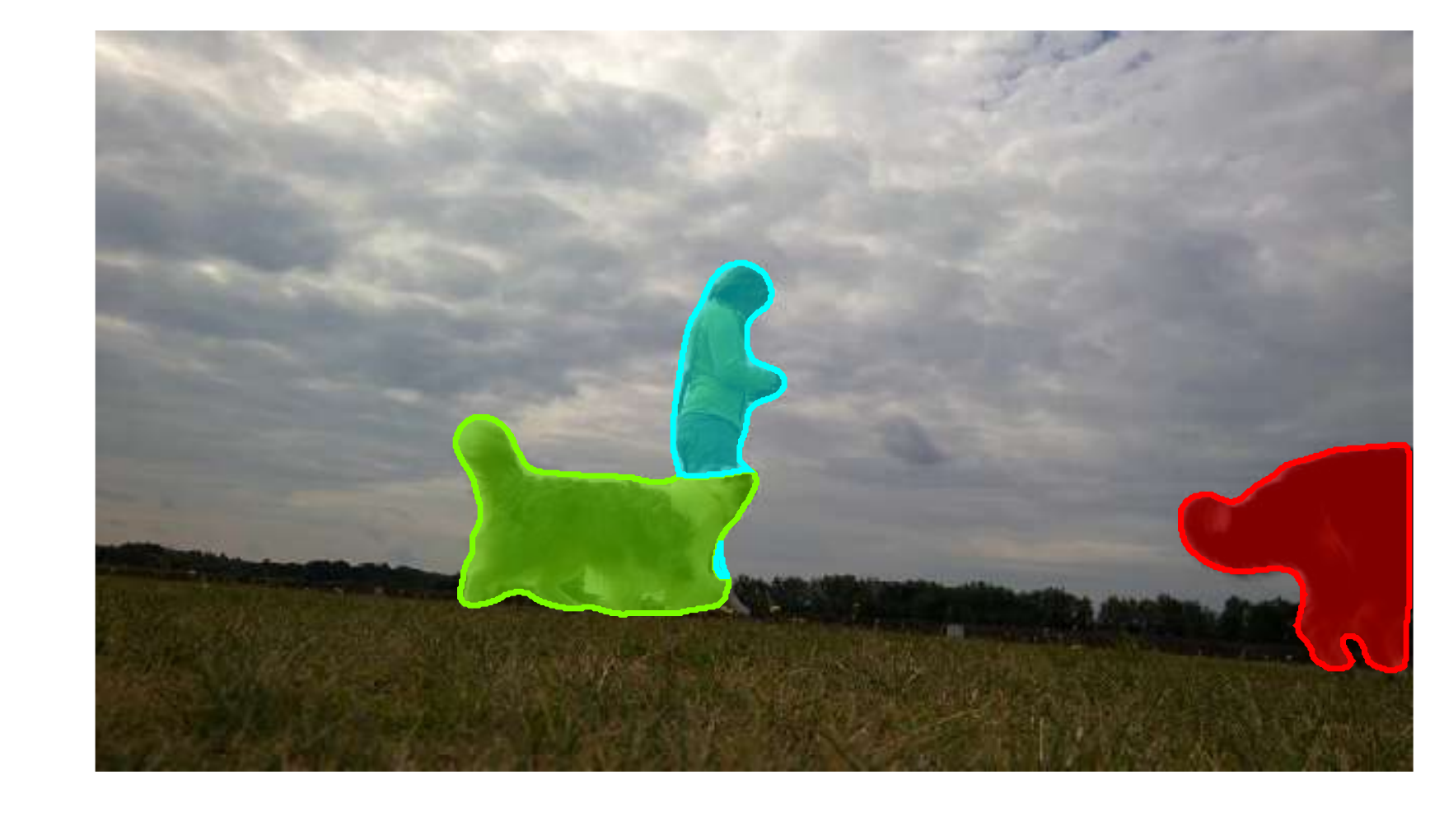}
& \includegraphics[trim={2.5cm 1cm 2.5cm 1cm},clip,width = 1.1in]{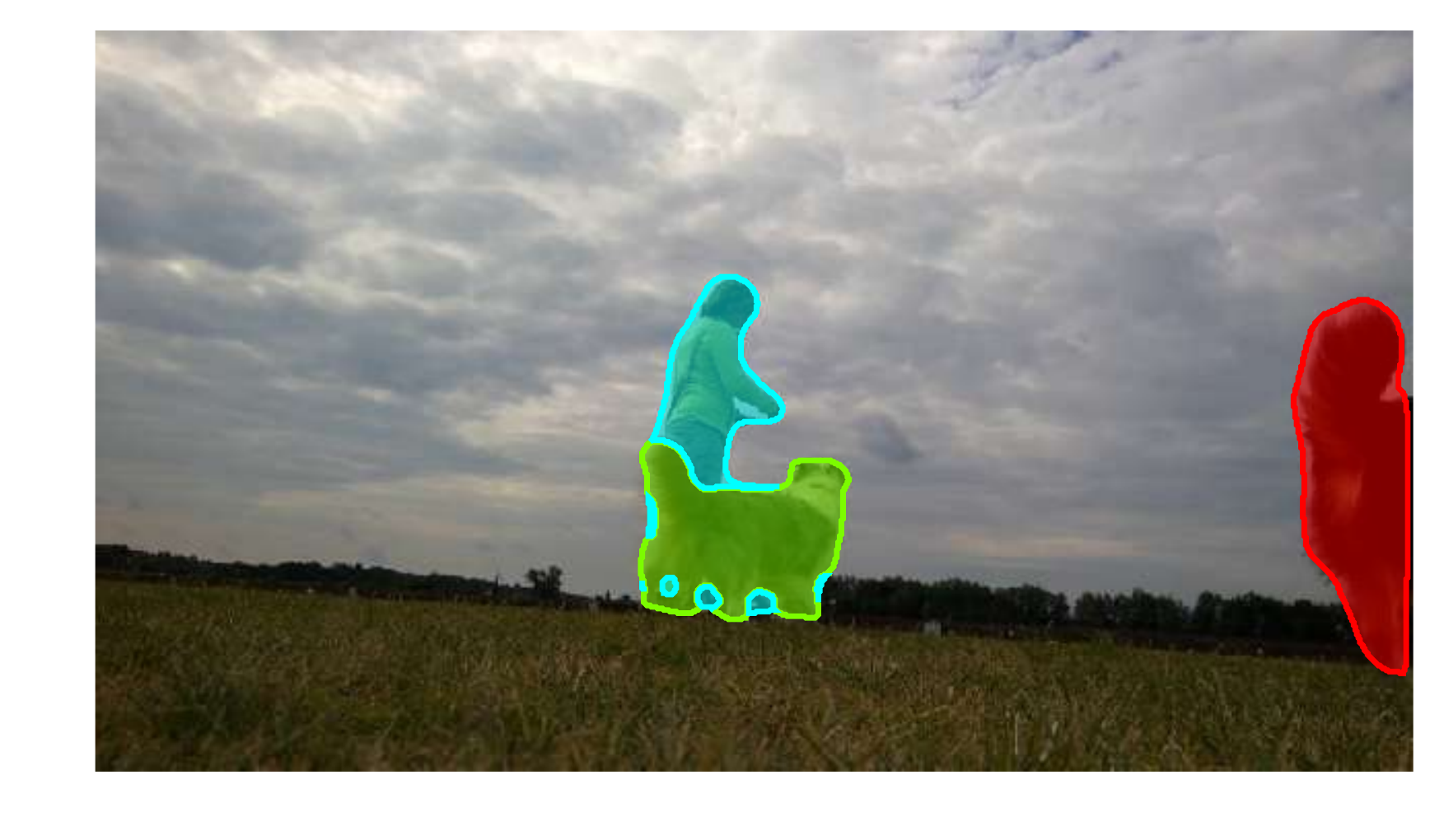}
& \includegraphics[trim={2.5cm 1cm 2.5cm 1cm},clip,width = 1.1in]{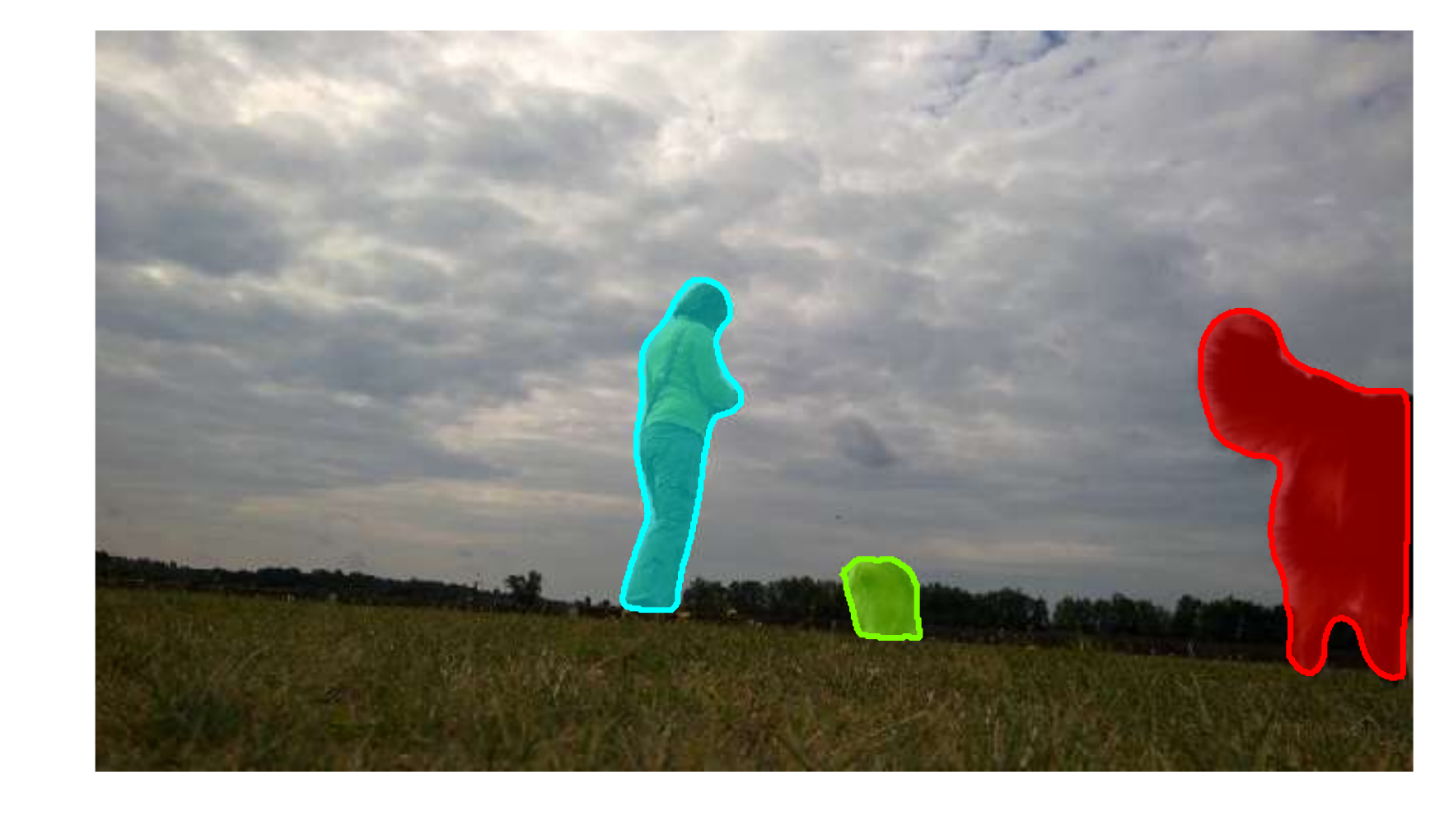}
\\
\mbox{\rotatebox[x=-0.55cm]{90}{\small{Pigs}}}
\includegraphics[trim={2.5cm 1cm 2.5cm 1cm},clip,width = 1.1in]{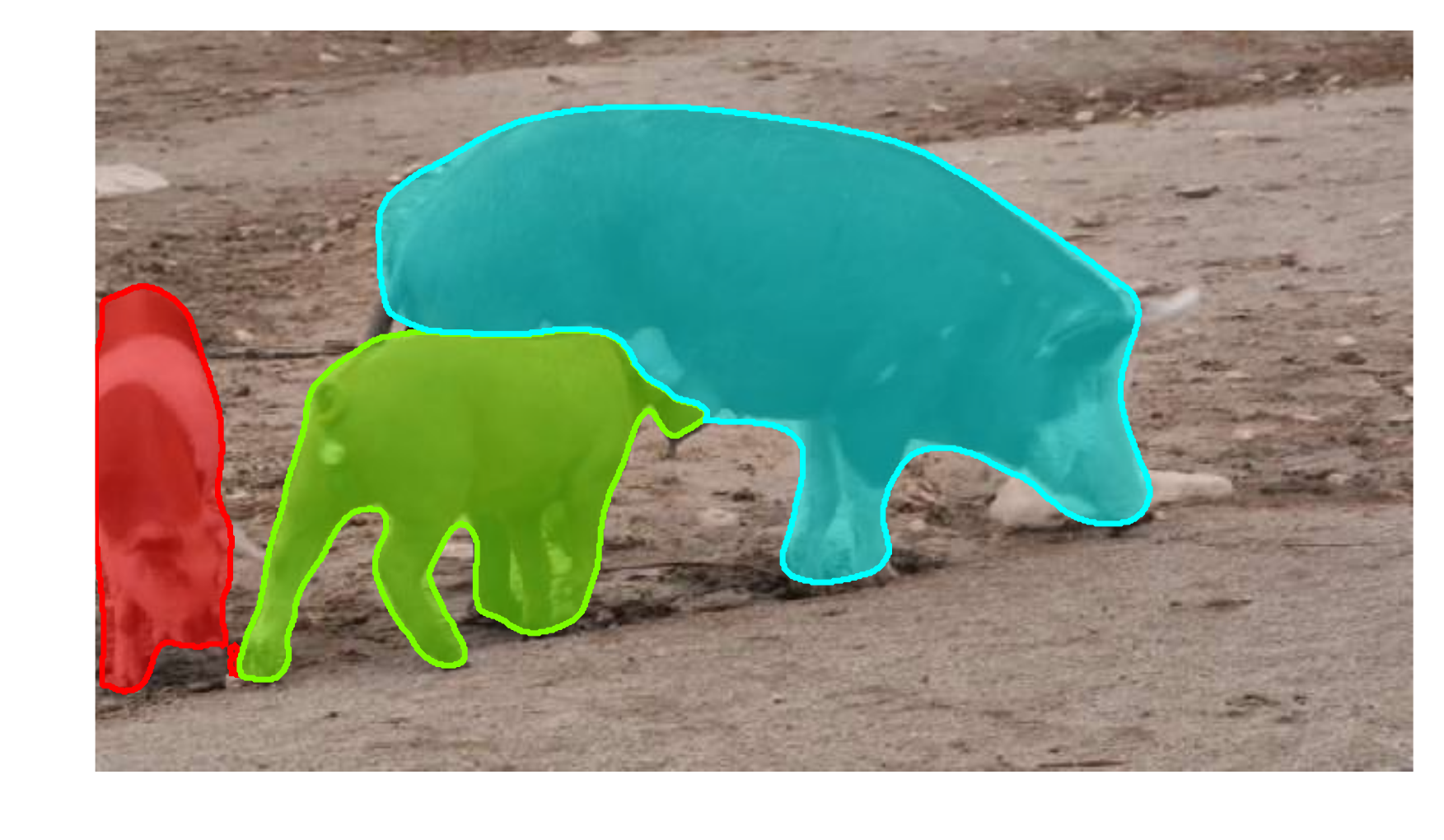}
&\includegraphics[trim={2.5cm 1cm 2.5cm 1cm},clip,width = 1.1in]{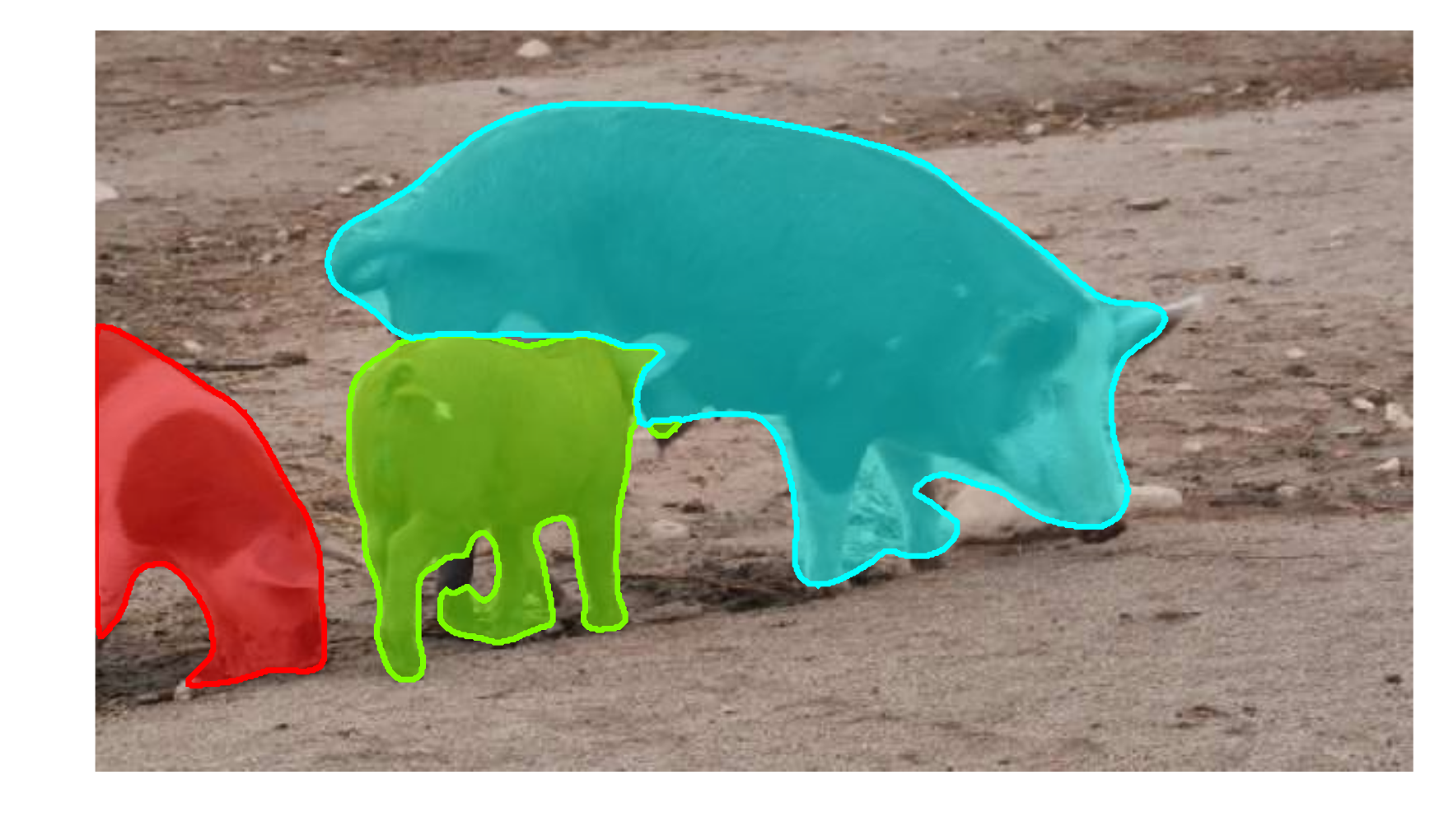}
& \includegraphics[trim={2.5cm 1cm 2.5cm 1cm},clip,width = 1.1in]{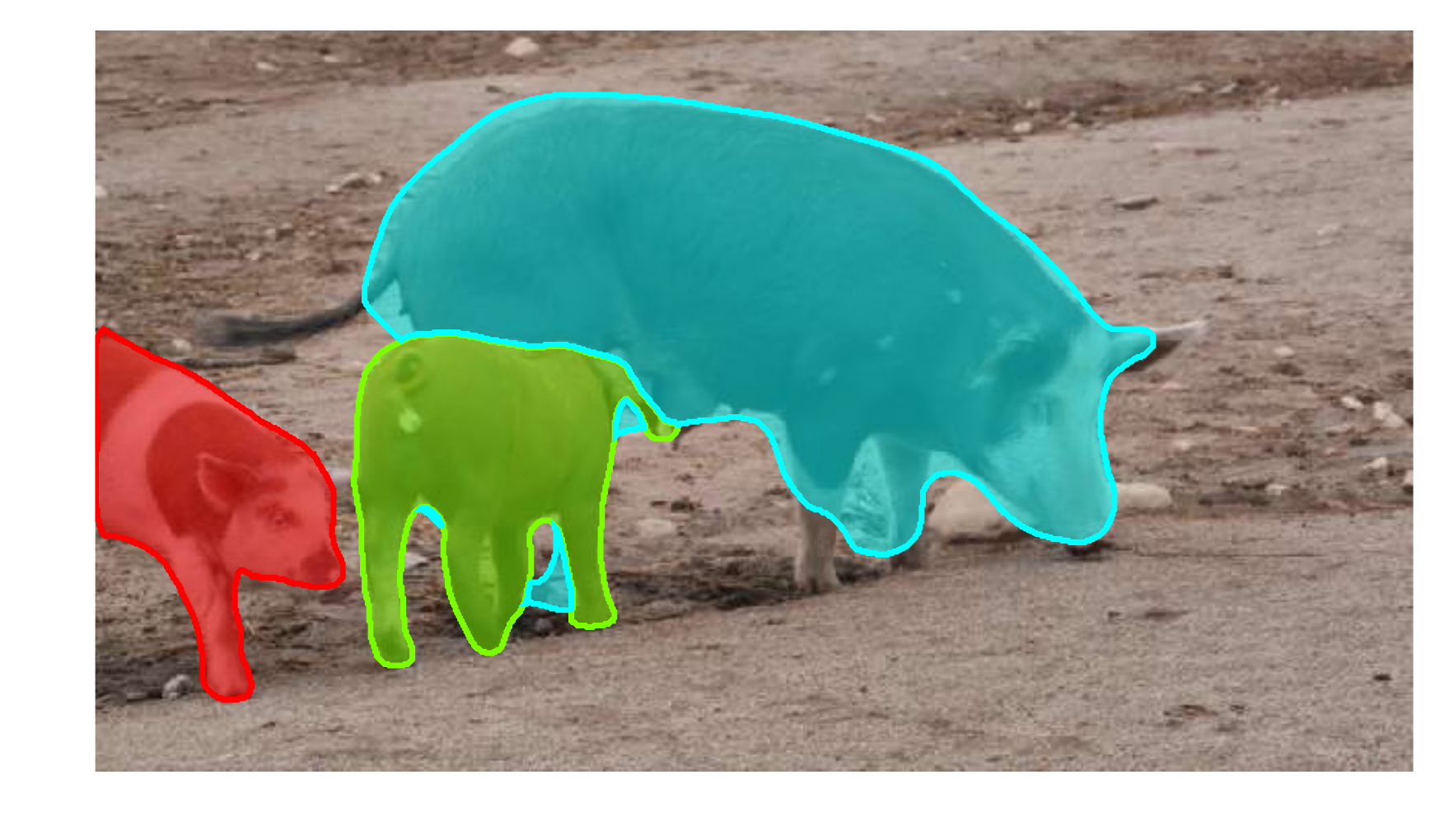}
& \includegraphics[trim={2.5cm 1cm 2.5cm 1cm},clip,width = 1.1in]{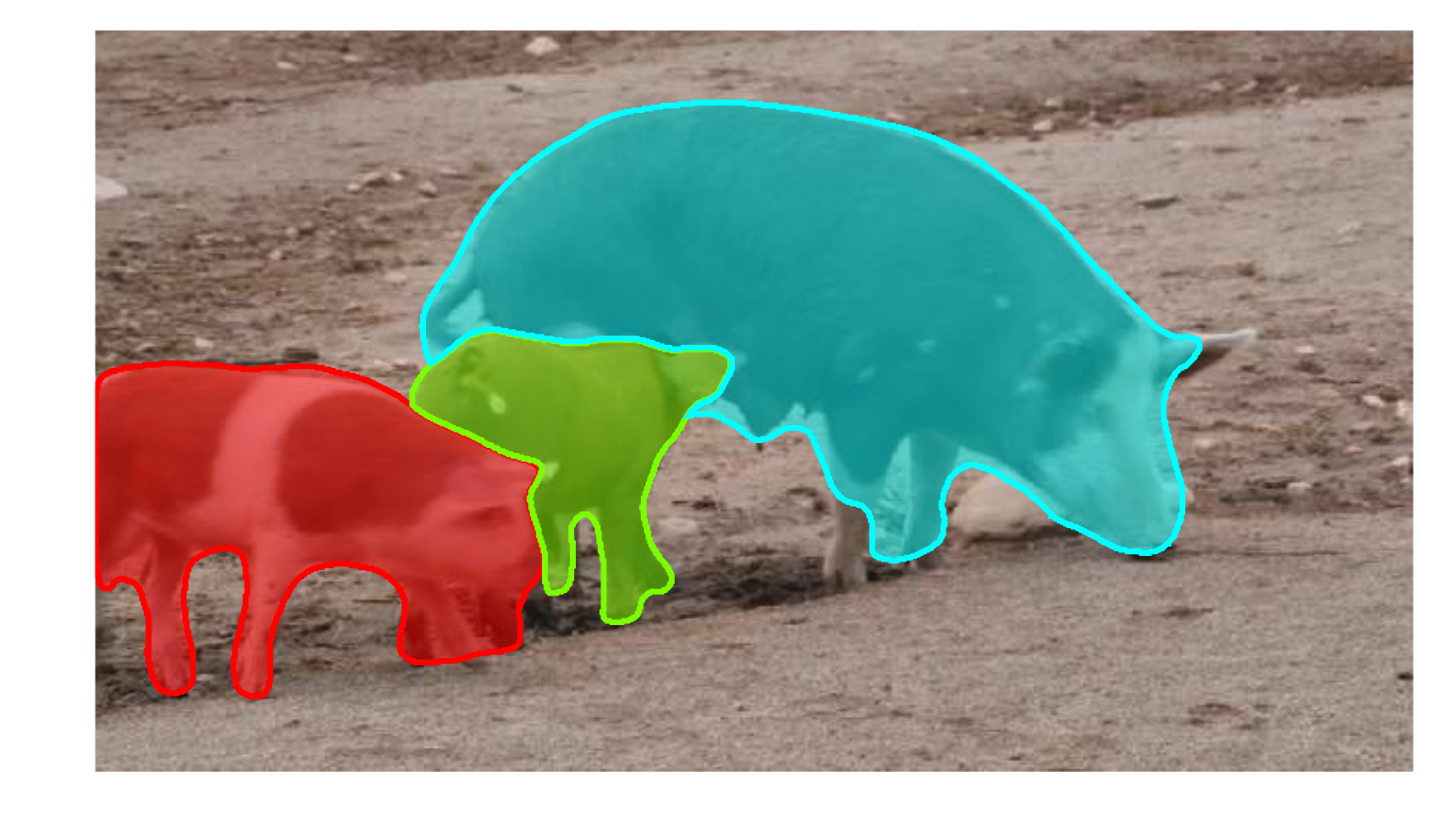}
& \includegraphics[trim={2.5cm 1cm 2.5cm 1cm},clip,width = 1.1in]{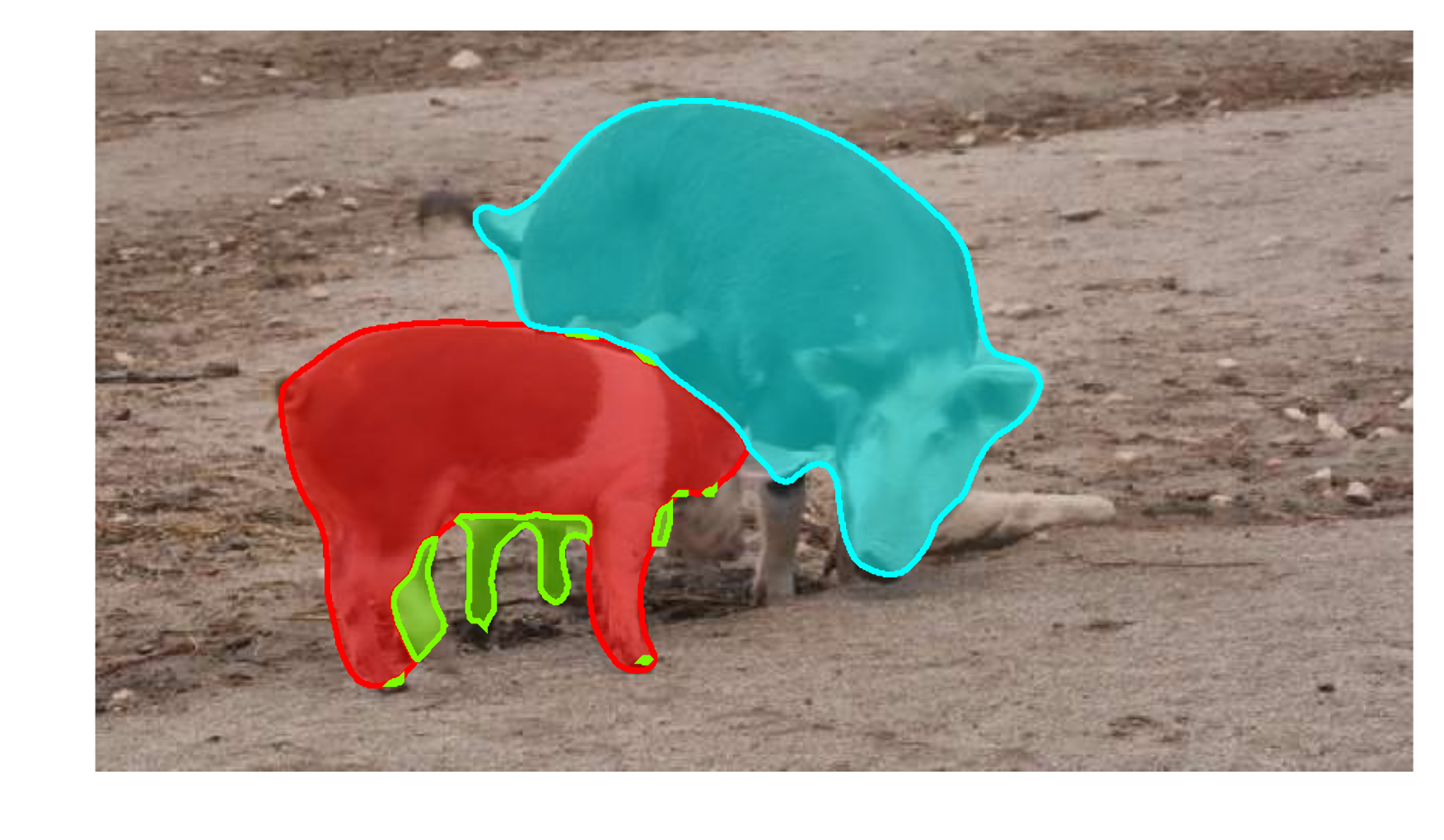}
& \includegraphics[trim={2.5cm 1cm 2.5cm 1cm},clip,width = 1.1in]{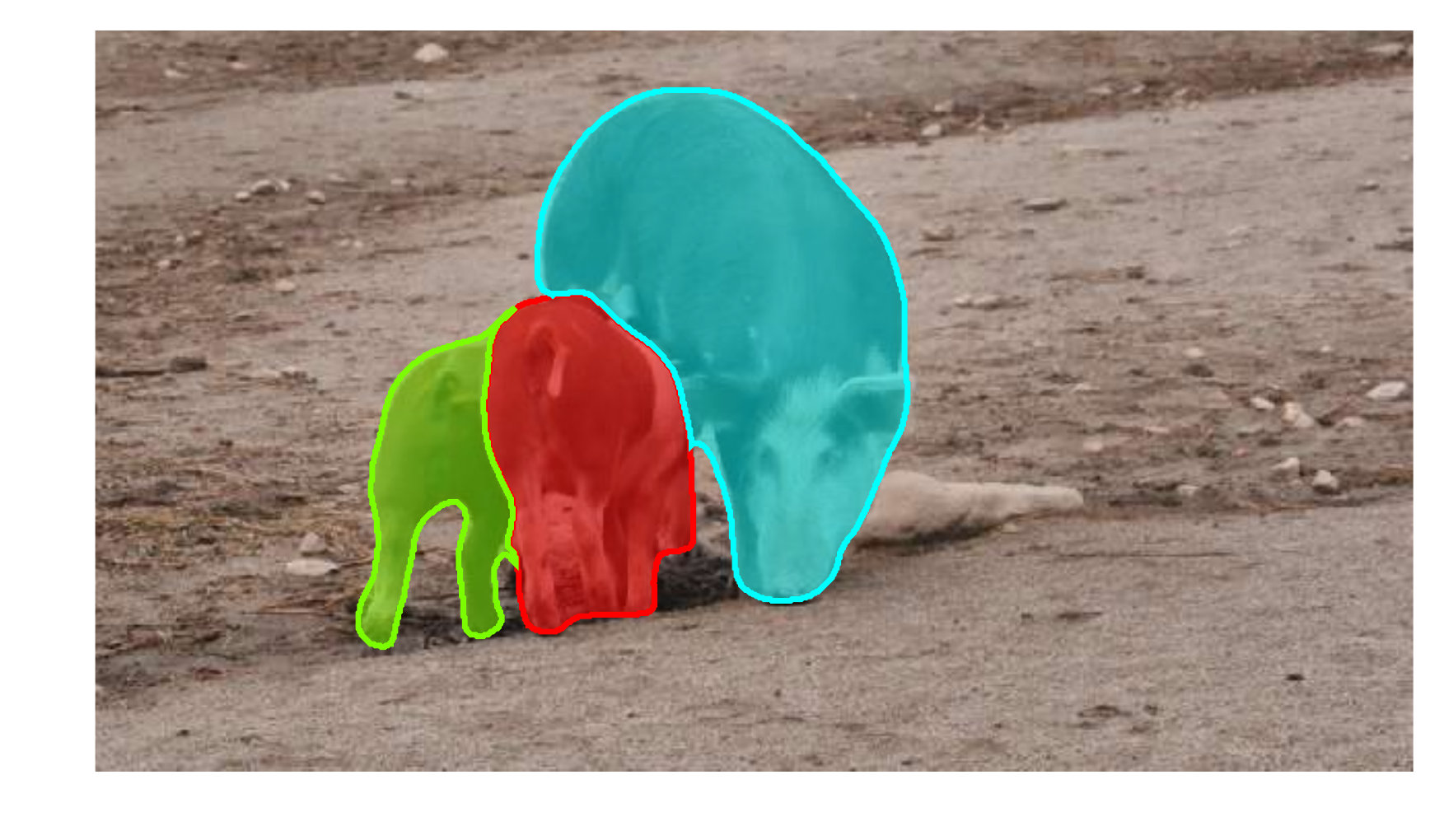}
\\

\end{tabular}

\caption{\bd{Qualitative results} of our method for sequences belonging to both object tracking and video object segmentation benchmarks.
\emph{Basketball} and \emph{Nature} are from VOT-2018~\cite{VOT2018}; \emph{Car-Shadow} is from DAVIS-2016~\cite{perazzi2016benchmark}; \emph{Dogs-Jump} and \emph{Pigs} are from DAVIS-2017~\cite{pont2017davis}.
Multiple masks are obtained from different inferences (with different initialisations).	
}	
\label{fig:davis16}
\vspace{-0.5cm}
\end{figure*}

Qualitative results of SiamMask for both VOT and DAVIS sequences are shown in Figure~\ref{fig:davis16}, \ref{fig:appendix_vot18} and~\ref{fig:appendix_davis16}.
Despite the high speed, SiamMask produces accurate segmentation masks even in presence of distractors.

\subsection{Further analysis}
In this section, we illustrate ablation studies, failure cases and timings of our methods.

\begin{table}[t]
\tablestyle{1.5pt}{1.2}
\begin{tabular}{l|x{16}x{16}|x{32}|x{22}x{22}|x{32}}

		& \texttt{AN}& \texttt{RN} & \texttt{EAO} $\uparrow$ & $\mathcal{J}_{\mathcal{M}\uparrow}$ & $\mathcal{F}_{\mathcal{M}\uparrow}$   & \texttt{Speed} \\[.1em]
\shline
SiamFC             & \cmark  &                   & 0.188   & -             & -             & 86    \\
SiamFC             &         & \cmark            & 0.251   & -             & -             & 40    \\
SiamRPN            & \cmark  &                   & 0.243   & -             & -             & \textbf{200}   \\
SiamRPN            &         & \cmark            & 0.359   & -             & -             & 76    \\ \hline 
SiamMask-2B w/o R        &         & \cmark            & 0.326   & 62.3          & 55.6          & 43    \\
SiamMask w/o R        &         & \cmark            & 0.375   & 68.6          & 57.8          & 58 \\ \hline
SiamMask-2B-score        &         & \cmark            & 0.265   & -             & -             & 40    \\
SiamMask-box        &         & \cmark            &  0.363   & -             & -             & 76    \\ \hline 
SiamMask-2B       &         & \cmark            & 0.334   & 67.4          & 63.5          & 60    \\
SiamMask       &         & \cmark            & \textbf{0.380} & \textbf{71.7} & \textbf{67.8} & 55    \\ 
\end{tabular}
\vspace{1mm}
\caption{Ablation studies on VOT-2018 and DAVIS-2016.}
\label{tab:arch}
\end{table}

\mypar{Network architecture.}
In Table~\ref{tab:arch}, \texttt{AN} and \texttt{RN} denote whether we use AlexNet or ResNet-50 as the shared backbone $f_{\theta}$ (Figure~\ref{fig:schematic}), while with ``w/o R'' we mean that the method does \emph{not} use  the refinement strategy of Pinheiro \etal~\cite{SharpMask}.
From the results of Table~\ref{tab:arch}, it is possible to make several observations.
(1) The first set of rows shows that, by simply updating the architecture of $f_{\theta}$, it is possible to achieve an important performance improvement.
However, this comes at the cost of speed, especially for SiamRPN.
(2) SiamMask-2B and SiamMask considerably improve over their baselines (with same $f_{\theta}$) SiamFC and SiamRPN.
(3) Interestingly, the refinement approach of Pinheiro \etal~\cite{SharpMask} is very important for the contour accuracy $\mathcal{F}_{\mathcal{M}}$, but less so for the other metrics.

\mypar{Multi-task training.}
We conducted two further experiments to disentangle the effect of multi-task training.
Results are reported in Table~\ref{tab:arch}.
To achieve this, we modified the two variants of SiamMask during inference so that, respectively, they report an axis-aligned bounding box from the score branch (SiamMask-2B-score) or the box branch (SiamMask-box).
Therefore, despite having been trained, the mask branch is \emph{not} used during inference.
We can observe how both variants obtain a modest but meaningful improvement with respect to their counterparts (SiamFC and SiamRPN): from 0.251 to 0.265 EAO for the \emph{two-branch} and from 0.359 to 0.363 for the \emph{three-branch} on VOT2018.

\mypar{Timing.} 
SiamMask operates online without any adaptation to the test sequence.
On a single NVIDIA RTX 2080 GPU, we measured an average speed of 55 and 60 frames per second, respectively for the \emph{two-branch} and \emph{three-branch} variants.
Note that the highest computational burden comes from the feature extractor $f_{\theta}$.

\begin{figure}[t]
\centering
\setlength{\tabcolsep}{0.25ex}

\begin{tabular}
{c cccc}
& \includegraphics[trim={2cm 0cm 2cm 0cm},clip,height = 1in]{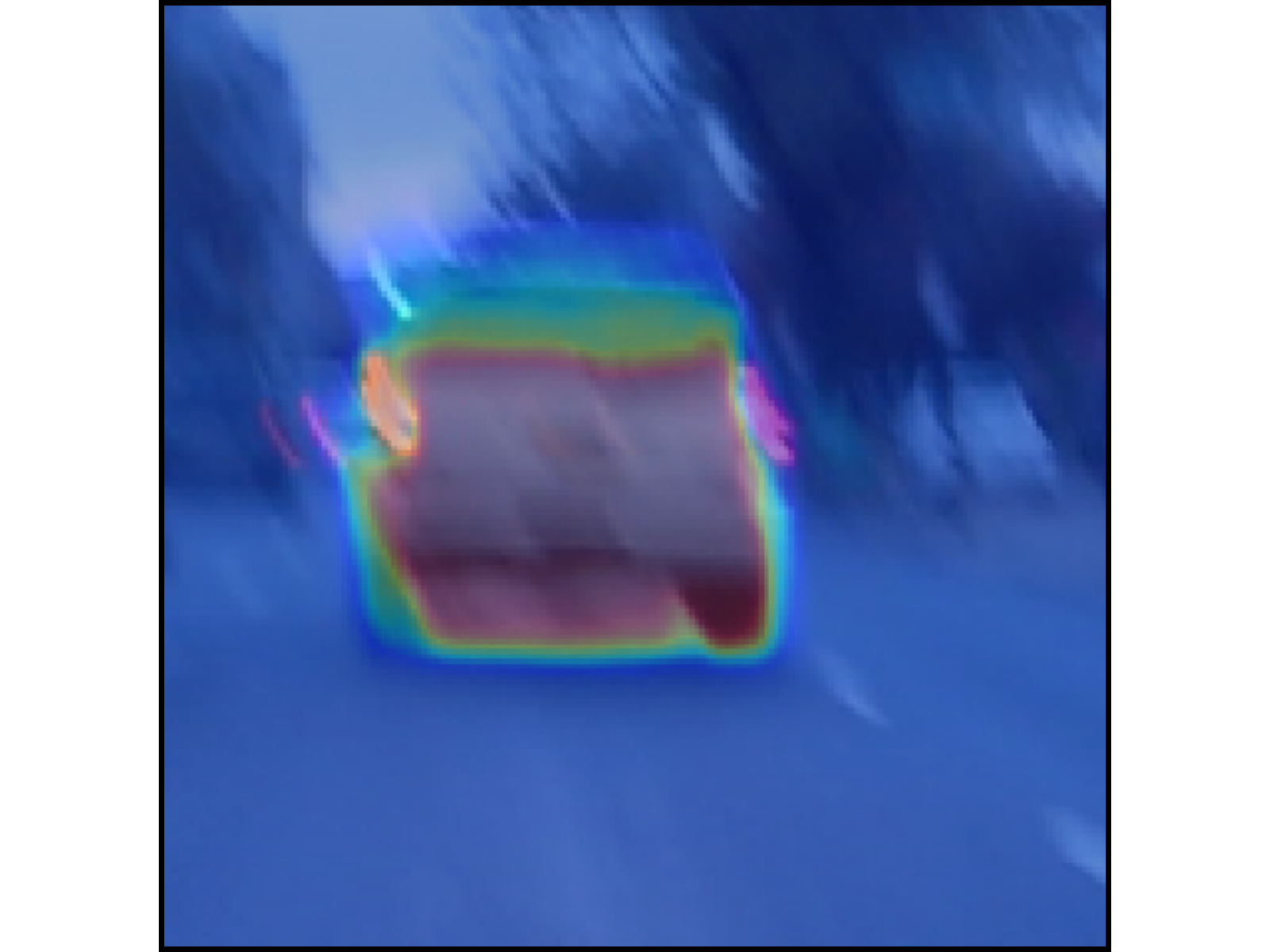}
& \includegraphics[trim={10cm 18.5cm 0.5cm 4.3cm},clip,height = 1in]{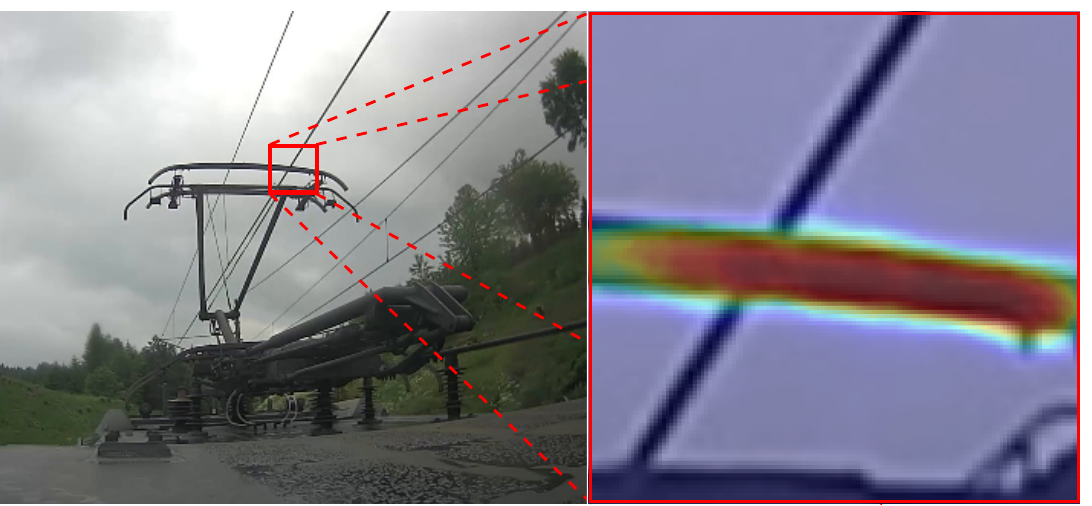}
\\

\end{tabular}

\caption{Failure cases: motion blur and ``non-object'' instance.}
\label{fig:fail}
\vspace{-0.2cm}
\end{figure}

\mypar{Failure cases.}
Finally, we discuss two scenarios in which SiamMask fails: motion blur and ``non-object'' instance (Figure~\ref{fig:fail}).
Despite being different in nature, these two cases arguably arise from the complete lack of similar training samples in a training sets, which are focused on objects that can be unambiguously discriminated from the foreground.

\section{Conclusion}
\label{sec:conclusion}
In this paper we introduced SiamMask, a simple approach that enables fully-convolutional Siamese trackers to produce class-agnostic binary segmentation masks of the target object.
We show how it can be applied with success to both tasks of visual object tracking \emph{and} semi-supervised video object segmentation, showing better accuracy than state-of-the-art trackers and, at the same time, the fastest speed among VOS methods.
The two variants of SiamMask we proposed are initialised with a simple bounding box, operate online, run in real-time and do not require any adaptation to the test sequence.
We hope that our work will inspire further studies that consider the two problems of visual object tracking and video object segmentation together.

\mypar{Acknowledgements.}
This work was supported by the ERC grant ERC-2012-AdG 321162-HELIOS, EPSRC grant Seebibyte EP/M013774/1 and EPSRC/MURI grant EP/N019474/1.
We would also like to acknowledge the support of the Royal Academy of Engineering and FiveAI Ltd.
Qiang Wang is partly supported by the NSFC (Grant No. 61751212, 61721004 and U1636218).

{\small
\bibliographystyle{ieee}
\bibliography{biblio.bib}
}
\clearpage
\appendix
\section{Architectural details}
\label{sec:appendix_architecture}
\mypar{Network backbone.} 
Table~\ref{tab:backbone} illustrates the details of our \emph{backbone} architecture ($f_{\theta}$ in the main paper).
For both variants, we use a ResNet-50~\cite{he2016deep} until the final convolutional layer of the 4-th stage.
In order to obtain a higher spatial resolution in deep layers, we reduce the output stride to 8 by using convolutions with stride 1. 
Moreover, we increase the receptive field by using dilated convolutions~\cite{chen2018deeplab}.
Specifically, we set the stride to 1 and the dilation rate to 2 in the $3{\times}3$ conv layer of \texttt{conv4\_1}. 
Differently to the original ResNet-50, there is no downsampling in \texttt{conv4\_x}.
We also add to the backbone an \emph{adjust} layer (a $1{\times}1$ convolutional layer with 256 output channels). 
Examplar and search patches share the network's parameters from \texttt{conv1} to \texttt{conv4\_x}, while the parameters of the \textit{adjust} layer are not shared.
The output features of the adjust layer are then depth-wise cross-correlated, resulting a feature map of size 17{$\times$}17.

\mypar{Network heads.} 
The network architecture of the branches of both variants are shows in Table~\ref{tab:three} and~\ref{tab:two}.
The \texttt{conv5} block in both variants contains a normalisation layer and ReLU non-linearity while \texttt{conv6} only consists of a $1{\times}1$ convolutional layer.

\begin{figure}
\begin{center}
\includegraphics[width=0.4\textwidth]{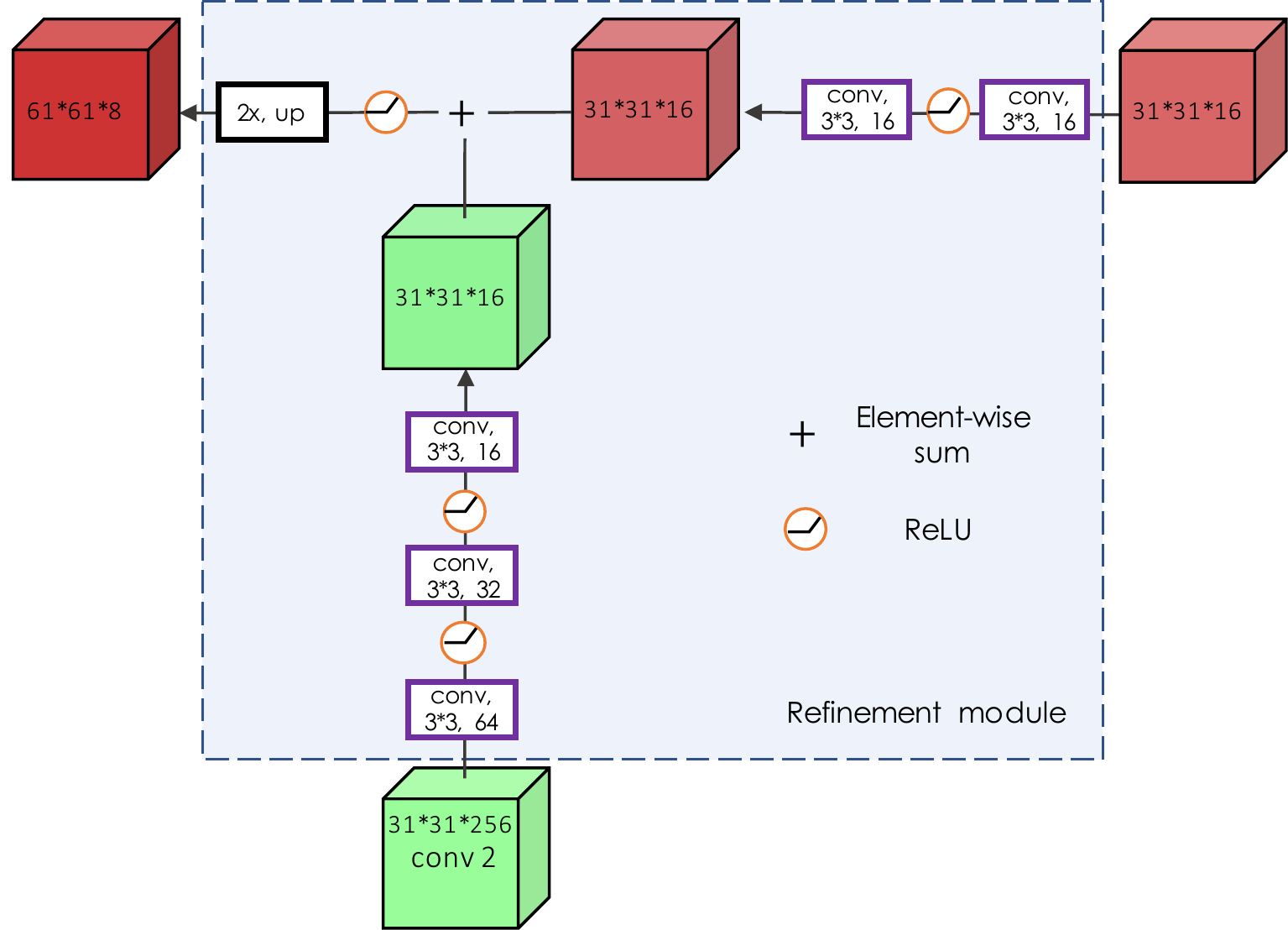}
\end{center}
\vspace{-0.2cm}
\caption{Example of a refinement module $U_3$.}
\label{fig:u}
\end{figure}

\begin{figure*}[h]
\begin{center}
\includegraphics[width=0.9 \textwidth]{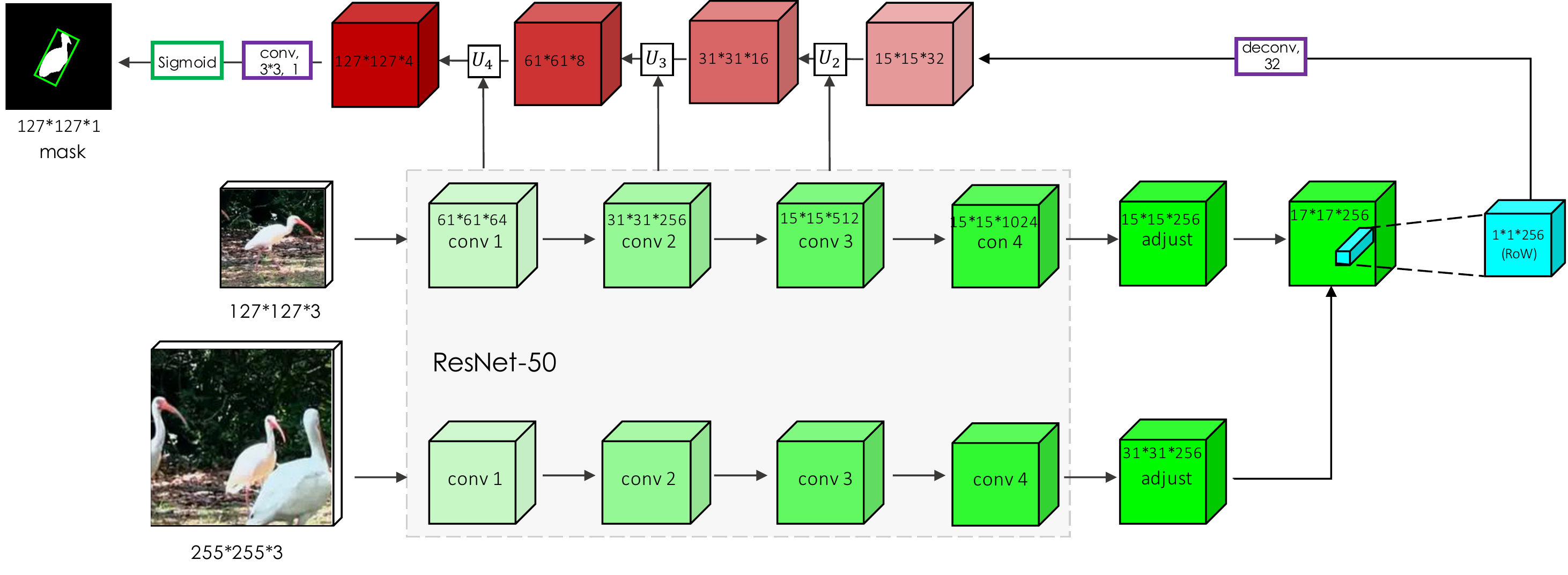}
\end{center}
\vspace{-0.2cm}
\caption{Schematic illustration of the stacked refinement modules.}
\label{fig:rm}
\end{figure*}

\mypar{Mask refinement module.} 
With the aim of producing a more accurate object mask, we follow the strategy of~\cite{SharpMask}, which merges low and high resolution features using multiple~\textit{refinement} modules made of upsampling layers and skip connections.
Figure~\ref{fig:u} gives an example of refinement module $U_3$, while Figure~\ref{fig:rm} illustrates how a mask is generated with stacked refinement modules.

\newcommand{\blocka}[2]{\multirow{3}{*}{\(\left[\begin{array}{c}\text{3$\times$3, #1}\\[-.1em] \text{3$\times$3, #1} \end{array}\right]\)$\times$#2}
}
\newcommand{\blockb}[3]{\multirow{3}{*}{\(\left[\begin{array}{c}\text{1$\times$1, #2}\\[-.1em] \text{3$\times$3, #2}\\[-.1em] \text{1$\times$1, #1}\end{array}\right]\)$\times$#3}
}
\renewcommand\arraystretch{1.1}
\setlength{\tabcolsep}{2pt}
\begin{table}[h]
\begin{center}
\resizebox{1\linewidth}{!}{
\begin{tabular}{c|c|c|c}
\hline
\textit{block} & \textit{examplar} output size  & \textit{search} output size & \textit{backbone} \\
\hline
conv1 & 61$\times$61 & 125$\times$125 & 7$\times$7, 64, stride 2\\
\hline
\multirow{4}{*}{conv2\_x} &\multirow{4}{*}{31$\times$31} & \multirow{4}{*}{63$\times$63} & 3$\times$3 max pool, stride 2 \\
  &  &  &\blockb{256}{64}{3} \\
  &  &  \\
  &  &  \\
\hline
\multirow{3}{*}{conv3\_x} &\multirow{3}{*}{15$\times$15}  &\multirow{3}{*}{31$\times$31}  & \blockb{512}{128}{4} \\
  &  &  \\
  &  &  \\
\hline
\multirow{3}{*}{conv4\_x} &\multirow{3}{*}{15$\times$15} &\multirow{3}{*}{31$\times$31}   & \blockb{1024}{256}{6} \\
  &  &   \\
  &  &   \\
\hline
		\textit{adjust} & $15{\times}15$ & $31{\times}31$ & $1{\times}1$, 256\\
\hline
xcorr&   \multicolumn{2}{c|}{17 $\times$ 17} & depth-wise \\
\hline
\end{tabular}
}
\end{center}
\caption{Backbone architecture. Details of each building block are shown in square brackets.
}
\label{tab:backbone}
\vspace{-0.2cm}
\end{table}

\renewcommand\arraystretch{1.1}
\setlength{\tabcolsep}{2pt}
\begin{table}
\begin{center}
\resizebox{0.75\linewidth}{!}{
\begin{tabular}{c|c|c|c}
\hline
\textit{block}  & score   & box  & mask\\
\hline
conv5& 1 $\times$ 1, 256  &  1 $\times$ 1, 256  & 1 $\times$ 1, 256  \\

\hline
conv6 & 1 $ \times $ 1, 2$k$ & 1 $\times$ 1, 4$k$  & 1 $\times$ 1, (63 $ \times $ 63) \\

\hline
\end{tabular}
}
\end{center}
\caption{Architectural details of the \textit{three-branch} head.
$k$ denotes the number of anchor boxes per RoW.
}
\label{tab:three}
\vspace{-0.2cm}
\end{table}

\renewcommand\arraystretch{1.1}
\setlength{\tabcolsep}{2pt}
\begin{table}
\begin{center}
\resizebox{0.6\linewidth}{!}{
\begin{tabular}{c|c|c}
\hline
\textit{block} & score  & mask\\
\hline
conv5& 1 $\times$ 1, 256  &  1 $\times$ 1, 256    \\

\hline
conv6 & 1 $ \times $ 1, 1 & 1 $\times$ 1, (63 $ \times $ 63) \\

\hline
\end{tabular}
}
\end{center}
\caption{Architectural details of the \textit{two-branch} head.
}
\label{tab:two}
\vspace{-0.2cm}
\end{table}

\section{Further qualitative results}
\label{sec:appendix_qualitative}
\begin{figure}
\begin{center}
\includegraphics[width=0.47 \textwidth]{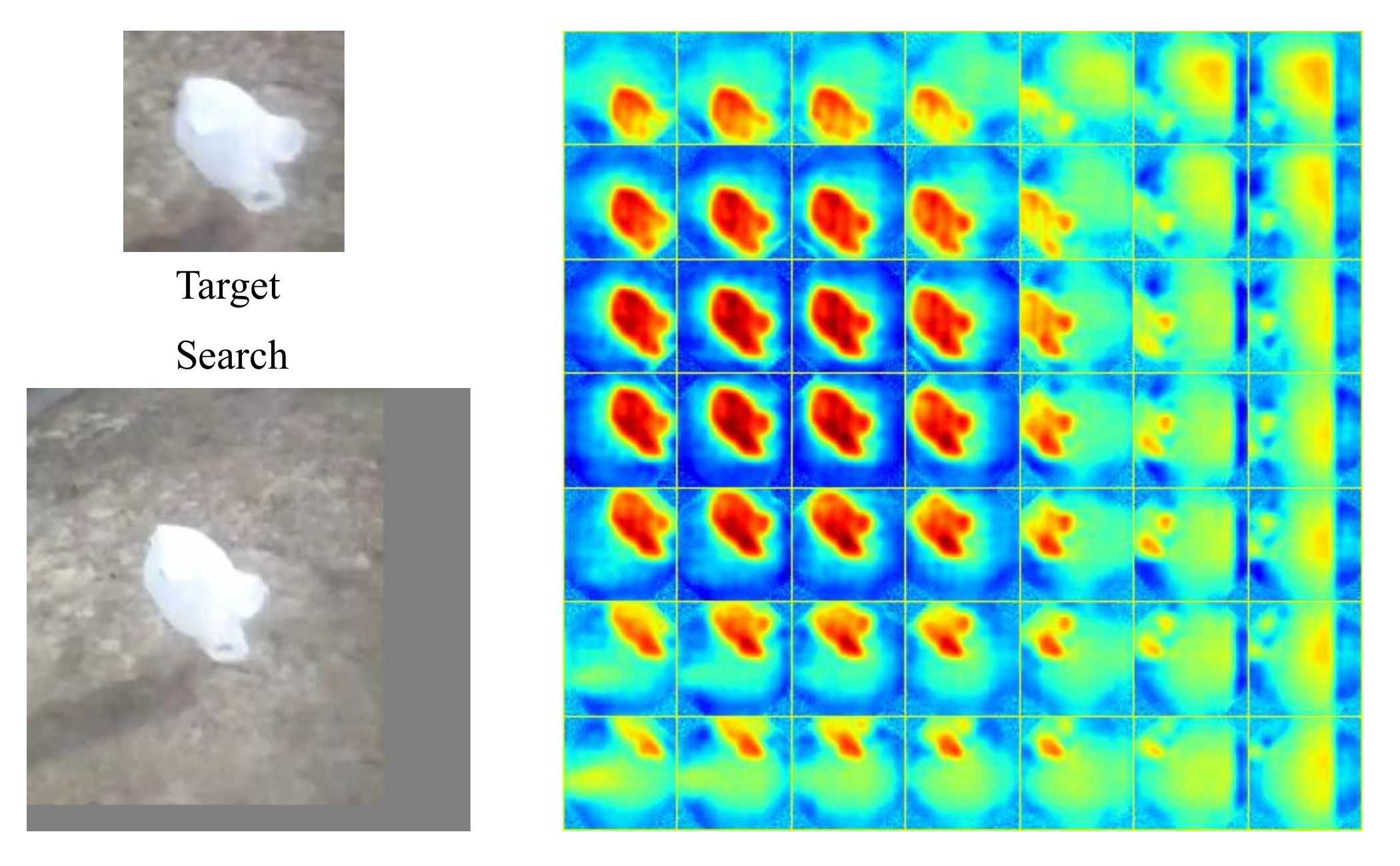}
\end{center}
\vspace{-0.2cm}
\caption{Score maps from the mask branch at different locations.}
\label{fig:map}
\end{figure}

\mypar{Different masks at different locations.}
Our model generates a mask for each RoW.
During inference, we rely on the score branch to select the final output mask (using the location attaining the maximum score).
The example of Figure~\ref{fig:map} illustrates the multiple output masks produced by the mask branch, each corresponding to a different RoW.

\begin{figure*}
\centering
\setlength{\tabcolsep}{0.25ex}

\begin{tabular}
{cccccc cccccc}
\mbox{\centering\rotatebox[x=-0.55cm]{90}{\small{butterfly}}}
\includegraphics[trim={2.5cm 2cm 2.5cm 0.5cm},clip,width = 1.1in]{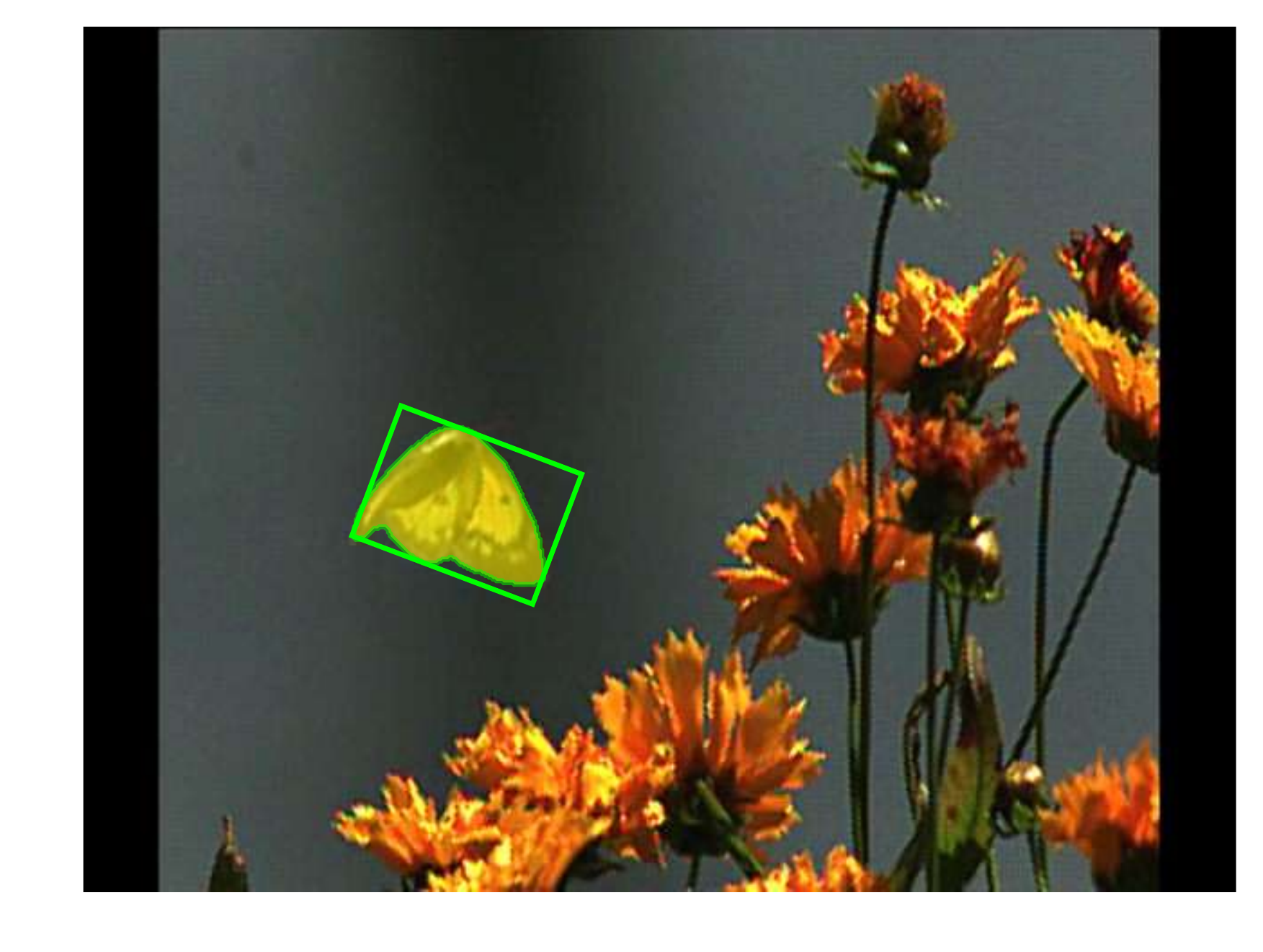}
&\includegraphics[trim={2.5cm 2cm 2.5cm 0.5cm},clip,width = 1.1in]{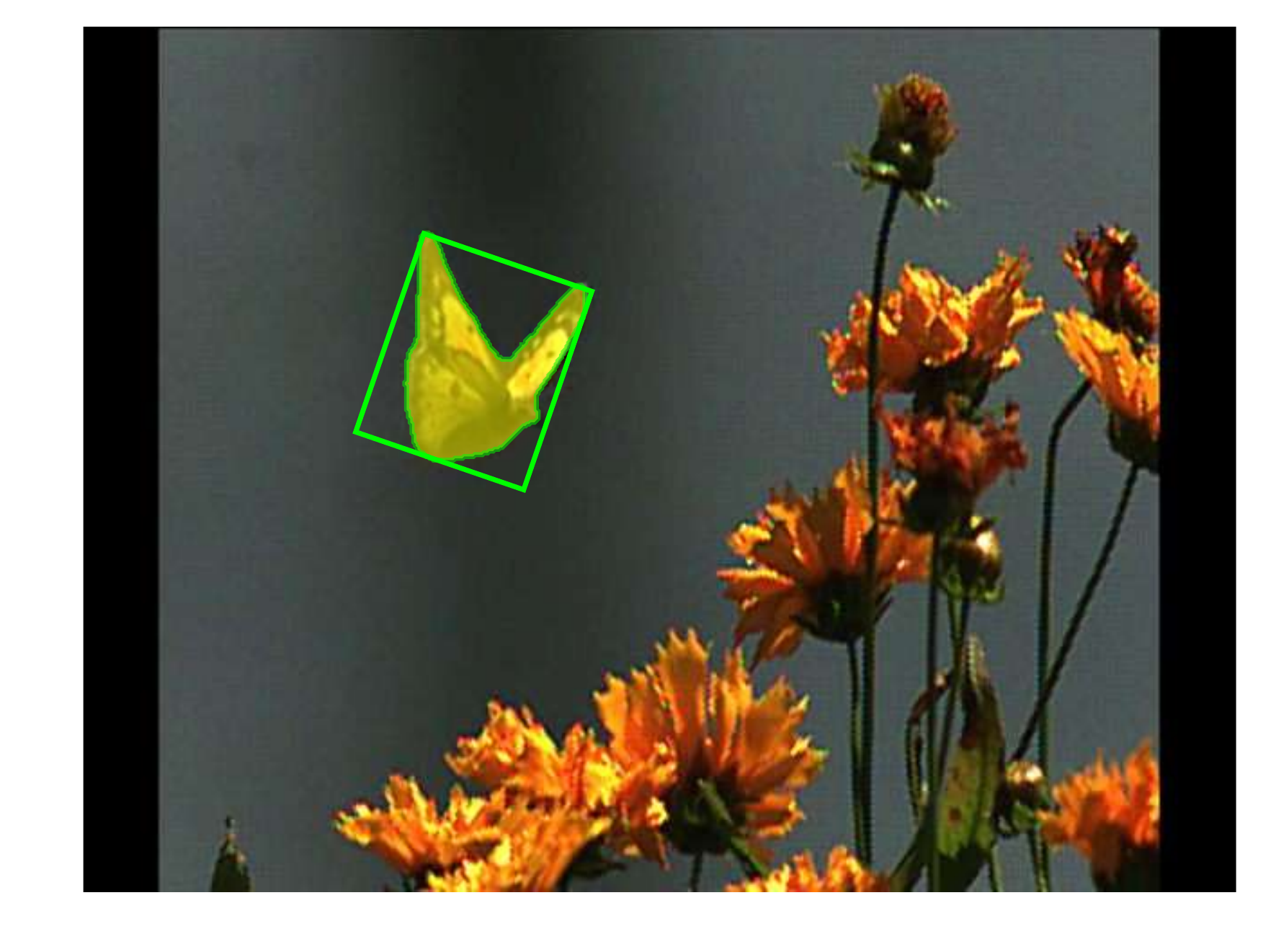}
& \includegraphics[trim={2.5cm 2cm 2.5cm 0.5cm},clip,width = 1.1in]{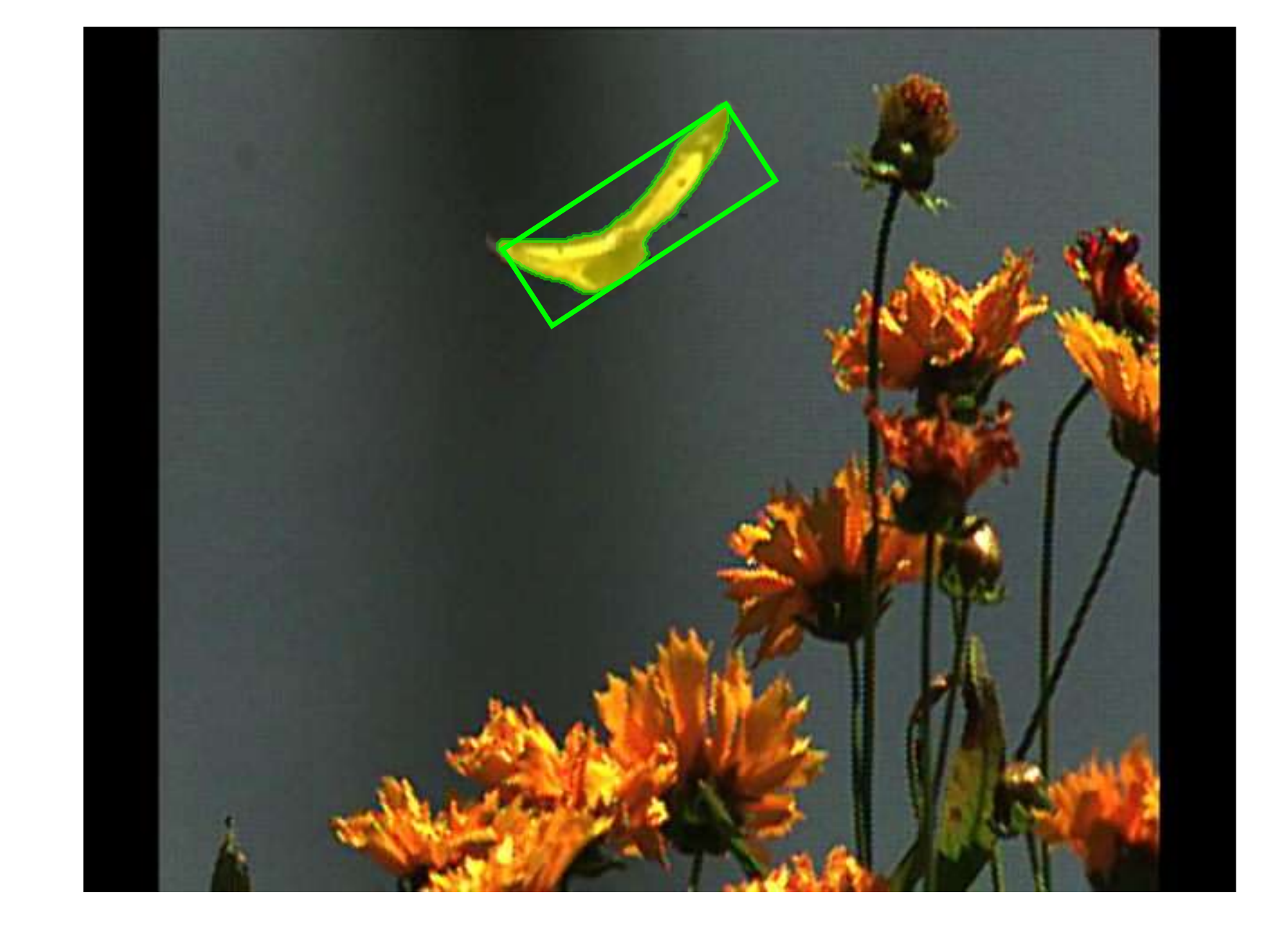}
& \includegraphics[trim={2.5cm 2cm 2.5cm 0.5cm},clip,width = 1.1in]{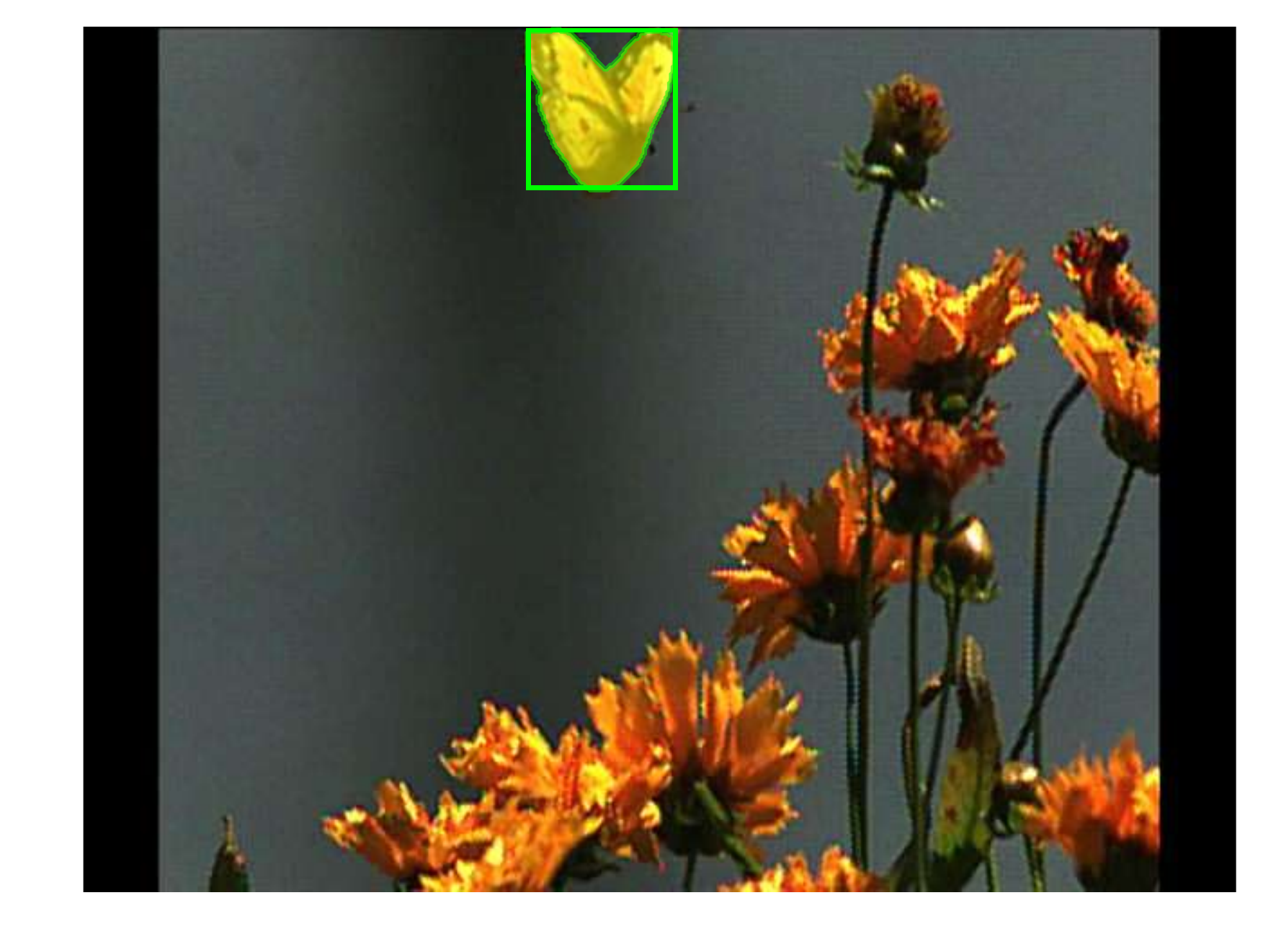}
& \includegraphics[trim={2.5cm 2cm 2.5cm 0.5cm},clip,width = 1.1in]{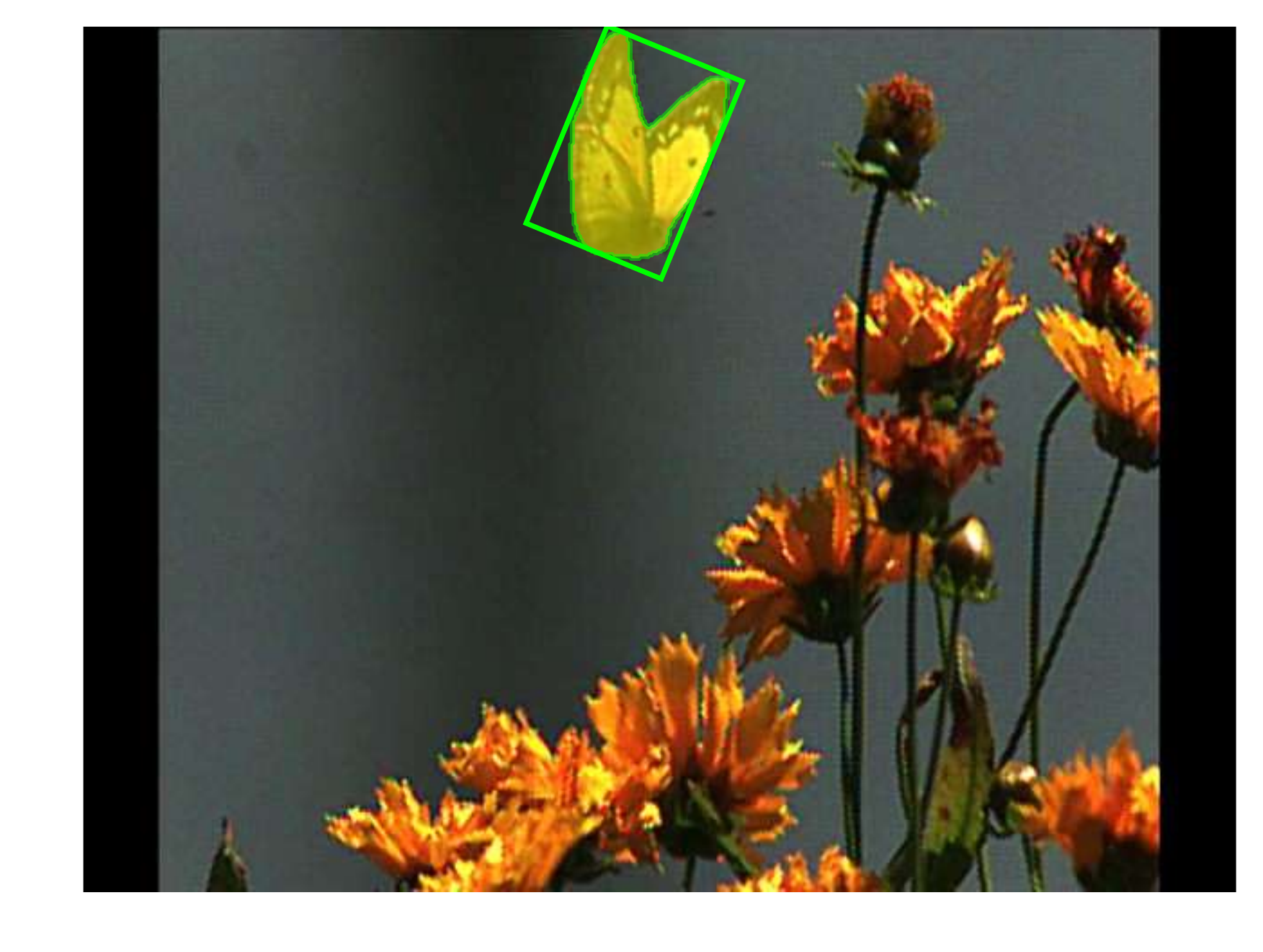}
& \includegraphics[trim={2.5cm 2cm 2.5cm 0.5cm},clip,width = 1.1in]{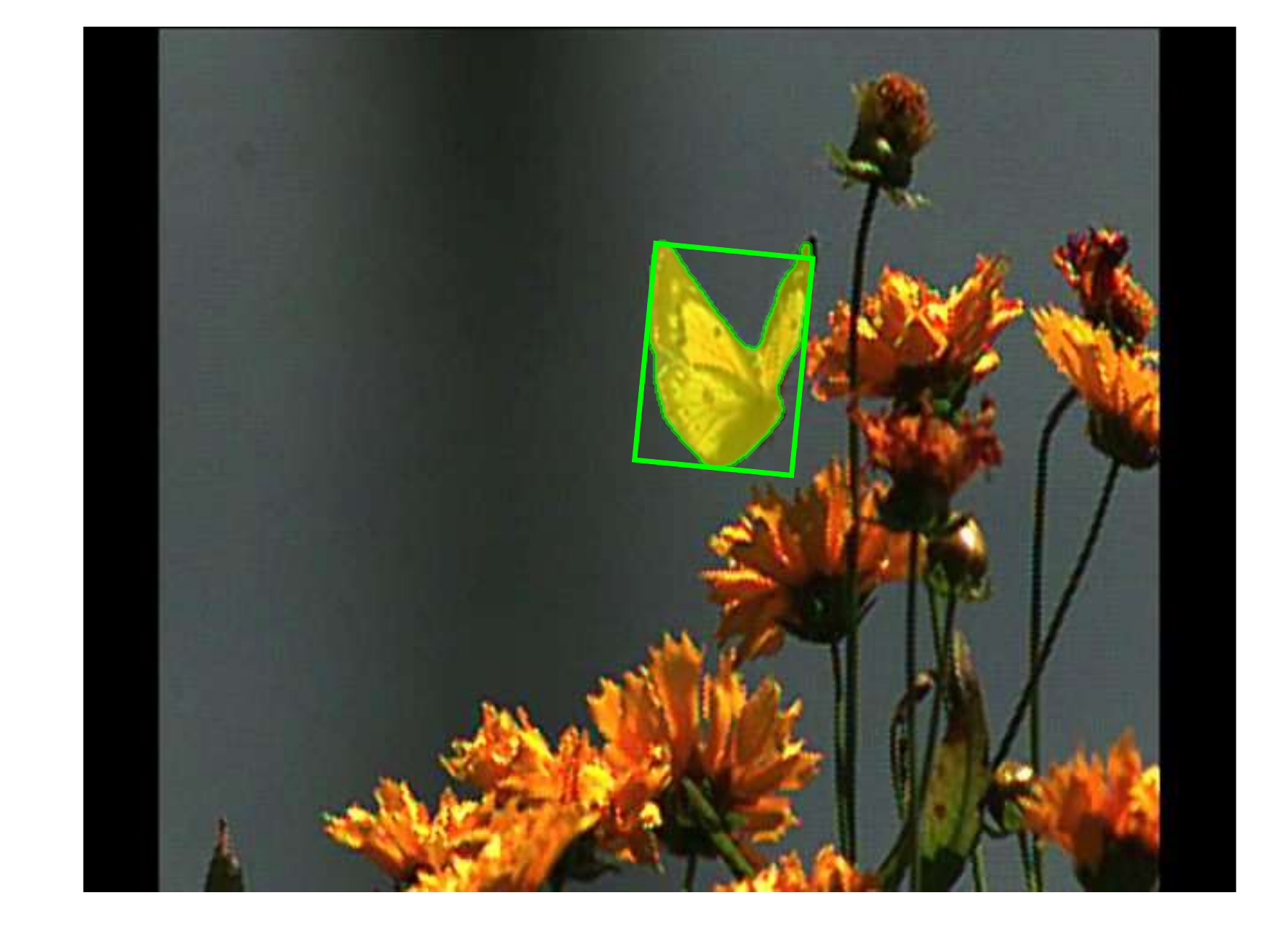}
\\
\mbox{\rotatebox[x=-0.55cm]{90}{\small{crabs1}}}
\includegraphics[trim={2.5cm 1cm 2.5cm 1cm},clip,width = 1.1in]{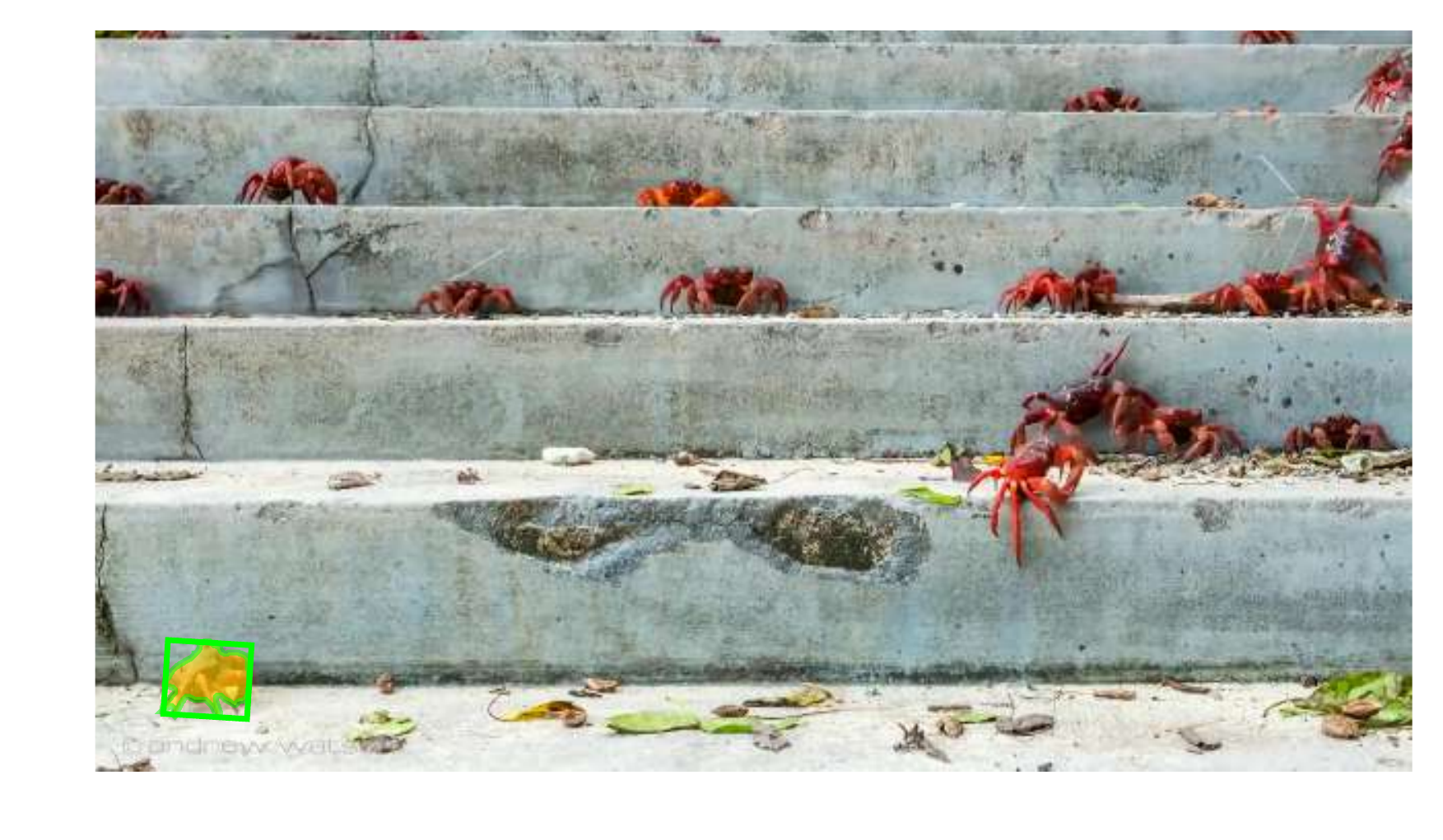}
&\includegraphics[trim={2.5cm 1cm 2.5cm 1cm},clip,width = 1.1in]{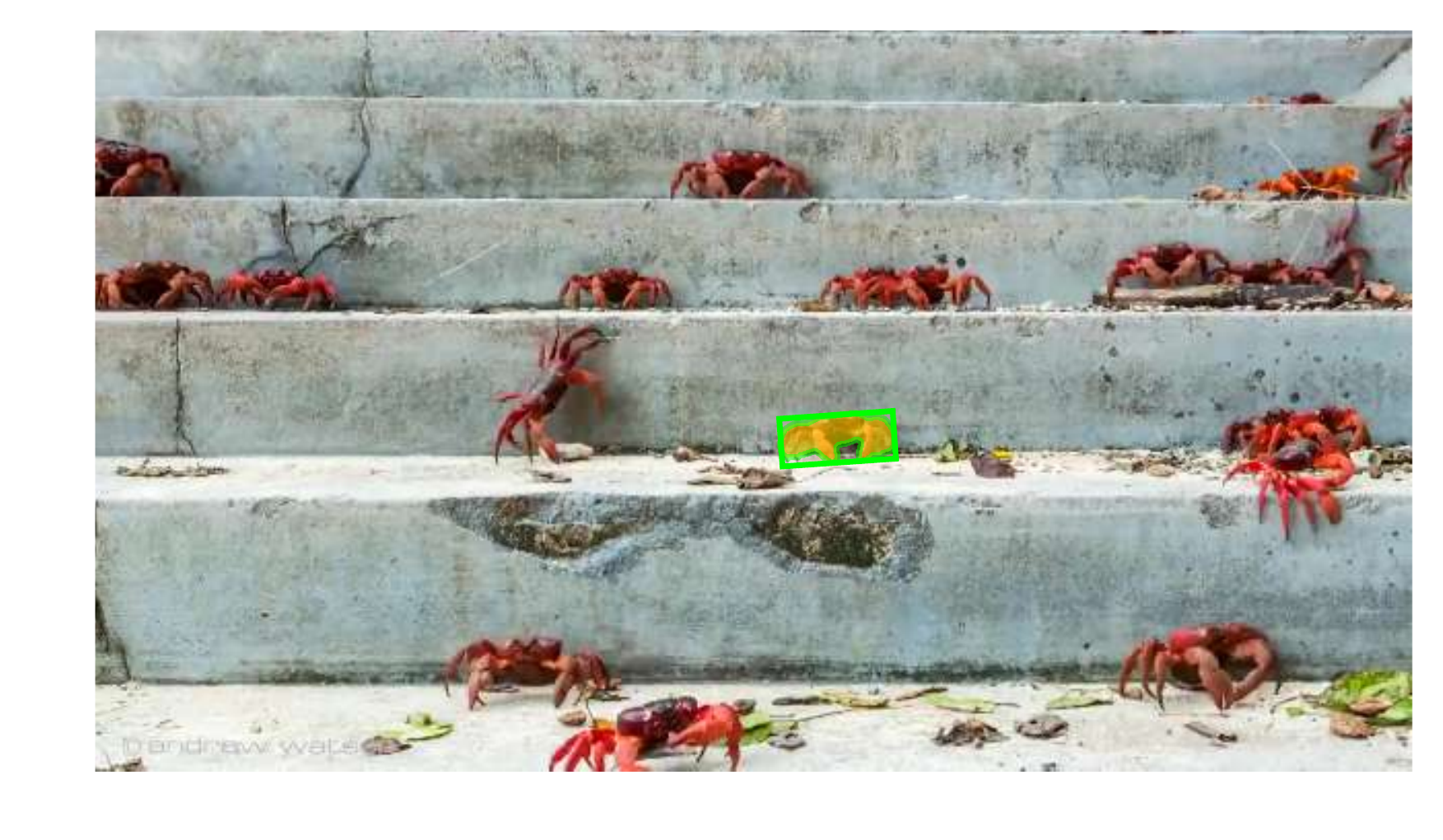}
& \includegraphics[trim={2.5cm 1cm 2.5cm 1cm},clip,width = 1.1in]{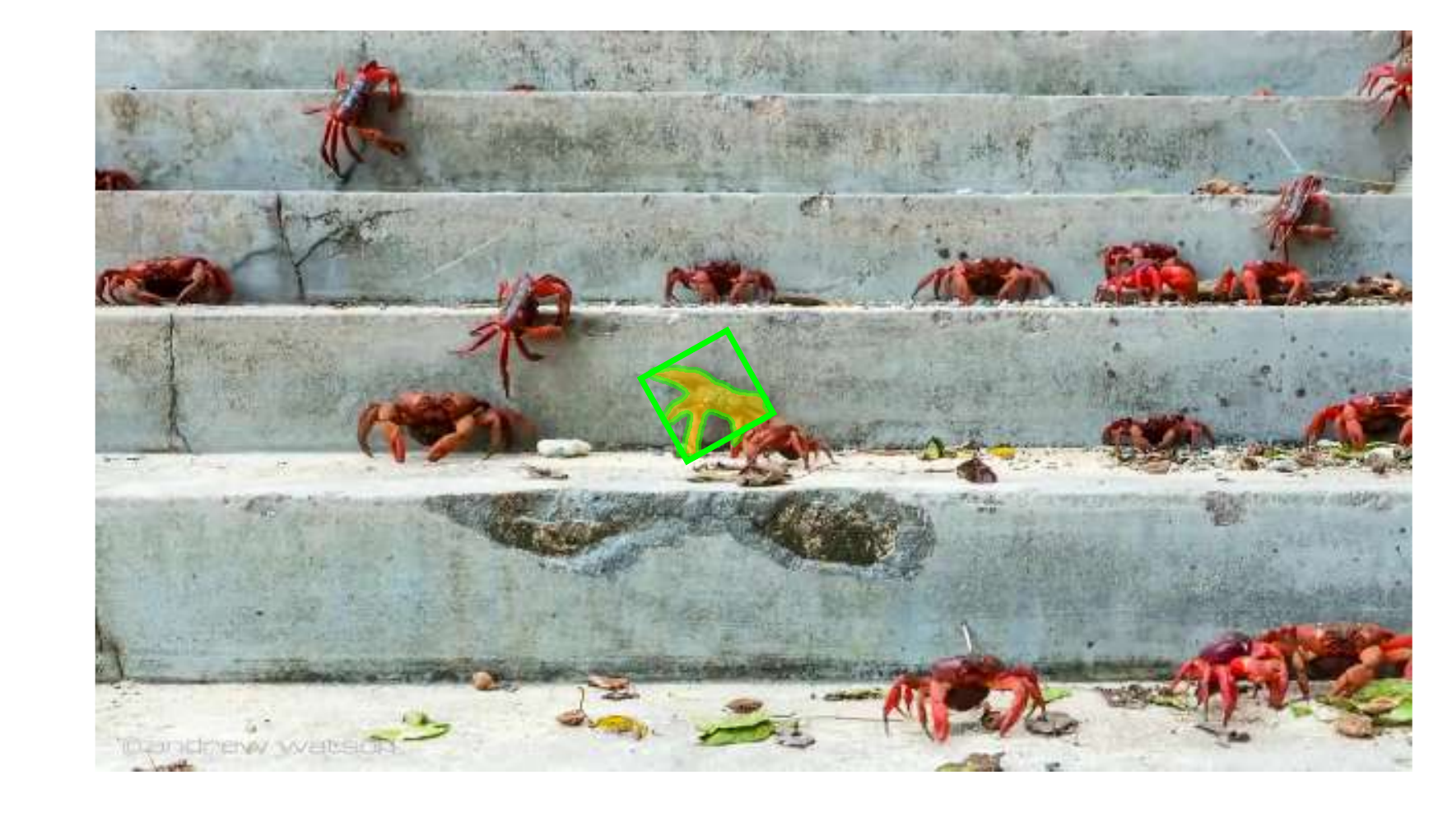}
& \includegraphics[trim={2.5cm 1cm 2.5cm 1cm},clip,width = 1.1in]{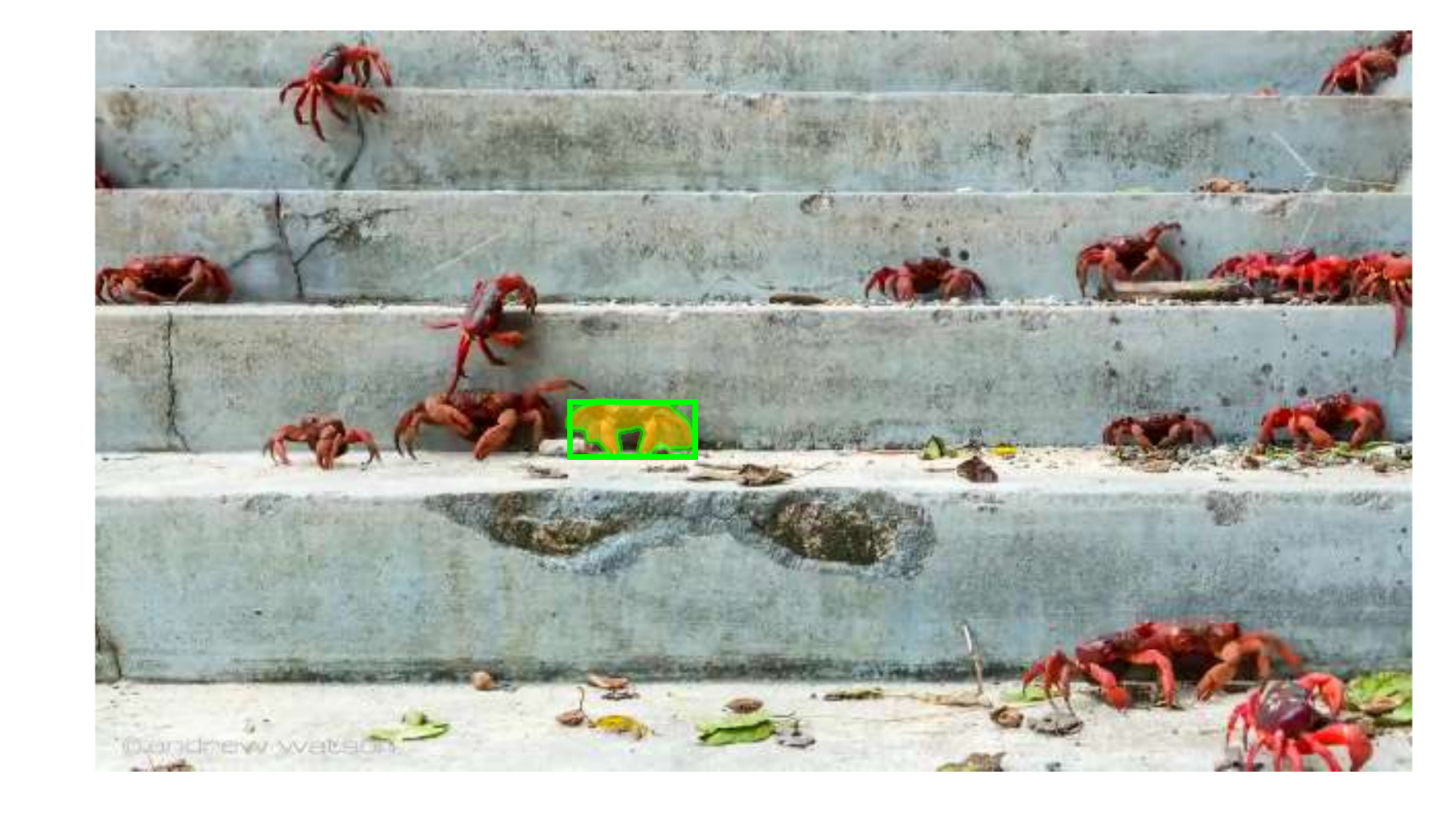}
& \includegraphics[trim={2.5cm 1cm 2.5cm 1cm},clip,width = 1.1in]{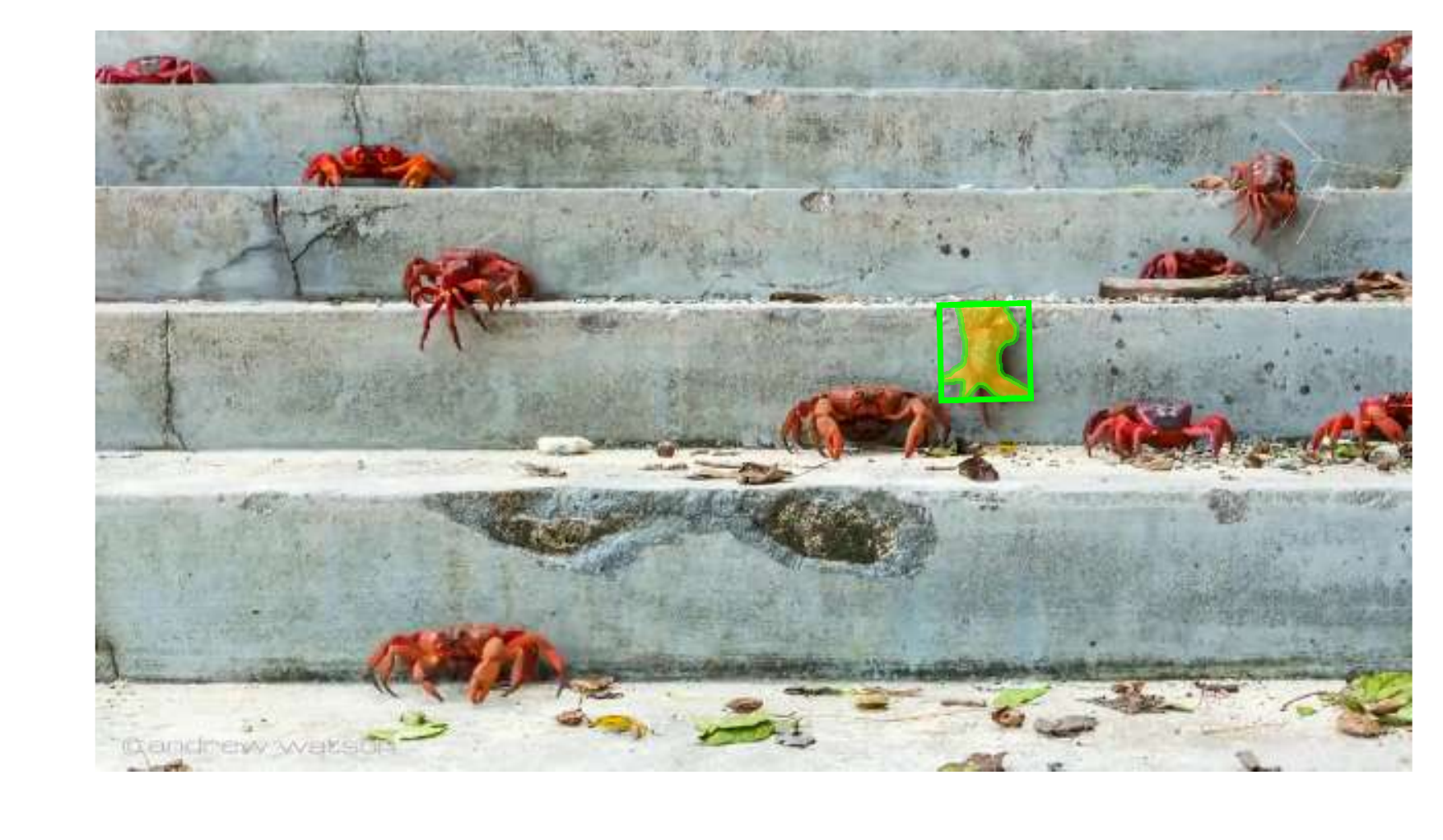}
& \includegraphics[trim={2.5cm 1cm 2.5cm 1cm},clip,width = 1.1in]{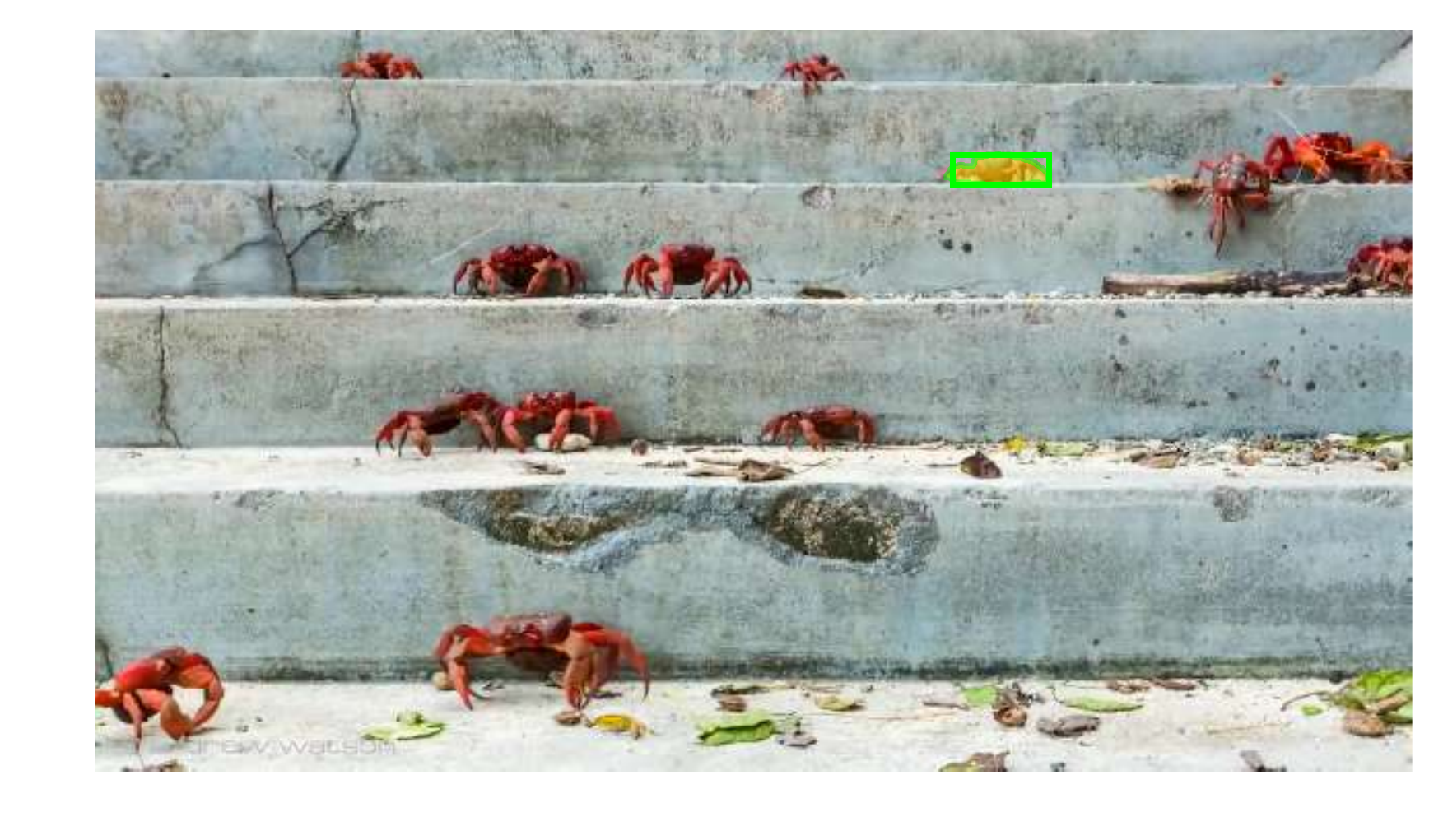}
\\
\mbox{\rotatebox[x=-0.0cm]{90}{\small{iceskater1}}}
\includegraphics[trim={2.5cm 1cm 2.5cm 1cm},clip,width = 1.1in]{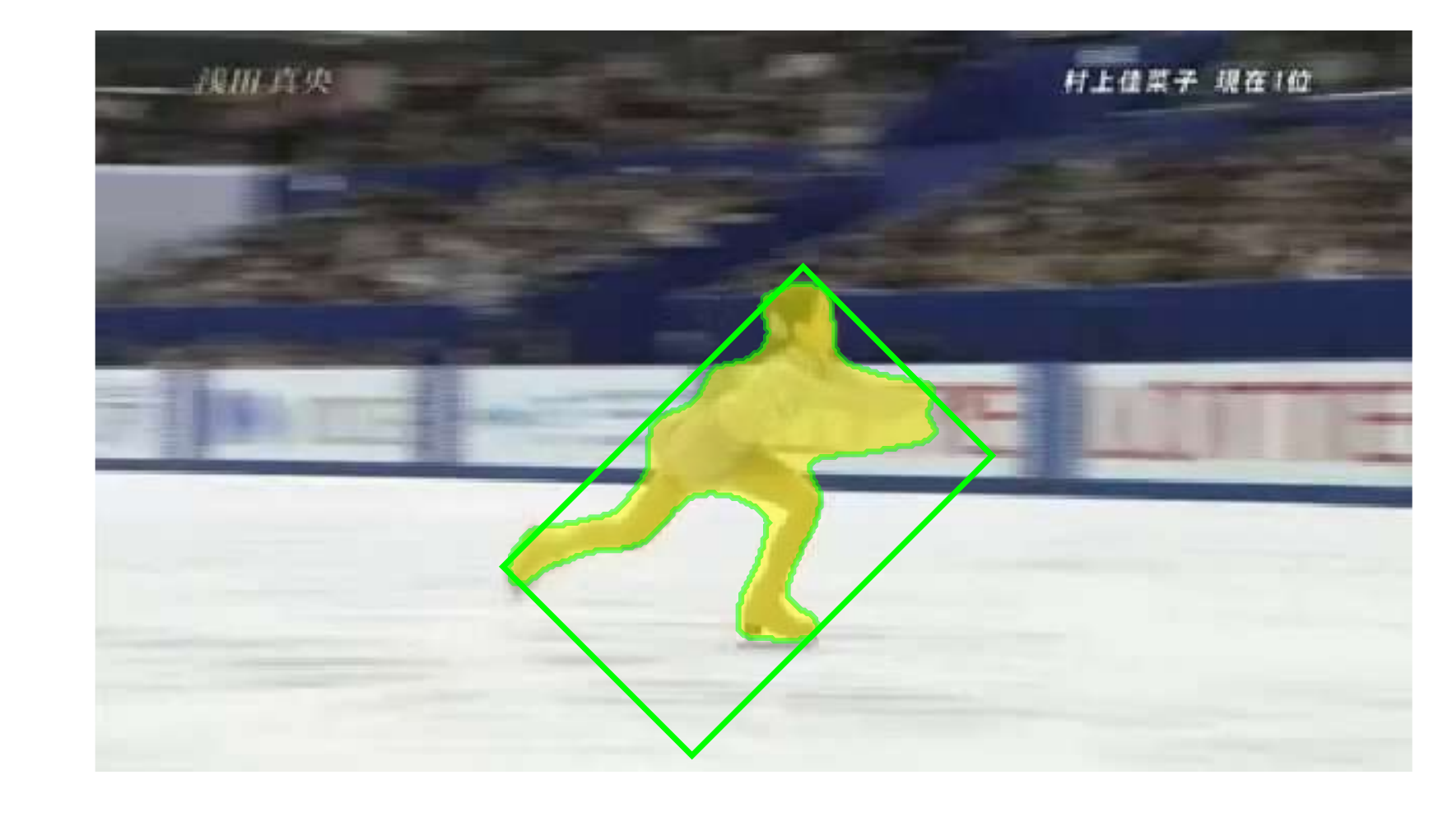}
&\includegraphics[trim={2.5cm 1cm 2.5cm 1cm},clip,width = 1.1in]{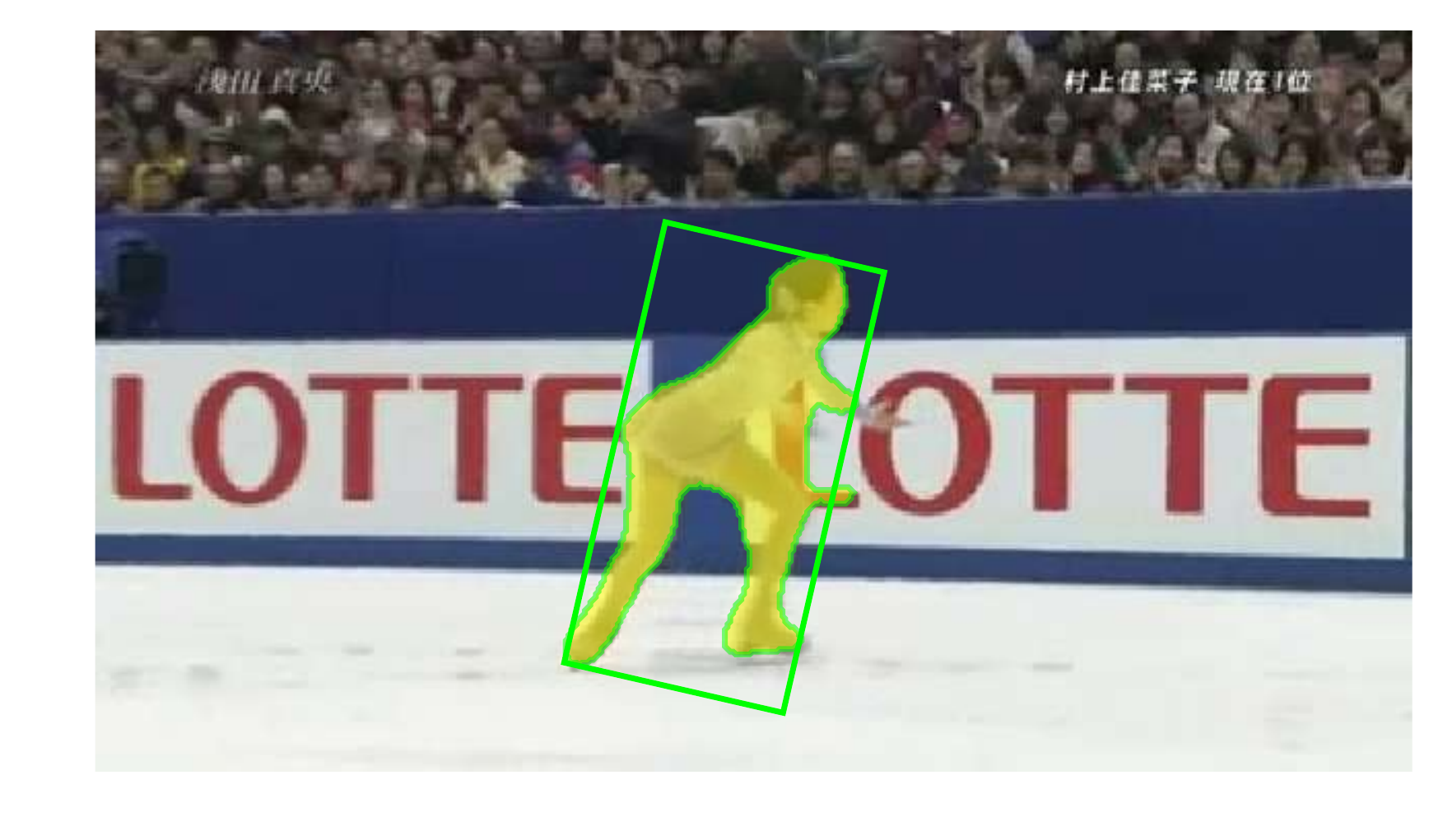}
& \includegraphics[trim={2.5cm 1cm 2.5cm 1cm},clip,width = 1.1in]{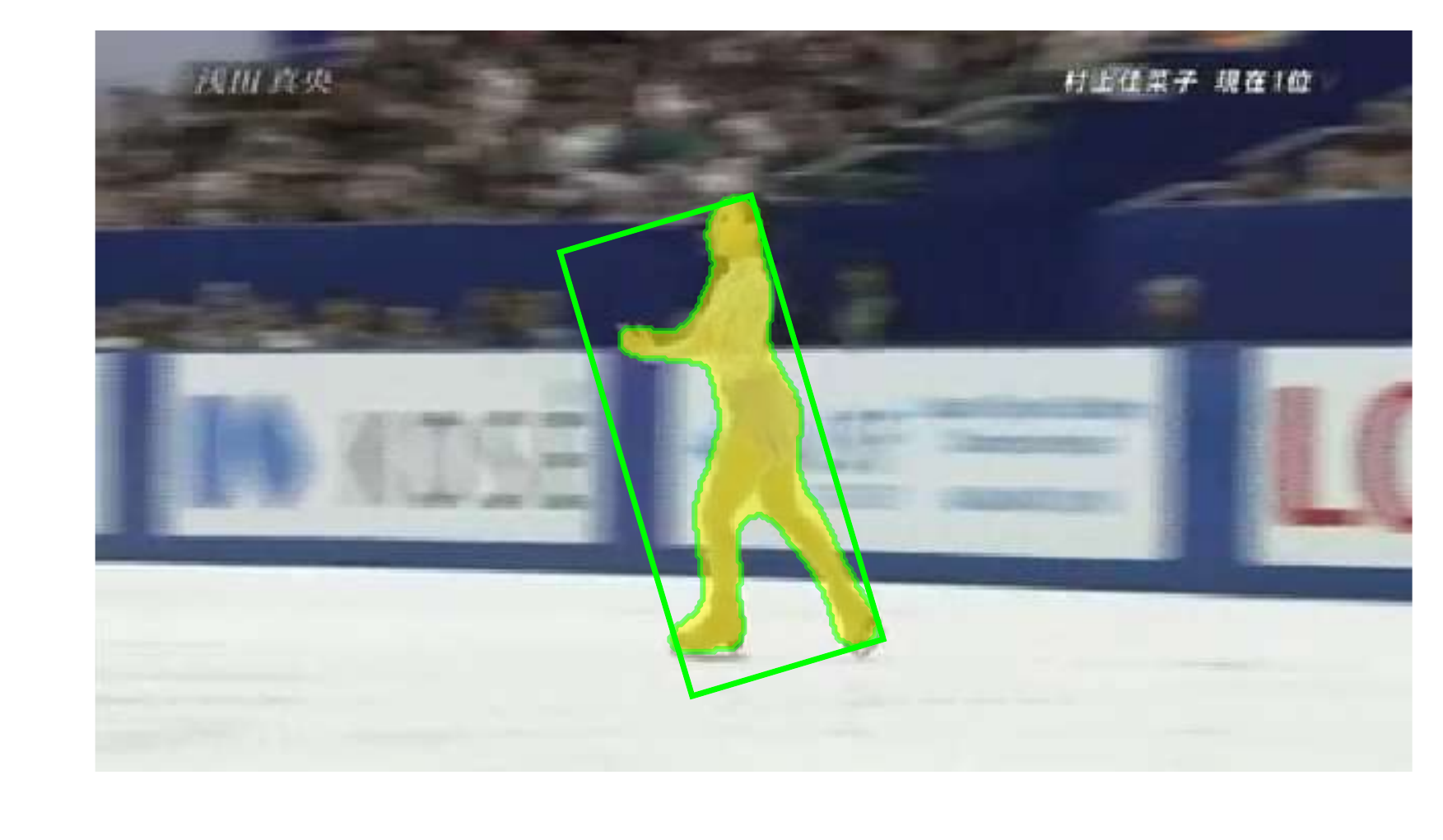}
& \includegraphics[trim={2.5cm 1cm 2.5cm 1cm},clip,width = 1.1in]{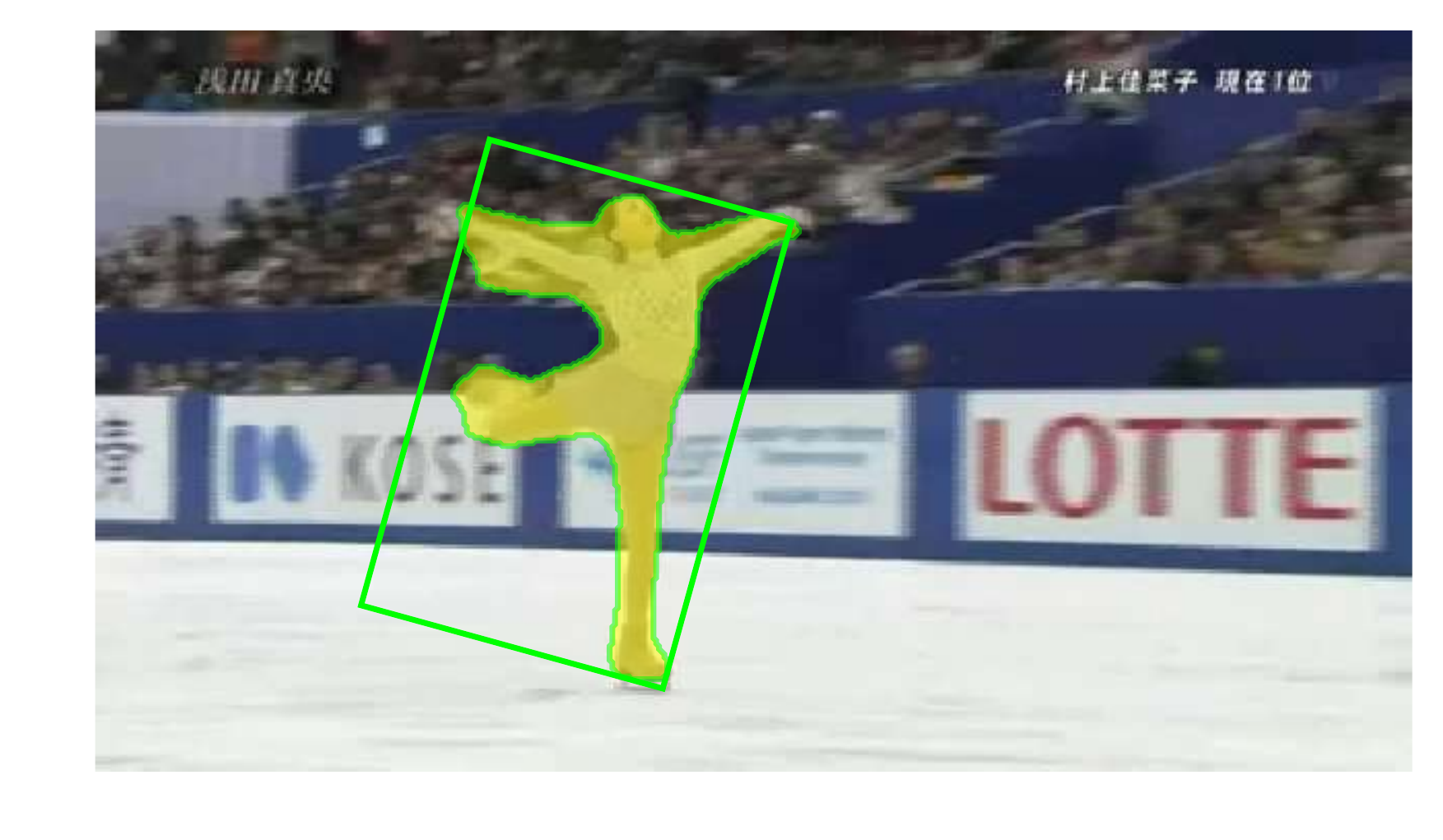}
& \includegraphics[trim={2.5cm 1cm 2.5cm 1cm},clip,width = 1.1in]{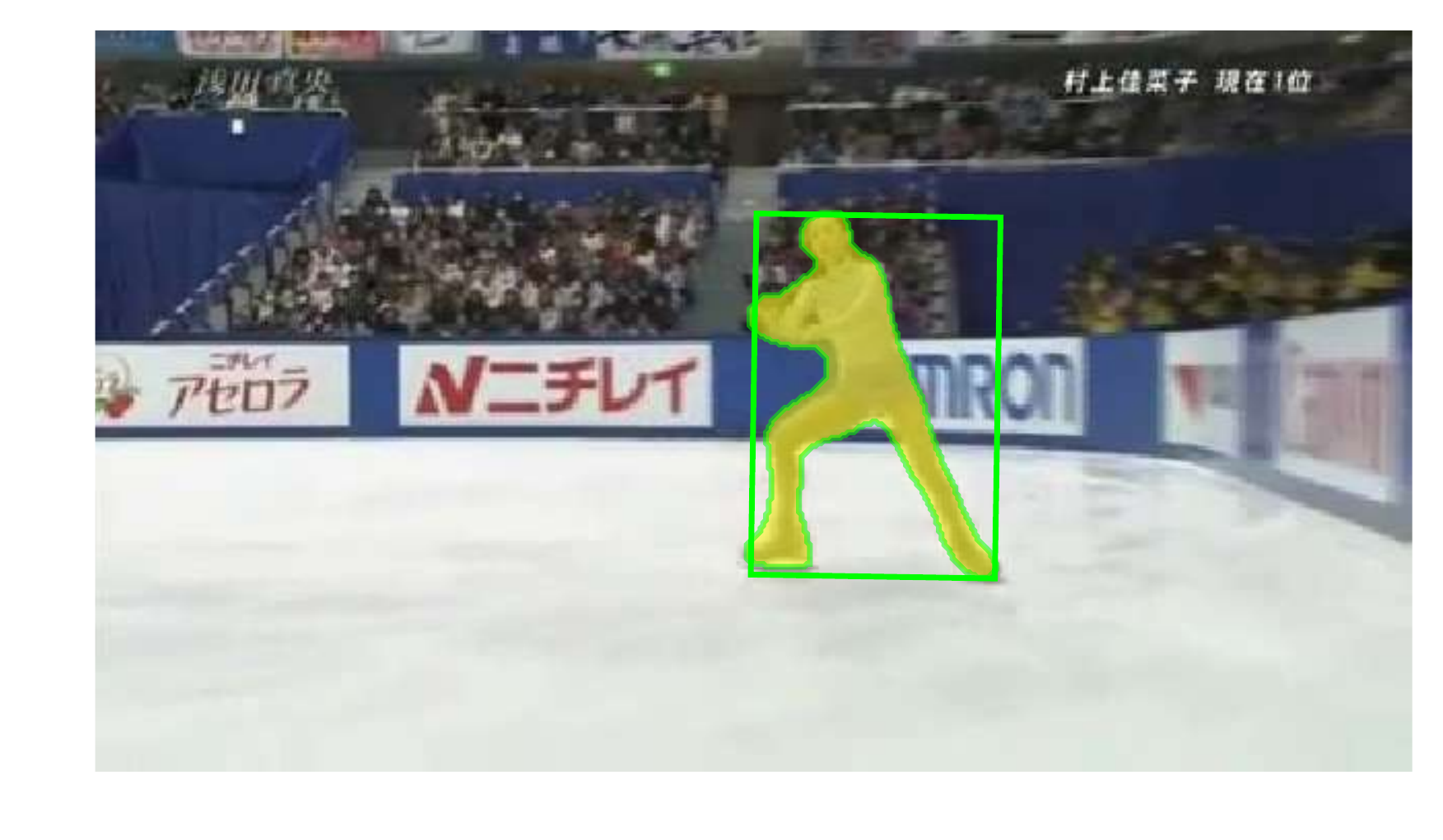}
& \includegraphics[trim={2.5cm 1cm 2.5cm 1cm},clip,width = 1.1in]{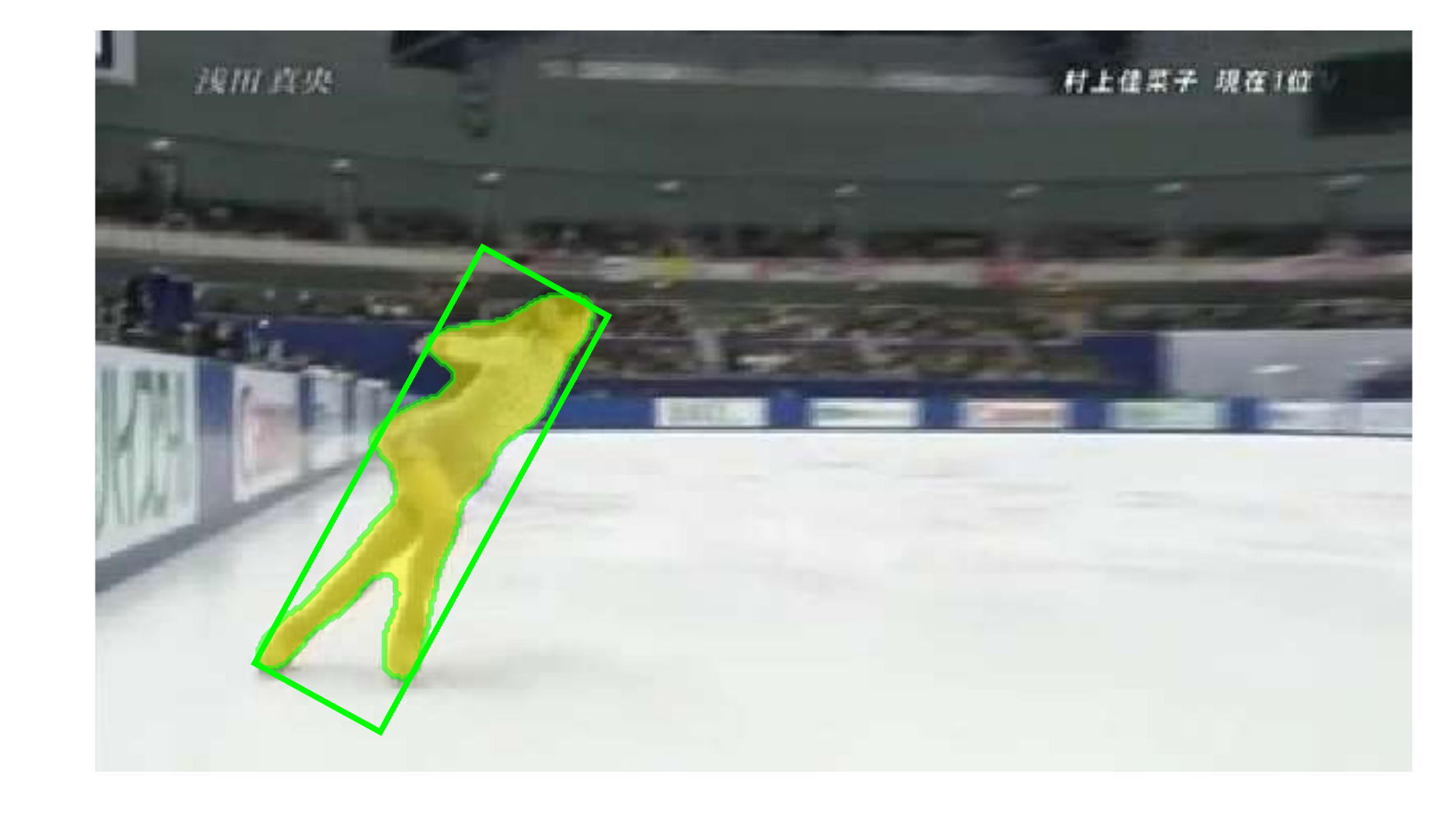}
\\
\mbox{\rotatebox[x=-0.15cm]{90}{\small{iceskater2}}}
\includegraphics[trim={2.5cm 1cm 2.5cm 1cm},clip,width = 1.1in]{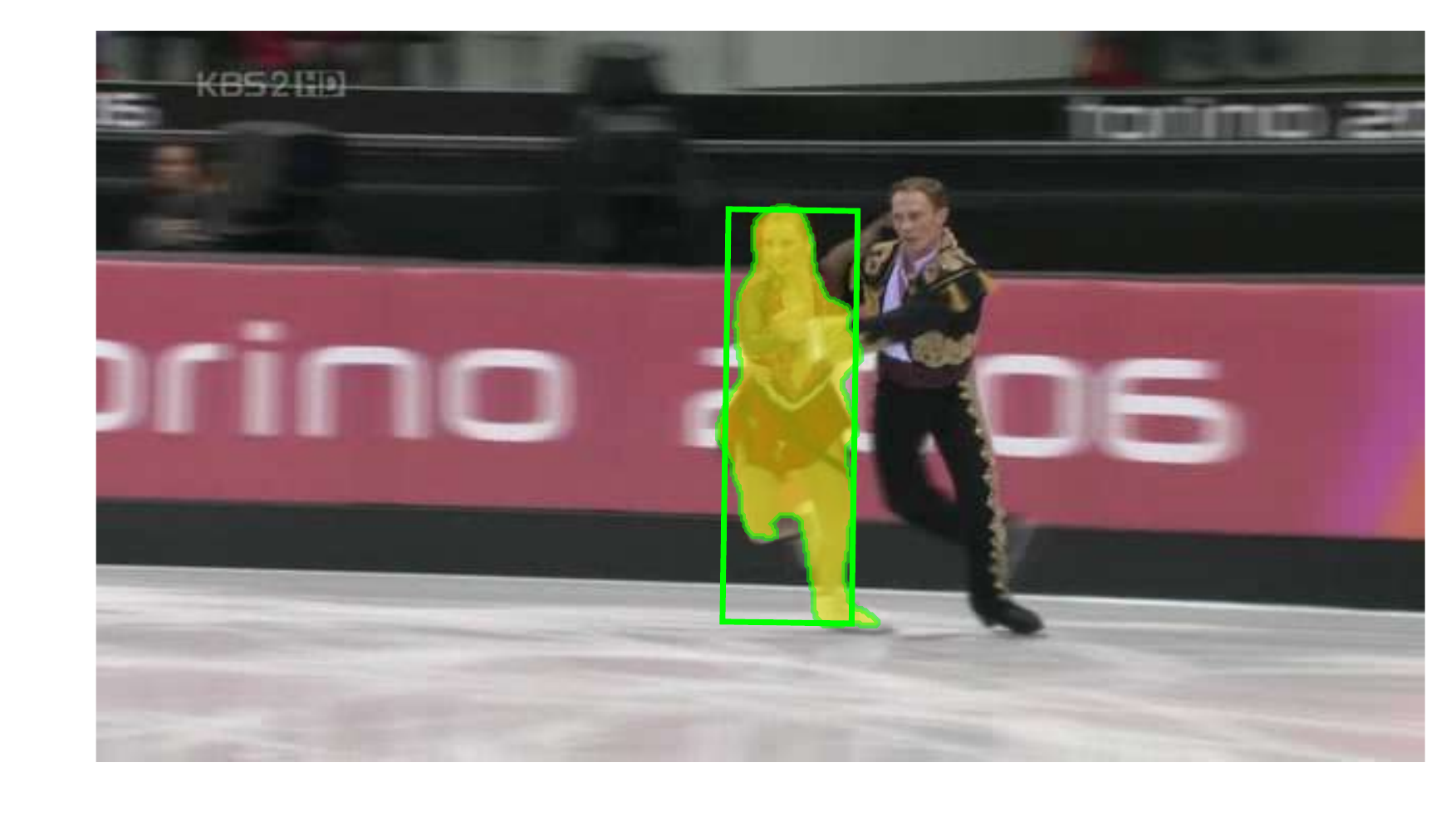}
&\includegraphics[trim={2.5cm 1cm 2.5cm 1cm},clip,width = 1.1in]{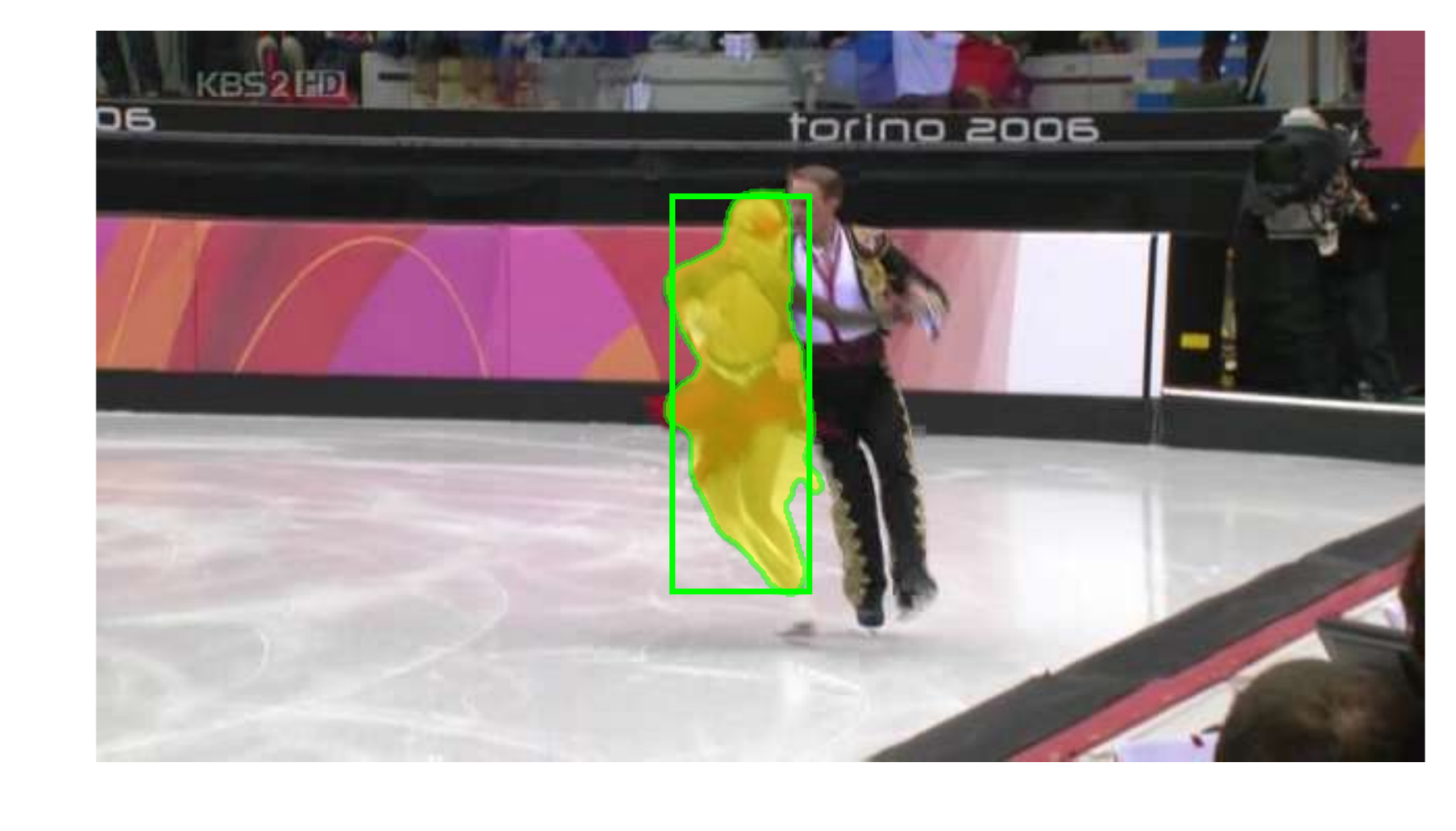}
& \includegraphics[trim={2.5cm 1cm 2.5cm 1cm},clip,width = 1.1in]{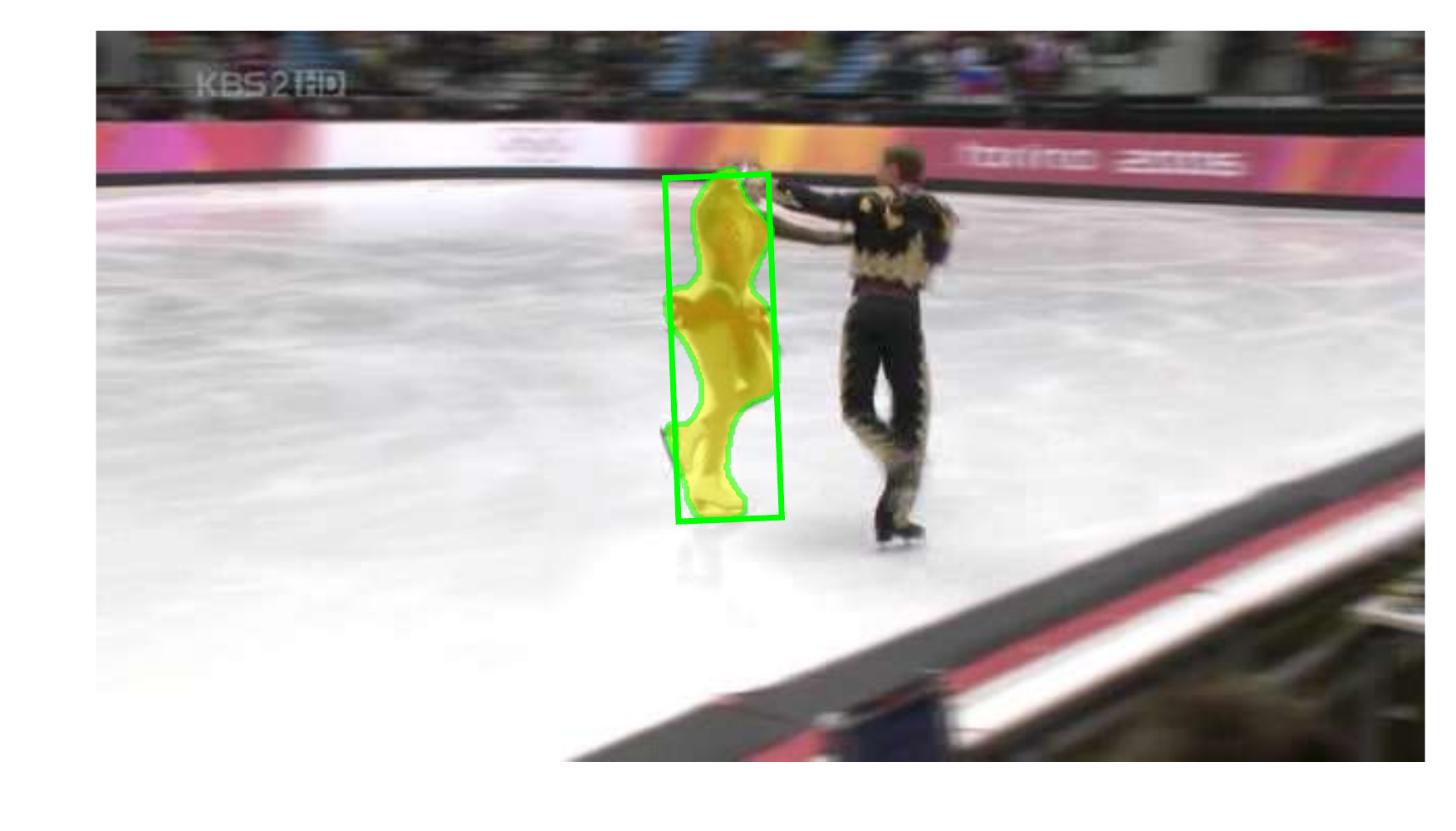}
& \includegraphics[trim={2.5cm 1cm 2.5cm 1cm},clip,width = 1.1in]{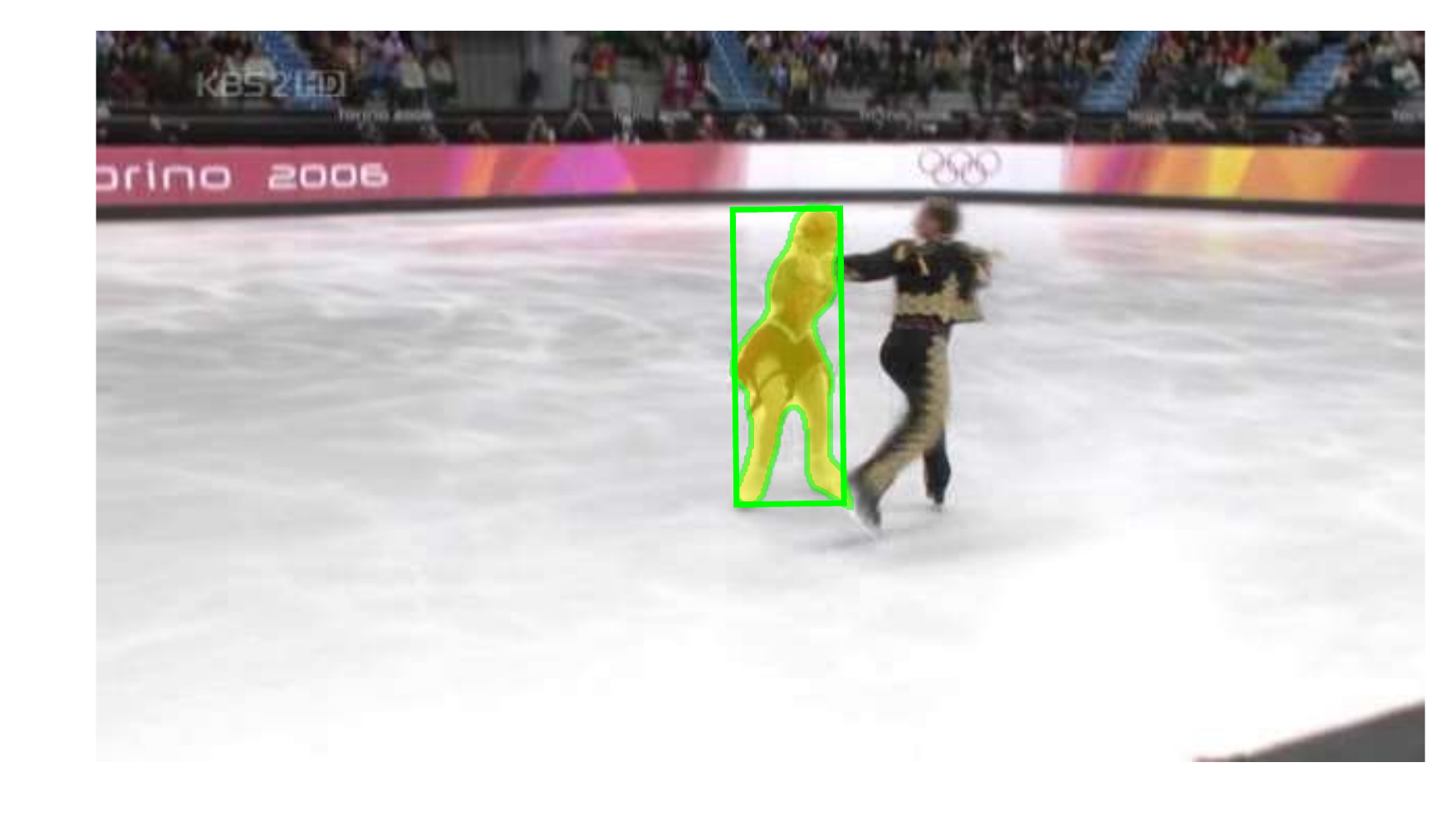}
& \includegraphics[trim={2.5cm 1cm 2.5cm 1cm},clip,width = 1.1in]{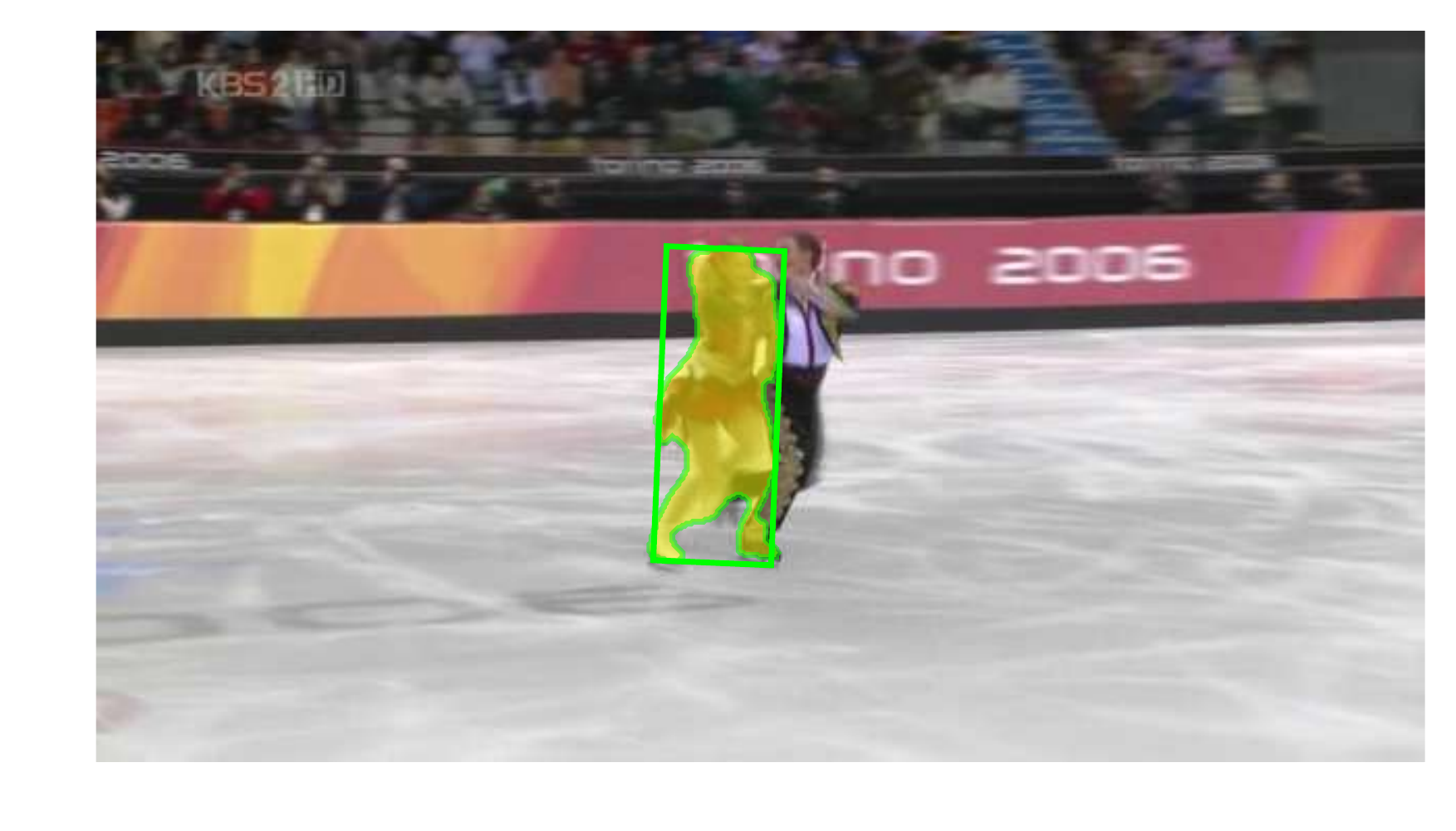}
& \includegraphics[trim={2.5cm 1cm 2.5cm 1cm},clip,width = 1.1in]{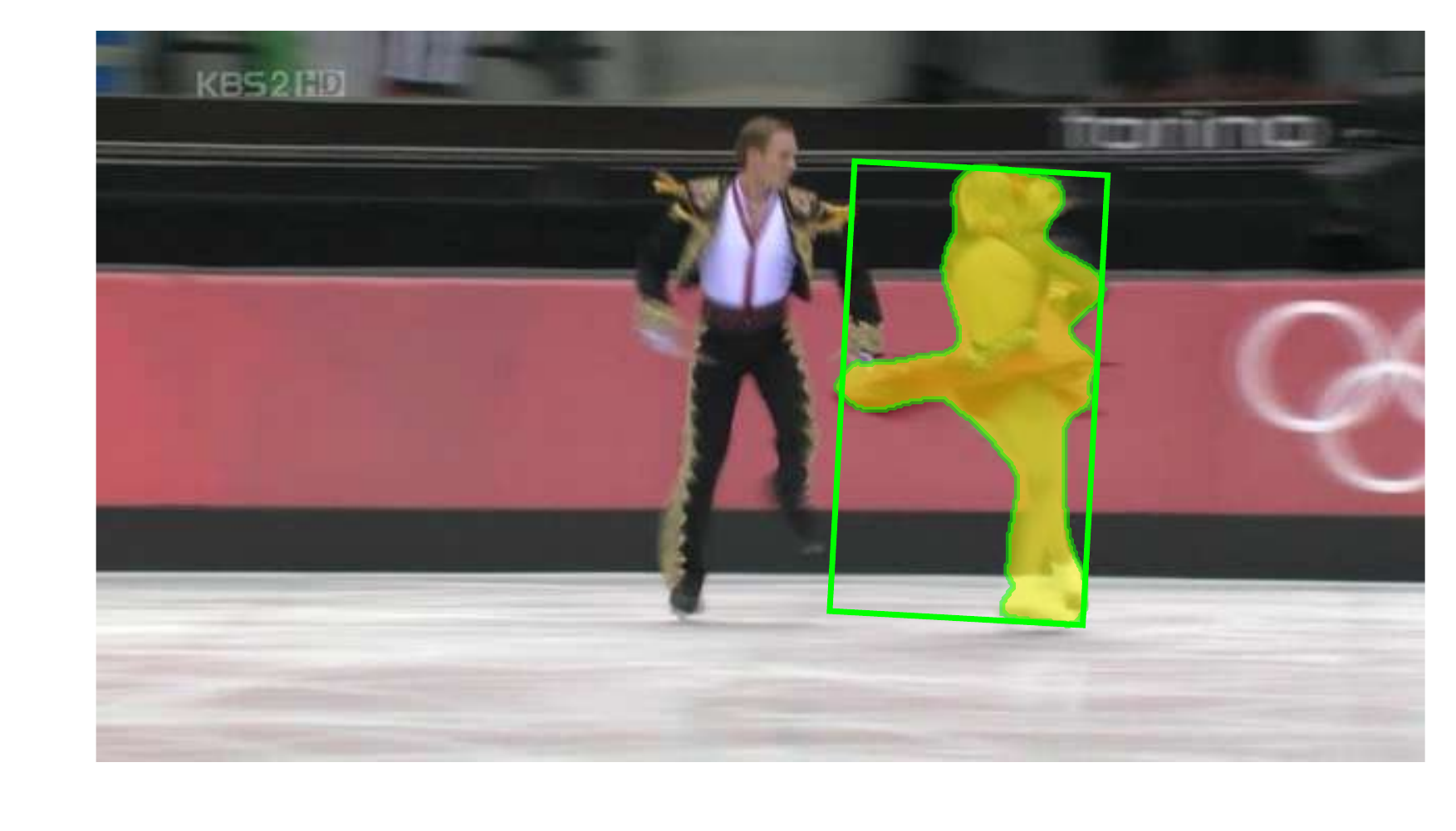}
\\
\mbox{\rotatebox[x=-0.1cm]{90}{\small{motocross1}}}
\includegraphics[trim={2.5cm 1cm 2.5cm 1cm},clip,width = 1.1in]{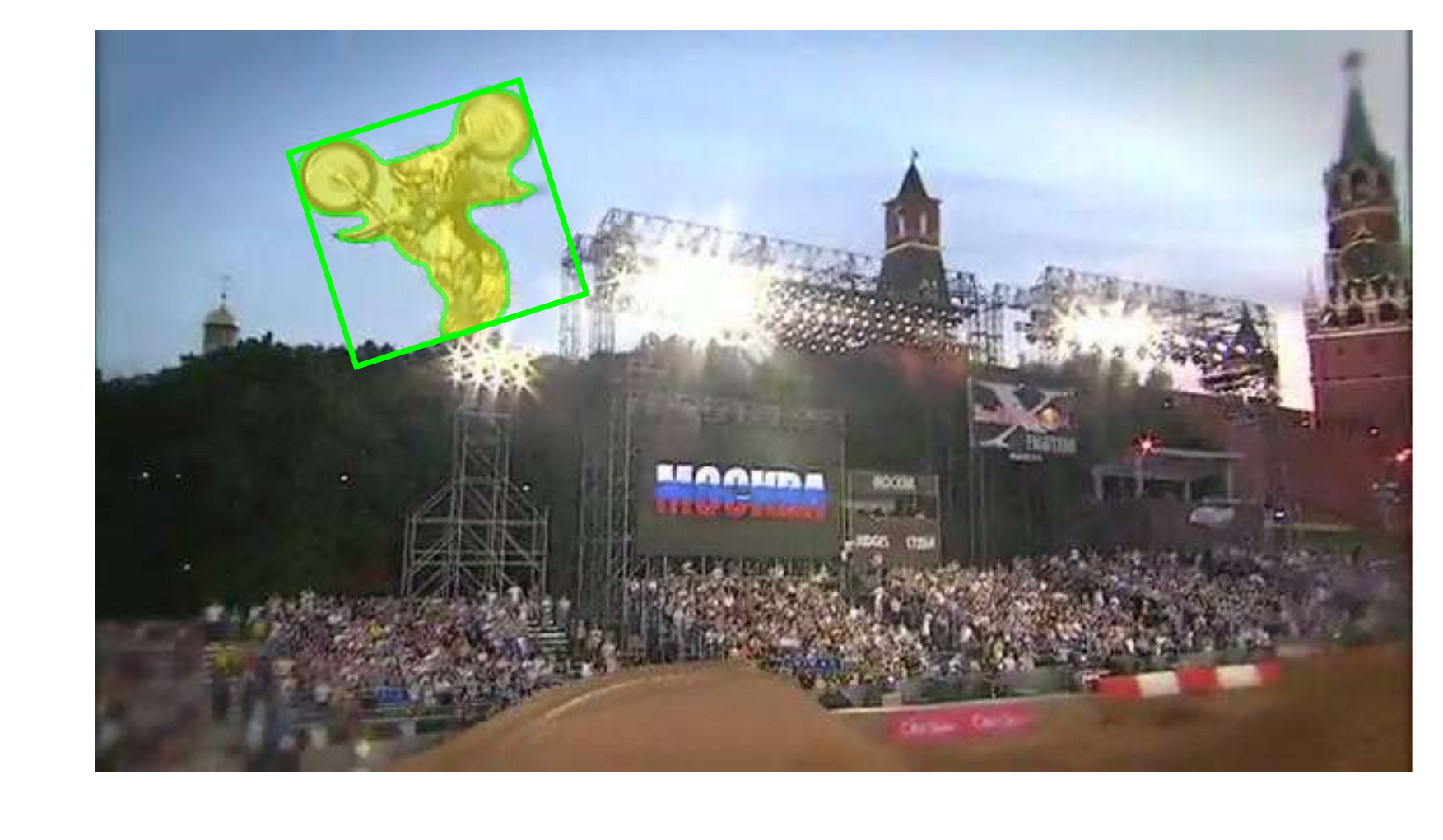}
&\includegraphics[trim={2.5cm 1cm 2.5cm 1cm},clip,width = 1.1in]{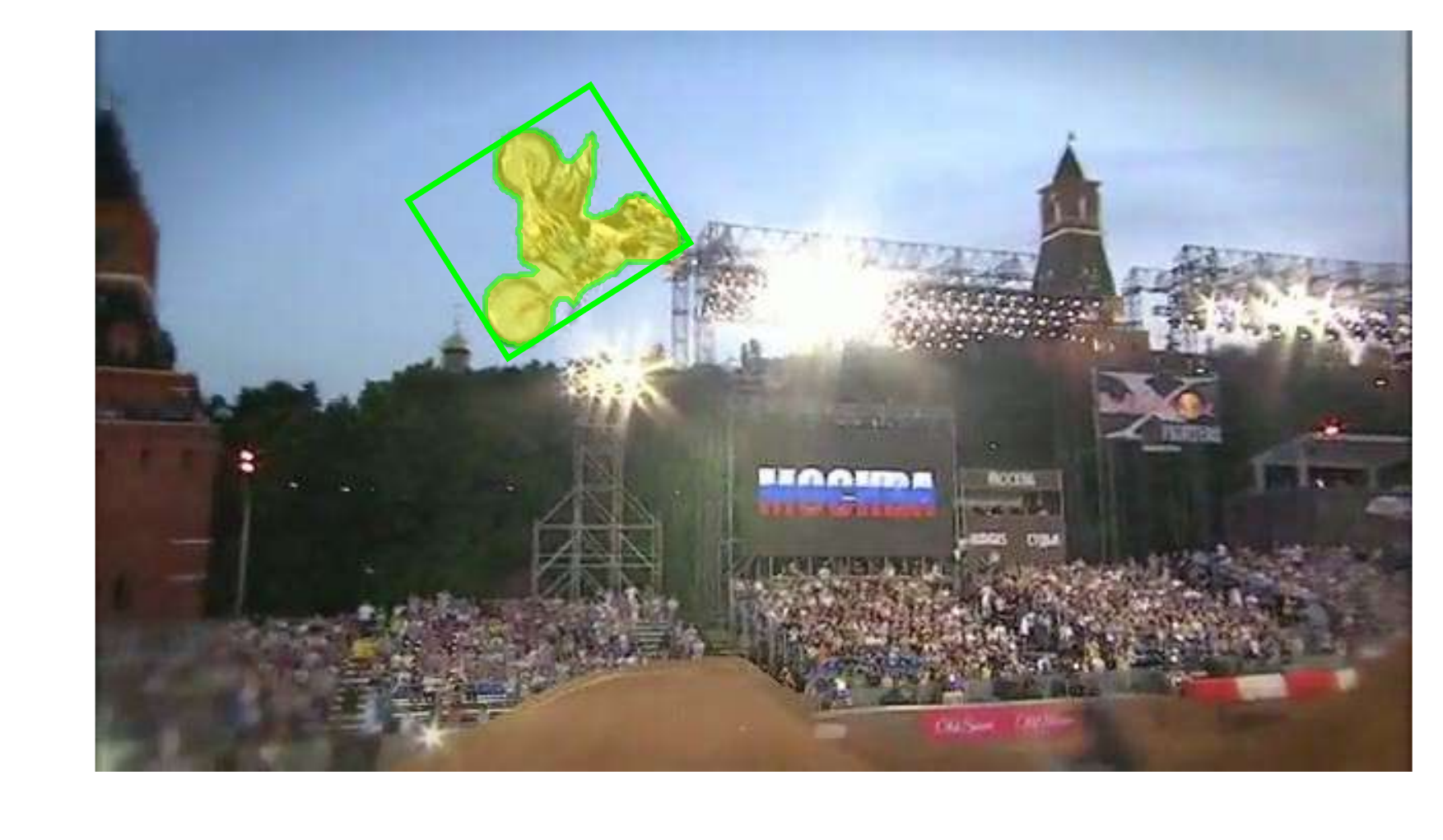}
& \includegraphics[trim={2.5cm 1cm 2.5cm 1cm},clip,width = 1.1in]{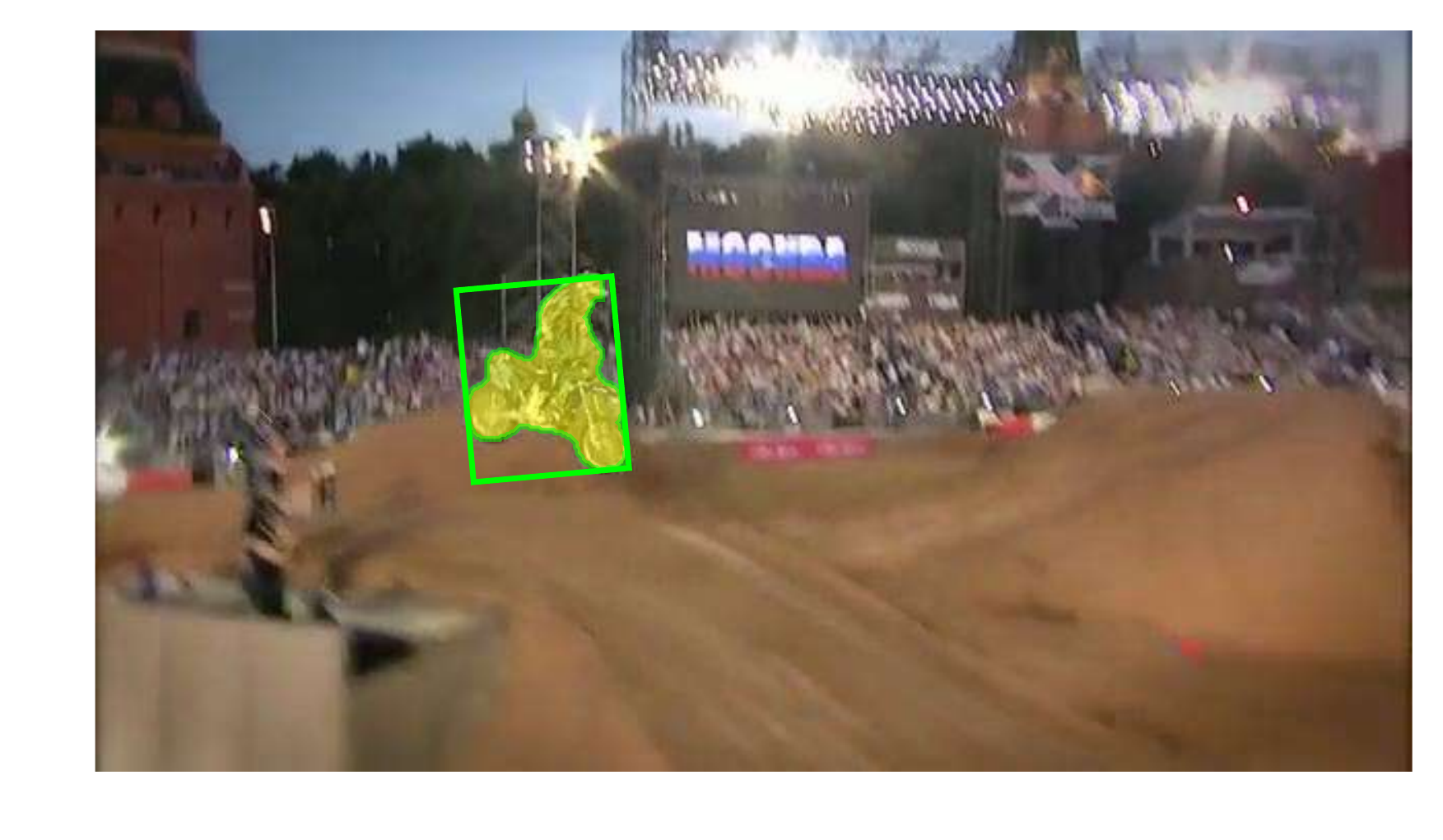}
& \includegraphics[trim={2.5cm 1cm 2.5cm 1cm},clip,width = 1.1in]{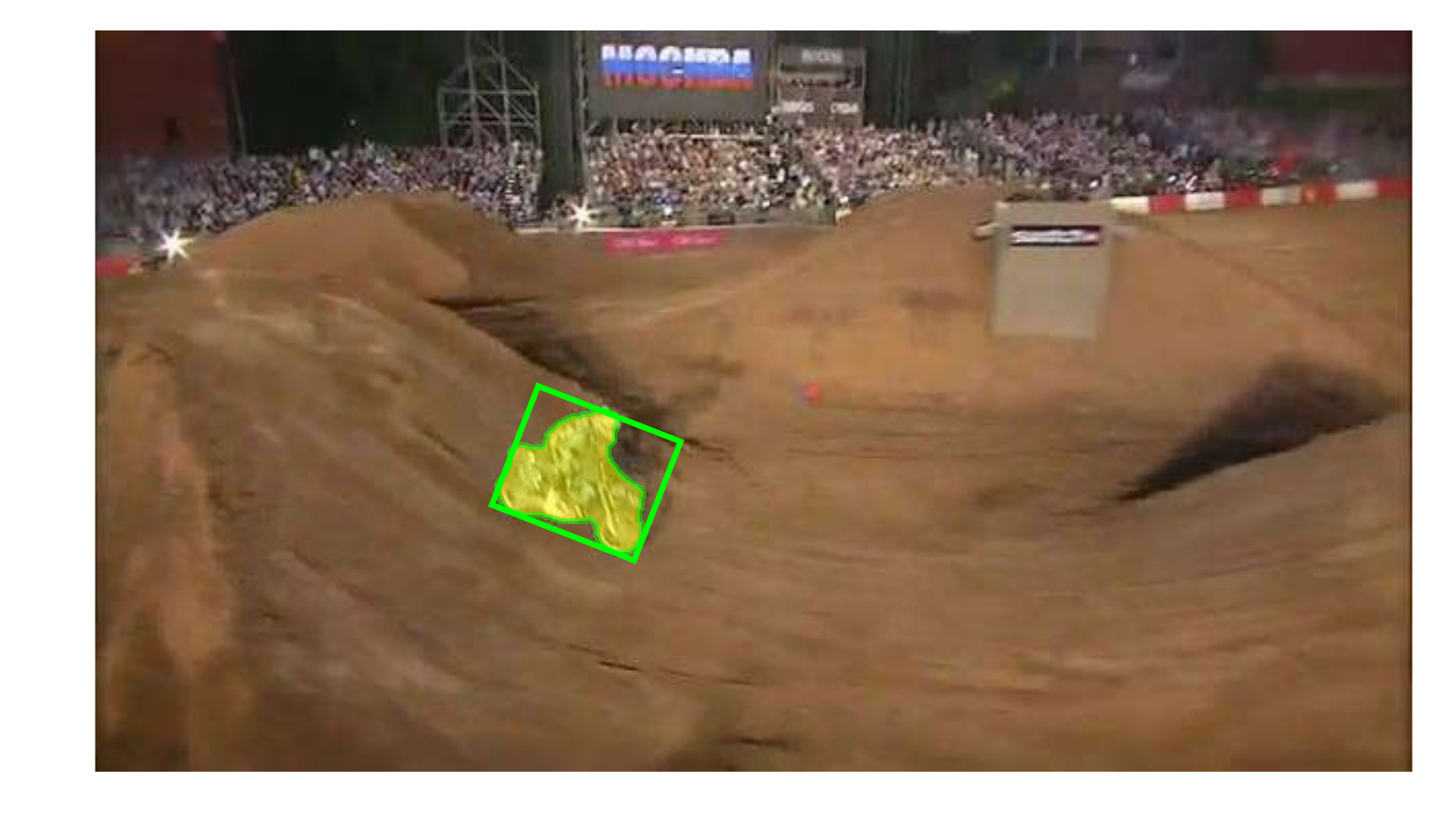}
& \includegraphics[trim={2.5cm 1cm 2.5cm 1cm},clip,width = 1.1in]{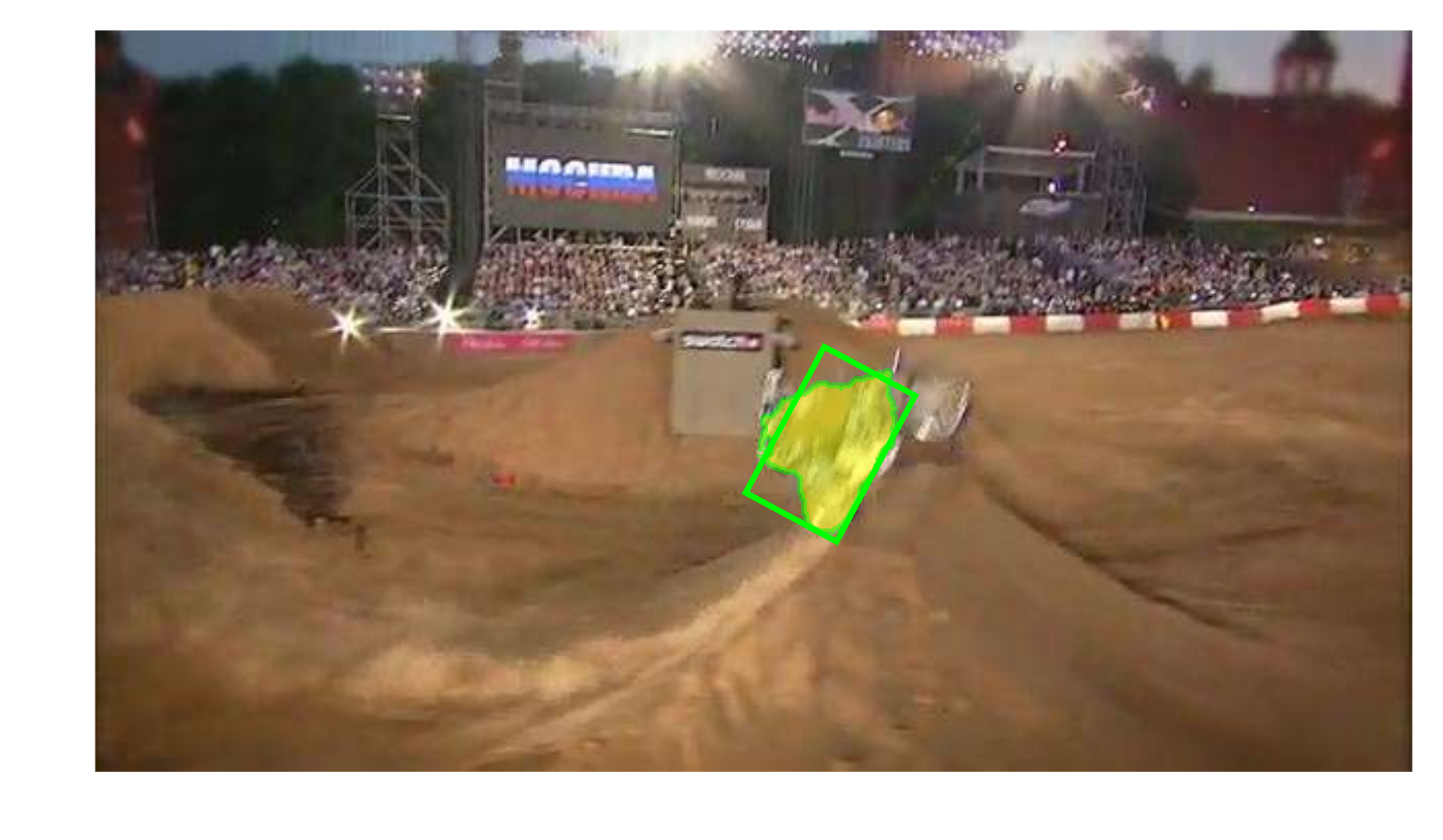}
& \includegraphics[trim={2.5cm 1cm 2.5cm 1cm},clip,width = 1.1in]{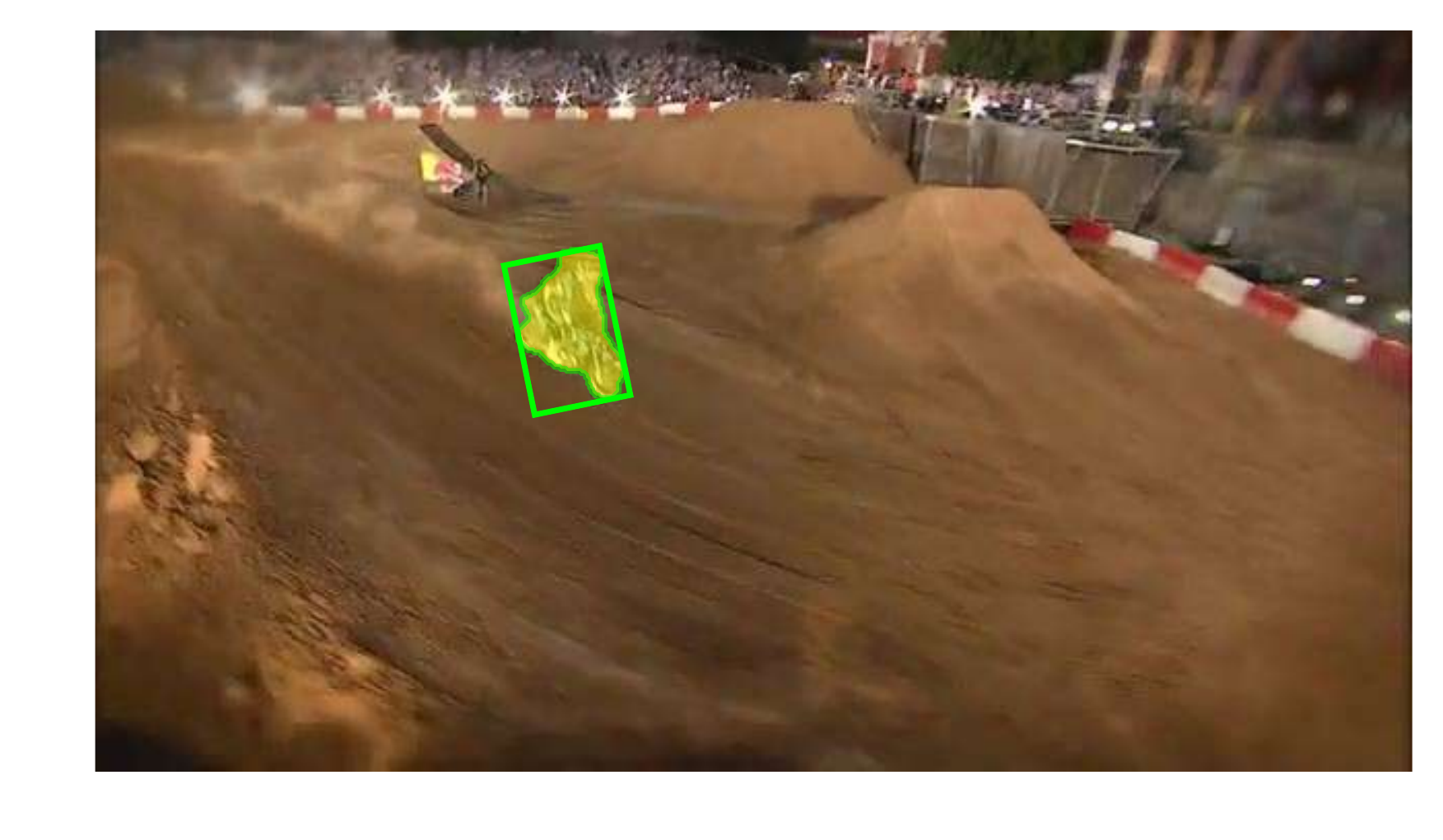}
\\
\mbox{\rotatebox[x=-0.2cm]{90}{\small{singer2}}}
\includegraphics[trim={2.5cm 1cm 2.5cm 1cm},clip,width = 1.1in]{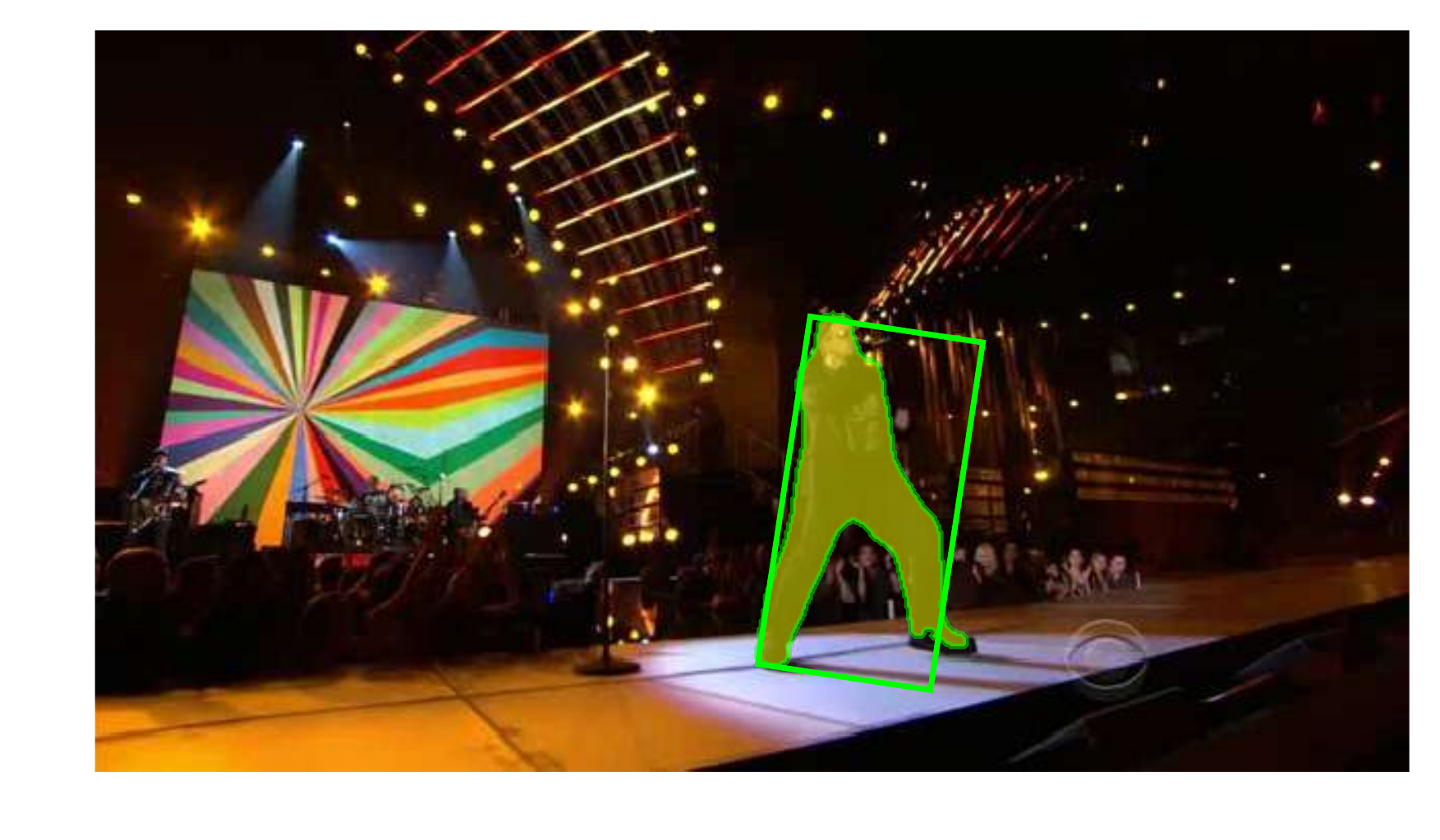}
&\includegraphics[trim={2.5cm 1cm 2.5cm 1cm},clip,width = 1.1in]{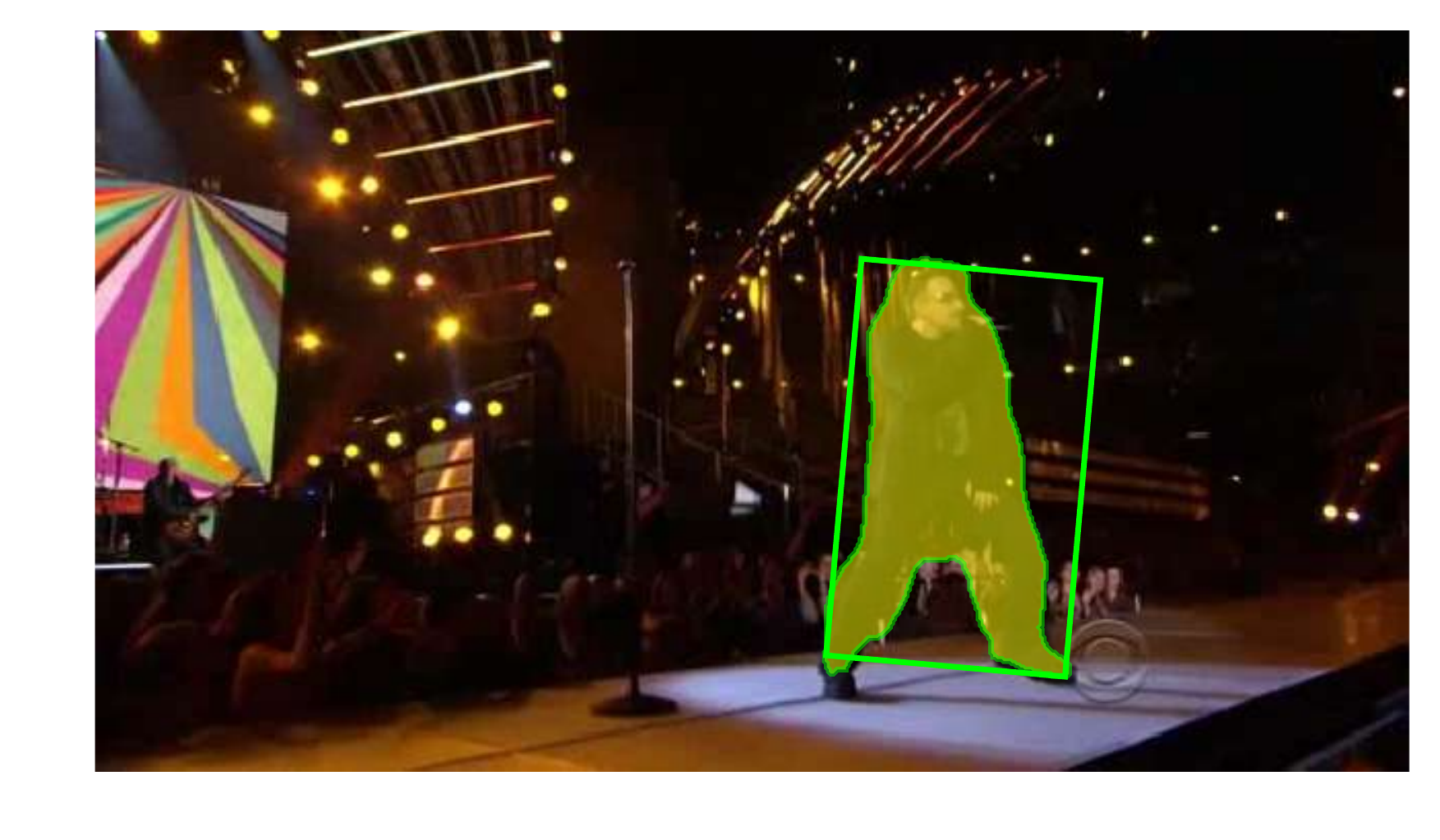}
& \includegraphics[trim={2.5cm 1cm 2.5cm 1cm},clip,width = 1.1in]{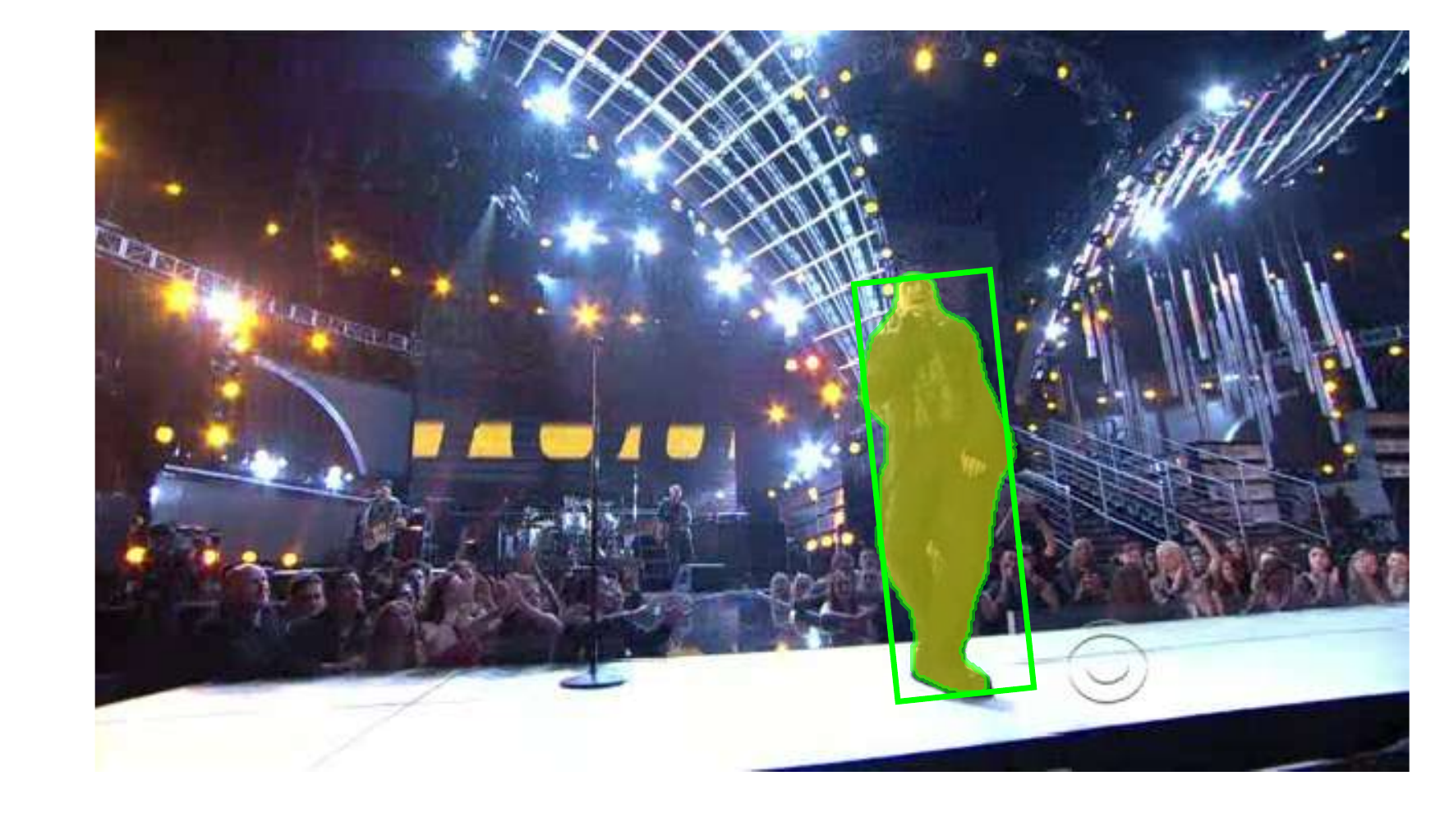}
& \includegraphics[trim={2.5cm 1cm 2.5cm 1cm},clip,width = 1.1in]{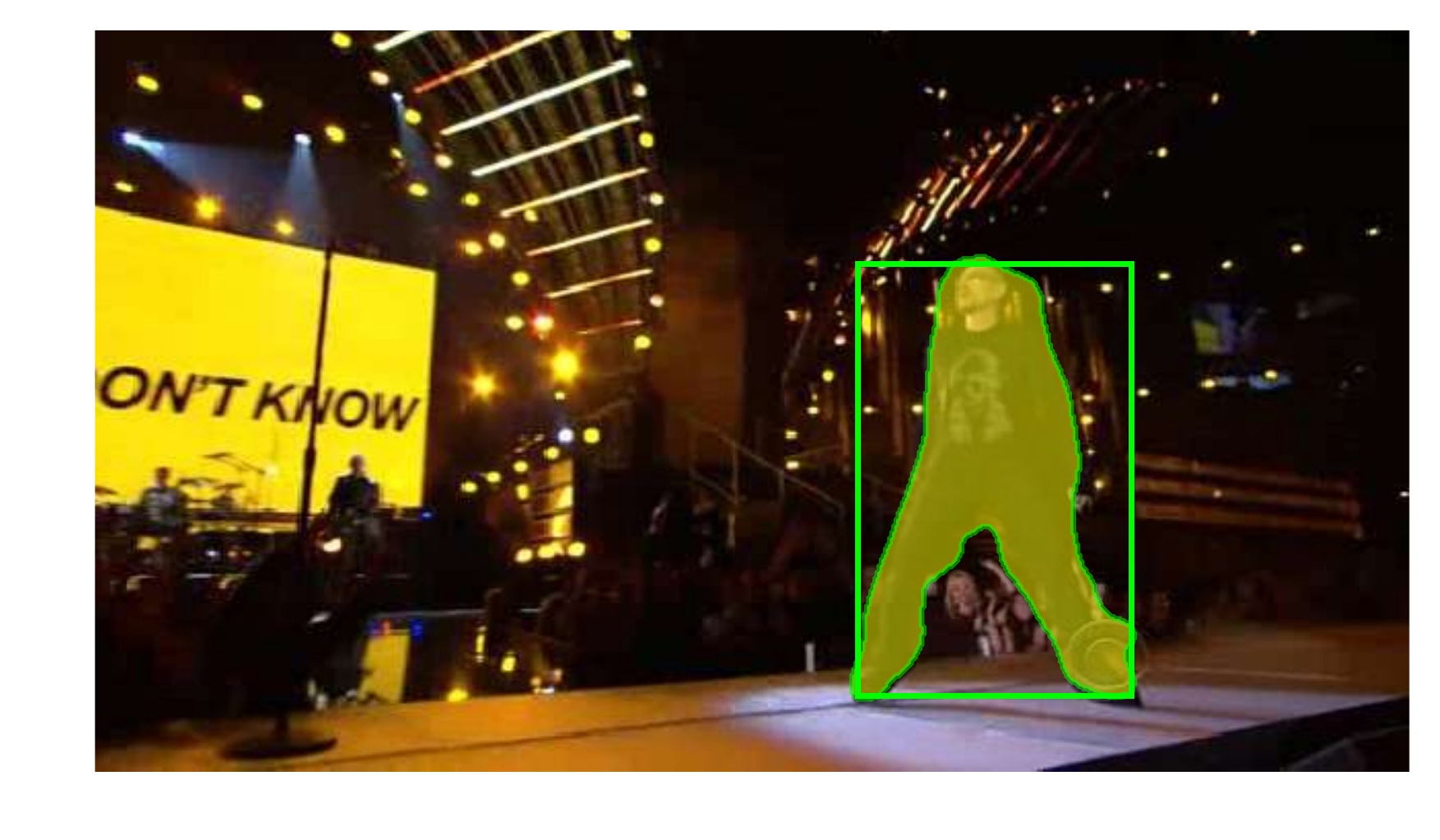}
& \includegraphics[trim={2.5cm 1cm 2.5cm 1cm},clip,width = 1.1in]{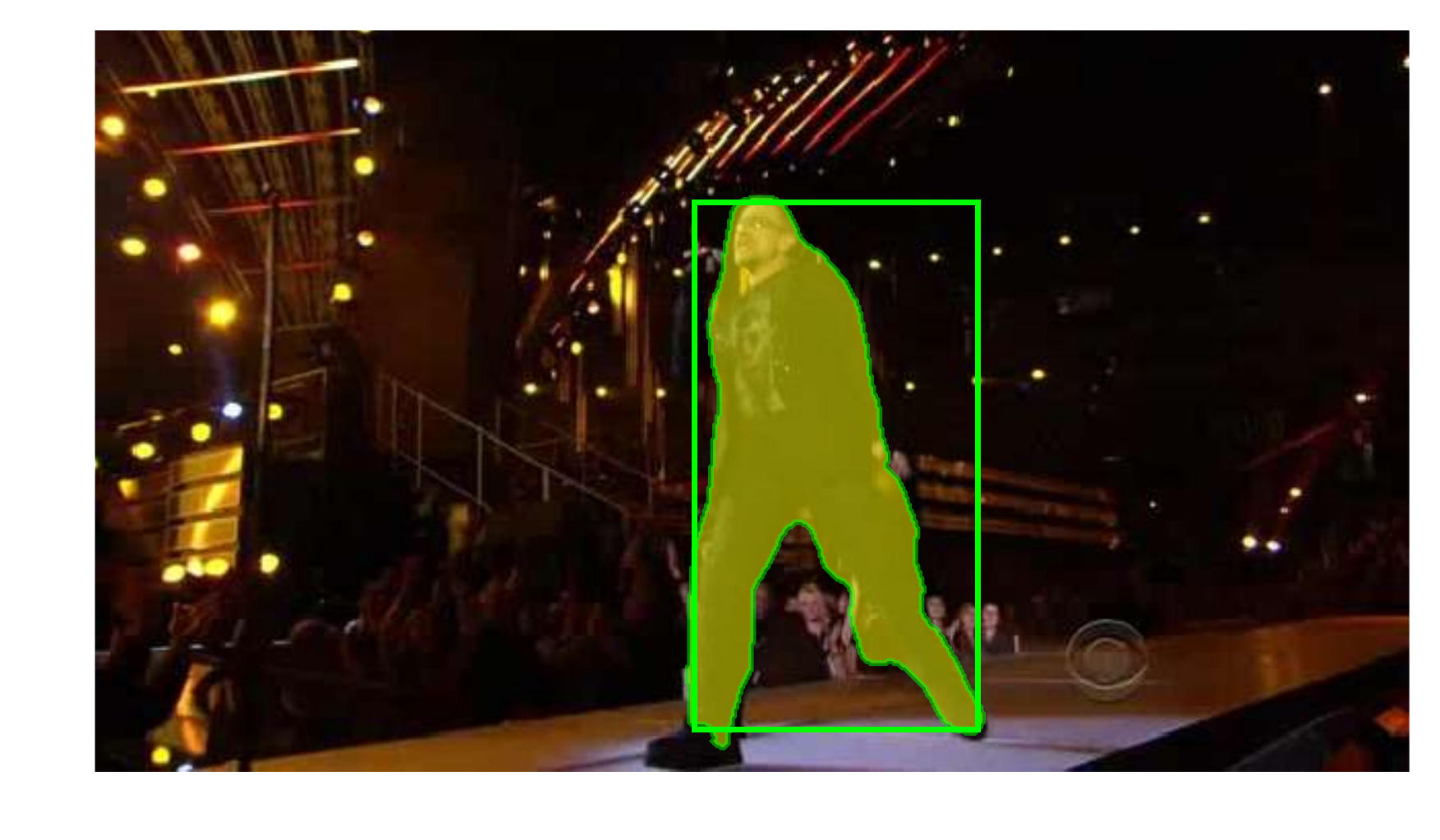}
& \includegraphics[trim={2.5cm 1cm 2.5cm 1cm},clip,width = 1.1in]{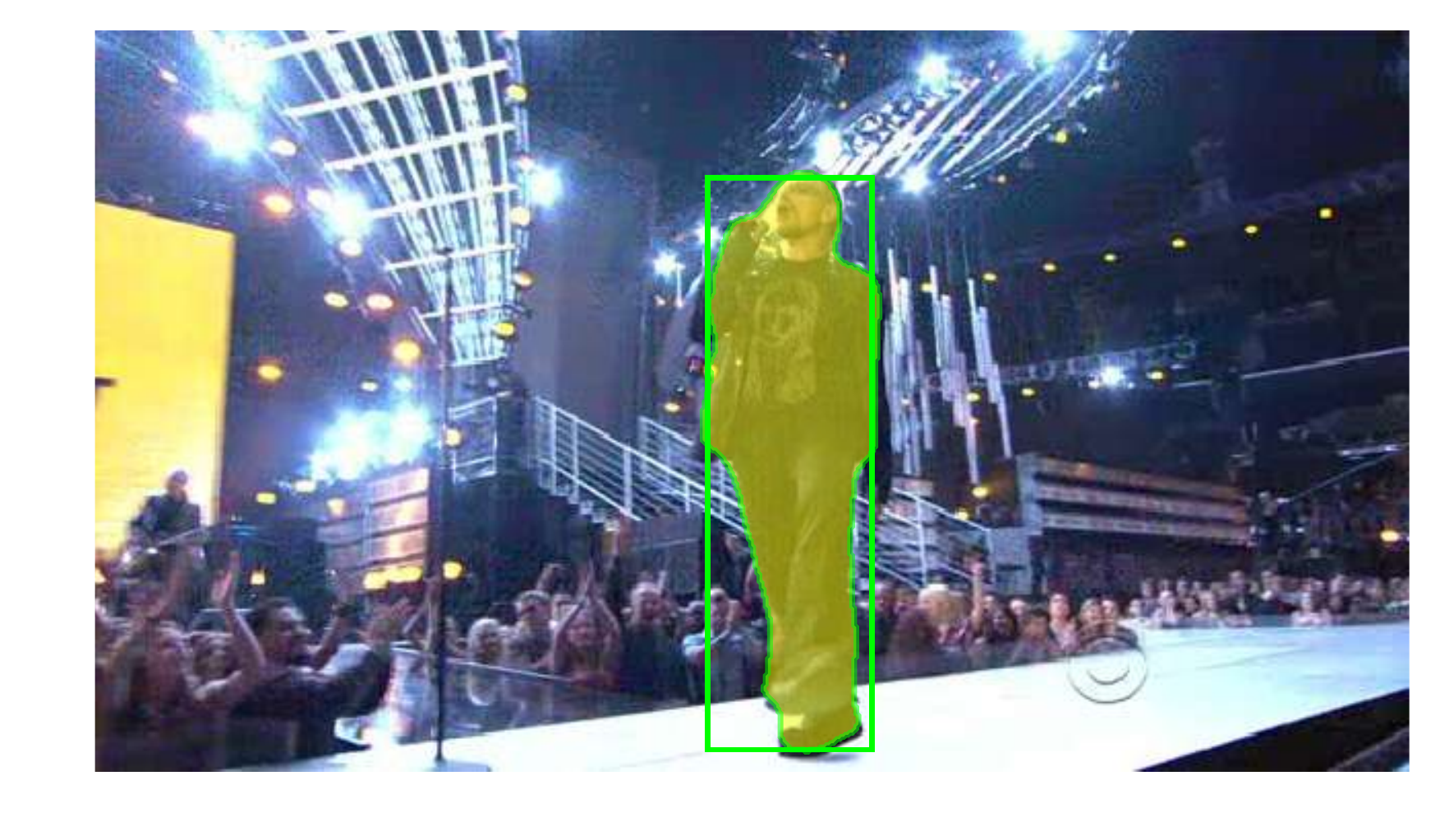}
\\
\mbox{\rotatebox[x=-0.55cm]{90}{\small{soccer1}}}
\includegraphics[trim={2.5cm 1cm 2.5cm 1cm},clip,width = 1.1in]{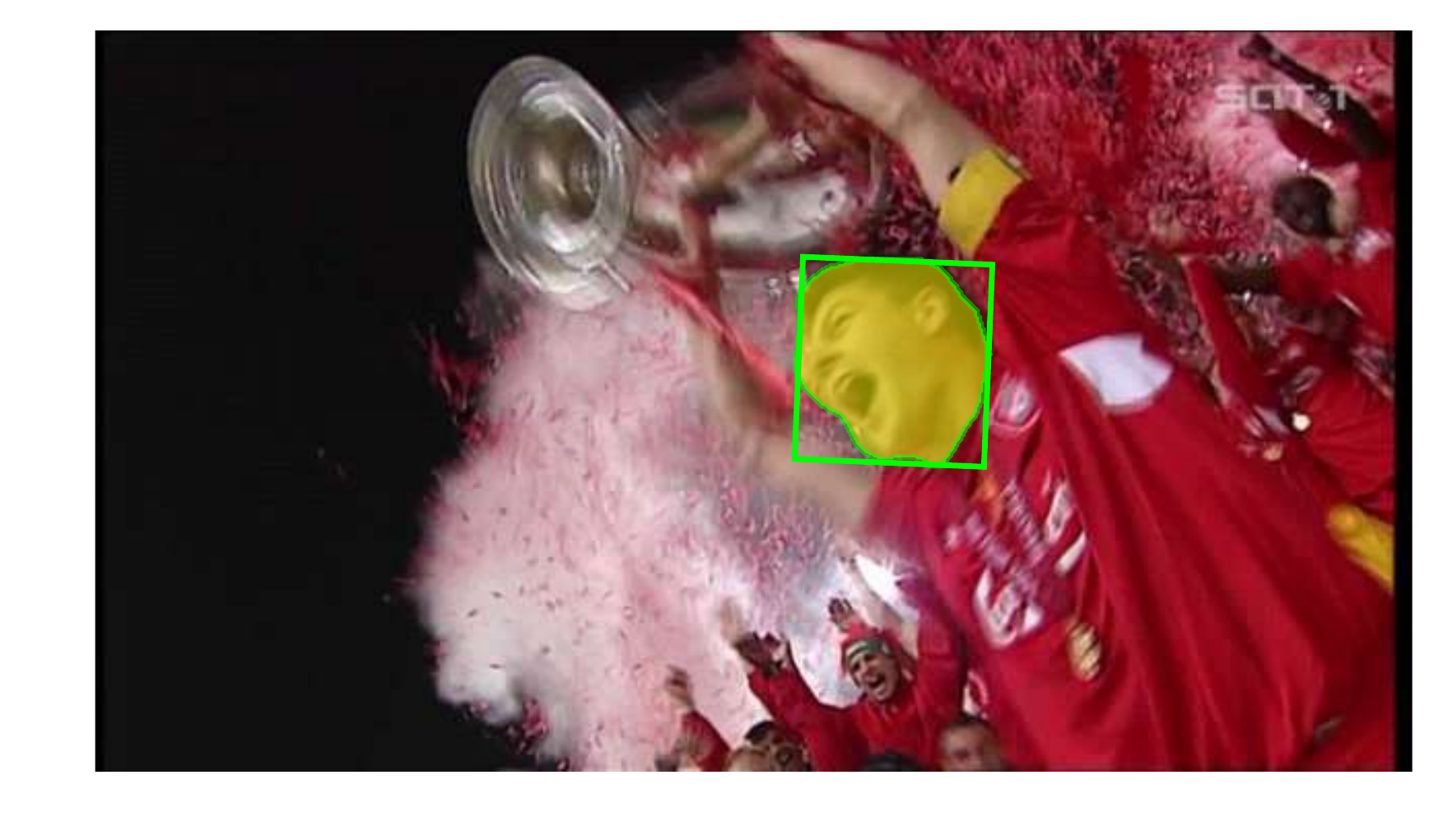}
&\includegraphics[trim={2.5cm 1cm 2.5cm 1cm},clip,width = 1.1in]{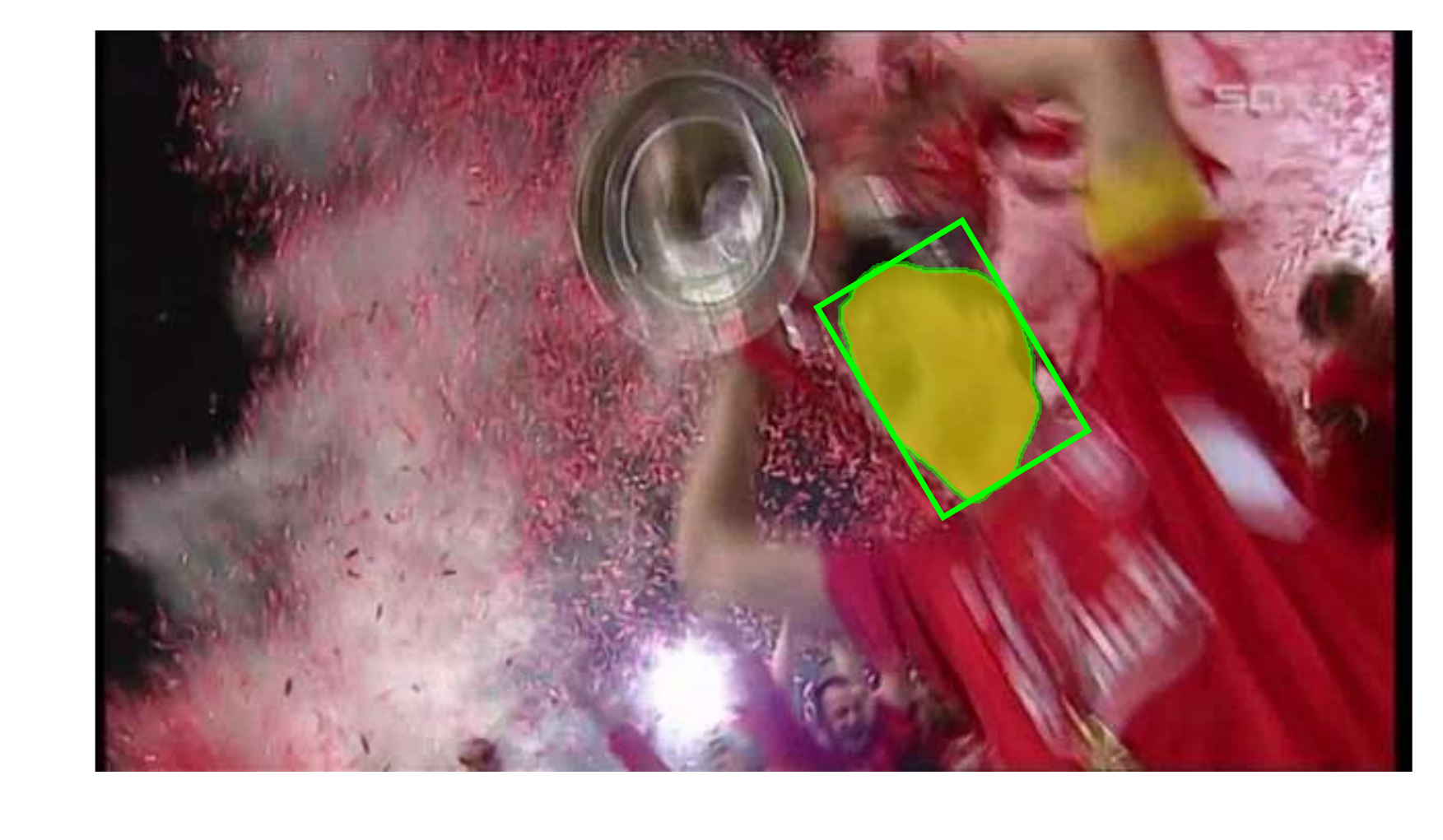}
& \includegraphics[trim={2.5cm 1cm 2.5cm 1cm},clip,width = 1.1in]{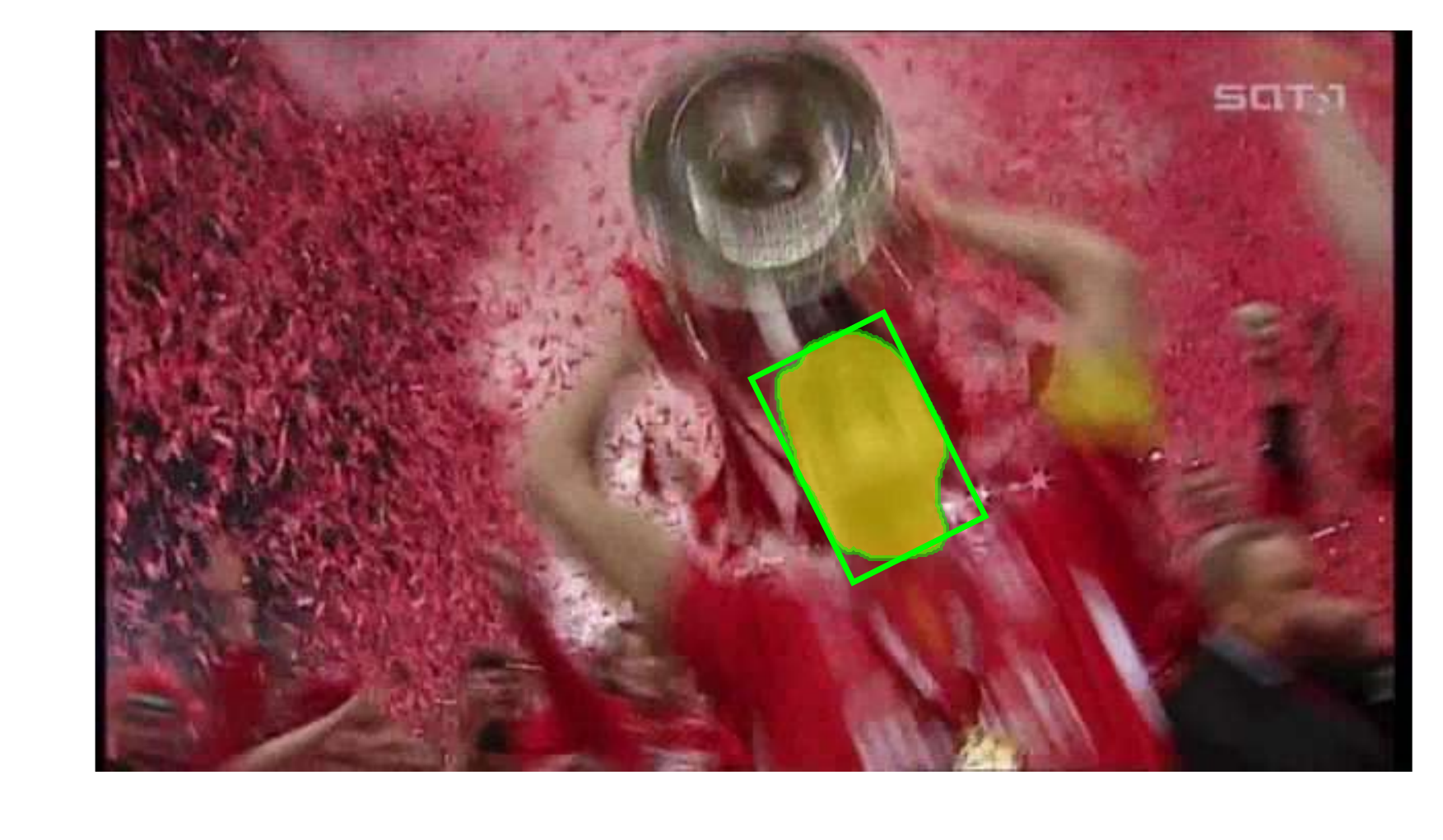}
& \includegraphics[trim={2.5cm 1cm 2.5cm 1cm},clip,width = 1.1in]{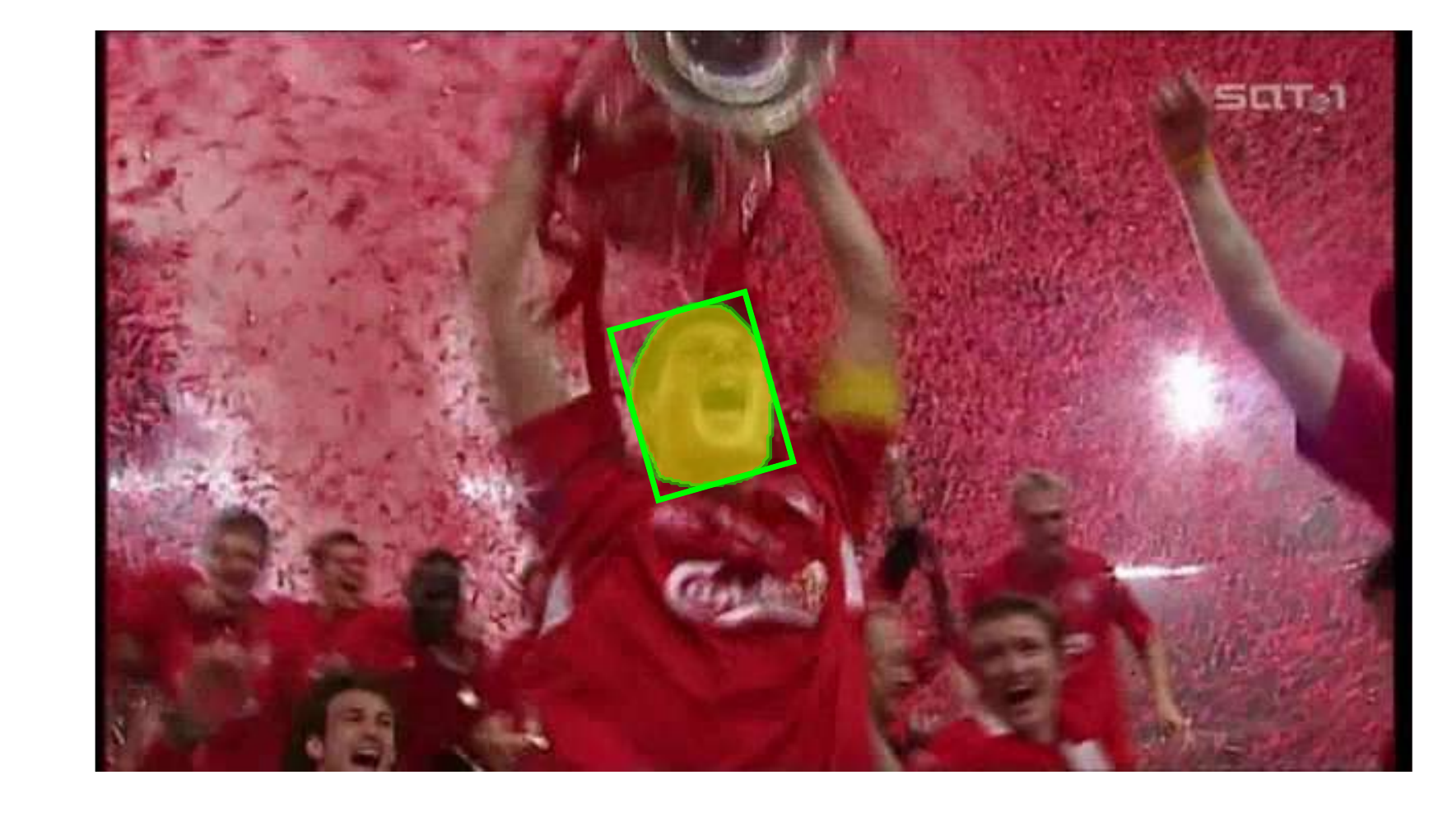}
& \includegraphics[trim={2.5cm 1cm 2.5cm 1cm},clip,width = 1.1in]{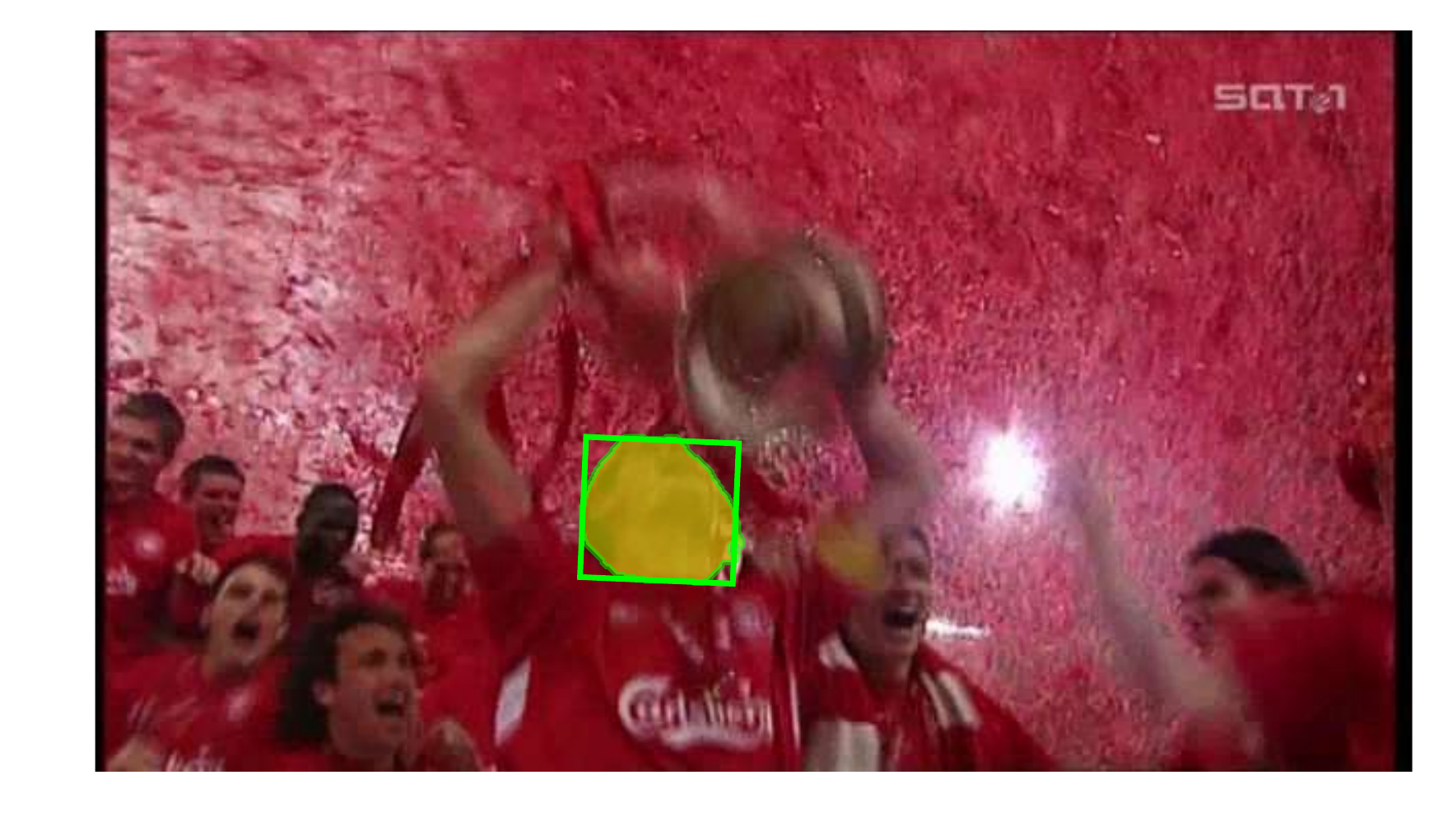}
& \includegraphics[trim={2.5cm 1cm 2.5cm 1cm},clip,width = 1.1in]{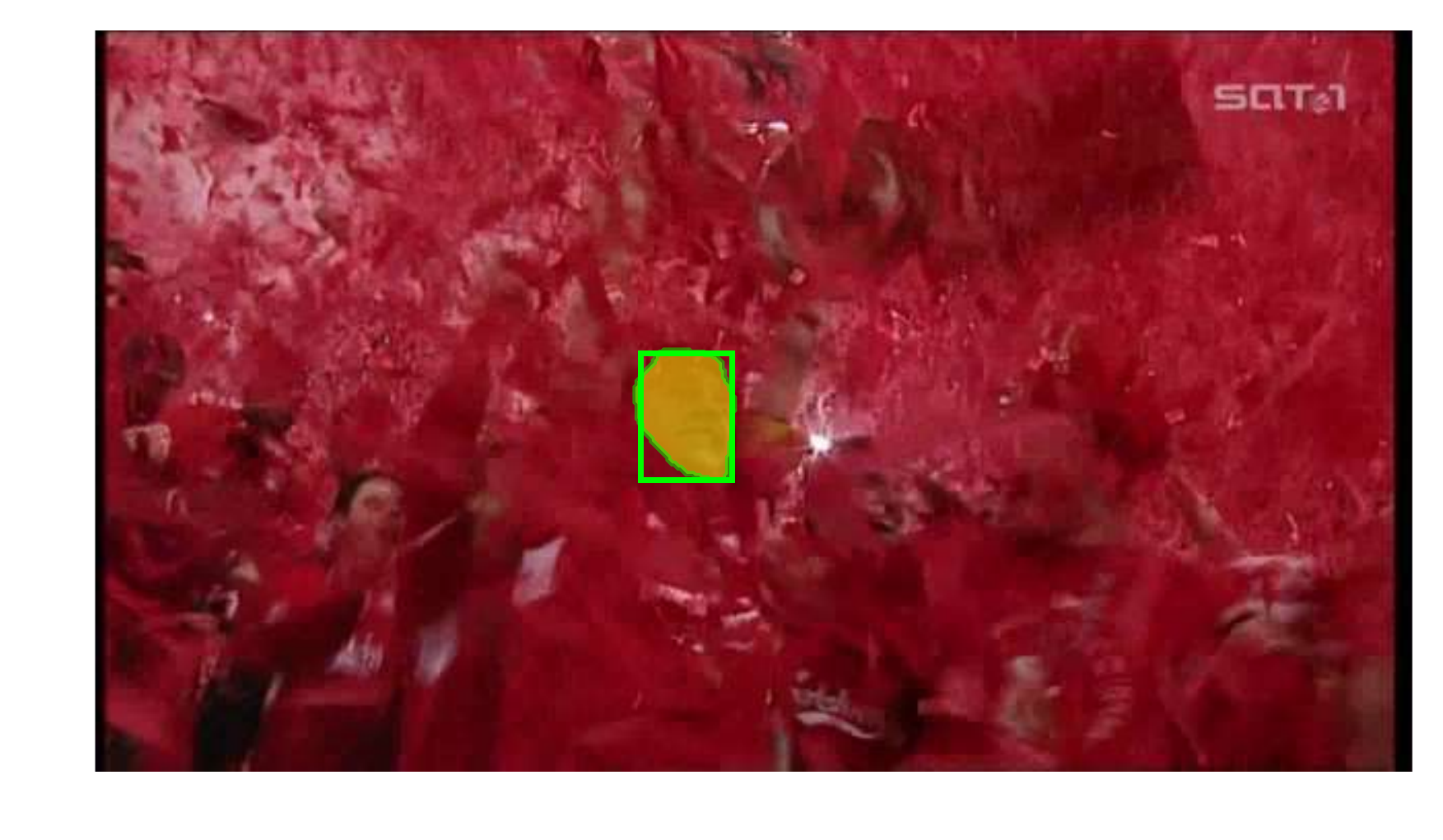}
\\
\end{tabular}

\vspace{-0.2cm}
\caption{Further qualitative results of our method on sequences from the visual object tracking benchmark VOT-2018~\cite{VOT2018}.}
\label{fig:appendix_vot18}
\end{figure*}

\begin{figure*}
\centering
\setlength{\tabcolsep}{0.25ex}

\begin{tabular}
{cccccc cccccc}

\mbox{\rotatebox[x=-0.55cm]{90}{\small{dog}}}
\includegraphics[trim={2.5cm 1cm 2.5cm 1cm},clip,width = 1.1in]{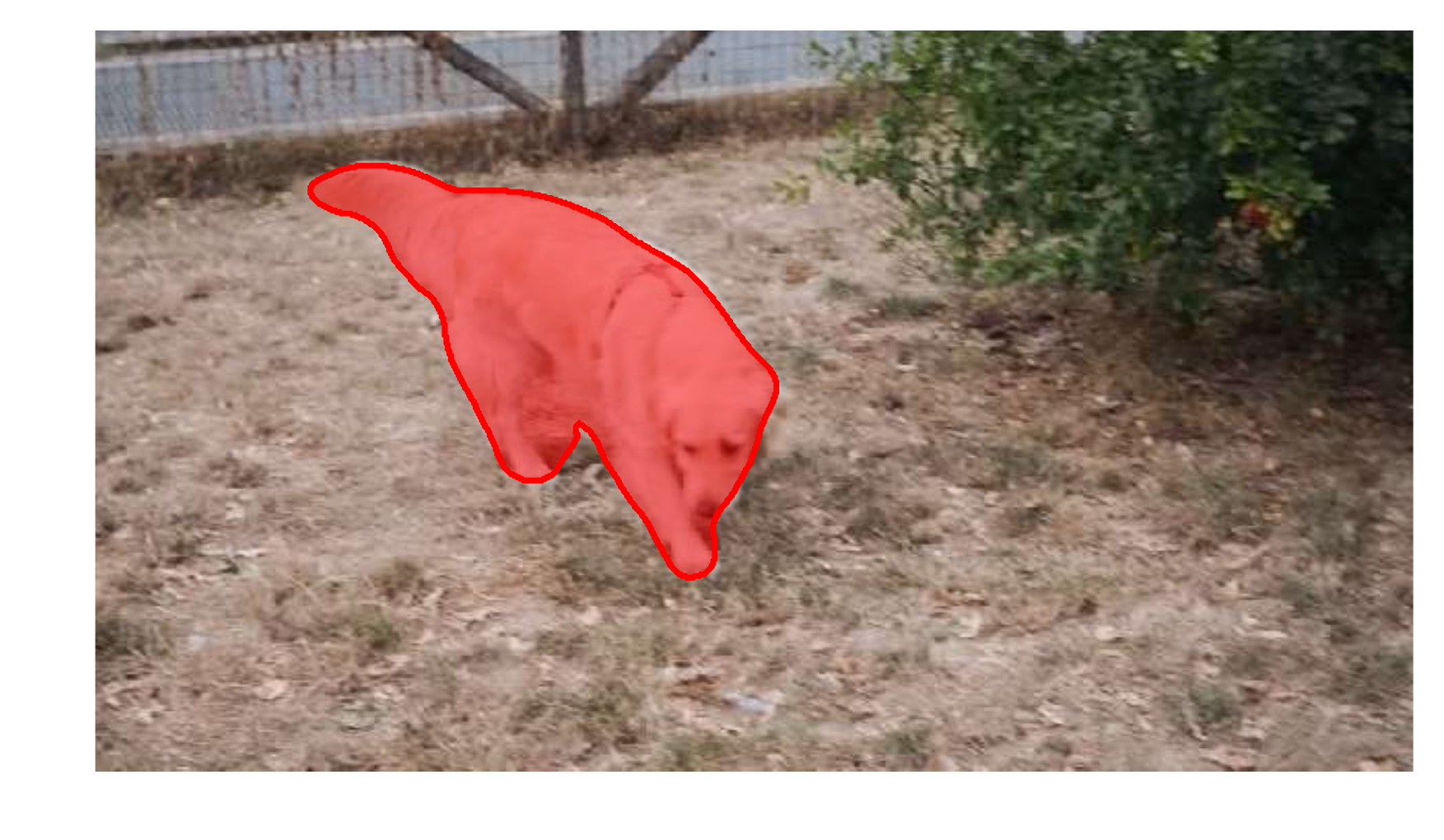}
&\includegraphics[trim={2.5cm 1cm 2.5cm 1cm},clip,width = 1.1in]{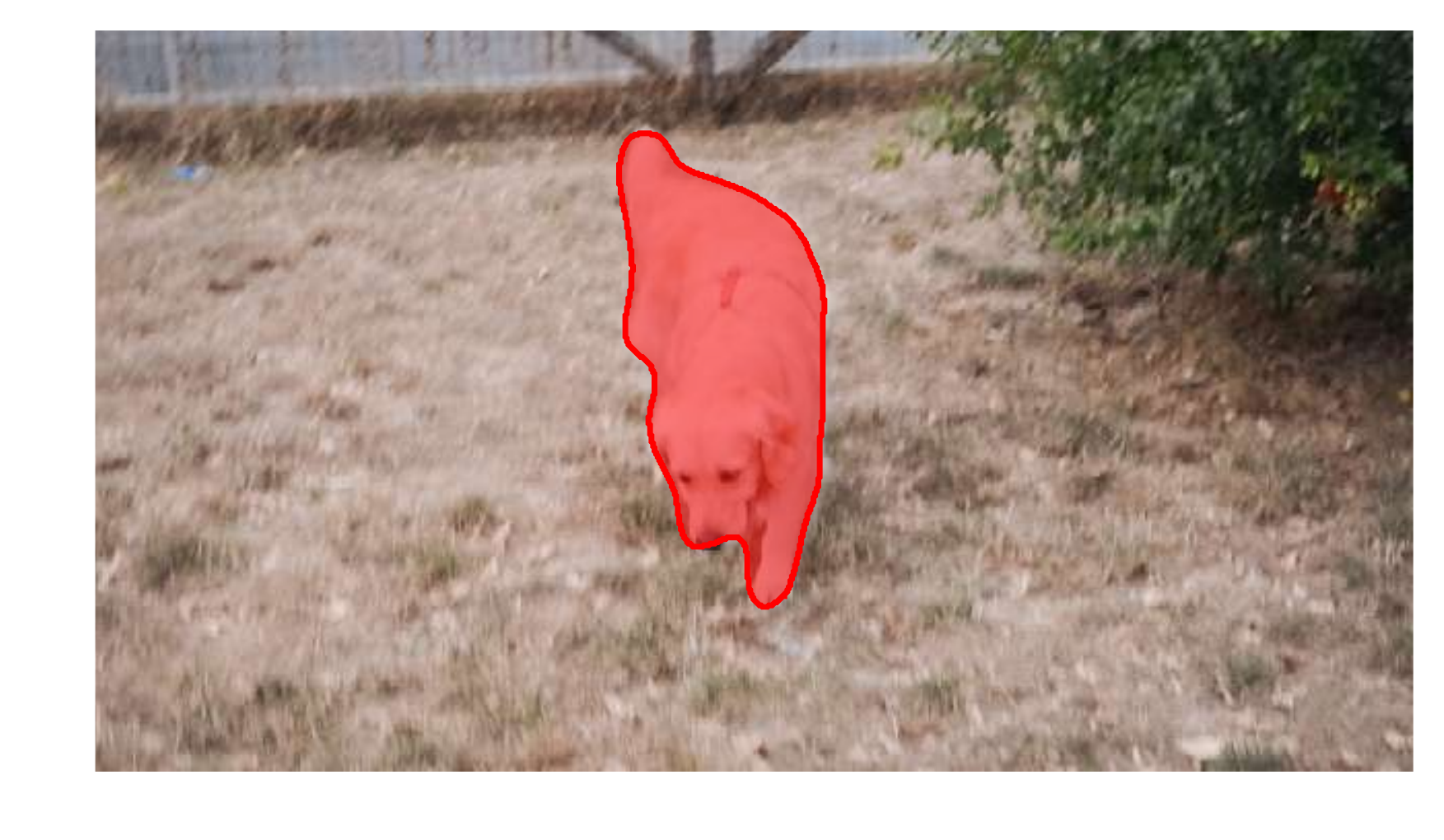}
& \includegraphics[trim={2.5cm 1cm 2.5cm 1cm},clip,width = 1.1in]{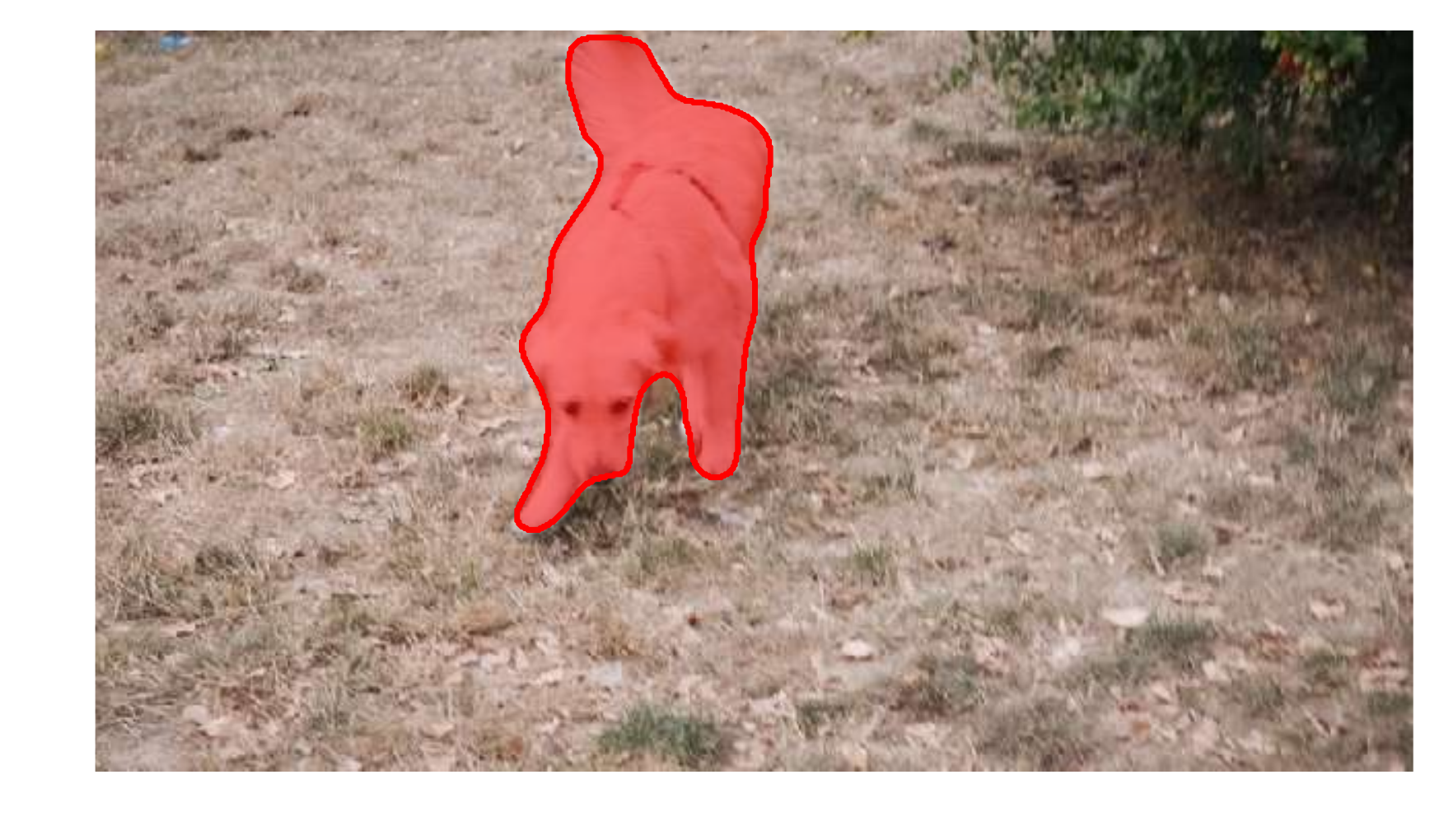}
& \includegraphics[trim={2.5cm 1cm 2.5cm 1cm},clip,width = 1.1in]{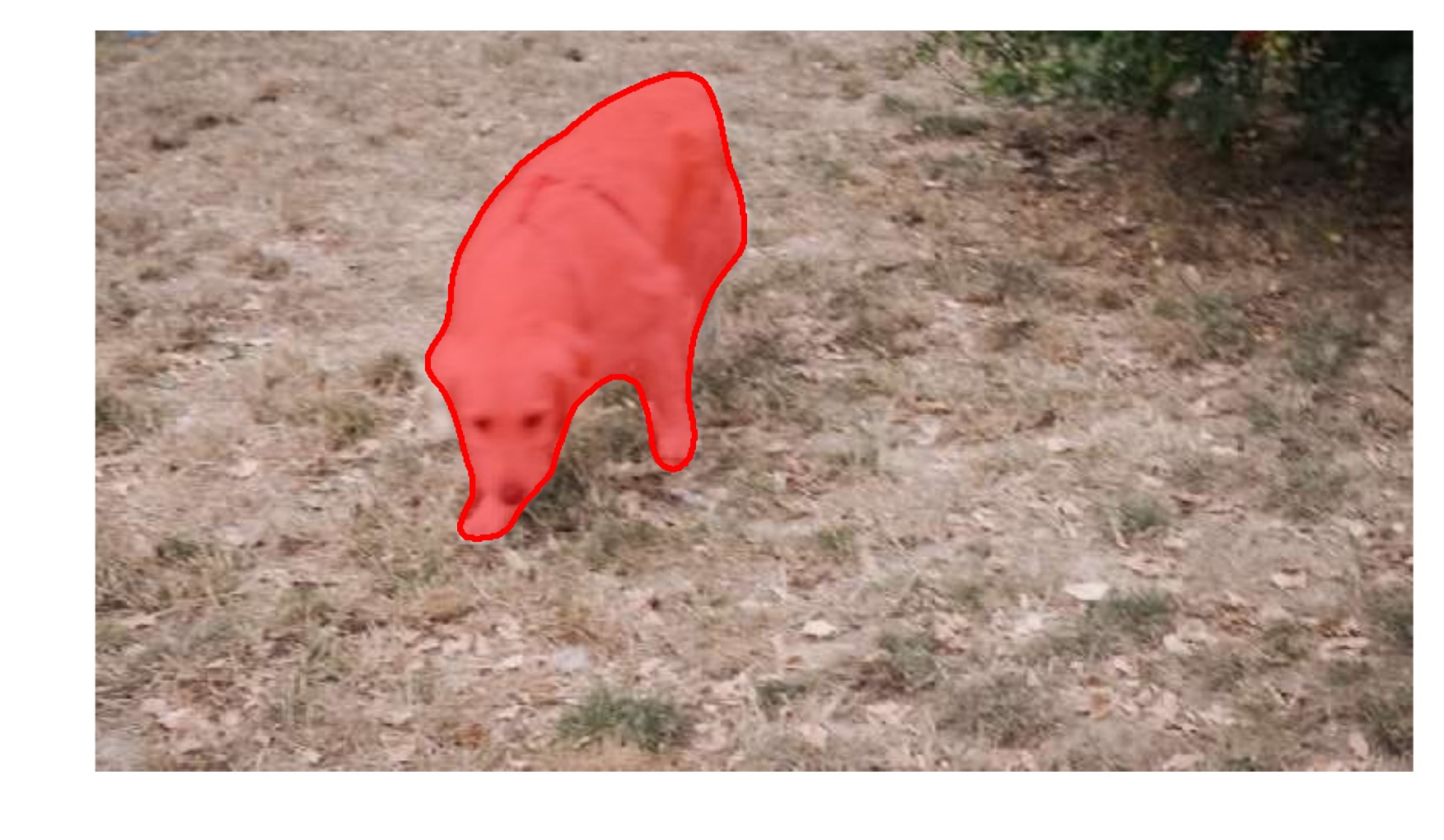}
& \includegraphics[trim={2.5cm 1cm 2.5cm 1cm},clip,width = 1.1in]{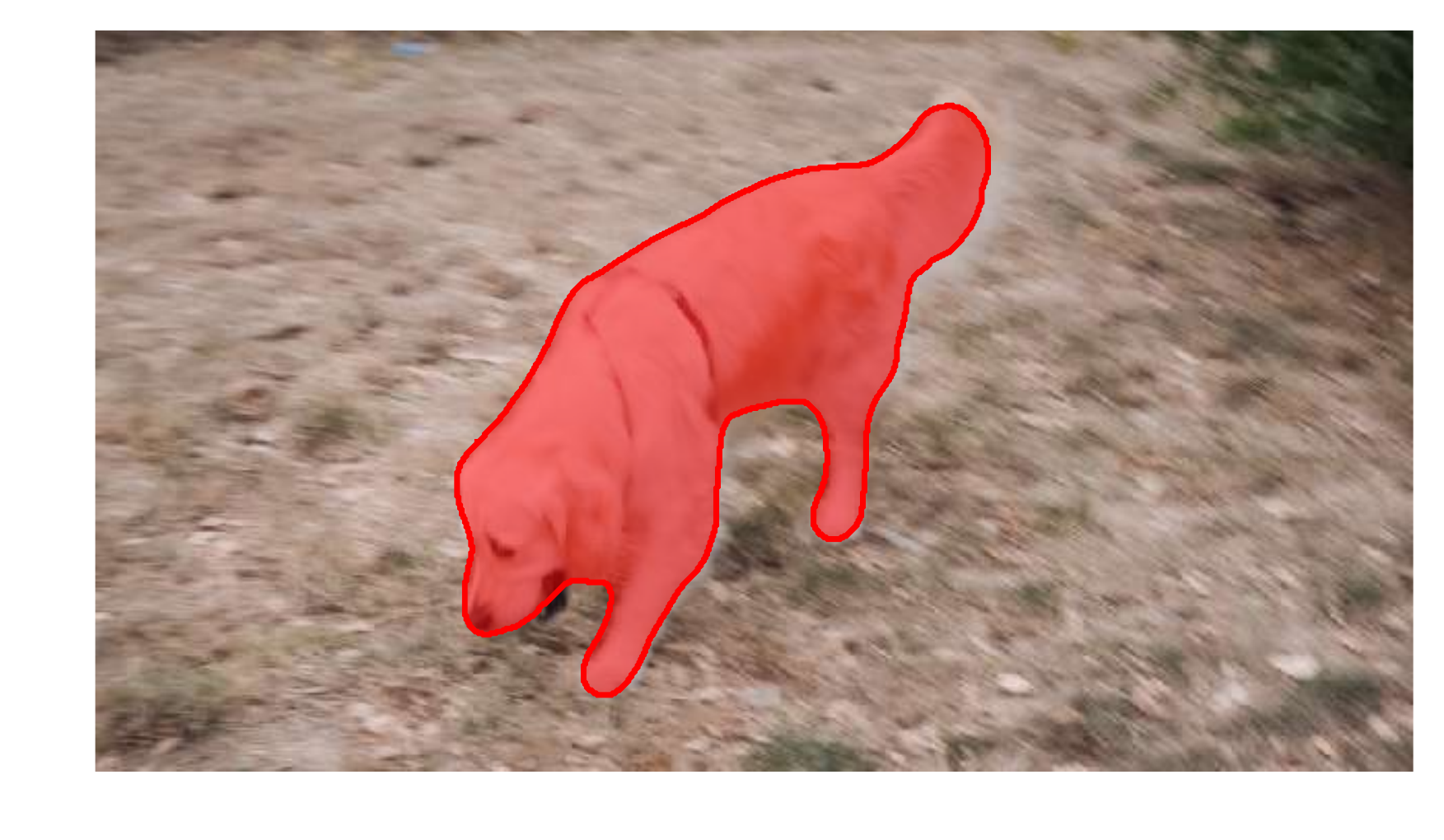}
& \includegraphics[trim={2.5cm 1cm 2.5cm 1cm},clip,width = 1.1in]{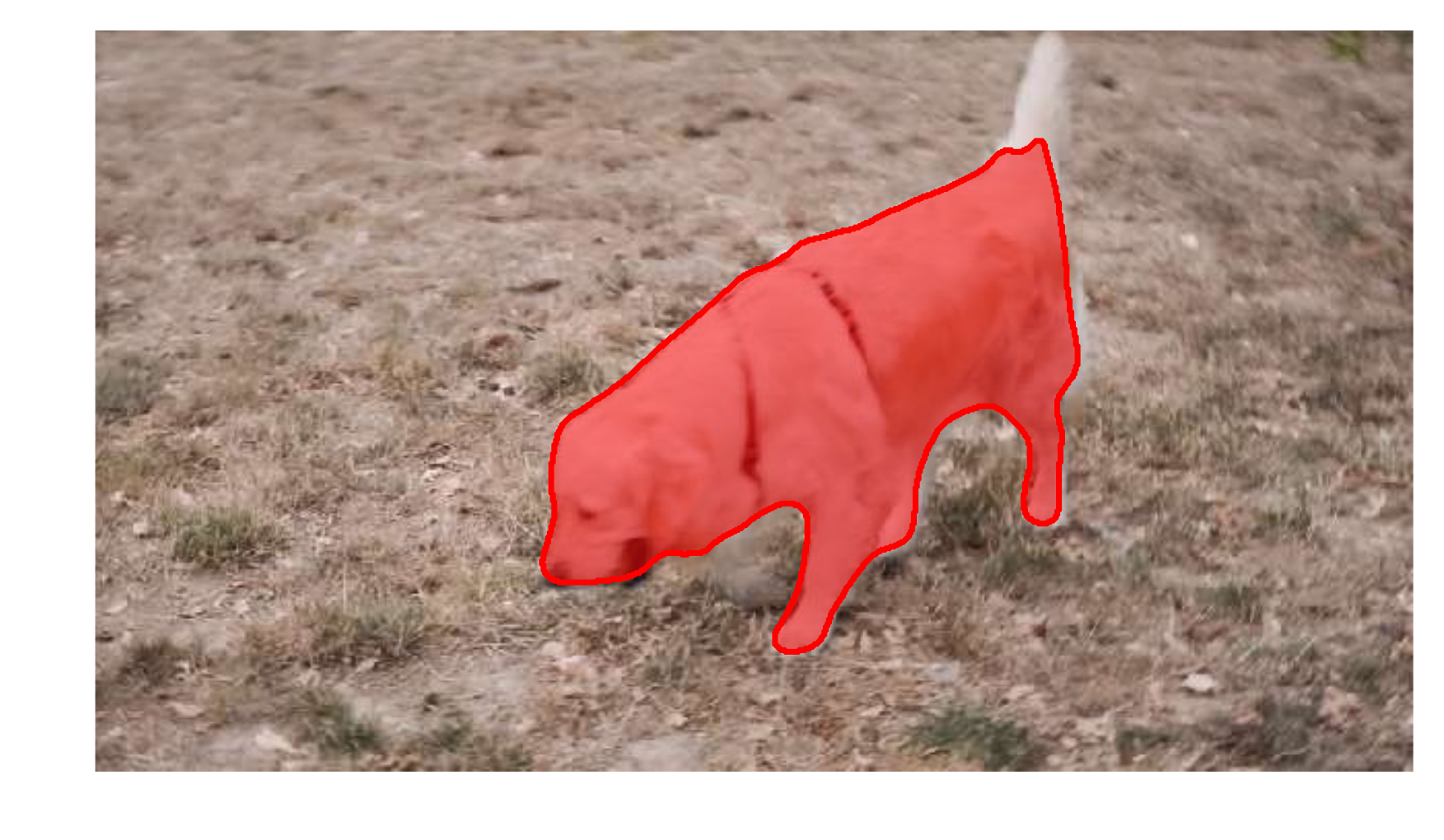}
\\

\mbox{\rotatebox[x=-0.cm]{90}{\small{drift-straight}}}
\includegraphics[trim={2.5cm 1cm 2.5cm 1cm},clip,width = 1.1in]{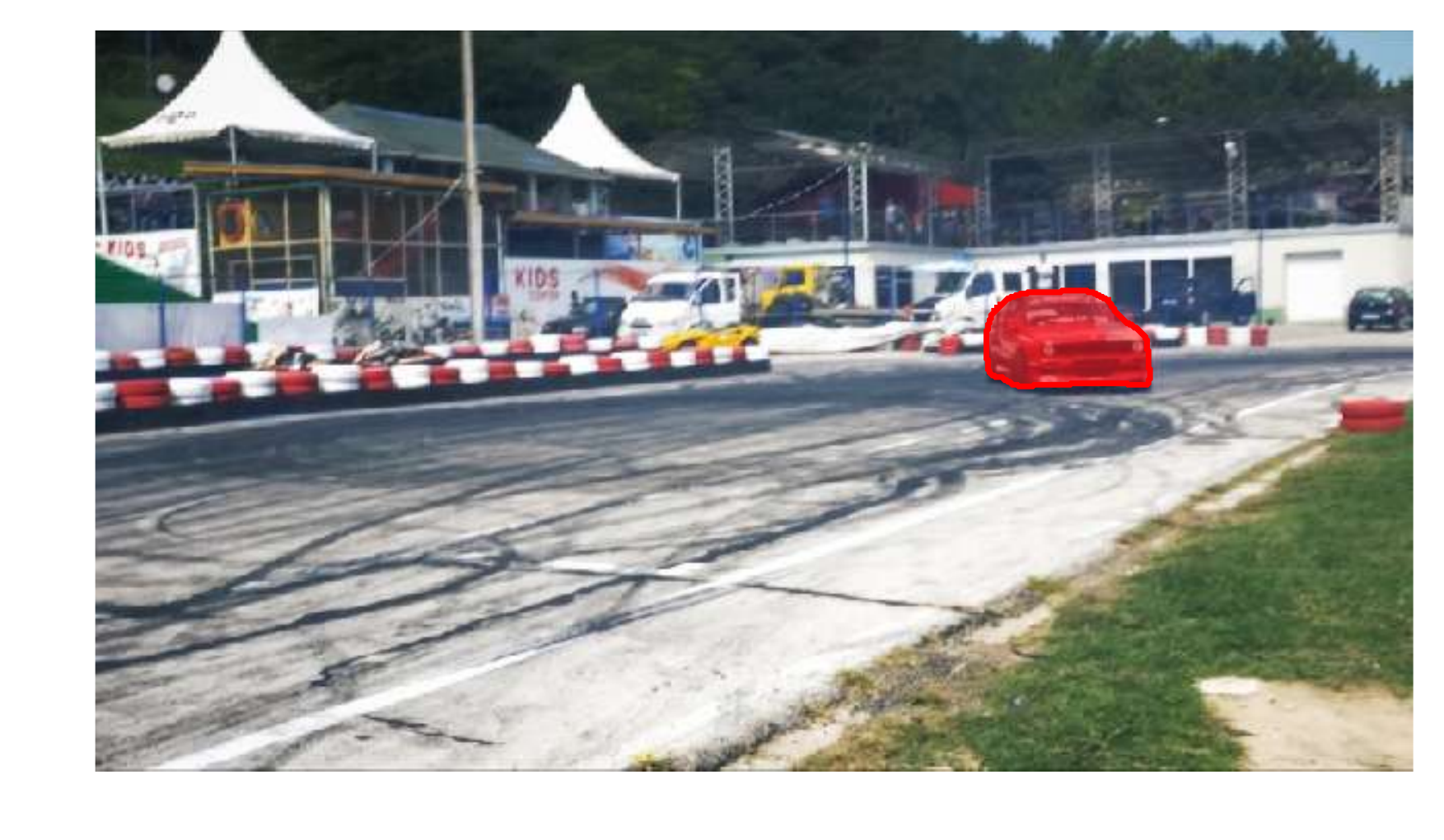}
&\includegraphics[trim={2.5cm 1cm 2.5cm 1cm},clip,width = 1.1in]{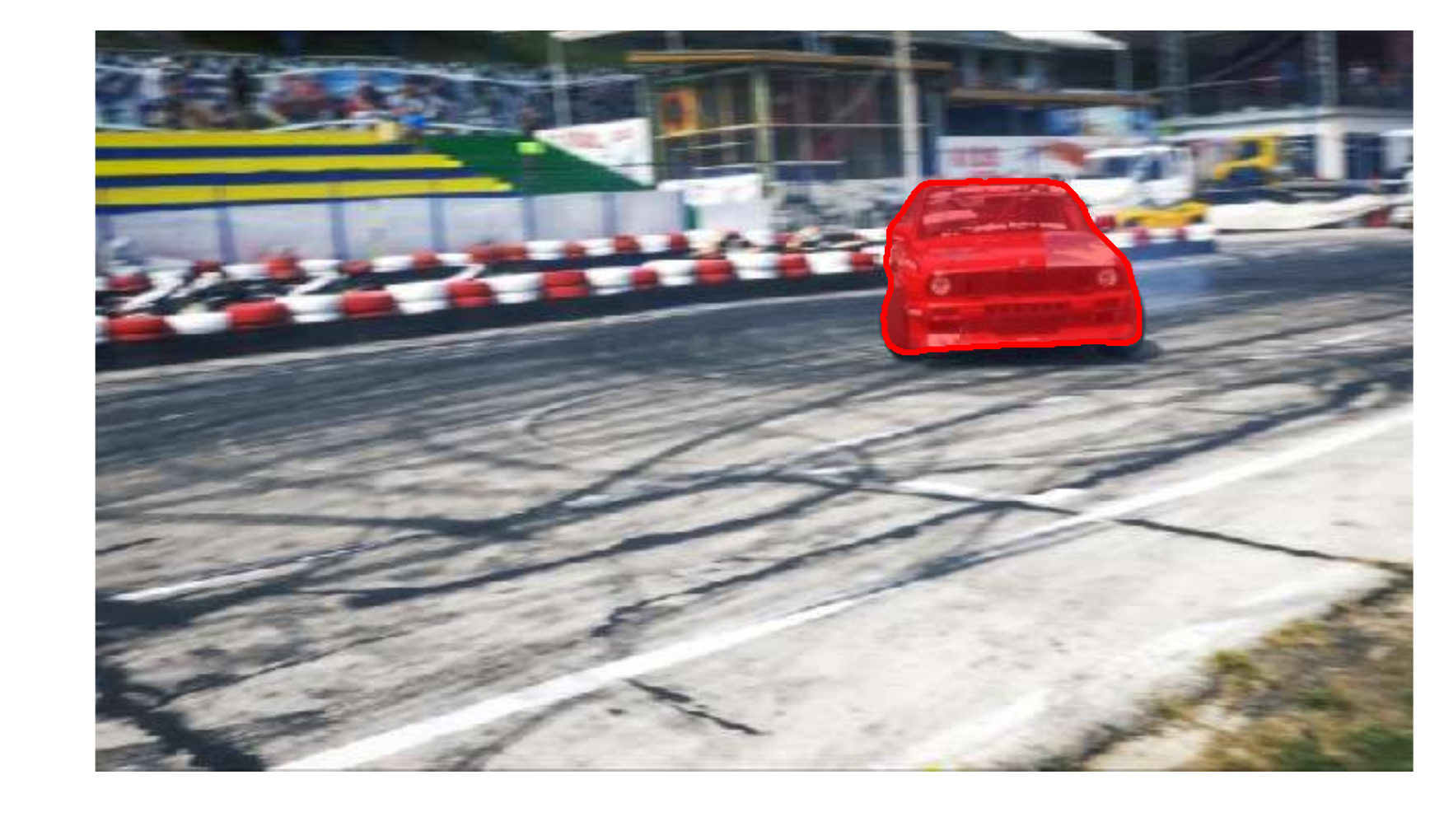}
& \includegraphics[trim={2.5cm 1cm 2.5cm 1cm},clip,width = 1.1in]{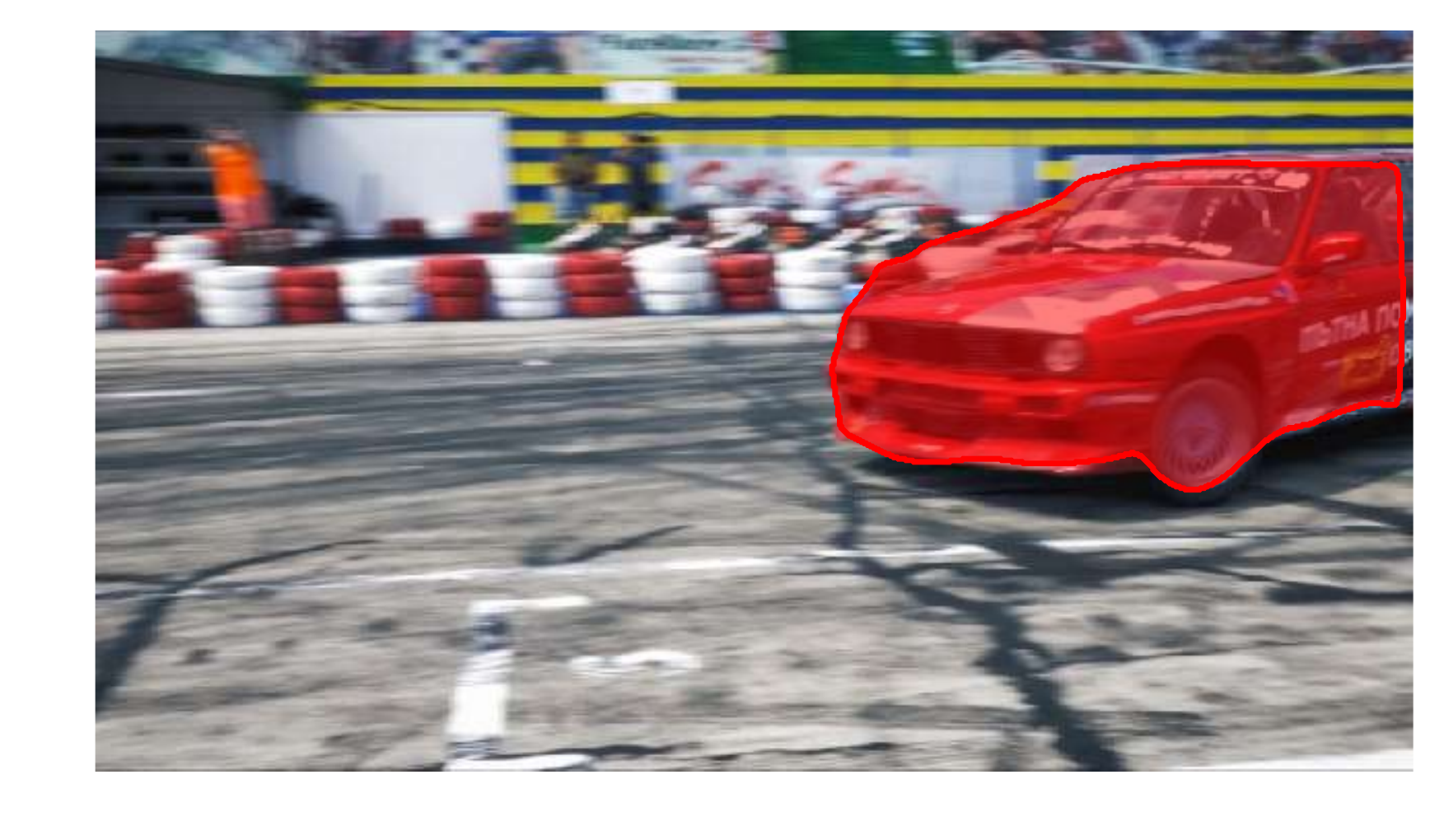}
& \includegraphics[trim={2.5cm 1cm 2.5cm 1cm},clip,width = 1.1in]{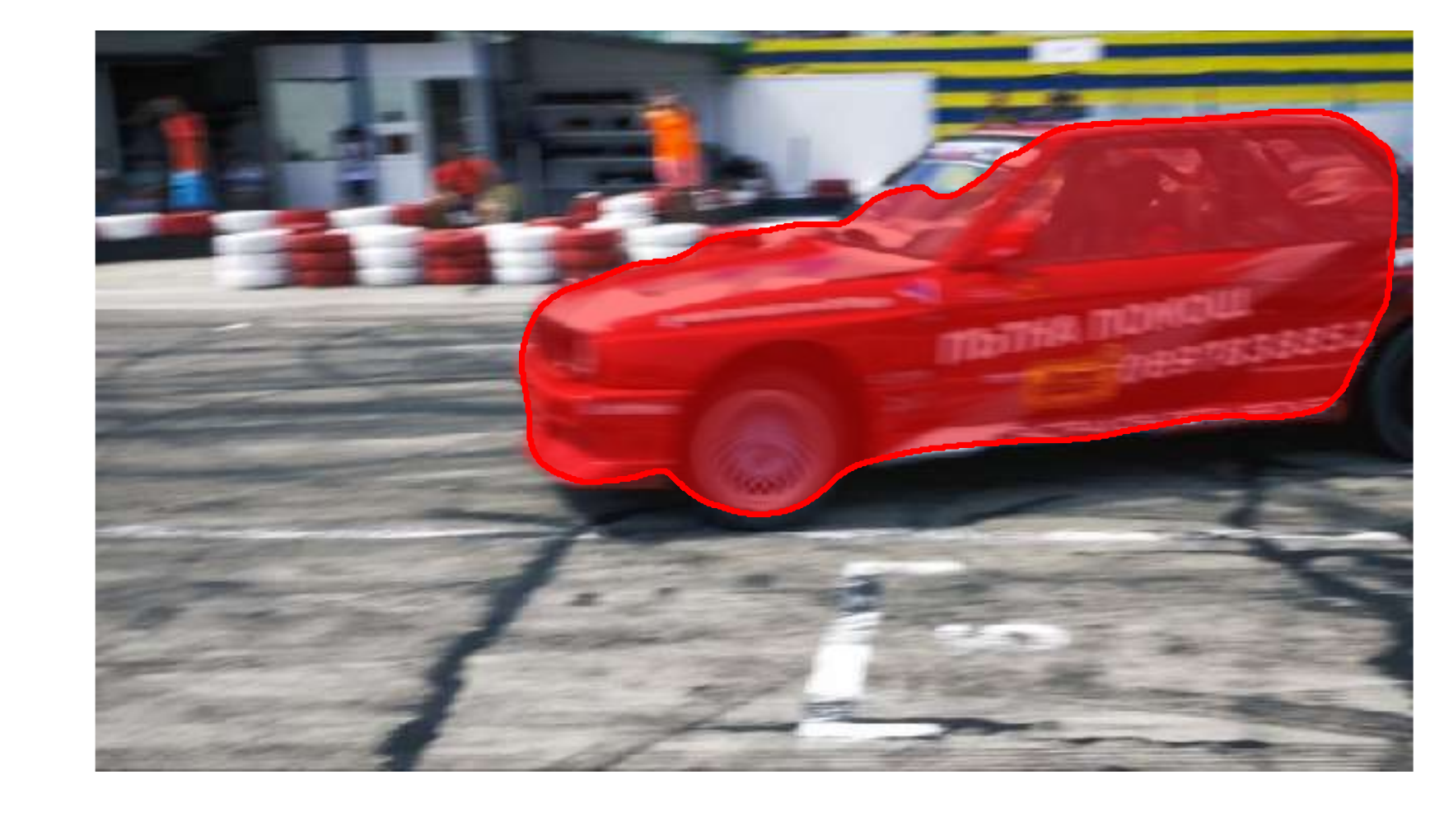}
& \includegraphics[trim={2.5cm 1cm 2.5cm 1cm},clip,width = 1.1in]{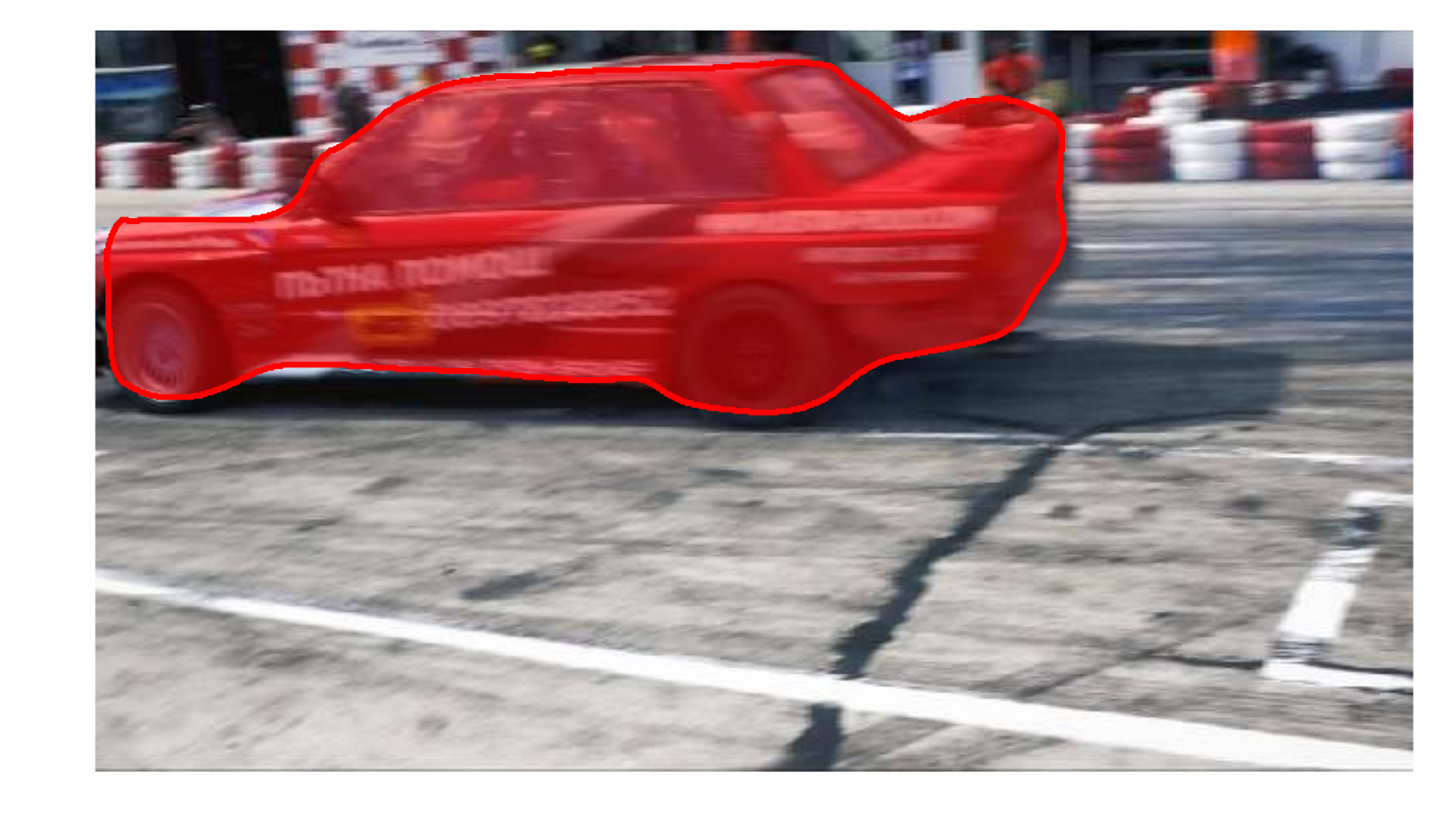}
& \includegraphics[trim={2.5cm 1cm 2.5cm 1cm},clip,width = 1.1in]{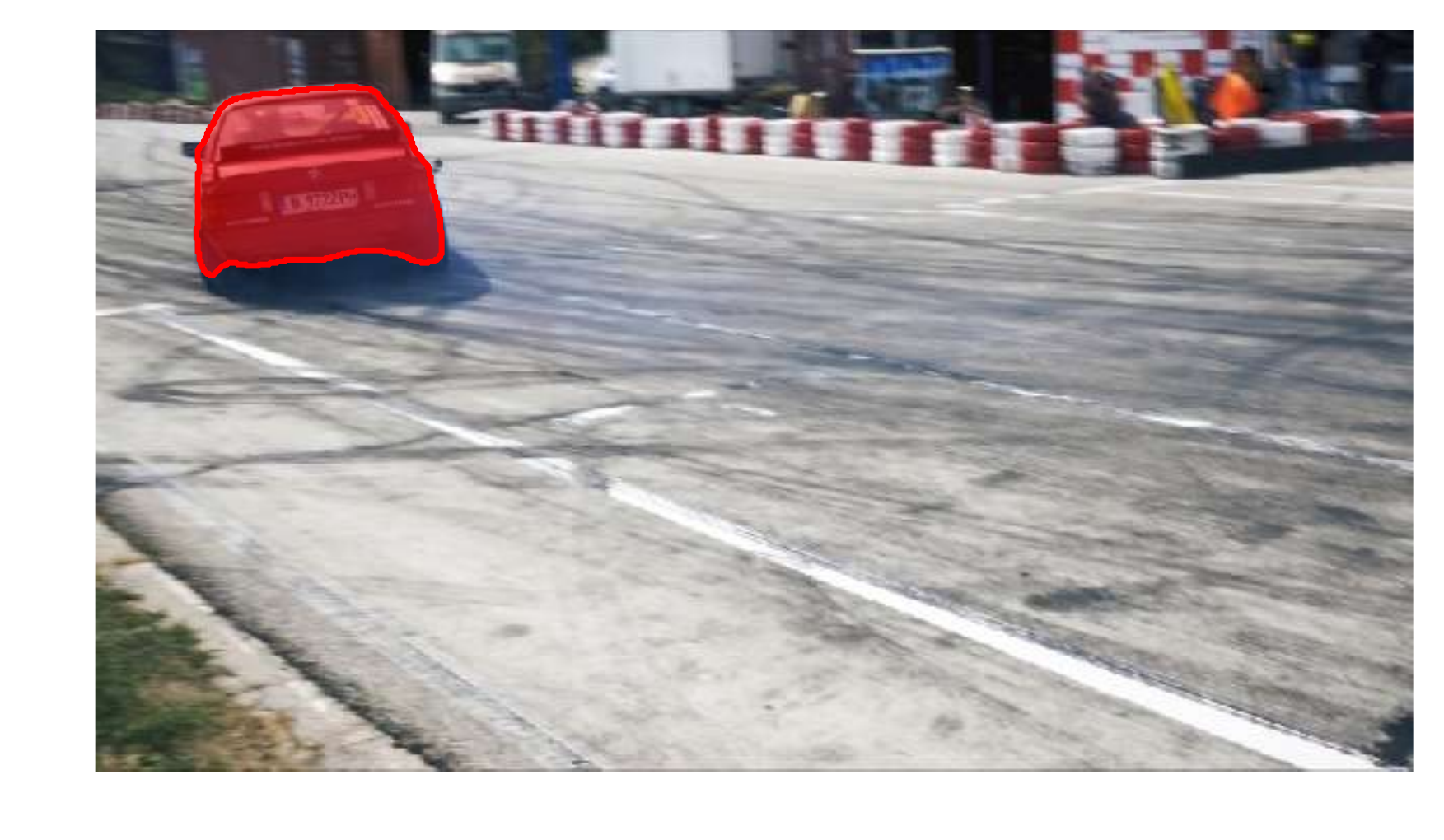}
\\

\mbox{\rotatebox[x=-0.55cm]{90}{\small{goat}}}
\includegraphics[trim={2.5cm 1cm 2.5cm 1cm},clip,width = 1.1in]{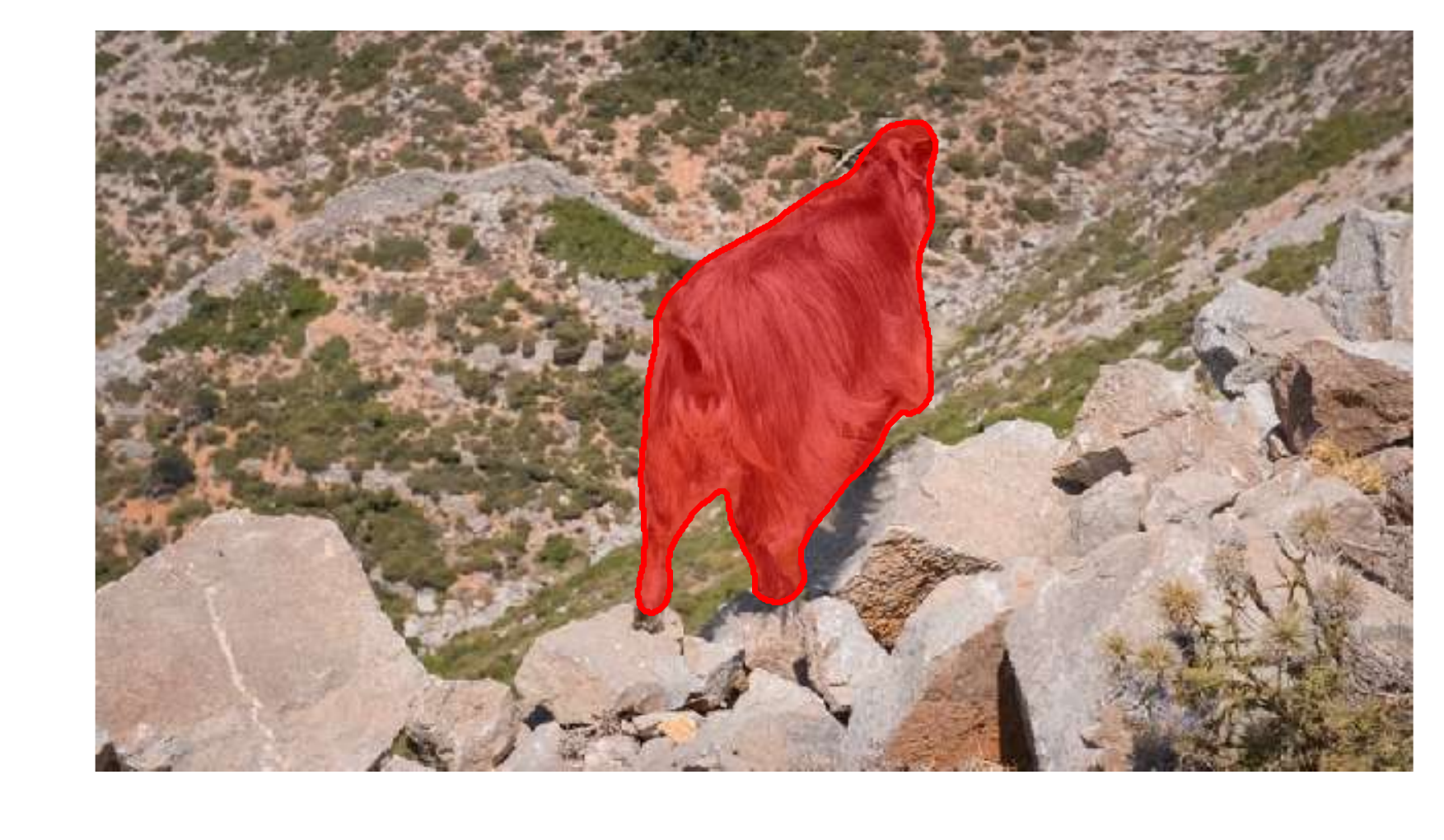}
&\includegraphics[trim={2.5cm 1cm 2.5cm 1cm},clip,width = 1.1in]{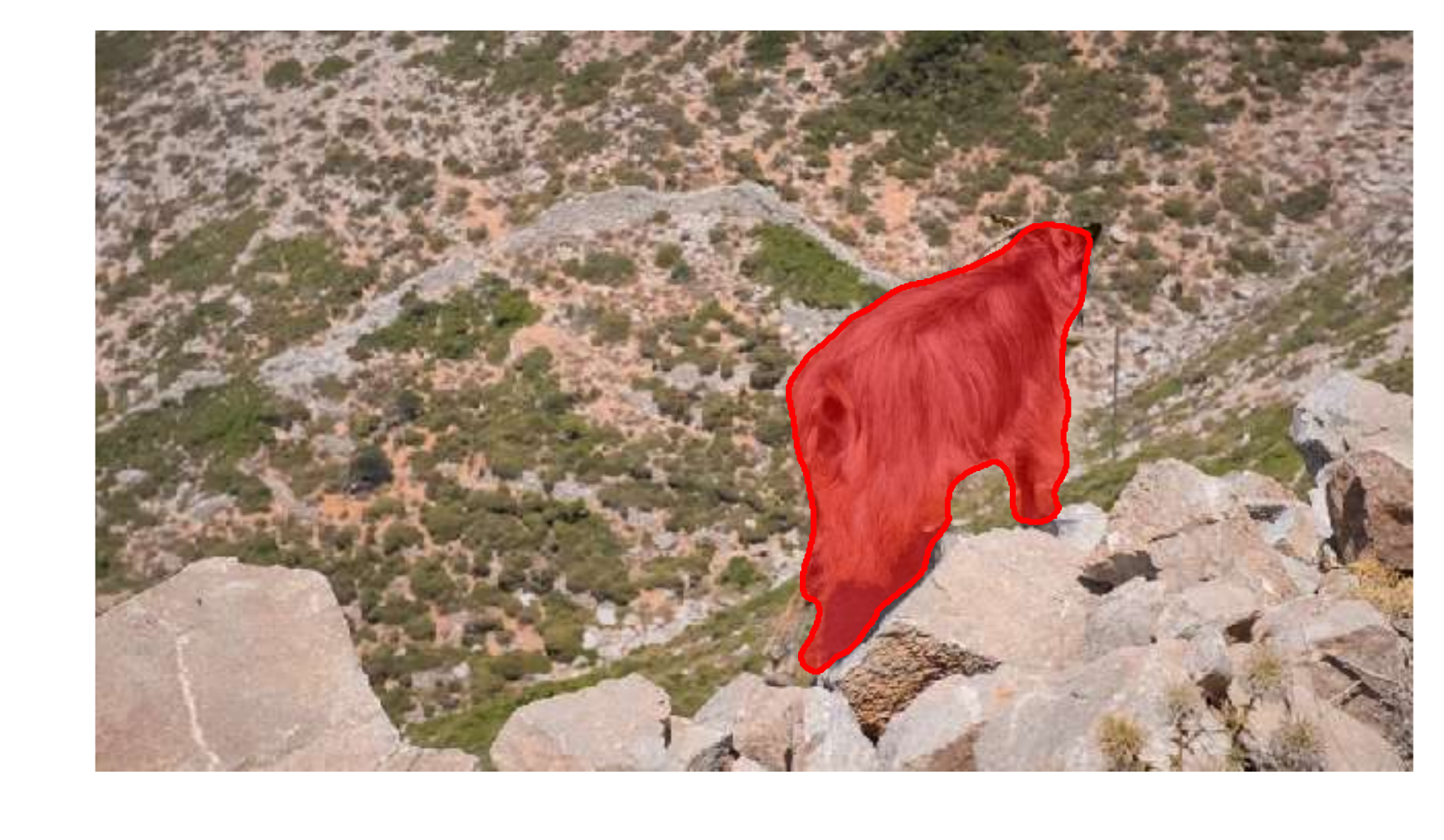}
& \includegraphics[trim={2.5cm 1cm 2.5cm 1cm},clip,width = 1.1in]{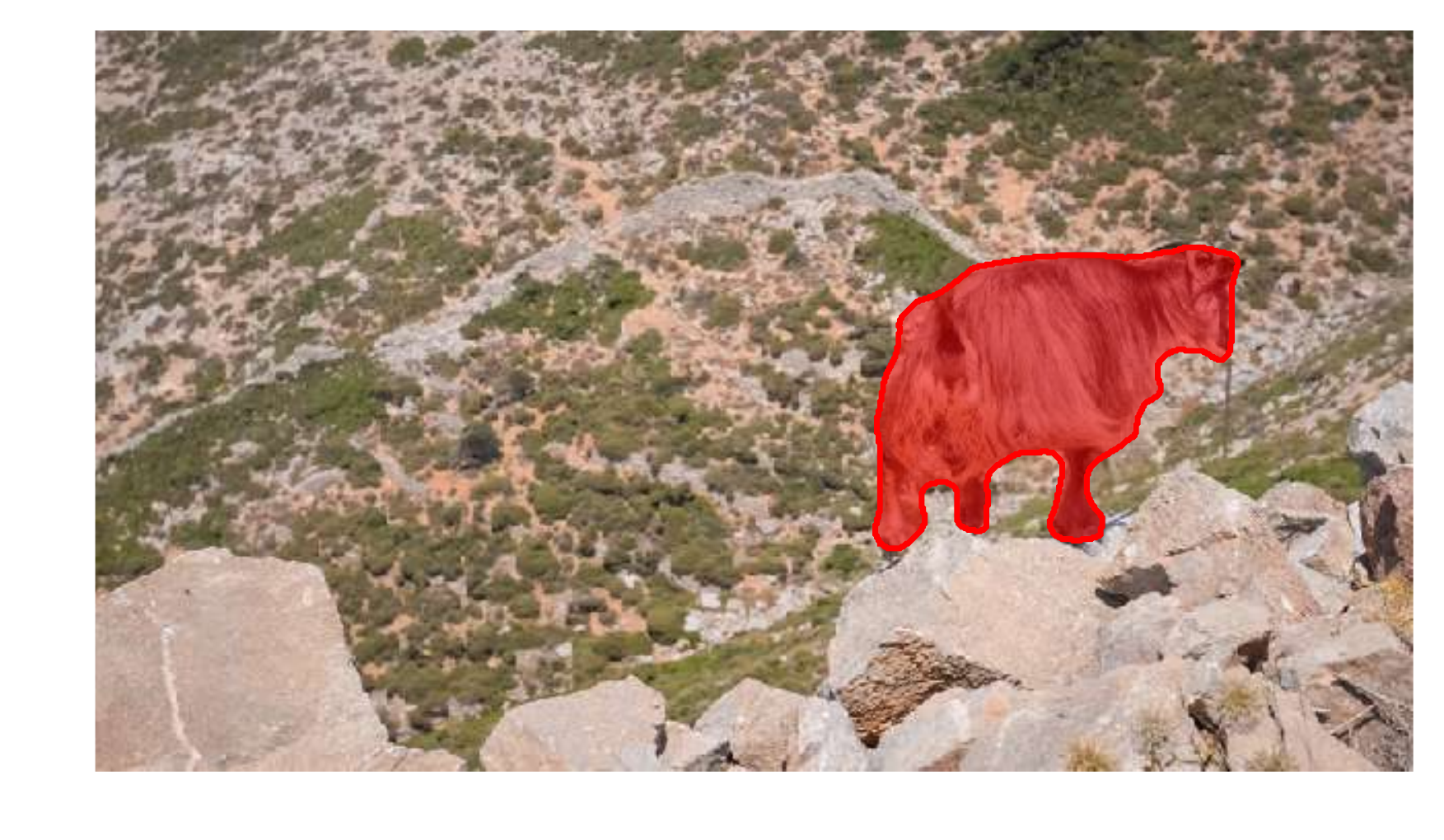}
& \includegraphics[trim={2.5cm 1cm 2.5cm 1cm},clip,width = 1.1in]{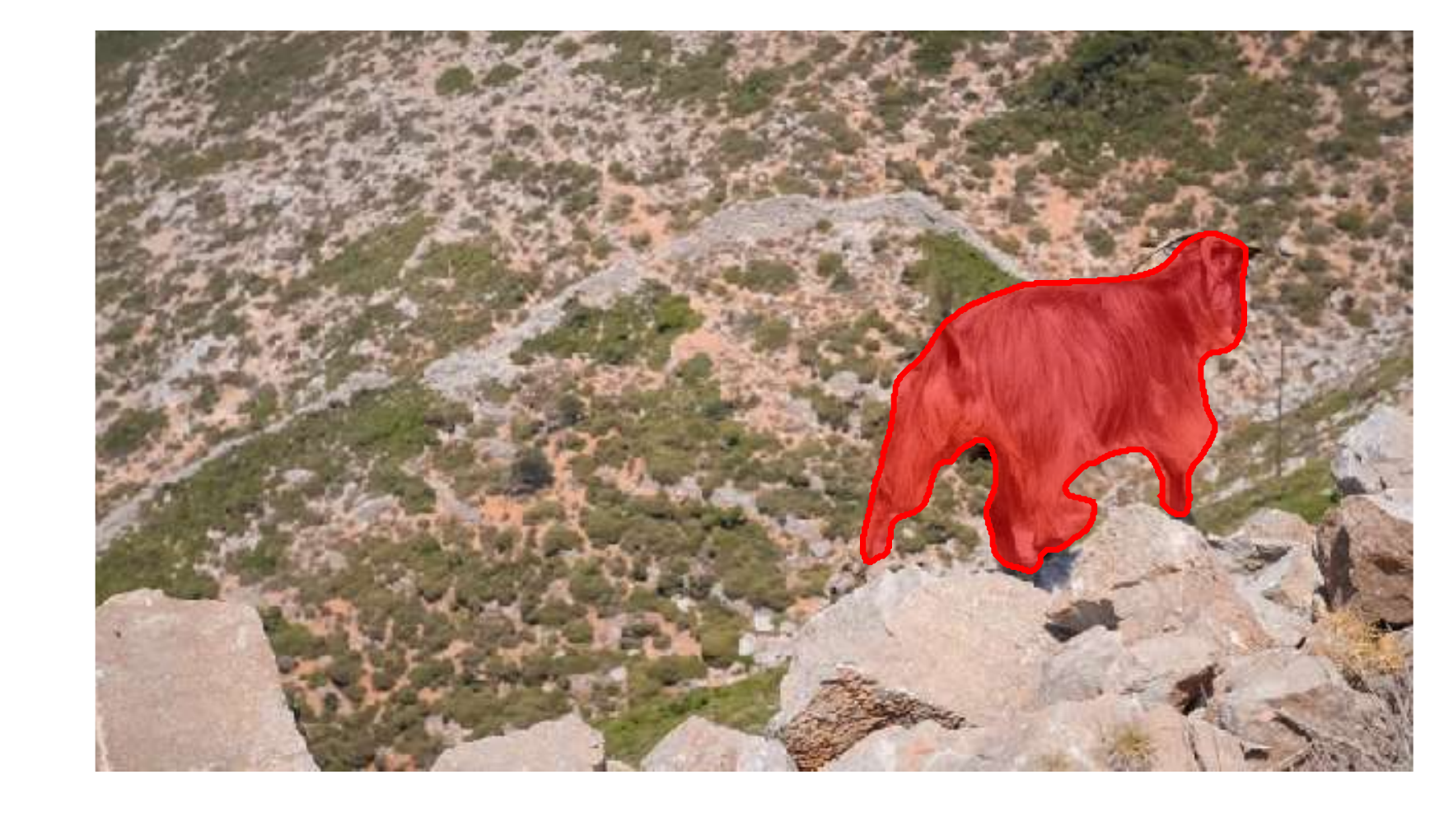}
& \includegraphics[trim={2.5cm 1cm 2.5cm 1cm},clip,width = 1.1in]{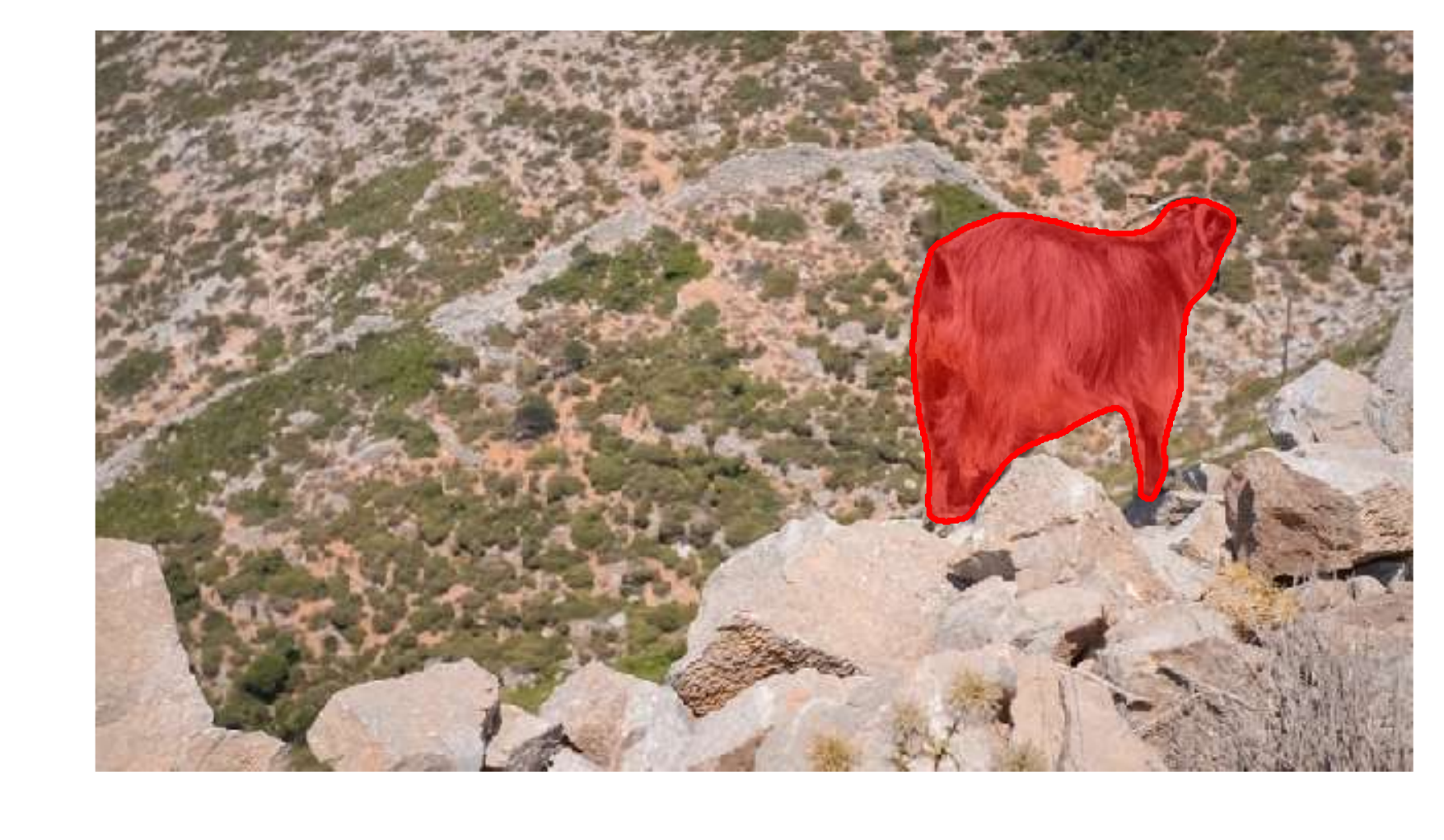}
& \includegraphics[trim={2.5cm 1cm 2.5cm 1cm},clip,width = 1.1in]{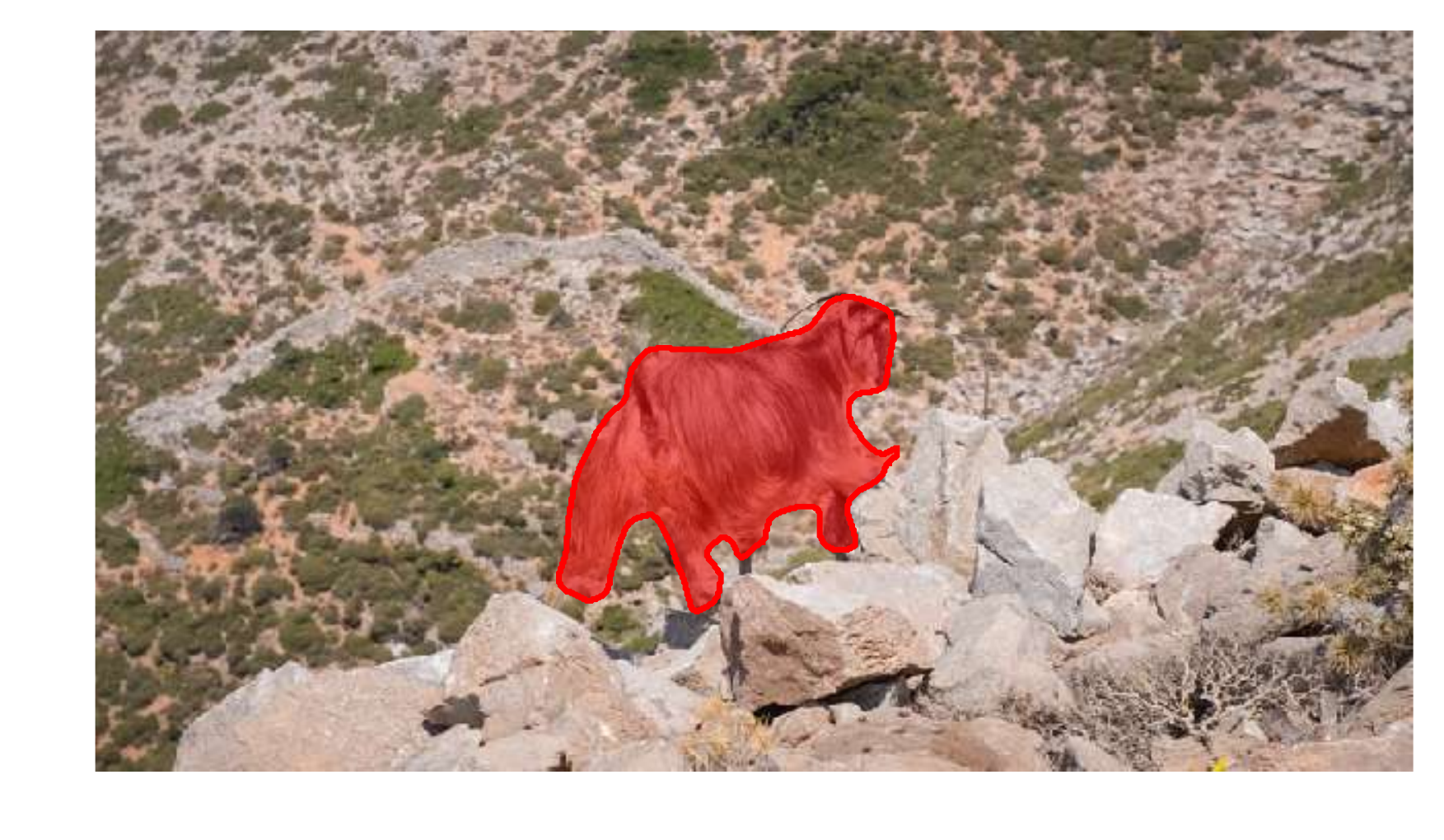}
\\

\mbox{\rotatebox[x=-0.55cm]{90}{\small{Libby}}}
\includegraphics[trim={2.5cm 1cm 2.5cm 1cm},clip,width = 1.1in]{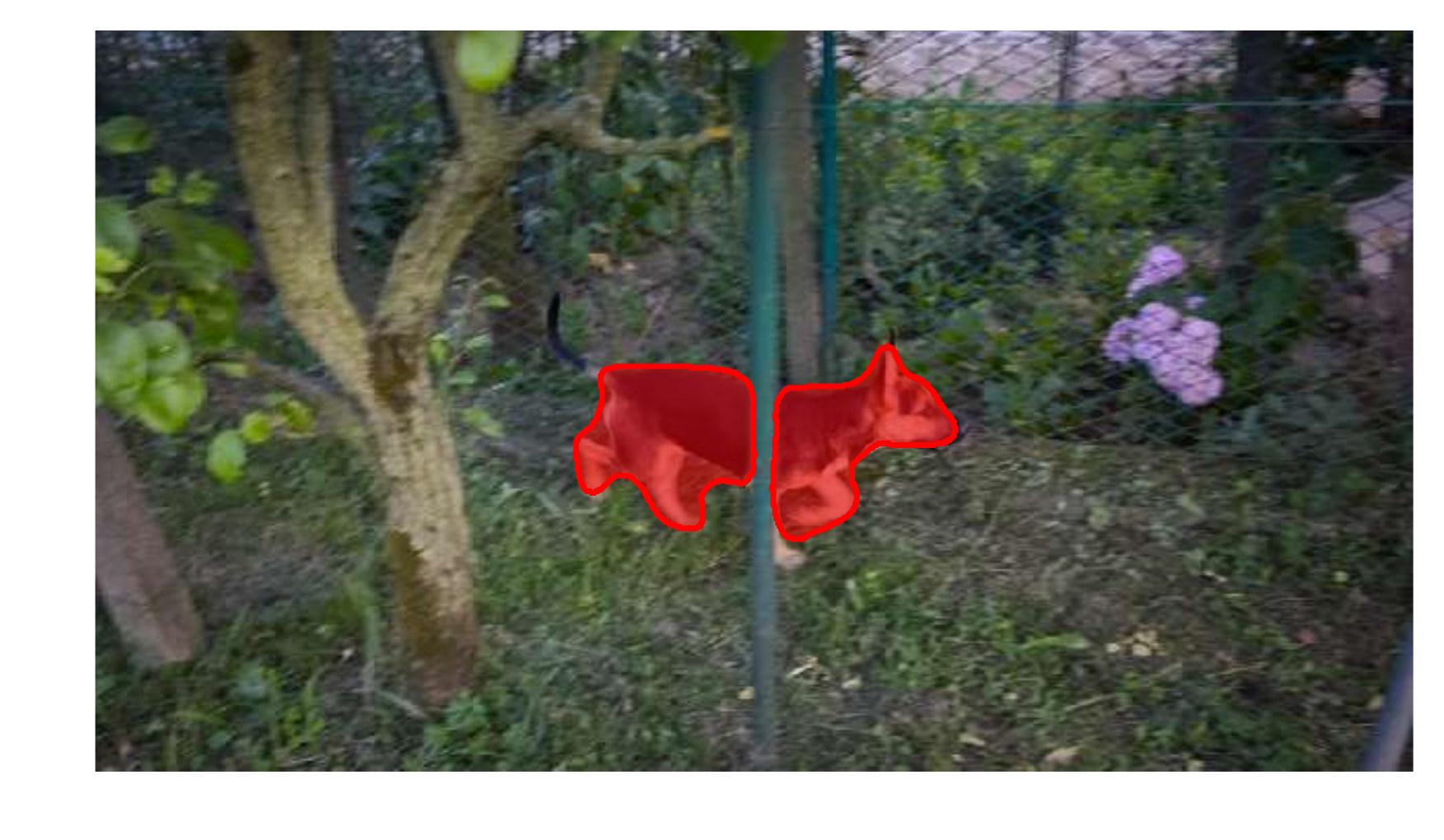}
&\includegraphics[trim={2.5cm 1cm 2.5cm 1cm},clip,width = 1.1in]{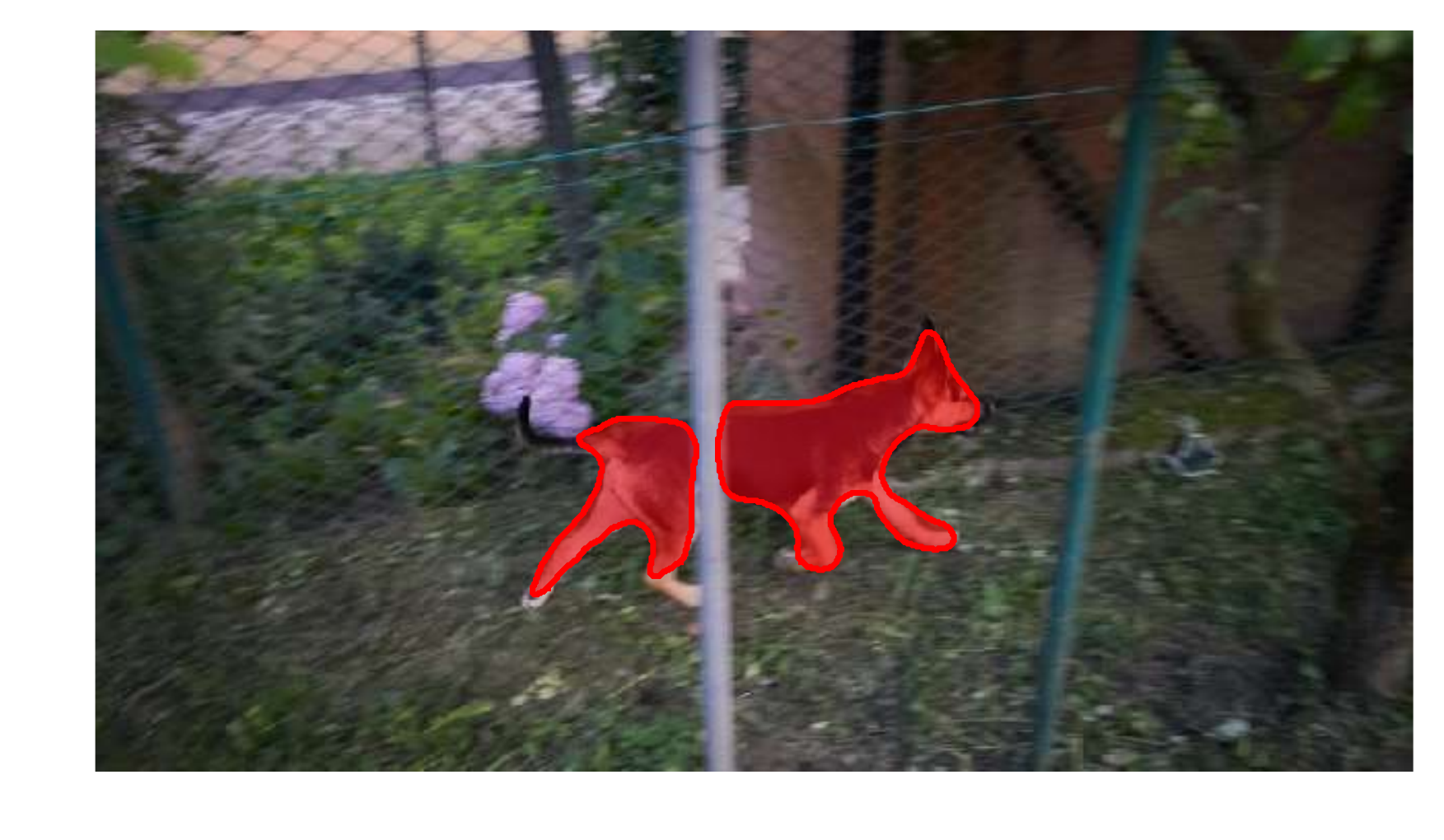}
& \includegraphics[trim={2.5cm 1cm 2.5cm 1cm},clip,width = 1.1in]{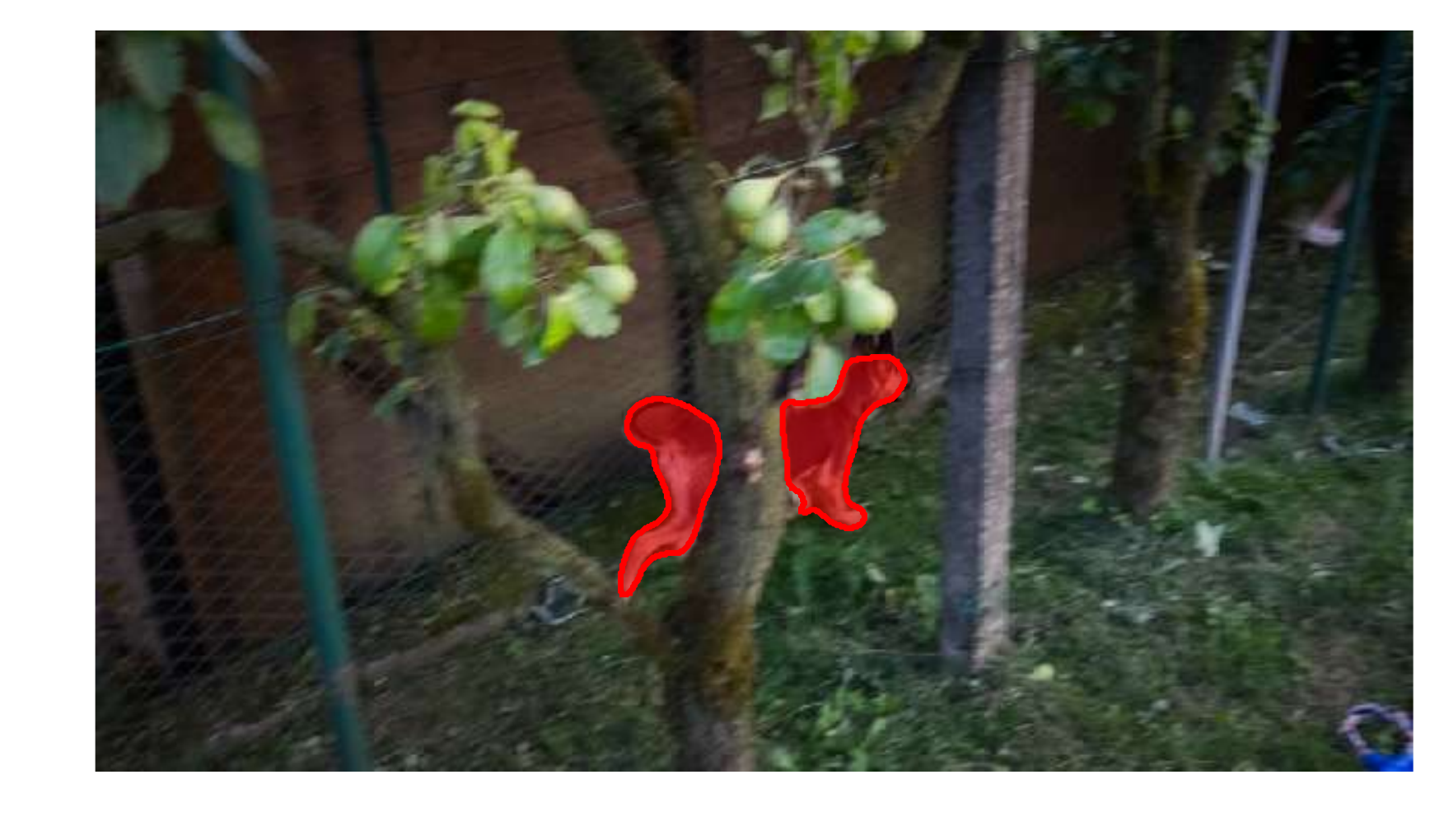}
& \includegraphics[trim={2.5cm 1cm 2.5cm 1cm},clip,width = 1.1in]{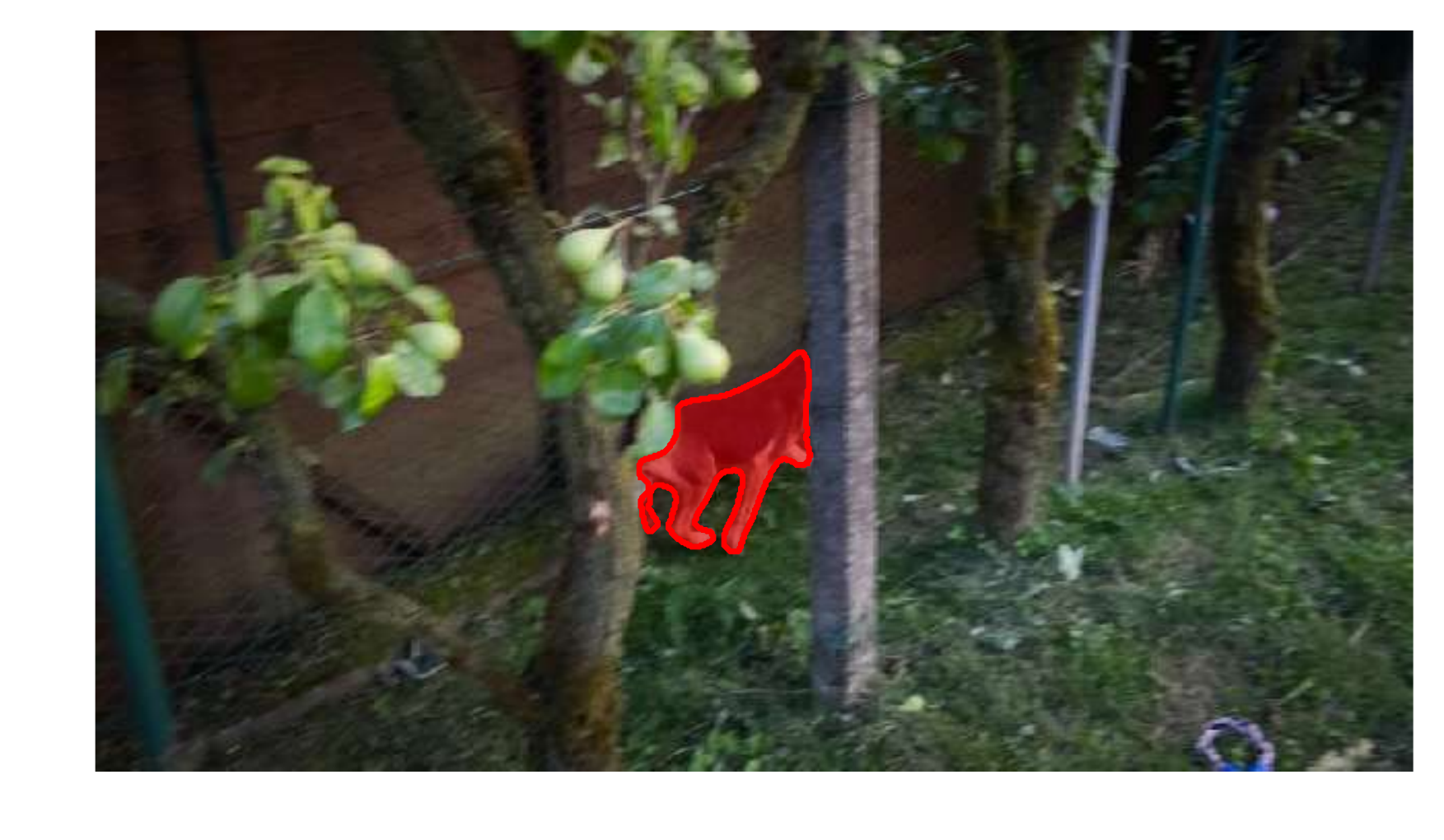}
& \includegraphics[trim={2.5cm 1cm 2.5cm 1cm},clip,width = 1.1in]{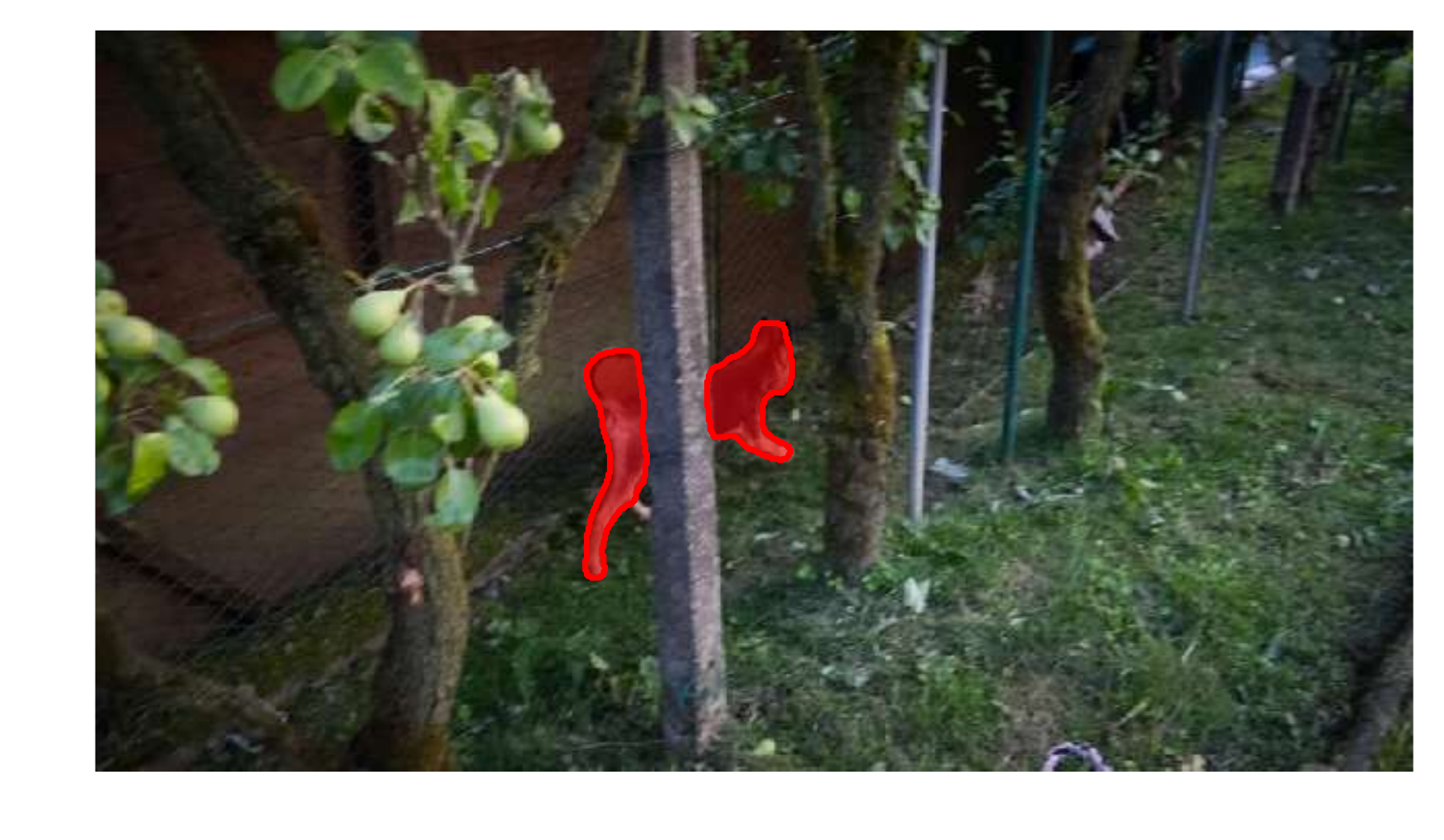}
& \includegraphics[trim={2.5cm 1cm 2.5cm 1cm},clip,width = 1.1in]{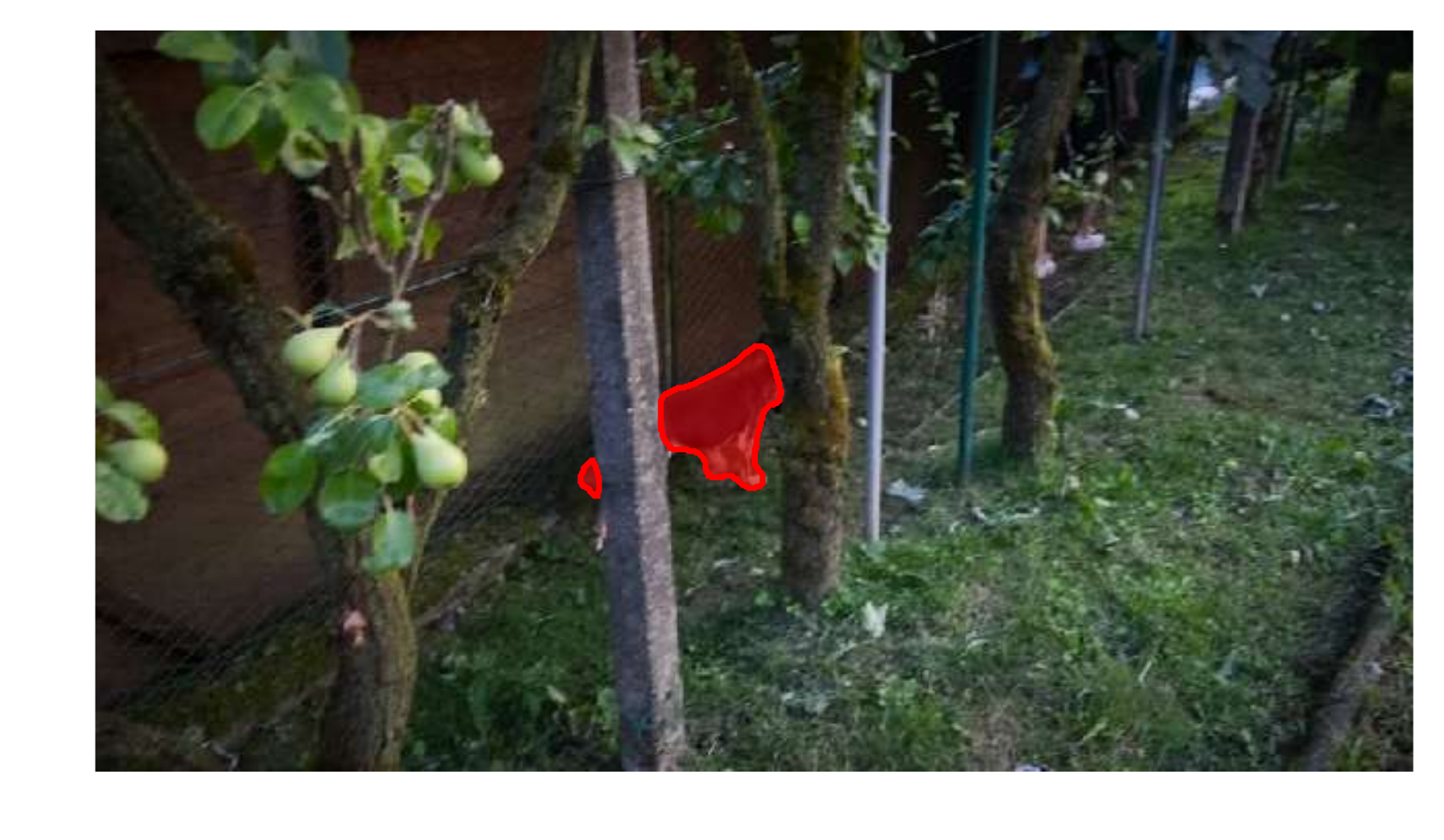}
\\

\mbox{\rotatebox[x=-0.cm]{90}{\scriptsize{motocross-jump}}}
\includegraphics[trim={2.5cm 1cm 2.5cm 1cm},clip,width = 1.1in]{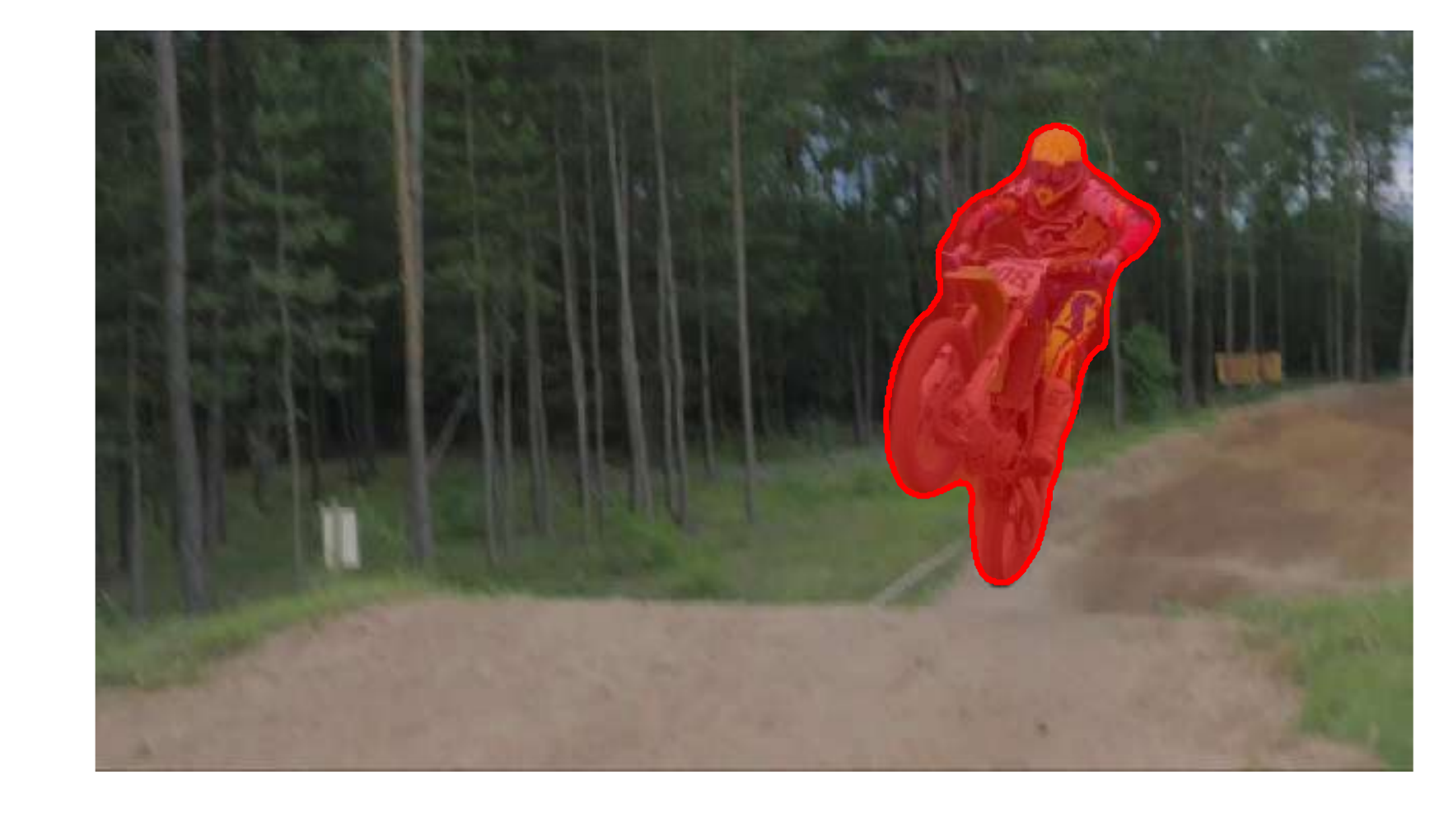}
&\includegraphics[trim={2.5cm 1cm 2.5cm 1cm},clip,width = 1.1in]{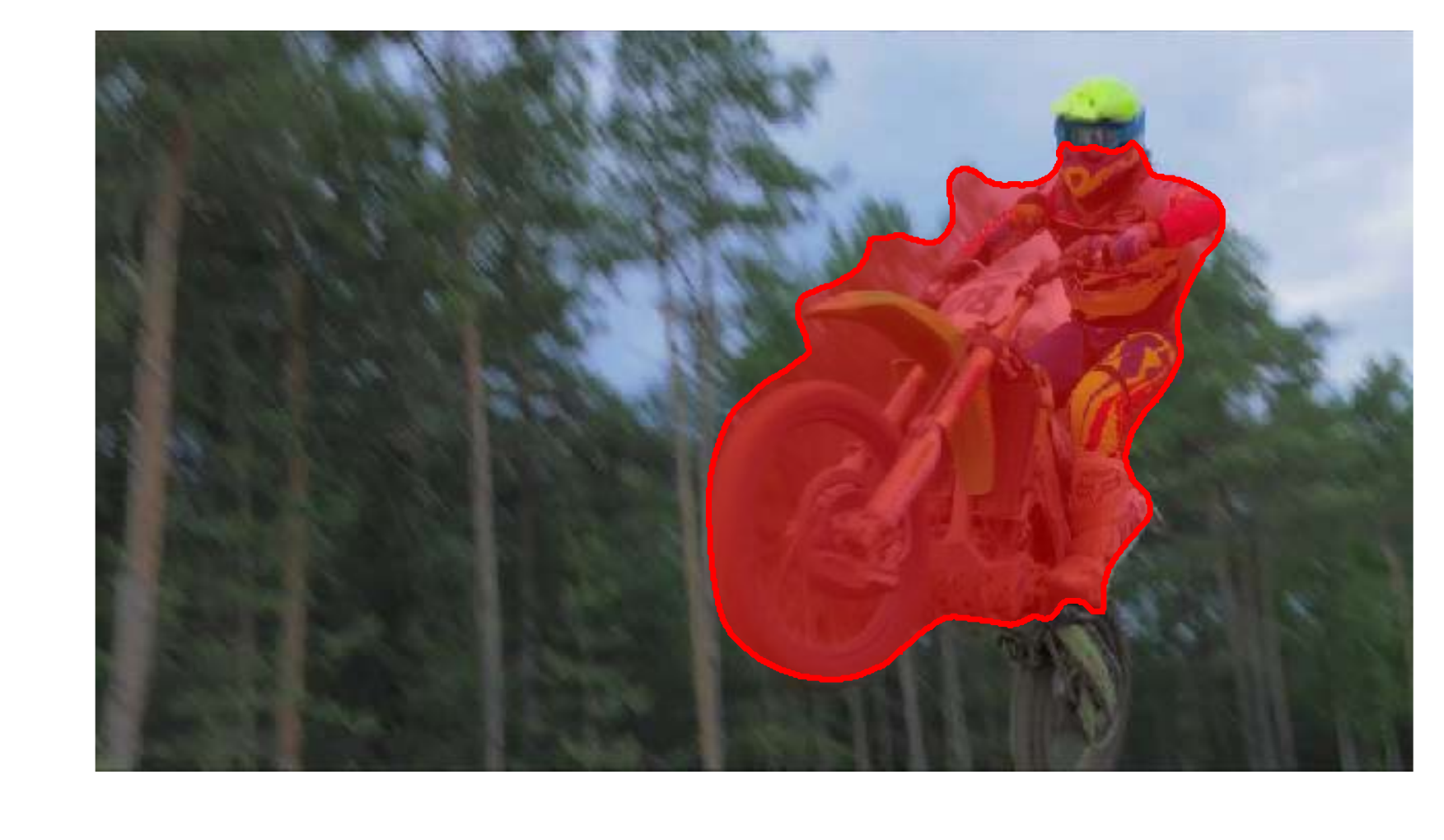}
& \includegraphics[trim={2.5cm 1cm 2.5cm 1cm},clip,width = 1.1in]{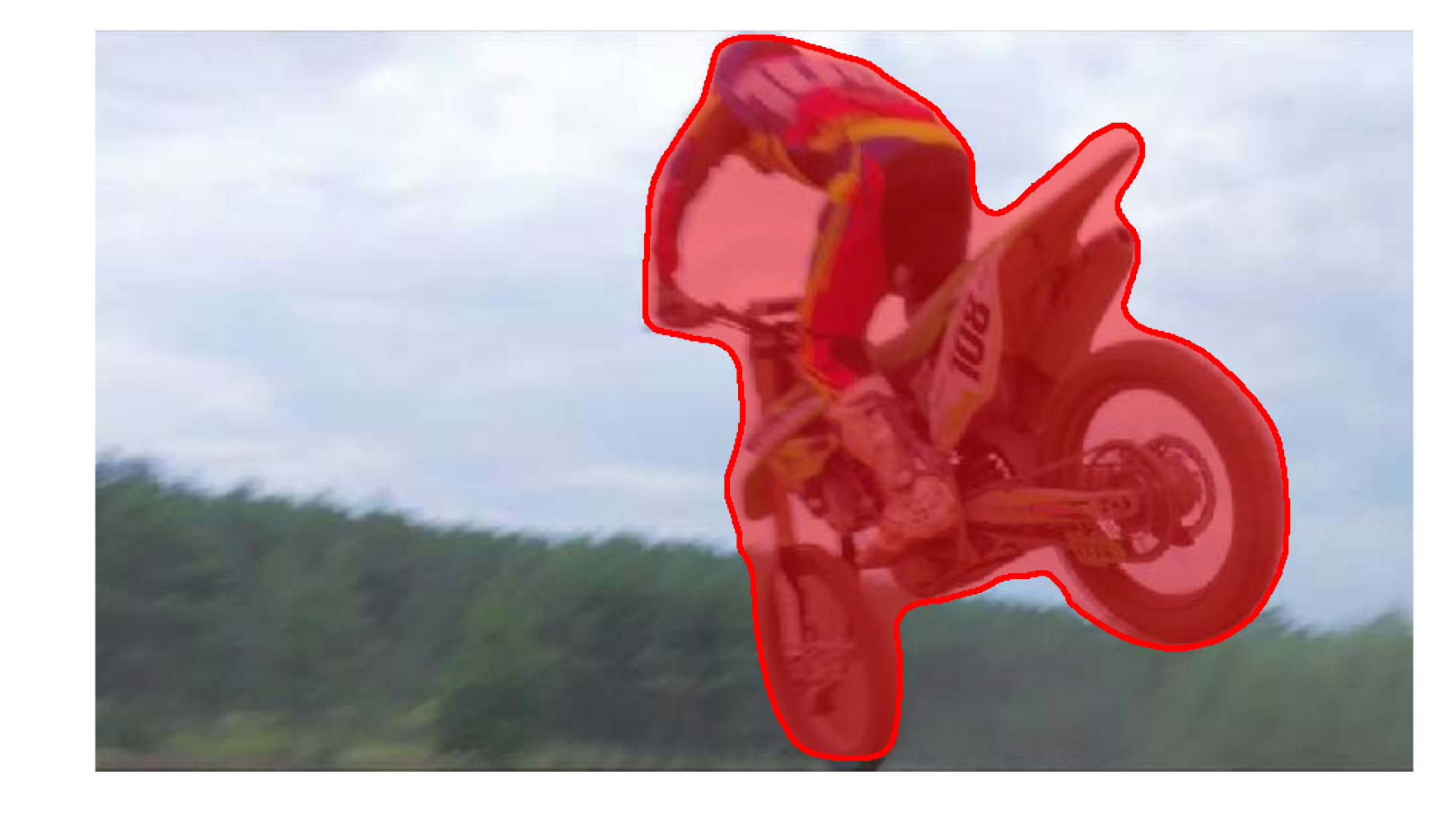}
& \includegraphics[trim={2.5cm 1cm 2.5cm 1cm},clip,width = 1.1in]{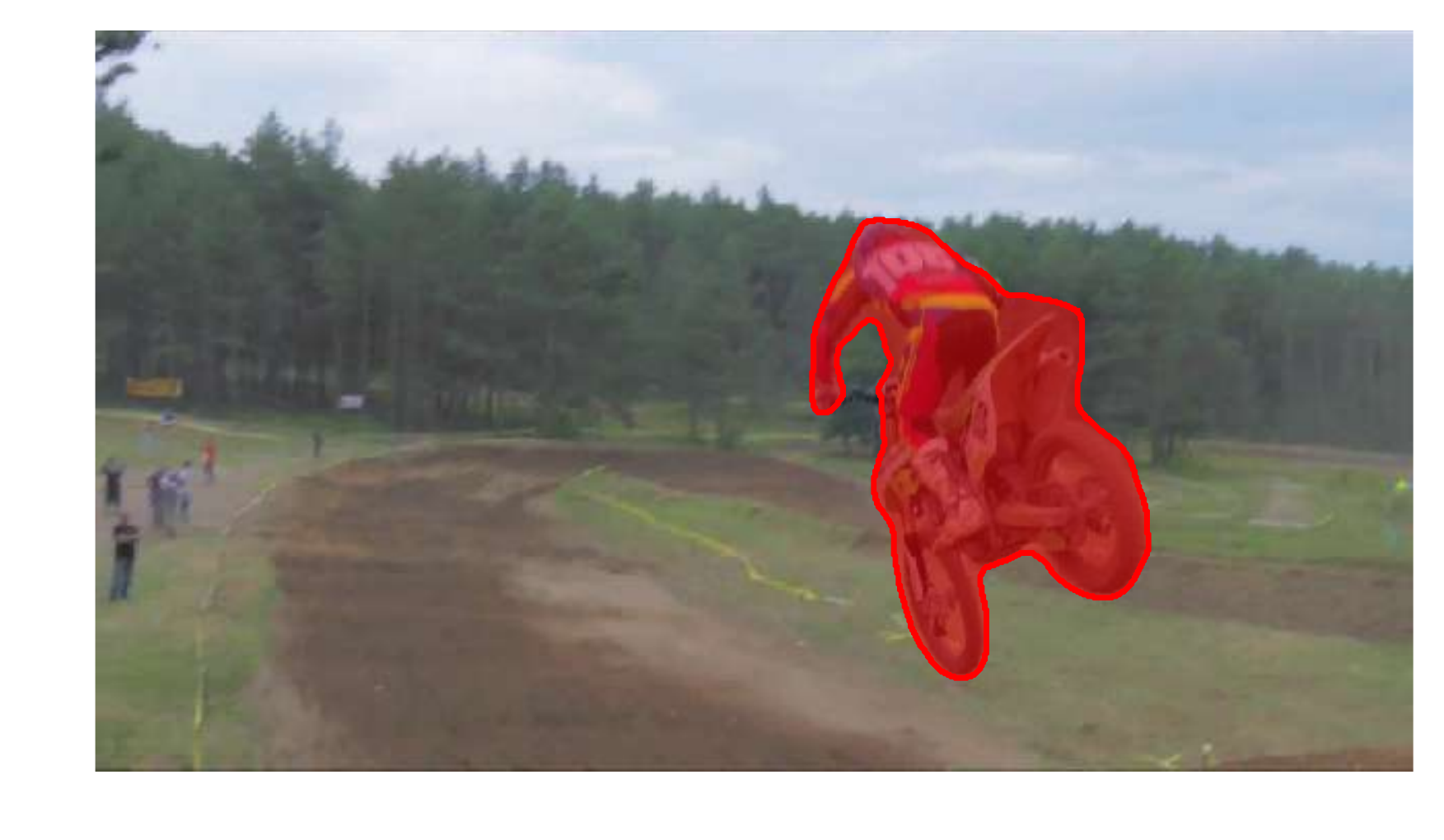}
& \includegraphics[trim={2.5cm 1cm 2.5cm 1cm},clip,width = 1.1in]{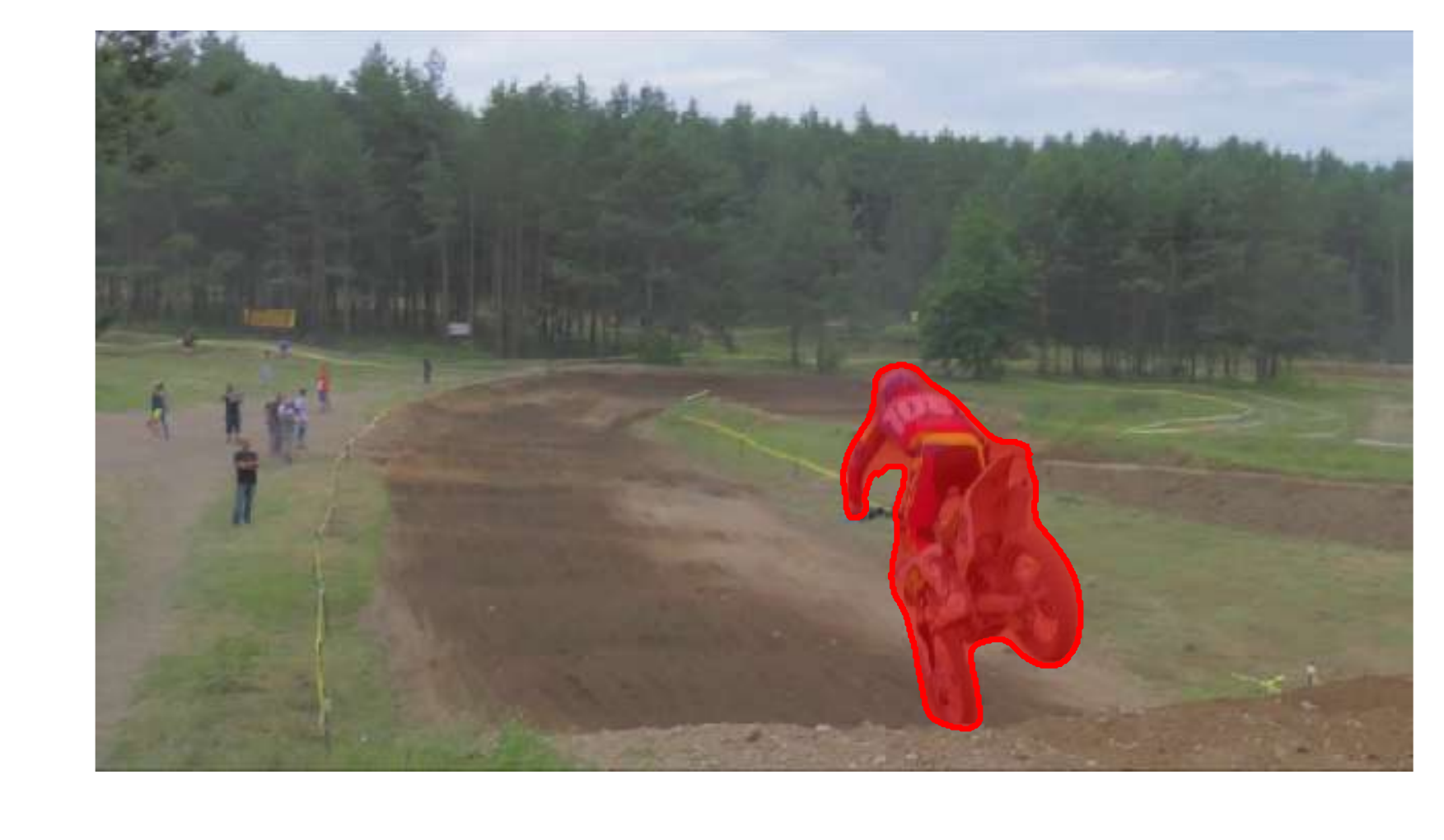}
& \includegraphics[trim={2.5cm 1cm 2.5cm 1cm},clip,width = 1.1in]{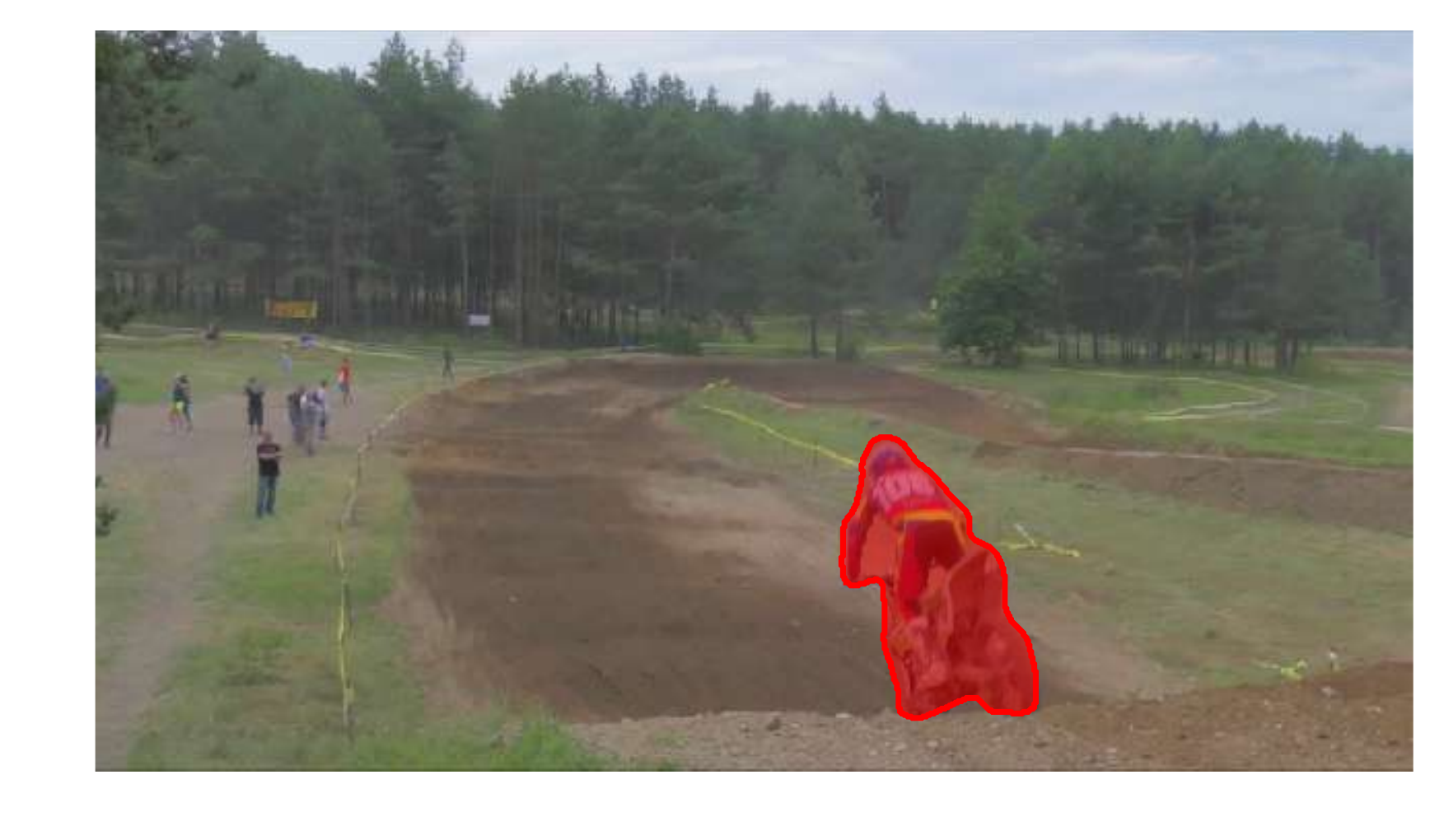}
\\

\mbox{\rotatebox[x=-0.55cm]{90}{\small{parkour}}}
\includegraphics[trim={2.5cm 1cm 2.5cm 0.5cm},clip,width = 1.1in]{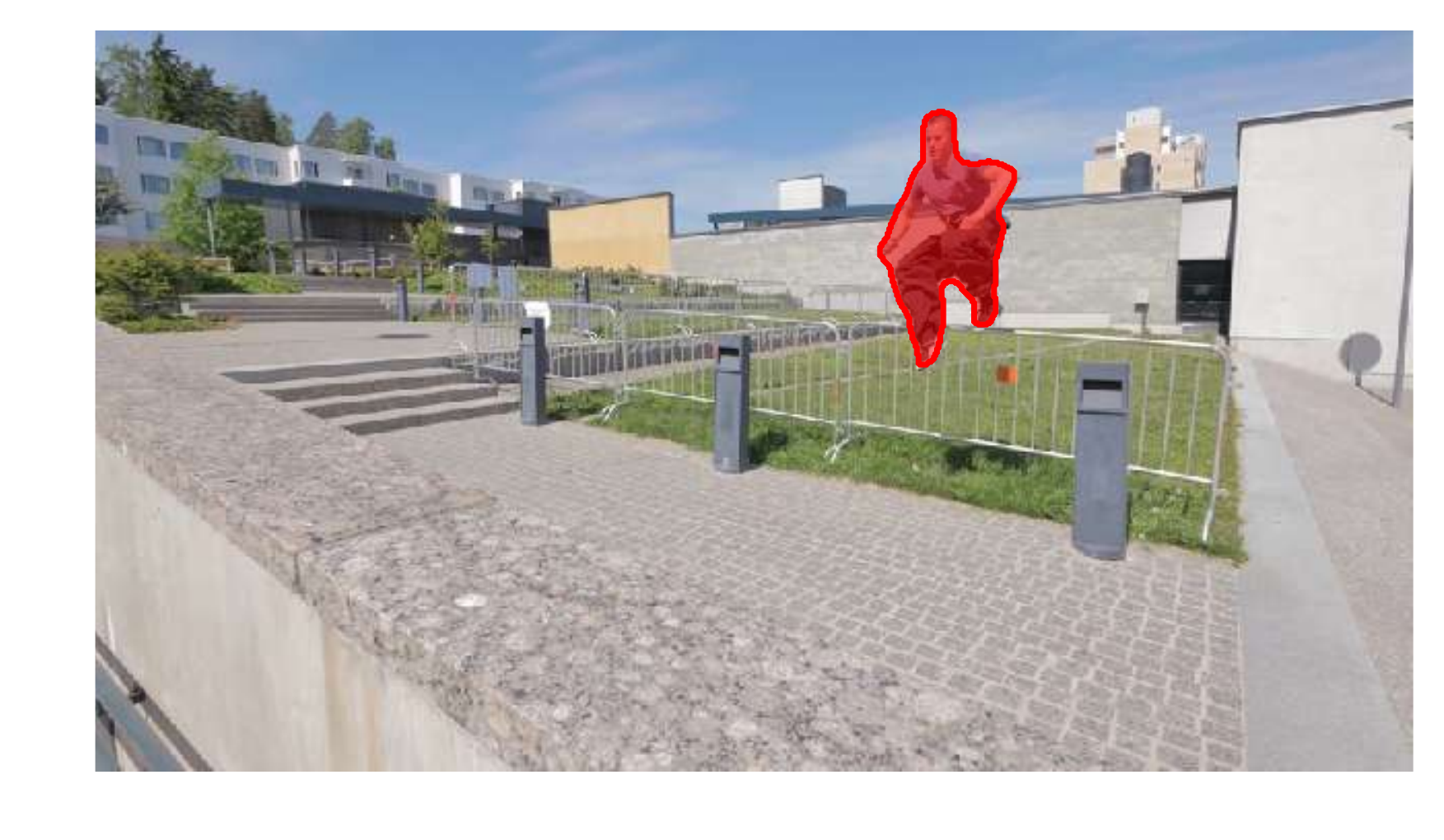}
&\includegraphics[trim={2.5cm 1cm 2.5cm 0.5cm},clip,width = 1.1in]{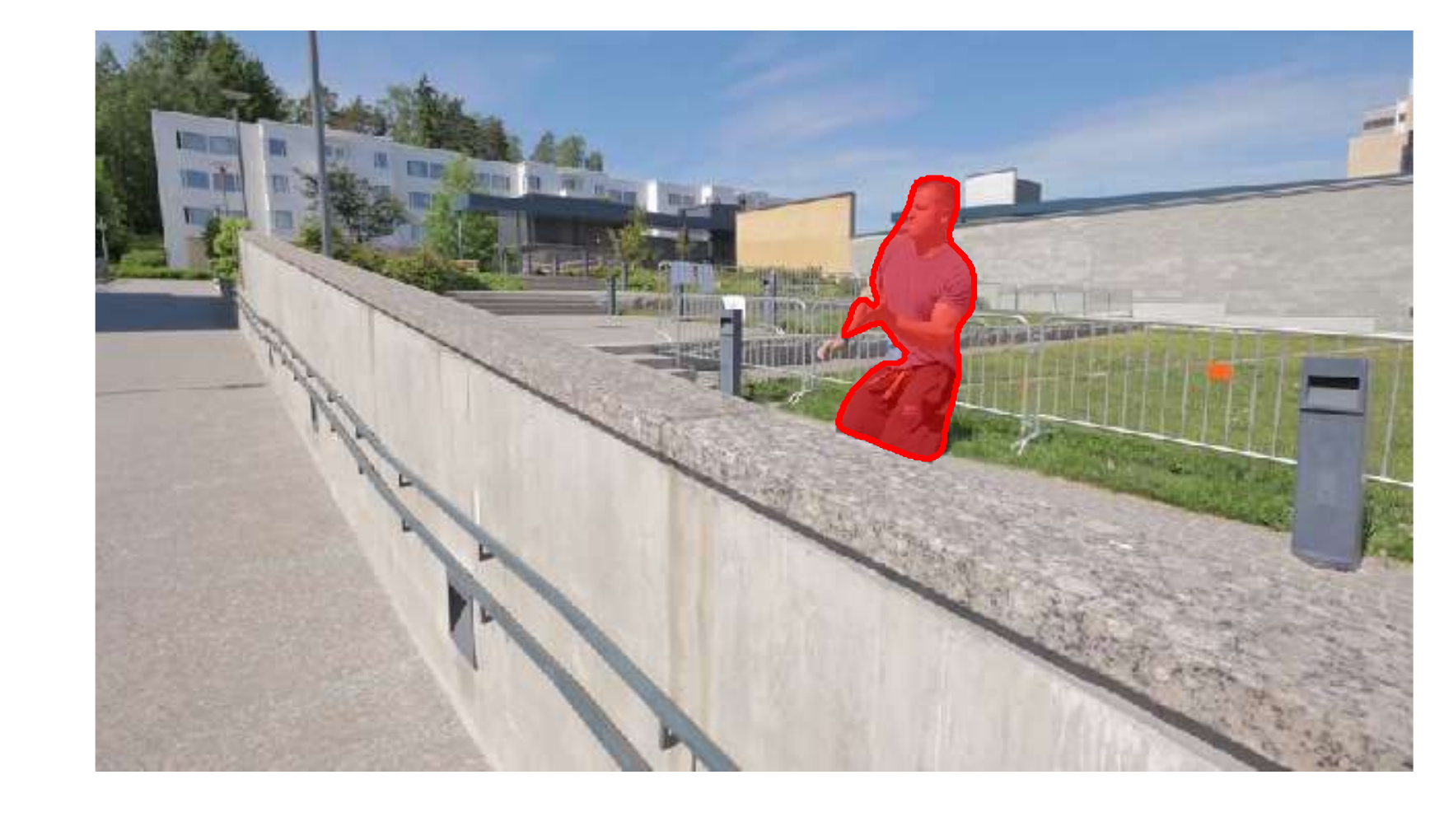}
& \includegraphics[trim={2.5cm 1cm 2.5cm 0.5cm},clip,width = 1.1in]{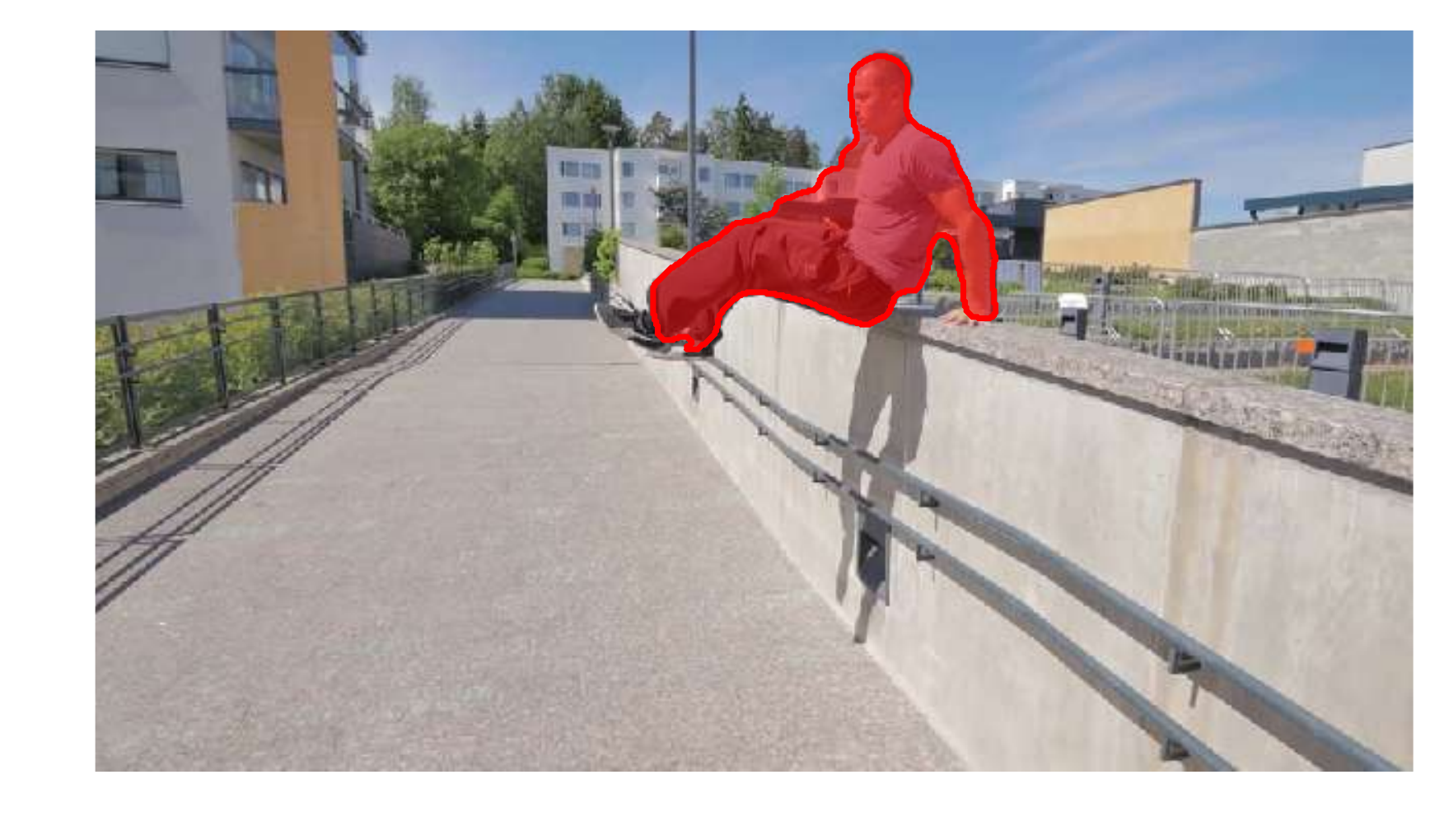}
& \includegraphics[trim={2.5cm 1cm 2.5cm 0.5cm},clip,width = 1.1in]{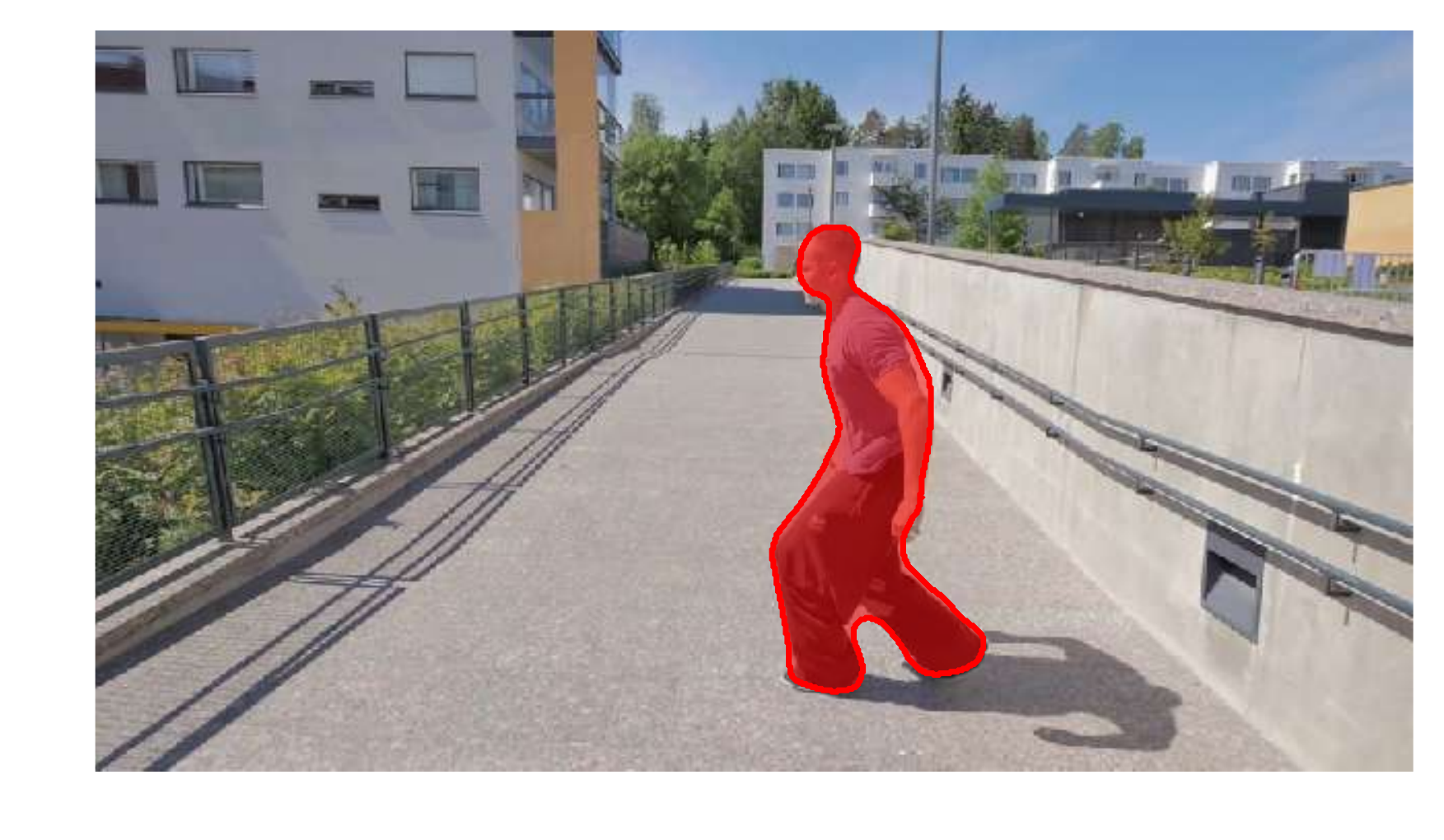}
& \includegraphics[trim={2.5cm 1cm 2.5cm 0.5cm},clip,width = 1.1in]{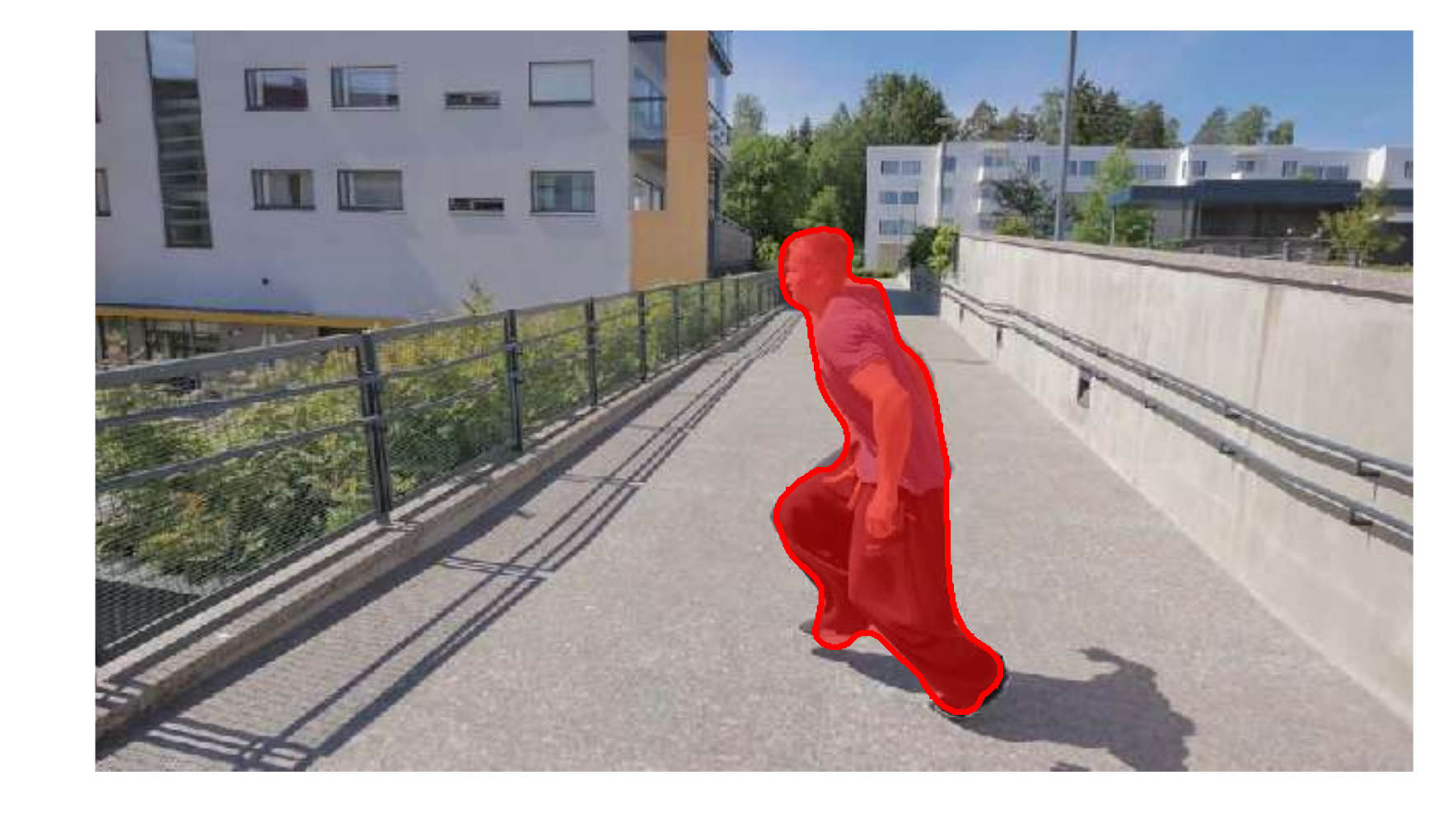}
& \includegraphics[trim={2.5cm 1cm 2.5cm 0.5cm},clip,width = 1.1in]{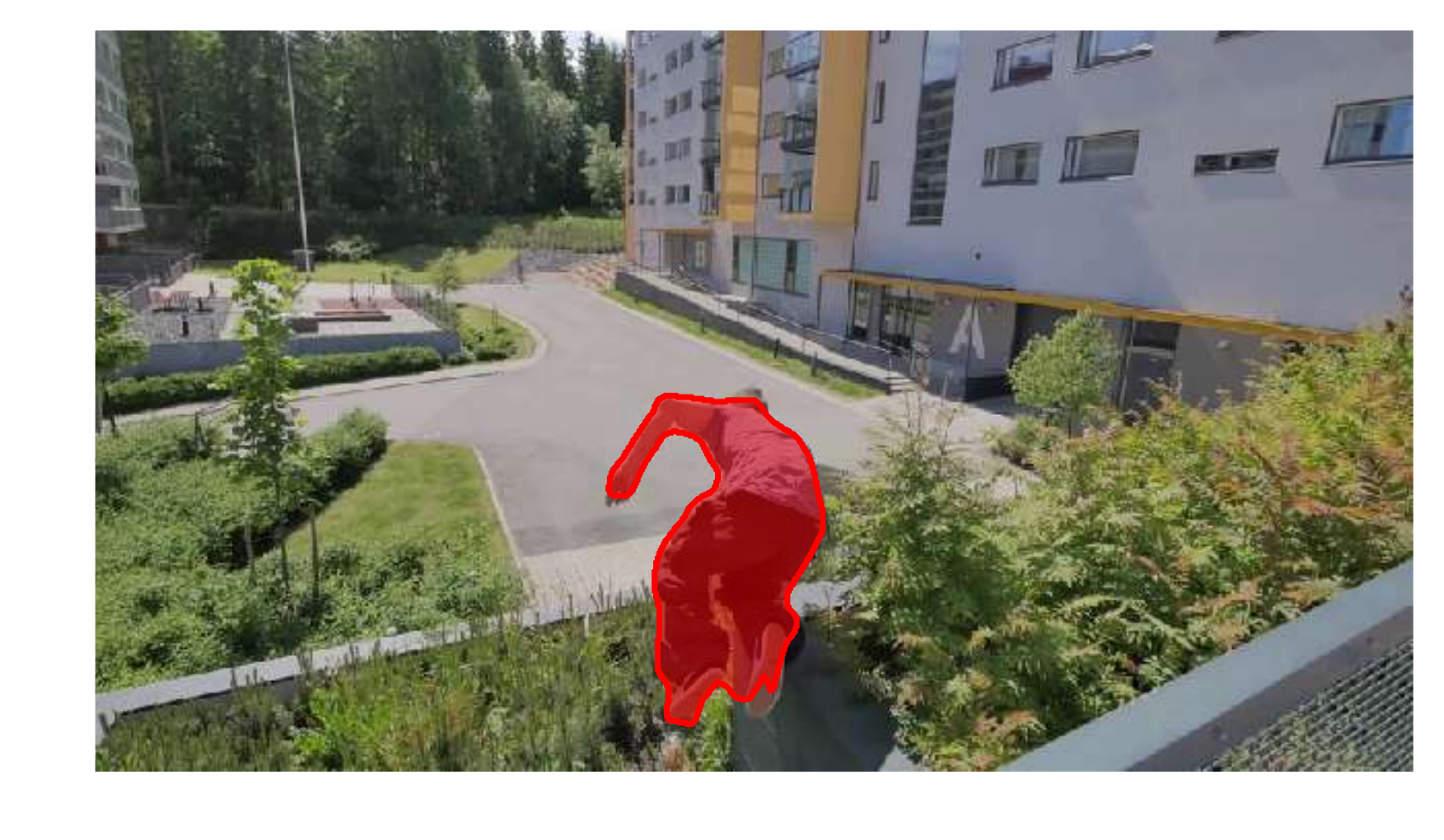}
\\
\mbox{\rotatebox[x=-0.0cm]{90}{\small{Gold-Fish}}}
\includegraphics[trim={2.5cm 1cm 2.5cm 1cm},clip,width = 1.1in]{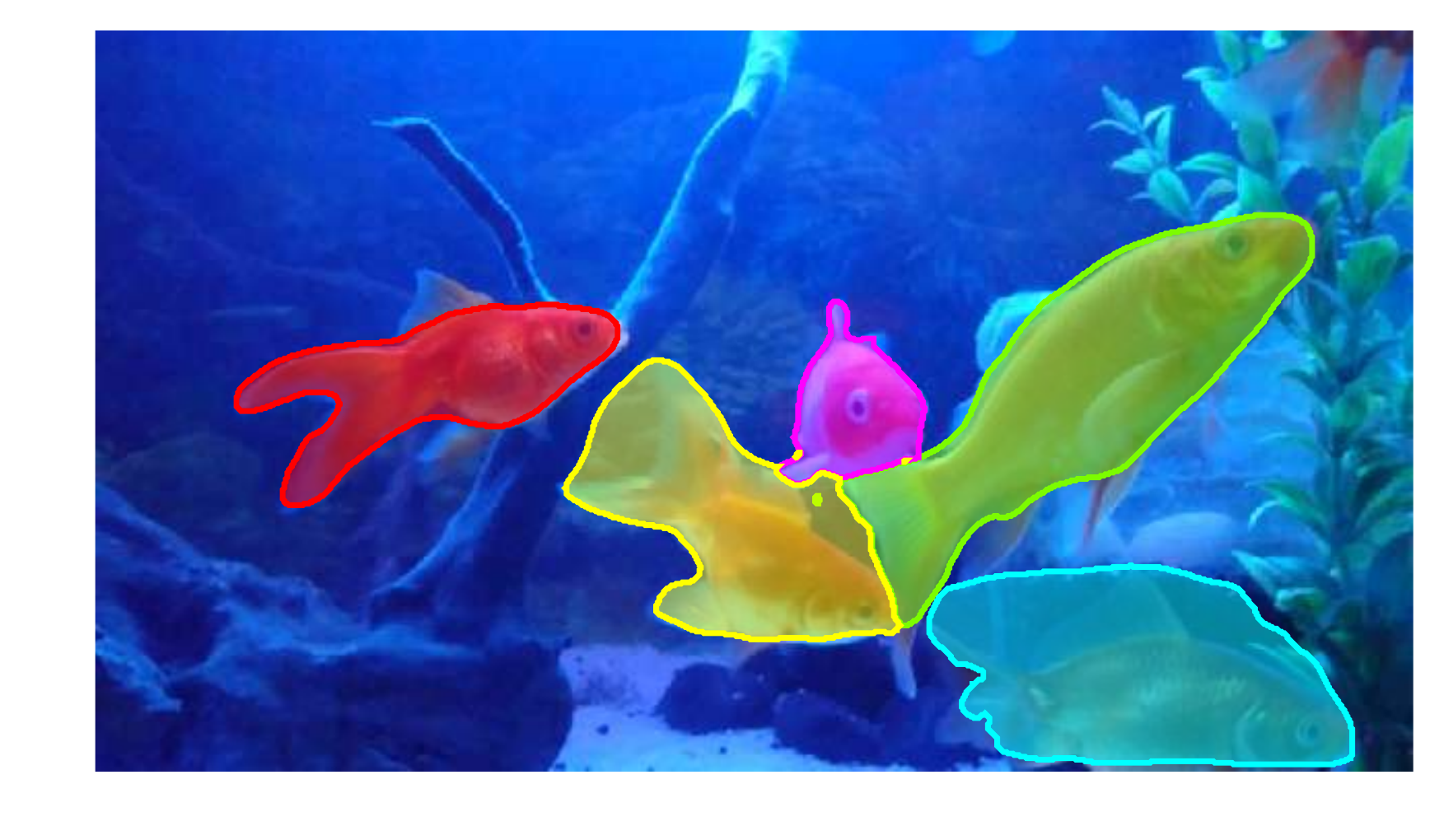}
&\includegraphics[trim={2.5cm 1cm 2.5cm 1cm},clip,width = 1.1in]{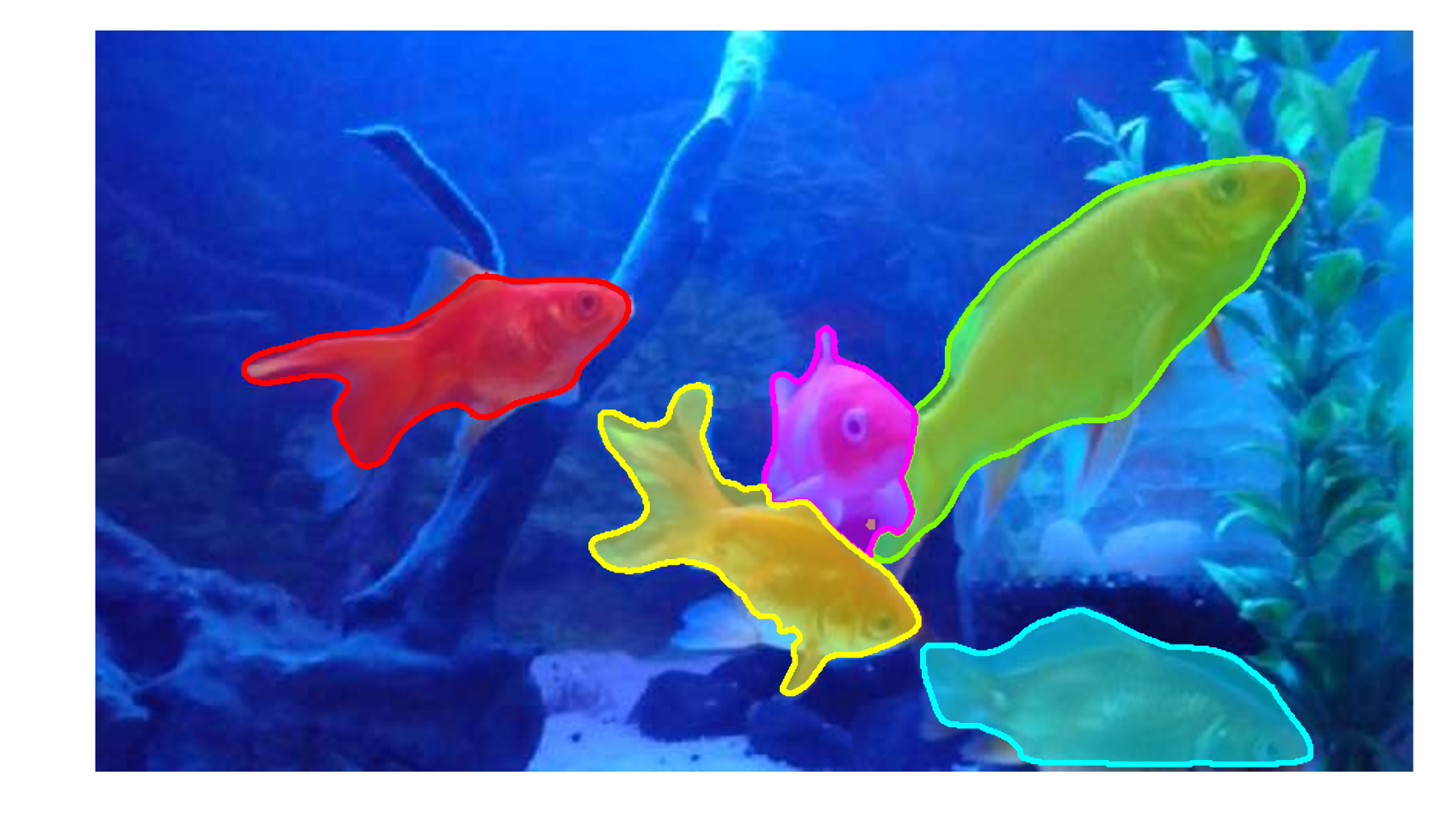}
& \includegraphics[trim={2.5cm 1cm 2.5cm 1cm},clip,width = 1.1in]{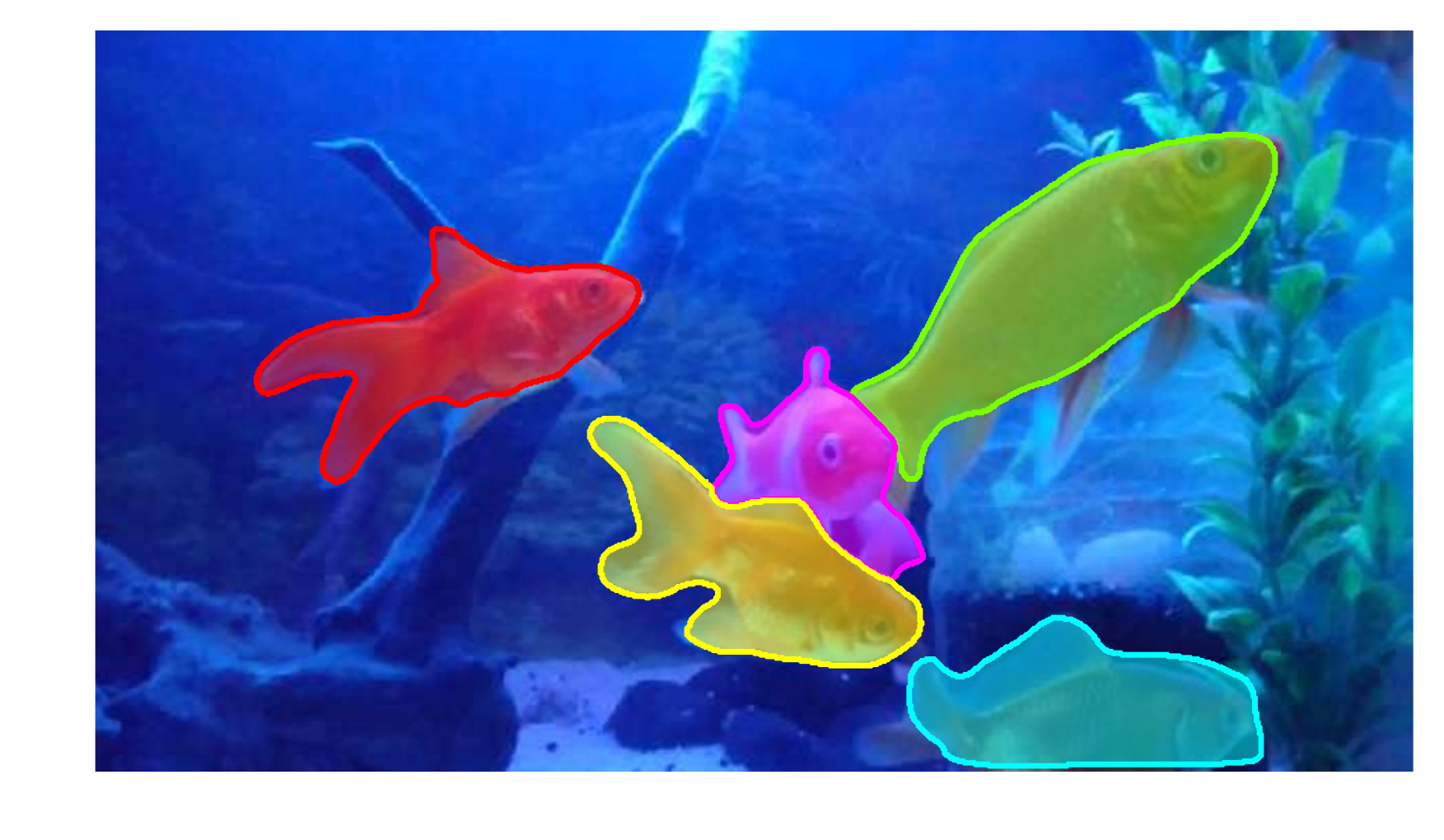}
& \includegraphics[trim={2.5cm 1cm 2.5cm 1cm},clip,width = 1.1in]{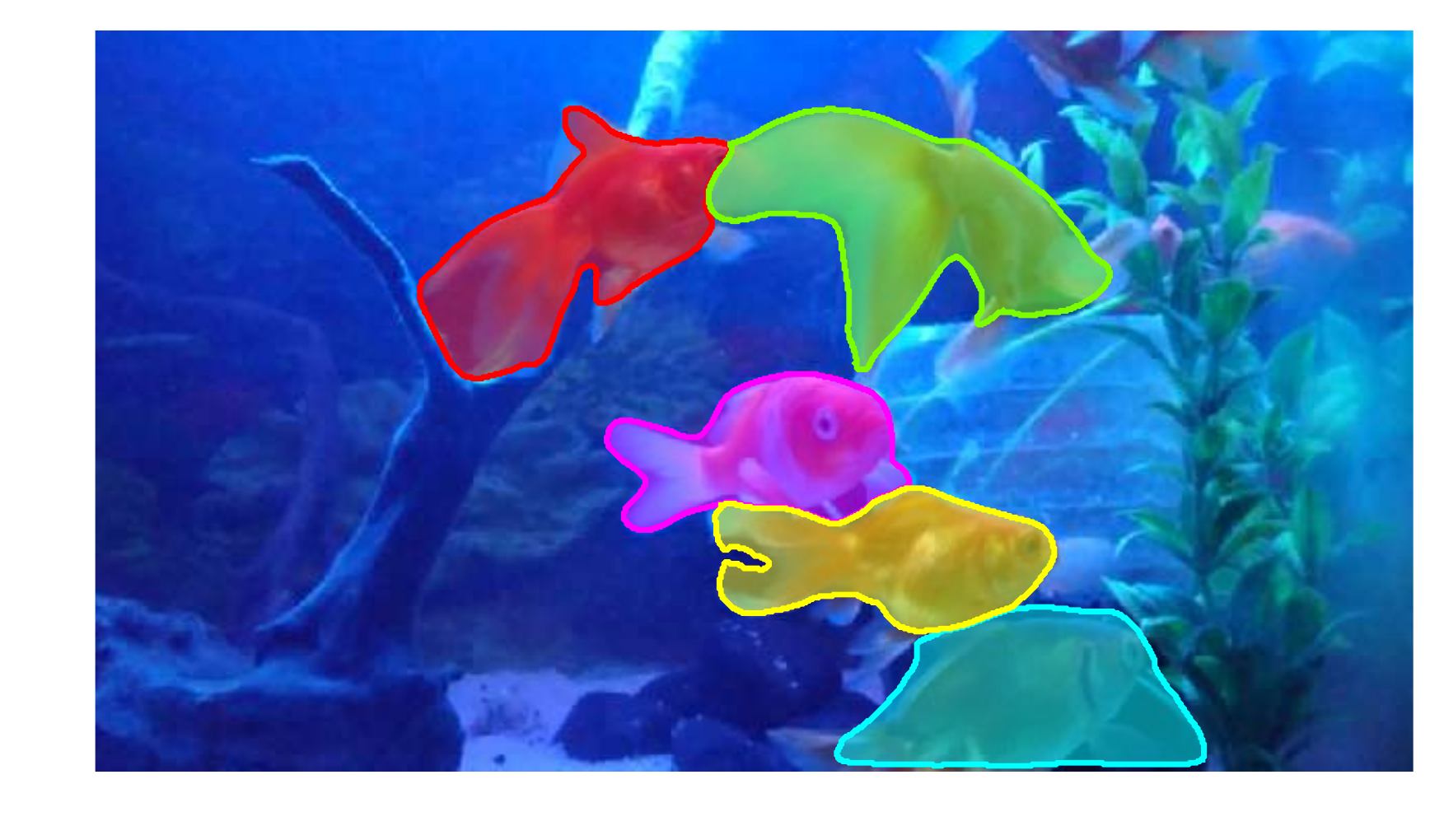}
& \includegraphics[trim={2.5cm 1cm 2.5cm 1cm},clip,width = 1.1in]{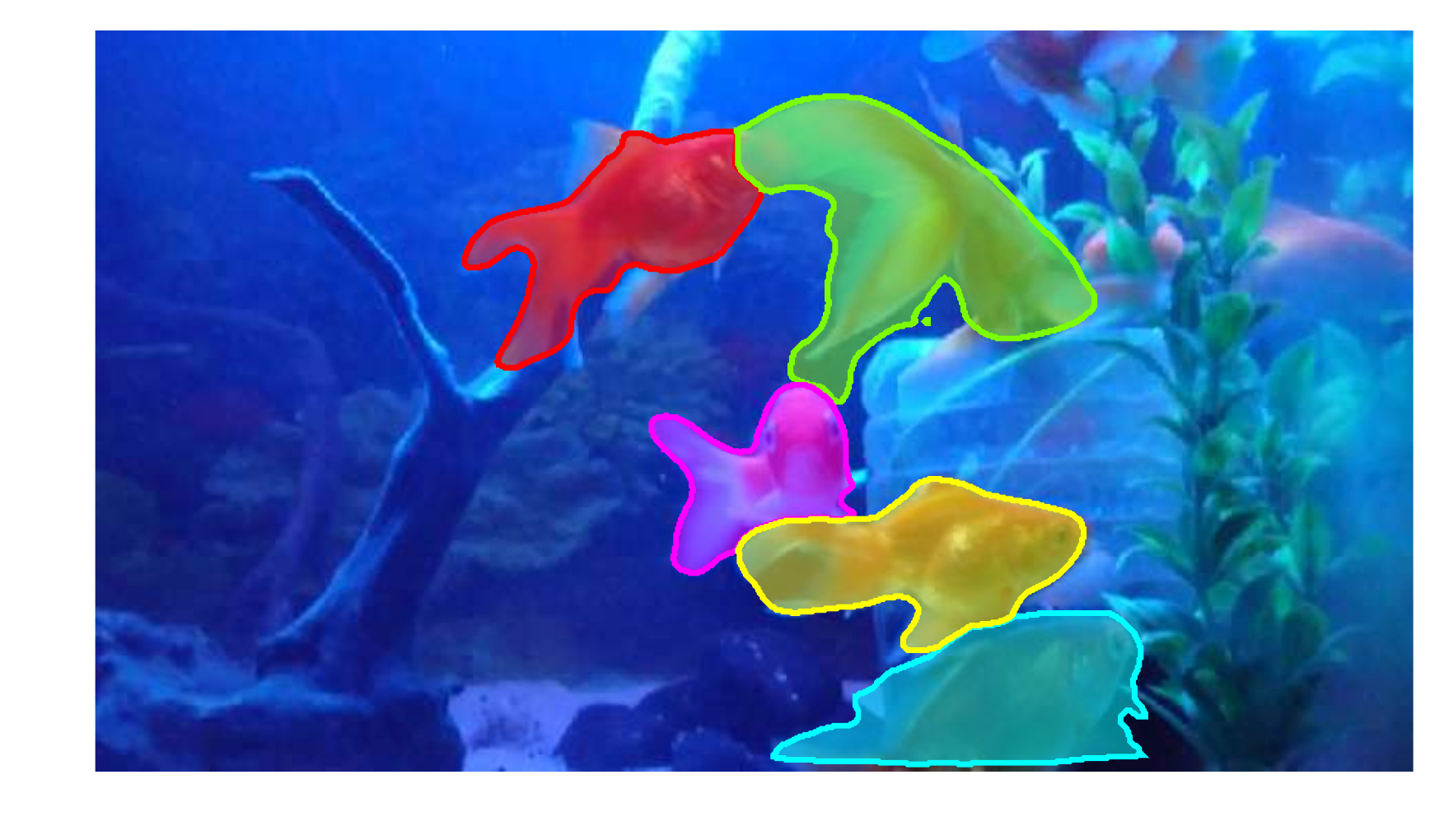}
& \includegraphics[trim={2.5cm 1cm 2.5cm 1cm},clip,width = 1.1in]{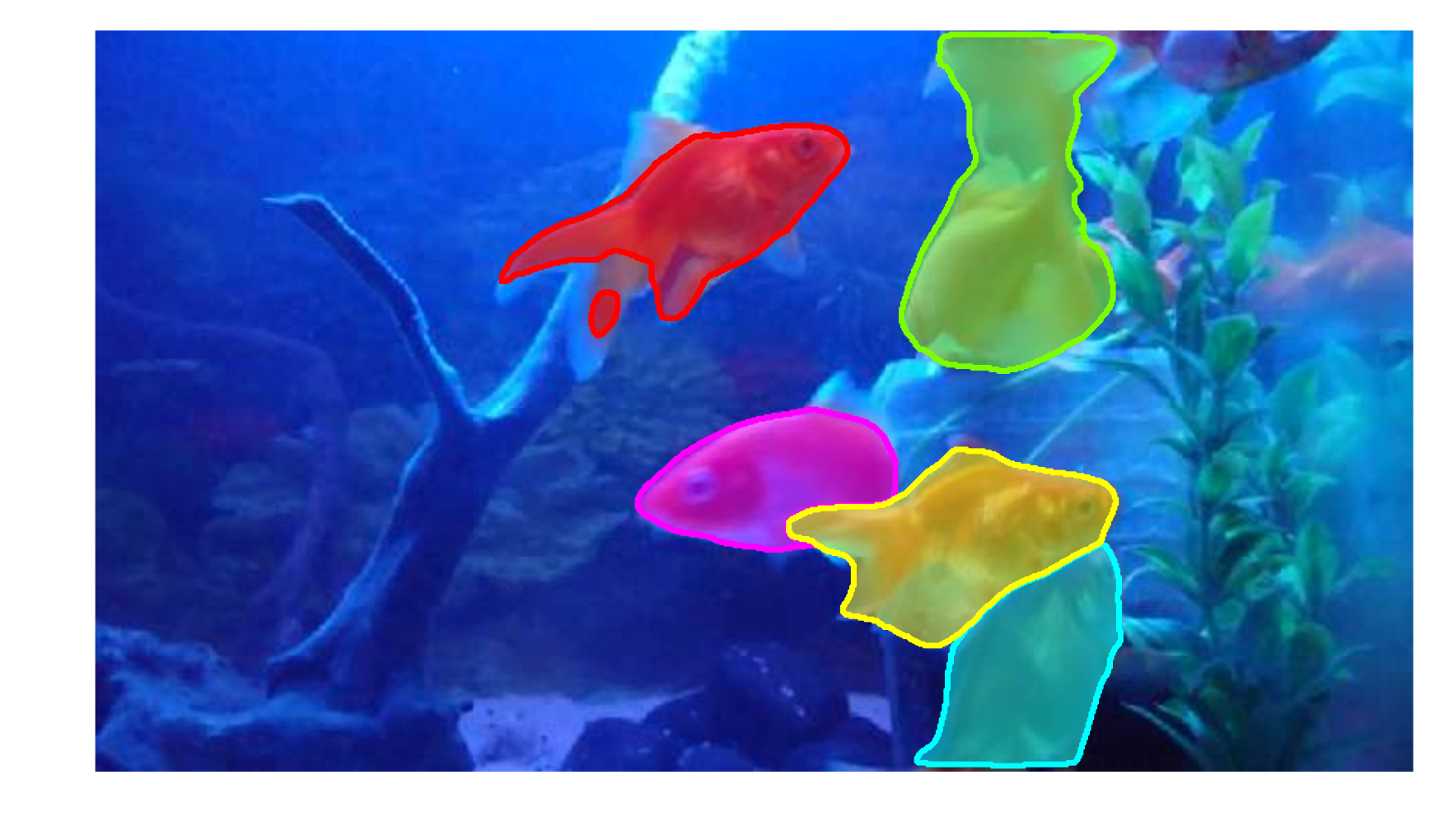}
\\

\end{tabular}

\caption{Further qualitative results of our method on sequences from the semi-supervised video object segmentation benchmarks DAVIS-2016~\cite{perazzi2016benchmark} and DAVIS-2017~\cite{pont2017davis}.
Multiple masks are obtained from different inferences (with different initialisations).
}
\label{fig:appendix_davis16}
\end{figure*}

\mypar{Benchmark sequences.}
More qualitative results for VOT and DAVIS sequences are shown in Figure~\ref{fig:appendix_vot18} and~\ref{fig:appendix_davis16}.

\end{document}